\documentclass[a4paper, 11pt]{book}

\title{Learning to Represent and Predict Sets with Deep Neural Networks}
\author{Yan Zhang}

\usepackage{import}
\usepackage[utf8]{inputenc} 
\usepackage[T1]{fontenc}    
\usepackage[pdfusetitle]{hyperref}       
\usepackage{url}            
\usepackage{booktabs}       
\usepackage{tabularx}       
\usepackage{amsmath,bm,amsthm,mathtools}
\usepackage{amsfonts}       
\usepackage{nicefrac}       
\usepackage{titling}

\usepackage{graphicx}

\usepackage[chapter]{algorithm}
\usepackage{algorithmicx}
\usepackage{algpseudocode}

\usepackage{xcolor}

\usepackage{float}

\usepackage[%
    backend=biber,
    natbib=true,
    style=ieee,
    citestyle=numeric,
    giveninits=false,
    sorting=nty,
    abbreviate=false,
    dashed=false,
    mincitenames=1,
    maxcitenames=1,
    maxbibnames=99,
]{biblatex}
\usepackage{csquotes}
\usepackage{capt-of}

\usepackage[english]{babel}
\usepackage{textcomp}

\usepackage[listings,theorems]{tcolorbox}

\makeatletter
\let\mytitle\@title
\let\myname\@author
\let\mydate\@date
\makeatother

\usepackage{acro}  
\acsetup{first-style=short, list/display = all}

\usepackage{setspace}
\usepackage{microtype}

\usepackage[ttscale=0.875]{libertine}
\usepackage[libertine]{newtxmath}

\usepackage[nottoc]{tocbibind}

\usepackage[
    left=40mm,
    right=40mm,
    top=15mm,
    bottom=15mm,
    bindingoffset=20mm,
    marginparwidth=0pt,
    marginparsep=0pt,
    includehead,
]{geometry}

\usepackage[parfill]{parskip}

\addto\captionsenglish{%

}

\tcbset{after={\parfillskip 0pt plus 1fil\par}}
\tcbuselibrary{theorems, skins}
\newtcolorbox[auto counter, number within=chapter, number freestyle={\noexpand\thechapter.\noexpand\arabic{\tcbcounter}}]{myexample}[1][]{%
    fonttitle=\bfseries,
    colframe=white,
    title filled=true,
    sharp corners=all,
    parbox=false,
    title=Example~\thetcbcounter,
    #1
}
\newtcolorbox[]{sidenote}[2][]{%
    fonttitle=\bfseries,
    colframe=white,
    title filled=true,
    sharp corners=all,
    parbox=false,
    title=#2,
    #1
}

\usepackage{fancyhdr}
\newlength{\myfigureoffset}
\newcommand{\myheaderstyle}{\fontsize{9}{11}\selectfont \slshape \nouppercase}
\fancypagestyle{main}{%
    \fancyhf{}
    \fancyhead[LE, RO]{\myheaderstyle\thepage} 
    \fancyhead[CO]{\myheaderstyle\leftmark} 
    \fancyhead[CE]{\myheaderstyle\rightmark} 
    \fancyhfoffset[LH,RH]{\myfigureoffset}  
}
\fancypagestyle{front}{%
    \fancyhf{}
    \fancyhead[LE, RO]{\myheaderstyle\thepage} 
    \fancyhfoffset[LH,RH]{\myfigureoffset}  
}
\fancypagestyle{blank}{%
    \fancyhf{}
}
\renewcommand{\headrulewidth}{0pt}  
\usepackage{etoolbox}
\patchcmd{\chapter}{\thispagestyle{plain}}{\thispagestyle{front}}{}{}
\appto\frontmatter{\pagestyle{front}}
\appto\mainmatter{\pagestyle{main}}
\appto\backmatter{\pagestyle{front}}

\renewbibmacro{in:}{%
  \ifentrytype{article}{}{\printtext{\bibstring{in}\intitlepunct}}}
\AtEveryBibitem{\clearfield{month}}
\AtEveryBibitem{\clearfield{day}}
\DefineBibliographyStrings{english}{%
    andothers = {{et al}\adddot},
    url = {Available at},
}

\newcommand{\centerfloat}[1]{\makebox[\textwidth]{#1}}

\newlength{\figurewidth}
\usepackage[
    width=\figurewidth, 
    tableposition=top,
    figureposition=bottom,
]{caption}

\usepackage{datetime}

\newdateformat{monthyeardate}{%
  \monthname[\THEMONTH], \THEYEAR}

\usepackage{xpatch}
\makeatletter
\AtBeginDocument{\xpatchcmd{\@thm}{\thm@headpunct{.}}{\thm@headpunct{~~}}{}{}}
\makeatother
\usepackage{tikz}
\usepackage{xcolor}
\usepackage{gensymb}

\usetikzlibrary{shapes, backgrounds, fit, calc, positioning, arrows.meta, bending, decorations.pathmorphing}

\definecolor{RdYlBu-4-1}{RGB}{215,25,28}
\definecolor{RdYlBu-4-2}{RGB}{253,174,97}
\definecolor{RdYlBu-4-3}{RGB}{171,217,233}
\definecolor{RdYlBu-4-4}{RGB}{44,123,182}
\definecolor{Blues-4-1}{RGB}{239,243,255}
\definecolor{Blues-4-2}{RGB}{189,215,231}
\definecolor{Blues-4-3}{RGB}{107,174,214}
\definecolor{Blues-4-4}{RGB}{33,113,181}
\definecolor{Oranges-4-1}{RGB}{254,237,222}
\definecolor{Oranges-4-2}{RGB}{253,190,133}
\definecolor{Oranges-4-3}{RGB}{253,141,60}
\definecolor{Oranges-4-4}{RGB}{217,71,1}
\definecolor{Purples-4-1}{RGB}{242,240,247}
\definecolor{Purples-4-2}{RGB}{203,201,226}
\definecolor{Purples-4-3}{RGB}{158,154,200}
\definecolor{Purples-4-4}{RGB}{106,81,163}

\definecolor{yes}{RGB}{0, 0, 0}
\definecolor{no}{RGB}{255, 255, 255}
\definecolor{maybe}{RGB}{128, 128, 128}  
\definecolor{grey}{RGB}{224, 224, 224}
\definecolor{bg}{RGB}{250, 250, 250}
\definecolor{blue2}{HTML}{004CDB}
\definecolor{lblue}{HTML}{74c9e5}
\definecolor{red2}{HTML}{FF4500}
\definecolor{yellow}{HTML}{FFb31a}
\definecolor{ccc1}{HTML}{3a913a}
\definecolor{ccc2}{HTML}{c03c3d}
\definecolor{ccc3}{HTML}{9371b2}
\definecolor{ccc4}{HTML}{e0802c}
\definecolor{cc1}{HTML}{1fad1f}
\definecolor{cc2}{HTML}{d42525}
\definecolor{cc3}{HTML}{9446dd}
\definecolor{cc4}{HTML}{ee8020}
\colorlet{c1}{cc1!80!white}
\colorlet{c2}{cc2!80!white}
\colorlet{c3}{cc3!80!white}
\colorlet{c4}{cc4!80!white}

\definecolor{colorOne}{HTML}{002e3b}    
\definecolor{colorTwo}{HTML}{005c84}    
\definecolor{colorThree}{HTML}{fcbc00}

\newcommand{\AxisRotator}[1][rotate=0]{%
    \tikz [x=1.1mm,y=1.1mm,line width=.1ex,>={Latex[length=.3em, bend]},#1] \draw (0,0) arc (-150:170:1);%
}

\def\centerarc[#1](#2)(#3:#4:#5)
    { \draw[#1] ($(#2)+({#5*cos(#3)},{#5*sin(#3)})$) arc (#3:#4:#5); }

\makeatletter
\newsavebox{\measure@tikzpicture}
\tikzset{
    every picture/.append style={x=5mm, y=5mm},
    circ/.style={circle, draw=black, fill=black, minimum size=2.5mm, inner sep=0mm},
    box/.style={rectangle,draw=black,thick, minimum size=5mm},
    object/.style={draw, dotted, inner sep=0.3em, fill=black!3, rounded corners=1mm},
    desc/.style={rectangle},
    sedge/.style={->, >=stealth},
    varname/.style={midway, above, opacity=1},
    ledge/.style={->, >=stealth, loop, looseness=7},
    dedge/.style={<->, >=stealth},
    imedge/.style={->, >=stealth, draw=yellow},
    node/.style={circle, draw, fill=grey},
    varname/.style={midway, above, opacity=1},
    count-object/.style={rounded corners=0.8em, draw, dotted, inner sep=1em, fill=bg},
    count-box/.style={
        draw=yellow, line width=0.35mm},
    block rect/.style={
        box, rectangle},
    block/.style={
        block rect, minimum height=0.8cm, minimum width=6em},
    from/.style args={#1 to #2}{
        above right={0cm of #1},
        /utils/exec=\pgfpointdiff
            {\tikz@scan@one@point\pgfutil@firstofone(#1)\relax}
            {\tikz@scan@one@point\pgfutil@firstofone(#2)\relax},
        minimum width/.expanded=\the\pgf@x,
        minimum height/.expanded=\the\pgf@y},
    w/.style = {circle, draw, fill=white, inner sep=0mm, minimum size=3.5mm, thick},
    params/.style = {rectangle, rounded corners=1mm, draw, fill=Blues-4-2, inner sep=1mm, thick},
    no params/.style = {rectangle, rounded corners=1mm, draw, fill=grey, inner sep=1mm, thick},
    symmetry/.pic = {
        \pic at (0, 0) {square};
        \begin{scope}[on background layer]
            \node (s1) [object, fit=(a) (b) (c) (d)] {};
        \end{scope}

        \pic at (13, 0) {square};
        \begin{scope}[on background layer]
            \node (s4) [object, fit=(a) (b) (c) (d)] {};
        \end{scope}

        \pic [rotate=-30] at (4, -3) {square};
        \begin{scope}[on background layer]
            \node (s2) [object, fit=(a) (b) (c) (d)] {};
        \end{scope}

        \pic [rotate=-30+90] at (9, -3) {square};
        \begin{scope}[on background layer]
            \node (s3) [object, fit=(a) (b) (c) (d)] {};
        \end{scope}

        \draw [sedge] (s1) -- (s4) node [above, midway] {\raisebox{-0.3mm}{\AxisRotator[xscale=-1]} $90\degree$};
        \draw [sedge, draw=red] (s2) -- (s3) node [below, midway] {\raisebox{-0.5mm}{\AxisRotator[xscale=-1]} $\epsilon$};
        \draw [sedge] (s1.south) to[bend right] node [below=1mm, midway] {\raisebox{-0.5mm}{\AxisRotator[xscale=-1]} $\alpha$} (s2.west);
        \draw [sedge, text width=1.8cm] (s3.east) to[bend right] node [below=4.5mm, pos=0.8] {\raisebox{-0.5mm}{\AxisRotator[xscale=-1]} $90\degree - \alpha - \epsilon$} (s4.south);
    },
    square/.pic = {
        \node (a) [circ, fill=RdYlBu-4-1] at (-0.8, 0.8) {};
        \node (b) [circ, fill=RdYlBu-4-2] at (0.8, 0.8) {};
        \node (c) [circ, fill=RdYlBu-4-3] at (0.8, -0.8) {};
        \node (d) [circ, fill=RdYlBu-4-4] at (-0.8, -0.8) {};
    },
    square2/.pic = {
        \node (a) [circ, fill=Purples-4-4!80!black] at (-0.8, 0.8) {};
        \node (b) [circ, fill=Purples-4-4!80!black] at (0.8, 0.8) {};
        \node (c) [circ, fill=Purples-4-4!80!black] at (0.8, -0.8) {};
        \node (d) [circ, fill=Purples-4-4!80!black] at (-0.8, -0.8) {};
    },
    square3/.pic = {
        \node (a) [circ, fill=Blues-4-4] at (-0.8, 0.8) {};
        \node (b) [circ, fill=Blues-4-4] at (0.8, 0.8) {};
        \node (c) [circ, fill=Blues-4-4] at (0.8, -0.8) {};
        \node (d) [circ, fill=Blues-4-4] at (-0.8, -0.8) {};
    },
    model/.pic = {
        \pic (inputs) at (0, 0) {inputs};
        \pic (sort) at (8, 0) {sorted};
        \pic [scale=1.25, every node/.style={transform shape}] (output) at (12.75, 0) {output};
        \pic (dweights) at (16, 0) {dweights};
        \pic (cweights) at (20.5, 0) {cweights};

        \node at (2.5, 1.5) {$\mX$};

        \node at (6.35, -1.2) {\scriptsize sort};
        \draw [sedge] (5.6,  0.0) -- (7.2,  0.0);
        \draw [sedge] (5.6,  0.7) -- (7.2,  0.7);
        \draw [sedge] (5.6, -0.7) -- (7.2, -0.7);

        \node at (8.75, 1.5) {$\mvX$};

        \draw [sedge] (10.2,  0.0) -- (12,  0.0);
        \draw [sedge] (10.2,  0.7) -- (12,  0.7);
        \draw [sedge] (10.2, -0.7) -- (12, -0.7);

        \node at (12.75, 1.4) {$\vy$};
        \node at (12.75, -1.35) {\scriptsize dot product};

        \draw [sedge] (15.2,  0.0) -- (13.5,  0.0);
        \draw [sedge] (15.2,  0.7) -- (13.5,  0.7);
        \draw [sedge] (15.2, -0.7) -- (13.5, -0.7);

        \node at (17.0, 1.4) {\scriptsize $f([\scriptscriptstyle 0, \frac{1}{3}, \frac{2}{3}, 1 \displaystyle], \mbW)$};

        \node at (19.2, -1.2) {\scriptsize discretise};

        \node at (23, 1.4) {\scriptsize $f(\cdot, \mbW)$};

        \draw [sedge] (20,  0.0) -- (18.2,  0.0);
        \draw [sedge] (20,  0.7) -- (18.2,  0.7);
        \draw [sedge] (20, -0.7) -- (18.2, -0.7);
    },
    sorted/.pic = {
        \node [box, scale=0.5, fill=Blues-4-4] at (0.0, 0.7) {};
        \node [box, scale=0.5, fill=Blues-4-3] at (0.5, 0.7) {};
        \node [box, scale=0.5, fill=Blues-4-2] at (1.0, 0.7) {};
        \node [box, scale=0.5, fill=Blues-4-1] at (1.5, 0.7) {};

        \node [box, scale=0.5, fill=Oranges-4-4] at (0.0, 0.0) {};
        \node [box, scale=0.5, fill=Oranges-4-3] at (0.5, 0.0) {};
        \node [box, scale=0.5, fill=Oranges-4-2] at (1.0, 0.0) {};
        \node [box, scale=0.5, fill=Oranges-4-1] at (1.5, 0.0) {};

        \node [box, scale=0.5, fill=Purples-4-4] at (0.0, -0.7) {};
        \node [box, scale=0.5, fill=Purples-4-3] at (0.5, -0.7) {};
        \node [box, scale=0.5, fill=Purples-4-2] at (1.0, -0.7) {};
        \node [box, scale=0.5, fill=Purples-4-1] at (1.5, -0.7) {};
    },
    inputs/.pic = {
        \node at (0, 0) {$\Bigg\{$};
        \node at (5, 0) {$\Bigg\}$};

        \node [box, scale=0.625, fill=Blues-4-3] at (1, 0.625) {};
        \node [box, scale=0.625, fill=Oranges-4-4] at (1, 0.0) {};
        \node [box, scale=0.625, fill=Purples-4-3] at (1, -0.625) {};

        \node [box, scale=0.625, fill=Blues-4-2] at (2, 0.625) {};
        \node [box, scale=0.625, fill=Oranges-4-2] at (2, 0.0) {};
        \node [box, scale=0.625, fill=Purples-4-1] at (2, -0.625) {};

        \node [box, scale=0.625, fill=Blues-4-1] at (3, 0.625) {};
        \node [box, scale=0.625, fill=Oranges-4-3] at (3, 0.0) {};
        \node [box, scale=0.625, fill=Purples-4-4] at (3, -0.625) {};

        \node [box, scale=0.625, fill=Blues-4-4] at (4, 0.625) {};
        \node [box, scale=0.625, fill=Oranges-4-1] at (4, 0.0) {};
        \node [box, scale=0.625, fill=Purples-4-2] at (4, -0.625) {};
    },
    dweights/.pic = {
        \node [box, scale=0.5, fill=Blues-4-3] at (0.0, 0.7) {};
        \node [box, scale=0.5, fill=Blues-4-3] at (0.5, 0.7) {};
        \node [box, scale=0.5, fill=Blues-4-3] at (1.0, 0.7) {};
        \node [box, scale=0.5, fill=Blues-4-3] at (1.5, 0.7) {};

        \node [box, scale=0.5, fill=Oranges-4-2] at (0.0, 0.0) {};
        \node [box, scale=0.5, fill=Oranges-4-3] at (0.5, 0.0) {};
        \node [box, scale=0.5, fill=Oranges-4-3] at (1.0, 0.0) {};
        \node [box, scale=0.5, fill=Oranges-4-2] at (1.5, 0.0) {};

        \node [box, scale=0.5, fill=Purples-4-1] at (0.0, -0.7) {};
        \node [box, scale=0.5, fill=Purples-4-1] at (0.5, -0.7) {};
        \node [box, scale=0.5, fill=Purples-4-3] at (1.0, -0.7) {};
        \node [box, scale=0.5, fill=Purples-4-4] at (1.5, -0.7) {};
    },
    cweights/.pic = {
        \draw [very thick] (0, 0) -- (5, 0);
        \draw [very thick, sedge] (0, -1) -- (0, 1.4);
        \draw [very thick] (5, -1) -- (5, 1.4);

        \draw [thick, draw=Blues-4-3] (0, 0.5) -- (2.5, 0.5) -- (5, 0.5);
        \draw [thick, draw=Oranges-4-3] (0, 0) -- (2.5, 1) -- (5, 0);
        \draw [thick, draw=Purples-4-3] (0, -0.5) -- (2.5, -0.5) -- (5, 1);

        \node [circle, scale=0.25, fill=Oranges-4-4] at (0, 0) {};
        \node [circle, scale=0.25, fill=Blues-4-4] at (0, 0.5) {};
        \node [circle, scale=0.25, fill=Purples-4-4] at (0, -0.5) {};
        \node [circle, scale=0.25, fill=Oranges-4-4] at (1.667, 0.666) {};
        \node [circle, scale=0.25, fill=Blues-4-4] at (1.667, 0.5) {};
        \node [circle, scale=0.25, fill=Purples-4-4] at (1.667, -0.5) {};
        \node [circle, scale=0.25, fill=Oranges-4-4] at (3.333, 0.666) {};
        \node [circle, scale=0.25, fill=Blues-4-4] at (3.333, 0.5) {};
        \node [circle, scale=0.25, fill=Purples-4-4] at (3.333, 0) {};
        \node [circle, scale=0.25, fill=Purples-4-4] at (5, 1) {};
        \node [circle, scale=0.25, fill=Blues-4-4] at (5, 0.5) {};
        \node [circle, scale=0.25, fill=Oranges-4-4] at (5, 0) {};
    },
    output/.pic = {
        \node [box, scale=0.5, fill=Blues-4-3!50!Blues-4-2] at (0, 0.5) {};
        \node [box, scale=0.5, fill=Oranges-4-3!50!Oranges-4-2] at (0, 0.0) {};
        \node [box, scale=0.5, fill=Purples-4-1] at (0, -0.5) {};
    },
    ae/.pic = {
        \node (l) at (0, 0) {\includegraphics[width=2cm]{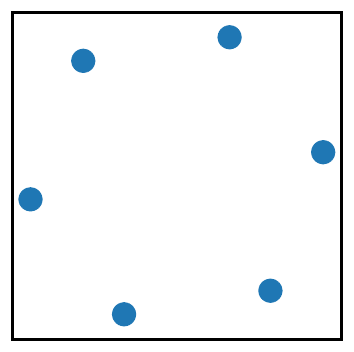}};
        \node (enc) [trapezium, draw, rotate=-90, inner ysep=3.5mm, trapezium angle=70] at (3.5, 0) {\scriptsize Encoder};
        \node [box, scale=0.5, fill=Blues-4-3] at (5.5, 0.5) {};
        \node (fv) [box, scale=0.5, fill=Oranges-4-3] at (5.5, 0) {};
        \node [box, scale=0.5, fill=Purples-4-3] at (5.5, -0.5) {};
        \node (dec) [trapezium, draw, rotate=90, inner ysep=3.5mm, trapezium angle=70] at (7.5, 0) {\scriptsize Decoder};
        \node (r) at (11, 0) {\includegraphics[width=2cm]{hexagon}};

        \draw [sedge] ([xshift=-2mm]l.east) -- (enc);
        \draw [sedge] (enc) -- (fv);
        \draw [sedge] (fv) -- (dec);
        \draw [sedge] (dec) -- ([xshift=2mm]r.west);

        \centerarc[<->,>=stealth](0.05,0)(-65+180:-05+180:1.3);
    },
    overview/.pic = {
        \pic at (0, 0) {perm fv};
        \pic at (6, 0) {ordercost};
        \pic at (12, 0) {total cost};
        \pic at (6, -5) {p progression};
        \pic at (19, -5) {result};

        \node [desc] at (0, -3.1) {$\mX$};
        \node [desc] at (7, -3.1) {$\mC$};

        \draw [sedge] (1, -1) -- (4, -1) node [varname, desc, inner sep=3] {$F$} node [midway, below, scale=0.7, inner sep=3] {(learnable)};
        \draw [sedge] (9, -1) -- (10.75, -1) {};
        \draw [sedge, purple, shorten <= 2mm] (c) to[out=-90, in=90, looseness=1.8] (0to1);
        \draw [sedge, purple, shorten <= 2mm] (c) to[out=-90, in=90, looseness=1.8] (1to2);

        \begin{scope}[on background layer]
            \node [object, fit=(p2) (fv result)] {};
        \end{scope}
    },
    result/.pic = {
        \node (dot) at (0, -1) {$\cdot$};
        \pic at (1, 0) {perm fv};
        \node [desc] at (1, -3.2) {$\mX$};
        \node (eq) at (2.5, -1) {$=$};
        \pic at (4, 0) {fv result};
        \node [desc] at (4, -3.2) {$\mY$};
    },
    p progression/.pic = {
        \pic (p0) at (0, 0) {p0};
        \pic (p1) at (5, 0) {p1};
        \begin{scope}[local bounding box=p2]
            \pic (p2) at (10, 0) {p2};
        \end{scope}

        \node at (-4.2, -0.5) {improve permutation};
        \node at (-4.2, -1.5) {by minimising cost:};

        \draw [sedge] (3, -1) -- (4, -1) node [midway] (0to1) {};
        \draw [sedge] (8, -1) -- (9, -1) node [midway] (1to2) {};

        \node [desc] at (1, -3.1) {$\mP^{(0)}$};
        \node [desc] at (6, -3.1) {$\mP^{(1)}$};
        \node [desc] at (11, -3.1) {$\mP^{(T)}$};
    },
    total cost/.pic = {
        \node [box, fill=black!10, draw=white, inner sep=3] (c) at (0, -1) {$c(\mP)$};
        \node [purple] at (1.75, -3) {$- \frac{\partial c(\mP)}{\partial \mP}$};
        \node at (5.5, -0.5) {function to measure cost};
        \node at (5.5, -1.5) {of permutation with};
    },
    ordercost/.pic = {
        \pic at (-1.33, 0) {fv2};
        \pic at (0, 1.33) {fvT};

        \node [box] at (0, 0) {};
        \node [box, fill=blue!40] at (1, 0) {$<$};
        \node [box, fill=blue!40] at (2, 0) {$<$};
        \node [box, fill=red!40] at (0, -1) {$>$};
        \node [box] at (1, -1) {};
        \node [box, fill=red!40] at (2, -1) {$>$};
        \node [box, fill=red!40] at (0, -2) {$>$};
        \node [box, fill=blue!40] at (1, -2) {$<$};
        \node [box] at (2, -2) {};
    },
    p0/.pic = {
        \node [box, fill=black!33] at (0, 0) {};
        \node [box, fill=black!33] at (1, 0) {};
        \node [box, fill=black!33] at (2, 0) {};
        \node [box, fill=black!33] at (0, -1) {};
        \node [box, fill=black!33] at (1, -1) {};
        \node [box, fill=black!33] at (2, -1) {};
        \node [box, fill=black!33] at (0, -2) {};
        \node [box, fill=black!33] at (1, -2) {};
        \node [box, fill=black!33] at (2, -2) {};
    },
    p1/.pic = {
        \node [box, fill=black!86] at (0, 0) {};
        \node [box, fill=black!14] at (1, 0) {};
        \node [box, fill=black!0] at (2, 0) {};
        \node [box, fill=black!0] at (0, -1) {};
        \node [box, fill=black!12] at (1, -1) {};
        \node [box, fill=black!88] at (2, -1) {};
        \node [box, fill=black!14] at (0, -2) {};
        \node [box, fill=black!75] at (1, -2) {};
        \node [box, fill=black!13] at (2, -2) {};
    },
    p2/.pic = {
        \node [box, fill=black!100] at (0, 0) {};
        \node [box, fill=black!0] at (1, 0) {};
        \node [box, fill=black!0] at (2, 0) {};
        \node [box, fill=black!0] at (0, -1) {};
        \node [box, fill=black!0] at (1, -1) {};
        \node [box, fill=black!100] at (2, -1) {};
        \node [box, fill=black!0] at (0, -2) {};
        \node [box, fill=black!100] at (1, -2) {};
        \node [box, fill=black!0] at (2, -2) {};
    },
    perm fv/.pic = {
        \node [box, fill=white] at (0, 0) {1};
        \node (fv) [box, fill=white] at (0, -1) {3};
        \node [box, fill=white] at (0, -2) {2};
    },
    fv result/.pic = {
        \node [box, fill=white] at (0, 0) {\textbf{1}};
        \node (fv result) [box, fill=white] at (0, -1) {\textbf{2}};
        \node [box, fill=white] at (0, -2) {\textbf{3}};
    },
    fv2/.pic = {
        \node (x) [box, fill=white, draw=white] at (0, 0) {1};
        \node (fv) [box, fill=white, draw=white] at (0, -1) {3};
        \node (z) [box, fill=white, draw=white] at (0, -2) {2};
        \draw (x.north east) -- (z.south east);
    },
    fvT/.pic = {
        \node (a) [box, fill=white, draw=white] at (0, 0) {1};
        \node (b) [box, fill=white, draw=white] at (1, 0) {3};
        \node (c) [box, fill=white, draw=white] at (2, 0) {2};
        \draw (a.south west) -- (c.south east);
    },
    arch classify/.pic = {
        \node (open) {\rotatebox{-90}{$\Bigg\{$}};
        \node [below = -2mm of open] (tiles) {\includegraphics[height=4cm]{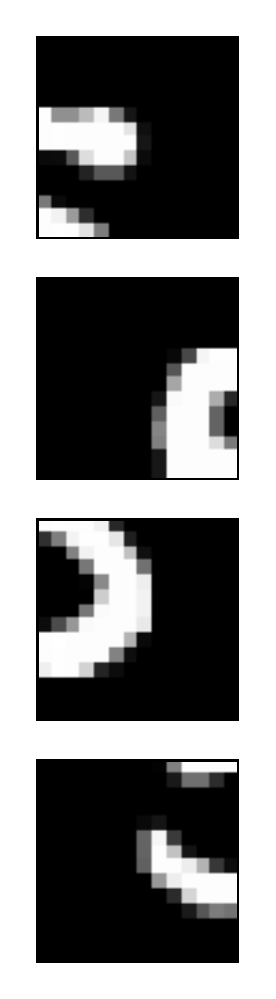}};
        \node [below = -2mm of tiles] (close) {\rotatebox{-90}{$\Bigg\}$}};

        \node (cnn1) [fill=yellow!10, trapezium, trapezium angle=80, minimum width=10mm, thick, draw, anchor=north, shape border rotate=270] at (4, -0.75) {CNN};
        \node (cnn2) [fill=yellow!10, trapezium, trapezium angle=80, minimum width=10mm, draw, anchor=north, shape border rotate=270, text=white, dashed] at (4, -2.625) {CNN};
        \node (cnn3) [fill=yellow!10, trapezium, trapezium angle=80, minimum width=10mm, draw, anchor=north, shape border rotate=270, text=white, dashed] at (4, -4.5) {CNN};
        \node (cnn4) [fill=yellow!10, trapezium, trapezium angle=80, minimum width=10mm, draw, anchor=north, shape border rotate=270, text=white, dashed] at (4, -6.375) {CNN};

        \node (F box) [rectangle, draw, minimum width=1.5cm, minimum height=3.5cm] at (9, -4.3) {};
        \node (mlp1) [scale=0.8, rectangle, draw, thick, fill=yellow!10] at (9, -2.5) {MLP};
        \node [above = 1mm of mlp1] {$F$};
        \node (mlp2) [scale=0.8, rectangle, draw, dashed, text=white, fill=yellow!10] at (9, -3.5) {MLP};
        \node [] at (9, -4.5) {\vdots};
        \node (mlp3) [scale=0.8, rectangle, draw, dashed, text=white, fill=yellow!10] at (9, -7) {MLP};

        \node (PO box) [rectangle, draw, minimum width=1.5cm, minimum height=2cm, very thick, fill=green!10] at (15.5, -4.3) {\textbf{PO}};

        \node (resnet) [trapezium, trapezium angle=80, minimum width=14mm, thick, draw, anchor=north, shape border rotate=270, fill=yellow!10] at (21, -3.3) {ResNet-18};

        \draw (tiles.east |- cnn1.west) -- (cnn1.west);
        \draw (tiles.east |- cnn2.west) -- (cnn2.west);
        \draw (tiles.east |- cnn3.west) -- (cnn3.west);
        \draw (tiles.east |- cnn4.west) -- (cnn4.west);

        \draw [sedge] (cnn1.east) -- (F box.west |- cnn1);
        \draw [sedge] (cnn2.east) -- (F box.west |- cnn2);
        \draw [sedge] (cnn3.east) -- (F box.west |- cnn3);
        \draw [sedge] (cnn4.east) -- (F box.west |- cnn4);

        \draw [gray] (F box.west |- cnn1) -- ([yshift=0.3mm]mlp1.west);
        \draw [gray] (F box.west |- cnn1) -- ([yshift=-0.3mm]mlp1.west);
        \draw [gray] (F box.west |- cnn1) -- ([yshift=0.3mm]mlp2.west);
        \draw [gray] (F box.west |- cnn2) -- ([yshift=-0.3mm]mlp2.west);
        \draw [gray] (F box.west |- cnn4) -- ([yshift=0.3mm]mlp3.west);
        \draw [gray] (F box.west |- cnn4) -- ([yshift=-0.3mm]mlp3.west);

        \draw [sedge] (F box.east |- cnn2) -- (PO box.west |- cnn2) node [varname] {\tiny $\mC_{\text{row}}$};
        \draw [sedge] (F box.east |- cnn3) -- (PO box.west |- cnn3) node [varname] {\tiny $\mC_{\text{col}}$};

        \draw [sedge] (PO box.east) -- (resnet.west) node [varname] {};

        \draw [sedge] (open.north) to [bend left=20] (PO box.north);

        \draw [sedge] (resnet.east) -- +(1, 0);
    },
    arch sort/.pic = {
        \node (cnn1) [] at (4, -1.75) {0.3};
        \node (cnn2) [] at (4, -3.50) {0.8};
        \node (cnn3) [] at (4, -5.25) {0.7};
        \node (cnn4) [] at (4, -7) {0.1};

        \node [above = 0mm of cnn1] (open) {\rotatebox{-90}{$\Bigg\{$}};
        \node [below = 0mm of cnn4] (close) {\rotatebox{-90}{$\Bigg\}$}};

        \node (a1) [] at (18, -1.75) {0.1};
        \node [] at (18, -3.50) {0.3};
        \node [] at (18, -5.25) {0.7};
        \node (a4) [] at (18, -7) {0.8};

        \node [above = 0mm of a1] {\rotatebox{-90}{$\Bigg[$}};
        \node [below = 0mm of a4] {\rotatebox{-90}{$\Bigg]$}};

        \node (F box) [rectangle, draw, minimum width=1.5cm, minimum height=3.5cm] at (9, -4.3) {};
        \node (mlp1) [fill=yellow!10, rectangle, draw, thick] at (9, -2.5) {\footnotesize MLP};
        \node [above = 1mm of mlp1] {$F$};
        \node (mlp2) [fill=yellow!10, rectangle, draw, dashed, text=white] at (9, -3.5) {\footnotesize MLP};
        \node [] at (9, -4.5) {\Large\vdots};
        \node (mlp3) [fill=yellow!10, rectangle, draw, dashed, text=white] at (9, -7) {\footnotesize MLP};

        \node (PO box) [rectangle, draw, minimum width=1.5cm, minimum height=2cm, very thick, fill=green!10] at (14, -4.3) {PO};

        \draw [sedge] ([xshift=3mm]cnn1.east) -- (F box.west |- cnn1);
        \draw [sedge] ([xshift=3mm]cnn2.east) -- (F box.west |- cnn2);
        \draw [sedge] ([xshift=3mm]cnn3.east) -- (F box.west |- cnn3);
        \draw [sedge] ([xshift=3mm]cnn4.east) -- (F box.west |- cnn4);

        \draw [gray] (F box.west |- cnn1) -- ([yshift=0.3mm]mlp1.west);
        \draw [gray] (F box.west |- cnn1) -- ([yshift=-0.3mm]mlp1.west);
        \draw [gray] (F box.west |- cnn1) -- ([yshift=0.3mm]mlp2.west);
        \draw [gray] (F box.west |- cnn2) -- ([yshift=-0.3mm]mlp2.west);
        \draw [gray] (F box.west |- cnn4) -- ([yshift=0.3mm]mlp3.west);
        \draw [gray] (F box.west |- cnn4) -- ([yshift=-0.3mm]mlp3.west);

        \draw [sedge] (F box.east) -- (PO box.west) node [varname] {$\mC$};

        \draw [sedge] (PO box.east) -- +(1, 0);

        \draw [sedge] (open.north) to [bend left=25] (PO box.north);
    },
    arch ban/.pic = {
        \node (v) at (0.5, -1.4) {Object proposals};
        \node (q) at (0.5, -2.65) {Question};

        \node (ban box) [rectangle, draw, minimum height = 2.5cm, minimum width = 10cm, thick, fill=blue2!7] at (14, -1) {};
        \node [below = 1mm of ban box.south] {BAN};
        \node (glimpse) [rectangle, draw, minimum height = 1cm, align=center, fill=white, inner sep=3] at (7.2, -2) {Attention glimpse};
        \node (glimpse2) [rectangle, draw, minimum height = 1cm, align=center, fill=white, inner sep=3] at (18, -2) {Attention glimpse};
        \node (dot) [right = 0cm of glimpse2, minimum height = 1cm, align=center, text width = 1.2cm] {$\cdots$};
        \node (dot2) [minimum height = 1cm, align=center, text width = 1.2cm] at (20.5, 2.8) {$\cdots$};

        \node (f) [rectangle, draw, minimum height = 8mm, minimum width = 10mm, very thick, fill=green!10, inner sep=4] at (10, 2.8) {$F$ $\to$ \textbf{PO}};
        \node (lstm) [rectangle, draw, minimum height = 8mm, minimum width = 10mm, thick, fill=yellow!10] at (13.5, 2.8) {LSTM};

        \draw [sedge, very thick, densely dotted] ([yshift=8mm]ban box.west) -- ([yshift=8mm] ban box.east) node [pos=0.25] (p1) {} node [pos=0.8] (p2) {} node [pos=0.67] (p3) {};

        \draw [very thick, densely dotted] (glimpse) to[out=90, in=180] (p1);
        \draw [very thick, densely dotted] (glimpse2) to[out=90, in=180] (p2);
        \draw [very thick, densely dotted] (lstm) to[out=0, in=180] (p3);

        \draw [sedge] ([xshift=-3mm]glimpse.north) to[out=90, in=180] ([yshift=-1mm]f.west);
        \draw [sedge] ([xshift=-3mm]glimpse2.north) to[out=90, in=180] ([yshift=-1mm]19.2, 2.8);
        \draw [sedge] (v.north) to[out=90, in=180] ([yshift=1mm]f.west);
        \draw [dotted] ([yshift=1mm]17, 2.8) -- +(0.5, 0);
        \draw [sedge] ([yshift=1mm]17.5, 2.8) -- ([yshift=1mm]19.2, 2.8);
        \draw [sedge] (f) -- (lstm);

        \draw [sedge] (v) -- (ban box.west |- v);
        \draw [sedge] (q) -- (ban box.west |- q);

    },
    nodes/.pic={
        \node [node, fill=yes] (tl) at (0, 0) {};
        \node [node, fill=no] (tr) at (0, -1) {};
        \node [node, fill=yes] (bl) at (1, 0) {};
        \node [node, fill=yes] (br) at (1, -1) {};

        \begin{scope}[on background layer]
            \node [count-object, fit=(bl) (br)] {};
            \node [count-object, fit=(tl)] {};
            \node [count-object, fit=(tr)] {};
        \end{scope}
    },
    a/.pic={
        \pic {nodes};
        \coordinate (aux1) at ($(tl)!0.46!(tr)$);
        \coordinate (aux2) at ($(tl)!0.54!(tr)$);
        \node at (-2, -0.5)  {
            \includegraphics[width=3.5cm]{COCO_val2014_000000121112}
        };
        \node (c11) [count-box, from={-1.55,-0.66 to -0.97,-0.13}] {};
        \node (c12) [count-box, from={-1.59,-0.70 to -0.93, -0.09}] {};
        \node (c2) [count-box, from={-2.72,-0.70 to -2.22, -0.16}] {};
        \node (k) [count-box, from={-2.45,-1.22 to -0.859, -0.83}] {};
        \draw [imedge] (k.east |- tr) to[out=0, in=180] (tr);
        \draw [imedge] (c2) to[out=45, in=180, looseness=0.3] (tl);
        \draw [draw=yellow] (c11.east |- aux1) to[out=0, in=180] (aux1);
        \draw [draw=yellow] (c12.east |- aux2) to[out=0, in=180] (aux2);
        \draw [imedge] (aux1) to[out=0, in=225] (bl);
        \draw [imedge] (aux2) to[out=0, in=135] (br);
        \draw [opacity=0] (tl.north) -- (bl.north) node [varname] {$\va$};
    },
    A/.pic={
        \pic {nodes};
        \draw [opacity=0] (tl.north) -- (bl.north) node [varname] {$\mA$};

        \draw [dedge] (tl) -- (bl);
        \draw [dedge] (tl) -- (br);
        \draw [dedge] (bl) -- (br);

        \draw [in=180, out=90, ledge] (tl) to (tl);
        \draw [in=0, out=90, ledge] (bl) to (bl);
        \draw [in=0, out=-90, ledge] (br) to (br);
    },
    Acolor/.pic={
        \pic {nodes};
        \draw [opacity=0] (tl.north) -- (bl.north) node [varname] {$\mA$};

        \draw [dedge, draw=blue2] (tl) -- (bl);
        \draw [dedge, draw=blue2] (tl) -- (br);
        \draw [dedge, draw=red2] (bl) -- (br);

        \draw [in=180, out=90, ledge] (tl) to (tl);
        \draw [in=0, out=90, ledge] (bl) to (bl);
        \draw [in=0, out=-90, ledge] (br) to (br);
    },
    O/.pic={
        \pic {nodes};
        \draw [opacity=0] (tl.north) -- (bl.north) node [varname] {$\mD$};

        \draw [dedge] (tl) -- (tr);
        \draw [dedge] (tl) -- (bl);
        \draw [dedge] (tl) -- (br);
        \draw [dedge] (tr) -- (bl);
        \draw [dedge] (tr) -- (br);
        \node [node, fill=yes] at (0, -1) {};
    },
    B/.pic={
        \pic {nodes};
        \draw [opacity=0] (tl.north) -- (bl.north) node [varname] {$\AO$};

        \draw [dedge] (tl) -- (bl);
        \draw [dedge] (tl) -- (br);
    },
    Bprime/.pic={
        \pic {nodes};
        \draw [opacity=0] (tl.north) -- (bl.north) node [varname] {$\AO'$};

        \draw [dedge] (tl) -- (bl);
        \draw [dedge] (tl) -- (br);
        \draw [in=180, out=90, ledge] (tl) to (tl);
        \draw [in=0, out=90, ledge] (bl) to (bl);
        \draw [in=0, out=-90, ledge] (br) to (br);
    },
    s/.pic={
        \pic {nodes};
        \draw [opacity=0] (tl.north) -- (bl.north) node [varname] {$\vs$};

        \node [node, fill=yes] at (0, -1) {};
        \node [node, fill=maybe] at (1, 0) {};
        \node [node, fill=maybe] at (1, -1) {};
    },
    C/.pic={
        \pic {nodes};
        \draw [opacity=0] (tl.north) -- (bl.north) node [varname] {$\mC$};
        \node [node, fill=maybe] at (1, 0) {};
        \node [node, fill=maybe] at (1, -1) {};

        \draw [in=180, out=90, ledge, draw=yes] (tl) to (tl);
        \draw [in=0, out=90, ledge, draw=maybe] (bl) to (bl);
        \draw [in=0, out=-90, ledge, draw=maybe] (br) to (br);

        \draw [dedge, draw=maybe] (tl) -- (bl);
        \draw [dedge, draw=maybe] (tl) -- (br);
    },
    C2/.pic={
        \node [node, fill=yes] (tl) at (0, 0) {};
        \node [node, fill=no, opacity=0] (tr) at (1, 0) {};
        \node [node, fill=no] (bl) at (0, -1) {};
        \node [node, fill=no, opacity=0] (br) at (1, -1) {};
        \node [node, fill=yes] (bm) at (1, -0.5) {};

        \begin{scope}[on background layer]
            \node [count-object, fit=(tr) (br)] {};
            \node [count-object, fit=(tl)] {};
            \node [count-object, fit=(bl)] {};
        \end{scope}

        \draw [in=180+22.5, out=90+22.5, ledge, draw=yes] (tl) to (tl);
        \draw [in=0+22.5, out=-90+22.5, ledge, draw=yes] (bm) to (bm);

        \draw [dedge] (tl) -- (bm);
    },
    count model/.pic={
        \node (img) {Image};
        \node (detect) [params, right = 5 mm of img, text width=1.6cm, align=center] {R-CNN};
        \node (l2) [no params, right = 3 mm of detect] {\tiny L2 norm};

        \node (quest) [below = 2 cm of img] {Question};
        \node (embed) [params, text width=1.6cm, align=center] at (detect |- quest) {Embedding};
        \node (rnn) [params, align=center] at (l2 |- embed) {GRU};

        \coordinate (atttop) at ($(l2)!0.25!(rnn)$);
        \coordinate (attmid) at ($(l2)!0.5!(rnn)$);
        \coordinate (attbottom) at ($(l2)!0.75!(rnn)$);
        \coordinate (attmid) at ($(l2)!0.5!(rnn)$);
        \node (attfuse) [no params, right = 7 mm of attmid] {\LARGE$\diamond$};
        \node (attw) [w, right = 2 mm of attfuse] {w};
        \coordinate (softpos) at (attw |- atttop);
        \node (softmax) [no params, right = 7 mm of softpos] {\tiny softmax};
        \node (sum) [no params] at (softmax |- l2) {\tiny$\sum$};

        \coordinate (wq) at (sum.east |- rnn);
        \coordinate (classmiddle) at (sum |- attmid);
        \node (classfuse) [no params, right = 8 mm of classmiddle] {\LARGE$\diamond$};
        \node (classadd) [no params, right = 2 mm of classfuse, draw=red2] {\tiny $+$};
        \node (classbn) [params, right = 2 mm of classadd] {\scriptsize BN};
        \node (class) [w, right = 2 mm of classbn] {w};
        \node (classsoft) [no params, right = 2 mm of class] {\tiny softmax};
        \node (end) [right = 2 mm of classsoft] {};

        \node (abovesum) [above = 7 mm of sum] {};
        \node (count) [params, draw=red2] at ([xshift=3mm]abovesum) {Count};
        \node (bn) [params, draw=red2] at (classadd |- atttop) {\scriptsize BN};
        \node (relu) [no params, draw=red2] at (classadd |- sum) {\tiny ReLU};

        \draw [thick, sedge] (img) to ([xshift=-1mm]detect.west);
        \draw [sedge] (detect) to (l2) to (sum);
        \draw [thick, sedge] (quest) to ([xshift=-1mm]embed.west);
        \draw (embed) to (rnn) to (wq);
        \draw (classfuse) to (classadd) to (classbn) to (class) to (classsoft) [sedge] to (end);
        \draw [ledge, out=-45, in=-135, looseness=4] (rnn) to (rnn);

        \draw [sedge] (l2.east) to [out=0, in=90] node [pos=0.5, w] {w} (attfuse);
        \draw [sedge] (rnn.east) to [out=0, in=-90] node [pos=0.5, w] {w} (attfuse);
        \draw (attfuse) to (attw);
        \draw ([yshift=0.3mm]attw.east) to [out=0, in=-90] ([xshift=-0.3mm]softmax.south);
        \draw ([yshift=-0.3mm]attw.east) to [out=0, in=-90] ([xshift=0.3mm]softmax.south);
        \draw [sedge] ([xshift=-0.3mm]softmax.north) to ([xshift=-0.3mm]sum.south);
        \draw [sedge] ([xshift=0.3mm]softmax.north) to ([xshift=0.3mm]sum.south);

        \draw [sedge] (sum) to [out=0, in=90] node [pos=0.5, w] {w} (classfuse);
        \draw [sedge] (wq) to [out=0, in=-90] node [pos=0.5, w] {w} (classfuse);

        \draw [color=red2] (count) to [out=0, in=90] node [pos=0.5, color=black, draw=red, w] {w} (relu);
        \draw [color=red2] (relu) to (bn) [sedge] to (classadd);

        \begin{scope}[on background layer]
            \draw [sedge, color=red2] ([yshift=0.3mm]attw.east) to [out=0, in=180, looseness=0.8] node [pos=0.7, color=black, draw=red2, no params] {\small$\sigma$} ([yshift=-0.3mm]count.west);
            \draw [sedge, color=red2] ([xshift=-0.4mm, yshift=-0.4mm]detect.north east) to [out=45, in=180, looseness=0.8] node [pos=0.5, above, rotate=13] {\tiny bounding boxes only} ([yshift=0.3mm]count.west);
            \coordinate (crossmid) at ($(attw)!0.4!(count)$);
            \node [fill=white, inner sep=0.8mm, text width=1cm] at (crossmid |- l2) { };
        \end{scope}

    },
    count model2/.pic={
        \clip (-0.5, -1.4) rectangle (7.4, 0.25);
        \node (img) {Image};
        \node (detect) [params, right = 7 mm of img, text width=1.6cm, align=center] {R-CNN};

        \node (quest) [below = 1.2 cm of img] {Question};
        \node (rnn) [params, text width=1.6cm, align=center] at (detect |- quest) {RNN};

        \coordinate (atttop) at ($(detect)!0.25!(rnn)$);
        \coordinate (attmid) at ($(detect)!0.5!(rnn)$);
        \coordinate (attbottom) at ($(detect)!0.75!(rnn)$);
        \node (attfuse) [no params, right = 9 mm of attmid] {\small$\diamond$};
        \node (attw) [params, right = 2 mm of attfuse] {Attention};
        \coordinate (softpos) at (attw |- atttop);
        \node (sum) [no params, scale=0.9] at ([xshift=13mm]attw |- detect) {\tiny$\sum$};

        \coordinate (wq) at (sum.east |- rnn);
        \coordinate (classmiddle) at (sum |- attmid);
        \node (classfuse) [no params, right = 8 mm of classmiddle] {\small$\diamond$};
        \node (classmlp) [params, right = 2 mm of classfuse] {MLP};
        \node (end) [right = 3 mm of classmlp, text width = 3 cm, scale=0.75] {Classify answer\\from 3000 most common answers};

        \node (abovesum) [above = 7 mm of sum] {};

        \draw [thick, sedge] (img) to ([xshift=-1mm]detect.west);
        \draw [sedge] (detect) to (sum);
        \draw [thick, sedge] (quest) to ([xshift=-1mm]rnn.west);
        \draw (rnn) to (wq);
        \draw (classfuse) to (classmlp) [sedge] to (end);
        \draw [ledge, out=45, in=135, looseness=4] (rnn) to (rnn);

        \draw [sedge] (detect.east) to [out=0, in=90] (attfuse);
        \draw [sedge] (rnn.east) to [out=0, in=-90] (attfuse);
        \draw (attfuse) to (attw);
        \draw [sedge] (attw.east) to [out=0, in=-90] (sum.south);

        \draw [sedge] (sum) to [out=0, in=90] (classfuse);
        \draw [sedge] (wq) to [out=0, in=-90] (classfuse);
    },
    dummyimg/.pic={
        \node [fill=black, inner sep = 0.5cm] (bg) at (0, 0) {};
        \node [below right = 4mm and 1mm of bg.north west, color=white, inner sep = 0.3mm, count-box] (star) {$\bigstar$};
    },
    fv/.pic={
        \node [rectangle, draw=black, thick, minimum size = 3mm, fill=c1] at (0, 0) {};
        \node [rectangle, draw=black, thick, minimum size = 3mm, fill=c2] at (0.6, 0) {};
        \node [rectangle, draw=black, thick, minimum size = 3mm, fill=c3] at (1.2, 0) {};
        \node [rectangle, draw=black, thick, minimum size = 3mm, fill=c4] at (1.8, 0) {};
    },
    probs/.pic={
        \node (img1) {\begin{tikzpicture}\pic{dummyimg};\end{tikzpicture}};
        \node [below = 0 mm of img1] (img2) {\begin{tikzpicture}\pic{dummyimg};\end{tikzpicture}};
        \node [below = 3.65mm of img2.center] (img3) {\begin{tikzpicture}\pic{dummyimg};\end{tikzpicture}};

        \node [left = 0mm of img1] {Image 1};
        \coordinate (img25) at ($(img2.west)!0.5!(img3.west)$);
        \coordinate (img26) at ($(img2.west)!0.2!(img3.west)$);
        \coordinate (img24) at ($(img2.west)!0.8!(img3.west)$);
        \node [left = 0mm of img25] {Image 2};

        \node [inner sep=0mm, right = 4mm of img1] (fv1) {\begin{tikzpicture}\pic{fv};\end{tikzpicture}};
        \node [inner sep=0mm,] (fv2) at (fv1 |- img26) {\begin{tikzpicture}\pic{fv};\end{tikzpicture}};
        \node [inner sep=0mm,] (fv3) at (fv1 |- img24) {\begin{tikzpicture}\pic{fv};\end{tikzpicture}};
        \coordinate (fv) at ($(fv1)!0.5!(fv3)$);

        \node [right = 2.5cm of fv1] (w1) {$w$};
        \node [right = 2.5cm of fv2] (w2) {$w$};
        \node [right = 2.5cm of fv3] (w3) {$w$};

        \draw [sedge] (fv1) -- (w1);
        \draw [sedge] (fv2) -- (w2);
        \draw [sedge] (fv3) -- (w3);

        \node [params, right = of fv, minimum height = 3cm] (att1) {Attention};

        \node [right = 10mm of w1] (v1) {$1.0$};
        \node [right = 10mm of w2] (v2) {$0.5$};
        \node [right = 10mm of w3] (v3) {$0.5$};

        \draw [sedge] (w1) -- (v1);
        \draw [sedge] (w2) -- (v2);
        \draw [sedge] (w3) -- (v3);

        \node [no params, right = 3.85cm of fv, anchor=north, rotate=-90, anchor=south,  minimum width = 3cm] (softm) {Softmax};

        \coordinate (v25) at ($(v2.east)!0.5!(v3.east)$);
        \node [no params, right = 3mm of v1, minimum height=1cm] (s1) {$\sum$};
        \node [no params, right = 3mm of v25, minimum height=1cm] (s2) {$\sum$};

        \draw [sedge] (v1) -- (s1);
        \draw [sedge] (v2) -- (s2.west |- v2);
        \draw [sedge] (v3) -- (s2.west |- v3);

        \node [right = 4mm of s1, inner sep=0mm] (out1) {\begin{tikzpicture}\pic{fv};\end{tikzpicture}};
        \node [right = 4mm of s2, inner sep=0mm] (out2) {\begin{tikzpicture}\pic{fv};\end{tikzpicture}};

        \coordinate (s15) at ($(s1)!0.5!(s2)$);
        \node (eq) at (out1 |- s15) {\small$=$};

        \draw [sedge] (s1) -- (out1);
        \draw [sedge] (s2) -- (out2);

        \node [above = 2mm of fv1, rotate=-90, anchor=center, scale=1.3, inner sep=0mm] (to) {$\Bigg\{$};
        \node [below = 2mm of fv1, rotate=-90, anchor=center, scale=1.3, inner sep=0mm] {$\Bigg\}$};
        \node [above = 2mm of fv2, rotate=-90, anchor=center, scale=1.3, inner sep=0mm] {$\Bigg\{$};
        \node [below = 2mm of fv3, rotate=-90, anchor=center, scale=1.3, inner sep=0mm] (bo) {$\Bigg\}$};

        \draw [sedge] (to.west) to[bend left=20] (s1.north west);
        \draw [sedge] (bo.east) to[bend right=20] (s2.south west);

        \draw [draw=yellow, sedge] (0, -0.1) to[in=180, out=0] (fv1);
        \draw [draw=yellow, sedge] (0, -2.6) to[in=180, out=0] (fv2);
        \draw [draw=yellow, sedge] (0, -4.6) to[in=180, out=0] (fv3);
    },
}

\makeatother

\theoremstyle{plain}
\newtheorem{thm}{Theorem}

\theoremstyle{definition}
\newtheorem{definition}{Definition}

\theoremstyle{plain}

\theoremstyle{plain}
\newtheorem{corollary}{Corollary}

\newcommand{\newterm}[1]{{\bf #1}}

\def\eqref#1{equation~\ref{#1}}

\def\1{\bm{1}}

\def\eps{{\epsilon}}

\def\va{{\bm{a}}}
\def\vb{{\bm{b}}}
\def\vc{{\bm{c}}}

\def\vo{{\bm{o}}}
\def\vp{{\bm{p}}}

\def\vs{{\bm{s}}}

\def\vw{{\bm{w}}}
\def\vbw{{\bm{\bar{w}}}}
\def\vx{{\bm{x}}}
\def\vy{{\bm{y}}}
\def\vyp{{\bm{y'}}}
\def\vhy{{\bm{\hat{y}}}}
\def\vz{{\bm{z}}}

\def\mA{{\bm{A}}}
\def\mB{{\bm{B}}}
\def\mC{{\bm{C}}}
\def\mD{{\bm{D}}}

\def\mG{{\bm{G}}}

\def\mL{{\bm{L}}}

\def\mO{{\bm{O}}}
\def\mP{{\bm{P}}}
\def\mQ{{\bm{Q}}}

\def\mS{{\bm{S}}}

\def\mW{{\bm{W}}}
\def\mbW{{\bm{\bar{W}}}}
\def\mX{{\bm{X}}}
\def\mvX{{\bm{\vec{X}}}}
\def\mY{{\bm{Y}}}
\def\mhY{{\bm{\hat{Y}}}}

\DeclareMathAlphabet{\mathsfit}{\encodingdefault}{\sfdefault}{m}{sl}
\SetMathAlphabet{\mathsfit}{bold}{\encodingdefault}{\sfdefault}{bx}{n}

\newcommand{\Ls}{\mathcal{L}}
\newcommand{\R}{\mathbb{R}}

\DeclareMathOperator*{\argmin}{arg\,min}

\def \enc {{g_{\text{enc}}}}
\def \dec {{g_{\text{dec}}}}
\def \Lrepr {{L_{\text{repr}}}}
\def \Lcha {{L_{\text{cha}}}}
\def \Lhun {{L_{\text{hun}}}}
\def \Lset {{L_{\text{set}}}}

\DeclareMathOperator*{\minimize}{minimize}
\newcommand{\m}[1]{\mathbf{#1}}

\newcommand{\AO}{\widetilde{\mA}}
\newcommand{\T}{\mathsf{T}}

\DeclareFontEncoding{LS1}{}{}
\DeclareFontSubstitution{LS1}{stix}{m}{n}
\newcommand{\bbone}{\text{\usefont{LS1}{stixbb}{m}{n}1}}
\newcommand{\norm}[1]{\left\lVert#1\right\rVert}

\begin{document}
\frontmatter
\centerfloat{
    \begin{minipage}[\textheight]{1.2\textwidth}
        \begin{titlepage}
            \begin{center}
                \Large
                University of Southampton

                \vspace{0.5cm}

                Faculty of Engineering and Physical Sciences
                
                \vspace{7cm}
                
                {\huge \bf \mytitle\par}

                \vspace{1.5cm}

                {\LARGE \bf \myname}

                \vfill

                \vspace{9cm}
                Thesis for the degree of Doctor of Philosophy

                \vspace{0.5cm}
                December, 2019
            \end{center}
        \end{titlepage}
    \end{minipage}
}
\cleardoublepage

\begin{center}
    UNIVERSITY OF SOUTHAMPTON

    \underline{ABSTRACT}

    \vspace{0.5cm}

    FACULTY OF ENGINEERING AND PHYSICAL SCIENCES

    Electronics and Computer Science

    \underline{Doctor of Philosophy}

    \vspace{0.5cm}

    \MakeUppercase{\mytitle}

    by \myname
    
    \vspace{0.5cm}
    
\end{center}
{
    \setlength{\parindent}{1em}
    In this thesis, we develop various techniques for working with \emph{sets} in machine learning.
    Each input or output is not an image or a sequence, but a set: an unordered collection of multiple objects, each object described by a feature vector.
    Their unordered nature makes them suitable for modeling a wide variety of data, ranging from objects in images to point clouds to graphs.
    Deep learning has recently shown great success on other types of structured data, so we aim to build the necessary structures for sets into deep neural networks.

    The first focus of this thesis is the learning of better set representations (sets as input).
    Existing approaches have bottlenecks that prevent them from properly modeling relations between objects within the set.
    To address this issue, we develop a variety of techniques for different scenarios and show that alleviating the bottleneck leads to consistent improvements across many experiments.

    The second focus of this thesis is the prediction of sets (sets as output).
    Current approaches do not take the unordered nature of sets into account properly.
    We determine that this results in a problem that causes discontinuity issues with many set prediction tasks and prevents them from learning some extremely simple datasets.
    To avoid this problem, we develop two models that properly take the structure of sets into account.
    Various experiments show that our set prediction techniques can significantly benefit over existing approaches.
}
\clearpage

\microtypesetup{protrusion=false}
\tableofcontents
\listoffigures
\listoftables
\microtypesetup{protrusion=true}

\chapter{Declaration of authorship}
    I, \myname, declare that the thesis entitled \emph{\mytitle} and the work presented in the thesis are both my own, and have been generated by me as the result of my own original research. I confirm that:
    
    \begin{itemize}
        \item this work was done wholly or mainly while in candidature for a research degree
        at this University;
        \item where any part of this thesis has previously been submitted for a degree or any
        other qualification at this University or any other institution, this has been clearly
        stated;
        \item where I have consulted the published work of others, this is always clearly
        attributed;
        \item where I have quoted from the work of others, the source is always given. With
        the exception of such quotations, this thesis is entirely my own work;
        \item I have acknowledged all main sources of help;
        \item where the thesis is based on work done by myself jointly with others, I have
        made clear exactly what was done by others and what I have contributed myself;
        \item parts of this work have been published as: \citep{Zhang2019DSPN, Zhang2019FSPool, Zhang2019PermOptim, Zhang2018Counting}.
    \end{itemize}
    \vspace{15.0mm}
    Signature: \hrulefill
    
    \vspace{3.0mm}
    Date: \hrulefill

\cleardoublepage

\chapter{Acknowledgements}
\begin{center}
    \Large
    \begin{equation*}
        \left\{
        \begin{array}{c}
            \text{Adam Pr\"ugel-Bennett}\\
            \text{Jonathon Hare}\\
            \text{Freddie Nash}\\
            \text{My family}
        \end{array}
        \right\}
    \end{equation*}
\end{center}

\chapter{List of abbreviations}
\DeclareAcronym{cnn}{
    short = CNN,
    long  = Convolutional neural network,
    tag = abbrev
}
\DeclareAcronym{rnn}{
    short = RNN,
    long  = Recurrent neural network,
    tag = abbrev
}
\DeclareAcronym{sgd}{
    short = SGD,
    long  = Stochastic gradient descent,
    tag = abbrev
}
\DeclareAcronym{mse}{
    short = MSE,
    long  = Mean squared error,
    tag = abbrev
}
\DeclareAcronym{vqa}{
    short = VQA,
    long  = Visual question answering,
    tag = abbrev
}
\DeclareAcronym{mlp}{
    short = MLP,
    long  = Multi-layer perceptron,
    tag = abbrev
}
\DeclareAcronym{nms}{
    short = NMS,
    long  = Non-maximum suppression,
    tag = abbrev
}
\DeclareAcronym{gpu}{
    short = GPU,
    long  = Graphics processing unit,
    tag = abbrev
}
\printacronyms[include=abbrev,name=Terminology,heading=section*]

\DeclareAcronym{vqa2}{
    short = VQA v1/v2,
    long  = VQA datasets \citep{Antol2015, Goyal2017},
    tag = dataset
}
\DeclareAcronym{lstm}{
    short = LSTM,
    long  = Long short-term memory RNN \citep{Hochreiter1997},
    tag = name
}
\DeclareAcronym{mnist}{
    short = MNIST,
    long  = Handwritten digits dataset \citep{Lecun1998},
    tag = dataset
}
\DeclareAcronym{clevr}{
    short = CLEVR,
    long  = Synthetic VQA dataset \citep{Johnson2017a},
    tag = dataset
}
\DeclareAcronym{rn}{
    short = RN,
    long  = Relation network \citep{Santoro2017},
    tag = name
}
\DeclareAcronym{rcnn}{
    short = Faster R-CNN,
    long  = Region-based CNN object detector \citep{Ren2015},
    tag = name
}
\DeclareAcronym{cifar10}{
    short = CIFAR-10,
    long  = Tiny image dataset \citep{Krizhevsky2009},
    tag = dataset
}
\DeclareAcronym{ban}{
    short = BAN,
    long  = Bilinear attention network \citep{Kim2018},
    tag = name
}
\DeclareAcronym{gin}{
    short = GIN,
    long  = Graph isomorphism network \citep{Xu2019},
    tag = name
}
\DeclareAcronym{ae}{
    short = AE-CD/AE-EMD,
    long  = Set auto-encoder with different set losses \citep{Achlioptas2018},
    tag = name
}
\printacronyms[include=dataset,name=Datasets,heading=section*]
\printacronyms[include=name,name=Models,heading=section*]

\mainmatter
\chapter{Sets in machine learning}\label{cha:introduction}
\sectionmark{Introduction}
Sets are collections of things without a natural ordering to them.
For example, the types of fruit that someone has eaten in the past week can be considered a \emph{set} of fruit.
While it may be possible to order them in some way (such as by name, date last eaten, tastiness, etc.), there is no inherently ``correct'' ordering.
Talking about the nutritional value of the collective $\{\text{apple}, \text{raspberry}, \text{blackberry}\}$ is exactly the same as talking about $\{\text{raspberry}, \text{blackberry}, \text{apple}\}$.
The order that they are written down in is different, but the set of fruits itself is the same.
This is in contrast to data that do have a natural order, like the words in a sentence -- a sentence can lose much of its meaning when the individual words are shuffled around.

Such sets are a natural way of describing different kinds of data in the real world.
There are many problems of interest to the machine learning community that can be described in terms of sets:
predicting the set of objects in an image is known as object detection;
Lidar scanners on self-driving cars produce a set of 3d points of their surroundings to find obstacles;
properties of molecules can be predicted based on the set of atoms and the set of bonds connecting those atoms.
In all of these examples, a machine learning model either takes a set as input or produces a set as output.

\begin{figure}
    \centering
    \centerfloat{
    \begin{tikzpicture}
        \fontsize{10}{10}\selectfont
        \node (vqa) at (0, 0) {\includegraphics[width=0.7\linewidth]{learn-to-count/440336000}};
        \node at (0, 6) {\textbf{(a)}};

        \node (open) at (14, 4) {};
        \node [below = -2mm of open] (tiles) {$\left\{\vcenter{\hbox{\includegraphics[height=4cm]{learn-to-permute/analysis/example-mnist-2}}}\right\}$};
        \node [below = -2mm of tiles] (close) {};
        \node (open) at (18, 4) {};
        \node [below = -2mm of open] (tiles) {$\left\{\vcenter{\hbox{\includegraphics[height=4cm]{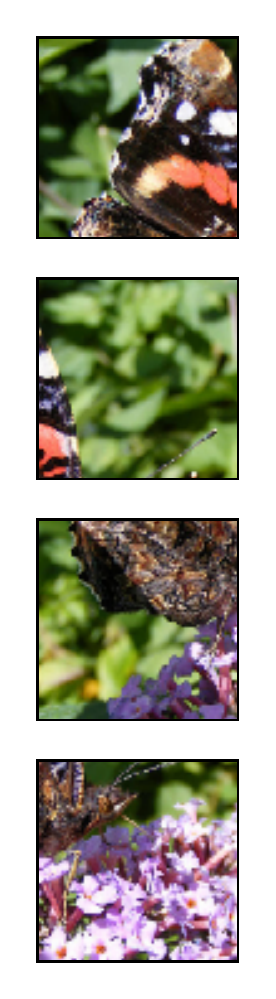}}}\right\}$};
        \node [below = -2mm of tiles] (close) {};
        \node at (16, 6) {\textbf{(b)}};
        
        \node at (-3.5, -11) {\includegraphics[width=0.3\linewidth, trim={3mm 3mm 3mm 3mm}, clip]{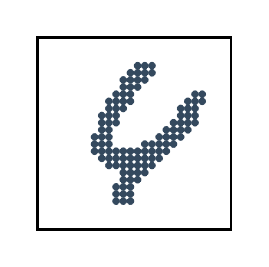}};
        \node at (3.5, -11) {\includegraphics[width=0.3\linewidth, trim={3mm 3mm 3mm 3mm}, clip]{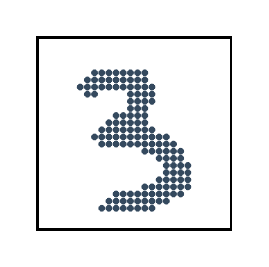}};
        \node at (0, -7) {\textbf{(c)}};

        \node at (16, -11) {%
            $
                \left\{
                    \def\arraystretch{1.4}
                    \begin{array}{l}
                        \text{large cyan rubber sphere},\\
                        \text{small blue rubber cylinder},\\
                        \text{small purple metal cube},\\
                        \text{small yellow metal cube},\\
                        \text{large cyan rubber cube},\\
                        \text{small green rubber sphere},\\
                    \end{array}
                \right\}
            $
        };
        \node at (16, -7) {\textbf{(d)}};
    \end{tikzpicture}
    }
    \caption[Examples of sets throughout the thesis]{Example sets that we will work with throughout the thesis.
    \textbf{(a)} set of object proposals (\autoref{cha:vqa}).
    \textbf{(b)} set of image tiles (\autoref{cha:perm-optim}).
    \textbf{(c)} set of points in a point cloud (\autoref{cha:fspool}).
    \textbf{(d)} set of object attributes (\autoref{cha:dspn}).
    }
    \label{intro fig:examples}
\end{figure}

In this thesis, we will focus on a specific type of machine learning method for working with sets: deep neural networks.
In recent years, these have stood out as one of the most successful approaches on structured data like images and text.
Part of this success is due to the building blocks that these deep neural networks are made of.
These building blocks (like convolutions for images) take advantage of the structure in the data and allow the neural networks to learn much more efficiently.

With sets, this structure is given by their unordered nature.
While there is existing work on neural network approaches to sets (which we briefly review in \autoref{cha:background}), the set domain has been studied to a much lesser extent than domains like images and text.
What building blocks are needed to adequately handle sets with neural networks?
This question is what this thesis explores by identifying shortcomings in existing techniques and developing new techniques for modeling sets with neural networks.

\thispagestyle{front}
We begin by motivating our work with a benchmark task for Machine Intelligence: visual question answering.
The task requires a machine to answer natural language questions about images, such as ``how many people are there?''  about the image in \autoref{intro fig:examples} (a).
These questions can be arbitrarily difficult, so the task probes the edge of what is possible with current machine learning techniques.
While this task was initially defined with images as input, extracting and using the \emph{set of objects} in the image has been found to lead to better results~\citep{Anderson2018}.
This makes visual question answering a task with set inputs.
We identify a problem with existing visual question answering models in \autoref{cha:vqa}, which means that \emph{counting questions} like the one above are virtually impossible to answer.
We then propose a building block that fixes this issue.
Our work on this task shows that sets can be quite interesting to work with and how existing methods for sets can fall short.
This is what motivated us to continue researching ways of modeling sets with neural networks.

We move on from this specific application of sets to more general set problems.
There are two main directions for this: a machine learning model can take a set as input or produce a set as output.

\paragraph{Sets as input}
In these tasks, the inputs are sets and the desired output is something else, like a classification or regression.
In visual question answering, the input is the set of object proposals (along with a question) and the output is an answer.
The objective is to extract as much relevant information out of the set as possible.
We worked on two different methods for doing this:

\begin{itemize}
    \item In \autoref{cha:perm-optim}, we propose a way of \emph{learning to permute} the elements of the set into a list, which is easier to work with than the set.
    This is based on the assumption that there are some orderings that are easier to learn from~\citep{Vinyals2015}, which our model tries to find.
    \item In \autoref{cha:fspool}, we propose a model that involves \emph{sorting} the features in the set numerically.
    This allows the model to learn about the \emph{distribution} of features, which can better capture relationships between set elements.
\end{itemize}

\paragraph{Sets as output}
In these tasks, we have some input and want to produce a set as output.
In object detection, the input is an image and output is the set of objects in that image.
In \autoref{cha:fspool}, we identify a problem that we call the \emph{responsibility problem} with existing set prediction techniques, which can significantly hinder the learning of the neural network.
We then propose a solution which works in the limited scenario of auto-encoders.
From what we learned through this, we develop a general set prediction algorithm in \autoref{cha:dspn} without the auto-encoder limitation.
This is the first neural network for predicting sets that properly respects the unordered nature of sets, and -- in our opinion -- it is the most significant contribution of this thesis.

We conclude the thesis by discussing potential future research directions in \autoref{cha:conclusion}.
Sets are a natural way of modeling object-based representations, which can be beneficial in a much wider variety of tasks and models than sets are currently used in.

\newpage
\section{List of contributions}
Concretely, this thesis contributes the following:

\paragraph{\autoref{cha:background}}
\begin{itemize}
    \item We provide an accessible introduction to modeling sets with neural networks.
    We cover how sets are represented (\autoref{sec:background-representation}), set encoders (\autoref{sec:background-encoders}), set decoders (\autoref{sec:background-decoders}), and how sets are used in other areas of machine learning (\autoref{sec:background-other}).
\end{itemize}

\paragraph{\autoref{cha:vqa}}
These contributions have been published as~\citep{Zhang2018Counting} in the International Conference on Learning Representations (ICLR) 2018 and have been presented at the Visual Question Answering Challenge workshop, hosted at the Conference on Computer Vision and Pattern Recognition (CVPR) 2018.

\begin{itemize}
    \item We review related work on counting objects in images with neural networks (\autoref{count sec:related}).
    \item We identify a problem in existing VQA models and explain why they struggle with counting questions as a result (\autoref{count sec:problems}).
    \item We propose a model for counting sets of objects that avoids this problem (\autoref{count sec:counting}).
    \item We evaluate our model on a toy dataset that we developed for testing counting ability in an isolated setting (\autoref{count sec:toy}), and on the full VQA v2 dataset (\autoref{count sec:vqa}). 
    \item We open-source the code to reproduce all our experiments at:\\ \url{https://github.com/Cyanogenoid/vqa-counting}\\ and the first publically-available code to reproduce a strong baseline by \citet{Kazemi2017} at:\\ \url{https://github.com/Cyanogenoid/pytorch-vqa}.
\end{itemize}

\paragraph{\autoref{cha:perm-optim}}
These contributions have been published as~\citep{Zhang2019PermOptim} in the International Conference on Learning Representations (ICLR) 2019.
\begin{itemize}
    \item We propose a model for learning permutations based on learning pairwise comparisons (\autoref{perm sec:model}).
    \item We review related work on learning permutations (\autoref{perm sec:related}).
    \item We evaluate our model on sorting numbers (\autoref{perm sec:numbersort}), assembling image mosaics explicitly (\autoref{perm sec:mosaic explicit}) or implicitly (\autoref{perm sec:mosaic implicit}), and VQA v2 (\autoref{perm sec:vqa}).
    \item We open-source the code to reproduce all experiments at:\\ \url{https://github.com/Cyanogenoid/perm-optim},\\ which also includes the first reproduction of the baseline results by \citet{Mena2018}.
\end{itemize}

\paragraph{\autoref{cha:fspool}}
These contributions have been presented as~\citep{Zhang2019FSPool} at the Sets \& Partitions workshop, hosted at the Neural Information Processing Systems (NeurIPS) 2019 conference, and have been submitted to the International Conference on Learning Representations (ICLR) 2020.
\begin{itemize}
    \item We identify a responsibility problem with existing set prediction methods, which results in discontinuities (\autoref{fspool sec:problem}).
    \item We propose a set encoder based on sorting features numerically and an associated set decoder for auto-encoding (\autoref{fspool sec:fsort}).
    \item We review related work on using sorting in neural networks (\autoref{fspool sec:related}).
    \item We evaluate our auto-encoder on a rotating polygon toy dataset (\autoref{fspool sec:polygons}) and a set version of MNIST (\autoref{fspool sec:mnist}).
    \item We evaluate our encoder on MNIST set classification (\autoref{fspool sec:mnist-class}), CLEVR (\autoref{fspool sec:clevr}), and graph classification (\autoref{fspool sec:graph}).
    \item We open-source the code to reproduce all experiments at:\\ \url{https://github.com/Cyanogenoid/fspool},\\ which also includes the first reproduction of the baseline graph classification results by \citet{Xu2019}.
\end{itemize}

\paragraph{\autoref{cha:dspn}}
These contributions have been published as~\citep{Zhang2019DSPN} in Advances in Neural Information Processing Systems 32 (NeurIPS) 2019 and have been presented at the Sets \& Partitions workshop, hosted at the Neural Information Processing Systems (NeurIPS) 2019 conference.
\begin{itemize}
    \item We propose a model for general set prediction (\autoref{dspn sec:model}) that avoids the responsibility problem.
    \item We review related work on predicting sets with neural networks (\autoref{dspn sec:related}).
    \item We evaluate our model on auto-encoding MNIST sets (\autoref{dspn sec:mnist}), predicting the set of bounding boxes in an image (\autoref{dspn sec:bbox}), and predicting the set of object attributes in an image (\autoref{dspn sec:state}).
    \item We open-source the code to reproduce all experiments at:\\ \url{https://github.com/Cyanogenoid/dspn}, which includes an implementation of the models by \citet{Achlioptas2018}.
\end{itemize}

\paragraph{\autoref{cha:conclusion}}
\begin{itemize}
    \item We discuss future research directions and open problems.
    We discuss ideas in the area of set encoders (\autoref{future sec:encoders}), set decoders (\autoref{future sec:decoders}), and using latent sets in neural networks (\autoref{future sec:latent}).
\end{itemize}

\chapter{Basics of set neural networks}\label{cha:background}
\chaptermark{Background}
In this chapter, we will introduce the basics of working with sets using neural networks.
This provides the necessary background from the set literature to understand this thesis, especially \autoref{cha:perm-optim}, \autoref{cha:fspool}, and \autoref{cha:dspn}.
This chapter is not an exhaustive review of the literature, but an accessible introduction into the foundational work on modeling sets with neural networks.
We leave the more detailed discussions of related work in the appropriate chapters, in particular \autoref{count sec:related}, \autoref{perm sec:related}, \autoref{fspool sec:related}, \autoref{dspn sec:related}.

\newcounter{texercise}[section]
\section{Overview}
First, let us be clear with what we mean when we talk about sets.
In essence, sets are \emph{unordered} collections of \emph{entities} or \emph{elements}.
These entities can be objects, people, atoms, symbols, and so on.
Since we are working in machine learning, it is useful to describe each entity in the set with a feature vector.
For example, we can have the set of points in a point cloud and store their 3d position ($xyz$ coordinates) as the set $\{[x_1, y_1, z_1]^\T, [x_2, y_2, z_2]^\T, \ldots\}$ with each $[x_i, y_i, z_i]^\T$ as the feature vector for a point in the set.
Because the order of the points does not matter -- swapping the order of any two points does not change the shape captured by the point cloud -- this is indeed a set.
Since the entities in the set typically correspond to real-world things, it does not make much practical sense to talk about infinitely many entities.
So, the sets we work with will always have a finite number of elements in them.
However, there is one major difference compared to sets in mathematics: duplicates are typically allowed, so when we talk about sets, we usually mean multisets (also known as bags).

Let us see how this fits into existing machine learning setups.
In traditional supervised learning, we have a dataset that consists of many input-output pairs.
The subject of interest in this thesis is when the input or the output of the model is a set of feature vectors.
These sets usually vary in size across the dataset, so any model for sets has to be able to work with varying set sizes.

\begin{figure}
    \centering
    \centerfloat{\resizebox{\figurewidth}{!}{%
        \sffamily
        \begin{tikzpicture}
            \pic at (-3, 0) {square3};
            \begin{scope}[on background layer]
                \node (s1) [object, fit=(a) (b) (c) (d)] {};
            \end{scope}

            \node (enc) [trapezium, draw, rotate=-90, inner ysep=3.5mm, trapezium angle=70, fill=Oranges-4-1] at (3.5, 0) {\scriptsize Encoder};
            \node [box, scale=0.5, fill=Blues-4-3] at (5.5, 0.5) {};
            \node (fv) [box, scale=0.5, fill=Oranges-4-3] at (5.5, 0) {};
            \node [box, scale=0.5, fill=Purples-4-3] at (5.5, -0.5) {};
            \node (dec) [trapezium, draw, rotate=90, inner ysep=3.5mm, trapezium angle=70, fill=Oranges-4-1] at (7.5, 0) {\scriptsize Decoder};

            \node [color=Blues-4-4] (i4) at (0,  1.5) {\small(-1, -1)};
            \node [color=Blues-4-4] (i3) at (0,  0.5) {\small(~1, -1)};
            \node [color=Blues-4-4] (i2) at (0, -0.5) {\small(~1, ~1)};
            \node [color=Blues-4-4] (i1) at (0, -1.5) {\small(-1, ~1)};
            \node [rotate=90] at (0, -2.2) {\Bigg\{};
            \node [rotate=-90] at (0, 2.2) {\Bigg\{};

            \node [color=Blues-4-4] (o4) at (11,  1.5) {\small(-1, ~1)};
            \node [color=Blues-4-4] (o3) at (11,  0.5) {\small(~1, ~1)};
            \node [color=Blues-4-4] (o2) at (11, -0.5) {\small(~1, -1)};
            \node [color=Blues-4-4] (o1) at (11, -1.5) {\small(-1, -1)};
            \node [rotate=90] at (11, -2.2) {\Bigg\{};
            \node [rotate=-90] at (11, 2.2) {\Bigg\{};

            \pic at (14, 0) {square3};
            \begin{scope}[on background layer]
                \node (s2) [object, fit=(a) (b) (c) (d)] {};
            \end{scope}

            \draw [sedge] (i1) -- (enc.south |- i1);
            \draw [sedge] (i2) -- (enc.south |- i2);
            \draw [sedge] (i3) -- (enc.south |- i3);
            \draw [sedge] (i4) -- (enc.south |- i4);
            \draw [sedge] (enc) -- (fv);
            \draw [sedge] (fv) -- (dec);
            \draw [sedge] (dec.south |- o1) -- (o1);
            \draw [sedge] (dec.south |- o2) -- (o2);
            \draw [sedge] (dec.south |- o3) -- (o3);
            \draw [sedge] (dec.south |- o4) -- (o4);
        \end{tikzpicture}
    }}
    \caption[Basic auto-encoder model]{A neural network for auto-encoding a set of four 2d points.
    The set encoder encodes the input set into a feature vector, which the set decoder decodes to predict a set.
    Notice how the output points are not necessarily in the same order as the input points.
    }
    \label{background fig:ae}
\end{figure}
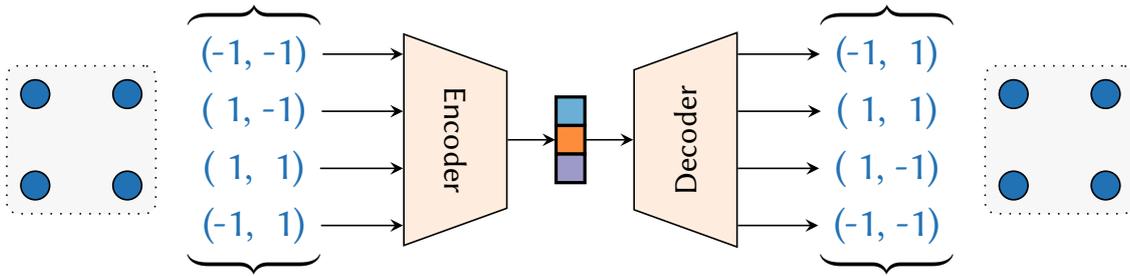

There are two main types of operations that we care about, which we visualise in \autoref{background fig:ae}.

\begin{enumerate}
    \item When the input is a set, we want to \emph{encode} the input set into a feature vector (\autoref{sec:background-encoders}).
    \item When the output is a set, we want to \emph{decode} a feature vector into the output set (\autoref{sec:background-decoders}).
\end{enumerate}

By relating sets of feature vectors with a single feature vector in this way, we can combine these operations on sets with other, well-established methods that operate on feature vectors.
If we can also make encoding and decoding differentiable, they can be included in deep neural networks.
This gives us a fully differentiable model for sets that can be trained with conventional methods for neural networks like stochastic gradient descent (SGD) and its variants.

This differentiability is one of the important foundations that much work in Deep Learning is built upon.
We will take care to build our models from differentiable primitives so that they easily fit into the usual gradient-based neural network training framework.
For more background on deep neural networks, we refer you to part two of the Deep Learning book~\citep{Goodfellow2016}.

\section{Representation in memory}\label{sec:background-representation}
To understand how set encoding and decoding works, we first have to understand how the set itself is represented.

Even though sets are orderless, we still have to store them in memory, which forces them to be ordered in some way.
Conveniently, the order of the set elements does not matter, so we can store the elements in the set in \emph{any} arbitrary order.
In essence, we treat the set of feature vectors as if it were just a list of feature vectors, which means that we can simply store it as a matrix.
A set of size $n$ wherein each feature vector has dimensionality $d$ can therefore be stored as an $\R^{d \times n}$ or $\R^{n \times d}$ matrix.
We will use one or the other in the upcoming chapters depending on what makes the narrative the clearest, but we will always clarify which one we are using in a chapter.
In this chapter, we store them as $\R^{d \times n}$ matrices: each column corresponds to one set element.

\begin{myexample}
    Let us look at an example. The set:
    
    \begin{equation}
        \left\{
            \left[
                \begin{array}{l}
                    1 \\
                    2 \\
                    3 \\
                \end{array}
            \right],
            \left[
                \begin{array}{l}
                    4 \\
                    5 \\
                    6 \\
                \end{array}
            \right]
        \right\}
    \end{equation}

    has two elements of dimensionality three.
    It can be stored as one of these two (equivalent) matrices:

    \begin{equation}
        \left[
            \begin{array}{ll}
                1 & 4 \\
                2 & 5 \\
                3 & 6 \\
            \end{array}
        \right]
        \text{ or }
        \left[
            \begin{array}{ll}
                4 & 1 \\
                5 & 2 \\
                6 & 3 \\
            \end{array}
        \right].
    \end{equation}
\end{myexample}

Storing the set as a matrix is memory efficient and easy to work with through the usual matrix operations like matrix multiplication.
However, this comes with the trade-off that there are now multiple representations of the same set, one for each possible permutation of the set elements.
Since these different representations are an artefact of the way the set is stored in memory, a key aspect when working with sets is to not rely on this arbitrary ordering of the elements.

There is one additional consideration for minibatch-wise training with sets, which is needed with stochastic gradient descent.
Since the set size typically varies across the dataset, we end up with sets of different sizes in a minibatch.
To make computation on a minibatch efficient, the typical approach is to pad all sets within a batch to a fixed size with elements of all zeros.
This fixed size is usually the size of the largest set in the minibatch or the size of the largest set in the dataset.
We then need to keep track of which elements are padding elements to not affect the result.
For example, if we want to compute a mean over the set elements, we cannot divide the sum by the number of columns in the matrix (which includes padding elements not part of the set), but we have to divide by the actual set size.

\section{Set encoders}\label{sec:background-encoders}
Now that we know how sets are stored, let us take a look at how such sets can be encoded into a feature vector.
We want to build a neural network that can take a set of feature vectors as input and produce a feature vector representation of that set as output.
Set encoders can for example be used for classifying what type of object the shape of a 3d point cloud forms, or answering natural language questions about a set of objects (visual question answering).

At this point, we have a matrix representation of the set and could just use a traditional neural network approach -- such as a multilayer perceptron (MLP) -- on it.
The $d \times n$ matrix can be flattened into a $dn$-dimensional vector and we could feed this into a normal fully-connected neural network.
This is essentially what \citet{Murphy2019} do, which we discuss in more detail in \autoref{perm sec:related}.
However, this has a major problem: different representations of the same set can give a different output.
There is no guarantee that $f([1, 2, 3]) = f([2, 3, 1]) = f([3, 2, 1]) = \ldots$, even though they all represent the same set $\{1, 2, 3\}$.

While a neural net as a universal approximator could learn to map all of these to the same output, this quickly becomes infeasible for non-trivial set sizes.
With $n$ elements in a set and thus $n!$ different representations of the same set, an MLP quickly runs into its modeling capabilities.
This is why \citet{Murphy2019} recommend sampling many permutations at test time and averaging the results to counteract this reliance on the arbitrary permutation.

We already know that order should not matter for sets.
So, let us be smarter about this: rather than letting the MLP learn freely, we can structure the neural network so that it is \emph{impossible} for it to rely on the arbitrary order.
This lets the neural network focus on extracting the pertinent information in the set, without having to learn to ignore the order at the same time.
Building the necessary structures into a neural network to properly work with sets is the key approach in this thesis.

\subsection{Properties}
To enforce a neural network to not rely on the ordering of the set elements, there are two properties that are important to know about.
In the context of deep neural networks, these were first defined by \citet{Zaheer2017}. 

\paragraph{Permutation-invariance}
The property of permutation-invariance says that the output of a function \emph{does not change} when its input is permuted.
If a permutation-invariant function is applied to a set, then we can guarantee that regardless of which arbitrary order we stored the elements in, we get the same output.
This ensures that the arbitrarily-chosen order of the set can never affect the result.
A permutation-invariant function can be thought of as serving the same purpose as pooling in convolutional neural networks (CNNs): it reduces multiple feature vectors to a single feature vector.

\begin{definition}
    A function $f: \R^{d \times n} \to \R^{c}$ is permutation-invariant iff it satisfies:

    \begin{equation}
        f(\mX) = f(\mX\mP)
    \end{equation}
    
    for all permutation matrices $\mP$.
\end{definition}

\begin{myexample}
    Take the set from the previous example:

    \begin{equation}
        \left\{
            \left[
                \begin{array}{l}
                    1 \\
                    2 \\
                    3 \\
                \end{array}
            \right],
            \left[
                \begin{array}{l}
                    4 \\
                    5 \\
                    6 \\
                \end{array}
            \right]
        \right\}
    \end{equation}

    Summing each feature over the set is permutation-invariant.
    Regardless of which order the set is stored in, we have:

    \begin{equation}
        \left[
            \begin{array}{ll}
                1 + 4 \\
                2 + 5 \\
                3 + 6 \\
            \end{array}
        \right]
        =
        \left[
            \begin{array}{ll}
                4 + 1 \\
                5 + 2 \\
                6 + 3 \\
            \end{array}
        \right]
        =
        \left[
            \begin{array}{ll}
                5 \\
                7 \\
                9 \\
            \end{array}
        \right].
    \end{equation}

    We can do this because summing (as well as other operations like calculating the mean, maximum, or the product) is commutative and associative.

    Another permutation-invariant function is numerical sorting.
    We will elaborate on how to make use of this in \autoref{cha:fspool}.
\end{myexample}

When building set encoders, this property of permutation-invariance is what is ultimately needed.
However, in the example, there are currently no weights to be learned and the possible operations are quite simplistic.
Using a single sum as set encoder can only express rather basic information about the set.
This is where the second property comes in, which adds in learnable weights.

\paragraph{Permutation-equivariance}
Permutation-equivariance says that the output of a function \emph{changes in a predictable way} when its input is permuted.
The difference to the permutation-invariance property is that the output is still a set, not a single feature vector.
Changing the order of the input should only change the order of the output, not any of its values.
Again, this property is to ensure that the arbitrary order of the set elements does not affect the result.

\begin{definition}
    A function $g: \R^{d \times n} \to \R^{c \times n}$ is permutation-equivariant iff it satisfies:

    \begin{equation}
        g(\mX\mP) = g(\mX) \mP
    \end{equation}

    for all permutation matrices $\mP$.
\end{definition}

\begin{myexample}
    The most important permutation-equivariant function is applying a function on each set element, i.e. on each feature vector.
    Since this function is applied on each set element individually, it does not take the order into account, so it is permutation-equivariant.
    Let us see what happens on our running example with the function $g(\vx) = \sum_i x_i $.

    \begin{equation}
        \left[
            \begin{array}{ll}
                1 & 4 \\
                2 & 5 \\
                3 & 6 \\
            \end{array}
        \right]
        \xrightarrow{\scalebox{0.7}{$g$}}
        \left[
            \begin{array}{ll}
                g\left(
                \begin{array}{l}
                    1 \\
                    2 \\
                    3 \\
                \end{array}
                \right)
                &
                g\left(
                \begin{array}{l}
                    4 \\
                    5 \\
                    6 \\
                \end{array}
                \right)
            \end{array}
        \right]
        =
        \left[
            \begin{array}{ll}
                6 & 15
            \end{array}
        \right]
    \end{equation}

    \begin{equation}
        \left[
            \begin{array}{ll}
                4 & 1 \\
                5 & 2 \\
                6 & 3 \\
            \end{array}
        \right]
        \xrightarrow{\scalebox{0.7}{$g$}}
        \left[
            \begin{array}{ll}
                g\left(
                \begin{array}{l}
                    4 \\
                    5 \\
                    6 \\
                \end{array}
                \right)
                &
                g\left(
                \begin{array}{l}
                    1 \\
                    2 \\
                    3 \\
                \end{array}
                \right)
            \end{array}
        \right]
        =
        \left[
            \begin{array}{ll}
                15 & 6
            \end{array}
        \right]
    \end{equation}

    Swapping the first and second element of the input also swaps the first and second element of the output.
    A widely-used choice for $g$ is a neural network, which lets the model \emph{learn} a transformation of the set elements.
    Another popular choice is concatenation of the same feature vector to every element, which is common in soft attention models~\citep{Ba2015,Mnih2014}.
\end{myexample}

With the two main properties in place, we can start combining functions with these properties to give us more complex functions.

\begin{corollary}
The composition of two permutation-equivariant functions $g$ and $h$ is permutation-equivariant.
\end{corollary}
\begin{proof}
    \begin{equation}
        h\big(g(\mX \mP)\big) = h\big(g(\mX) \mP\big) = h\big(g(\mX)\big) \mP
    \end{equation}
\end{proof}

\begin{corollary}
The composition of a permutation-equivariant function $g$ with a permutation-invariant function $f$ is permutation-invariant.
\end{corollary}
\begin{proof}
    \begin{equation}
        f\big(g(\mX\mP)\big) = f\big(g(\mX) \mP\big) = f\big(g(\mX)\big)
    \end{equation}
\end{proof}

This means that we can compose equivariant and invariant functions to build learnable and more powerful set encoders while still being permutation-invariant on the whole.

\subsection{Specific set encoders}
Let us look at some specific examples of set encoder architectures used in the literature.

\paragraph{Deep Sets}
One of the simplest set encoders is the Deep Sets model by \citet{Zaheer2017}.
At its core, it is defined as:

\begin{equation}
    h(\mX) = p\left(\sum_{\vx \in \mX} g(\vx)\right)
\end{equation}

We use $\vx \in \mX$ to refer to the columns in the matrix $\mX$, which correspond to the different set elements.
The set encoder $h$ applies a neural network $g$ on every set element $\vx$ (permutation-equivariant), then sums over the transformed set (permutation-invariant).
The feature vector result of this is fed into another neural network $p$ to perform the task at hand, such as classification.
While this seems quite basic, it turns out that it is a universal approximator for permutation-invariant functions~\citep{Zaheer2017,Wagstaff2019}.

\paragraph{Relation Networks}
Of course, universal approximation does not mean that the set encoder problem is solved, since it says nothing about learnability and generalisation.
Just like how there are improvements to neural networks with a single hidden layer (which are already capable of universal approximation~\citep{Cybenko1989}), there are improvements to the Deep Sets model as well.
One such improvement are Relation Networks~\citep{Santoro2017}:

\begin{equation}
    h(\mX) = p\left(\sum_{\vx, \vy \in \mX} g(\vx, \vy)\right)
\end{equation}

The difference here is that the sum is not over the transformed set elements, but over the transformed \emph{pairs} of set elements.
Expanding the set into the set of all pairs first is permutation-equivariant.
To model relations between different elements, the Deep Sets model can only let information from the set elements interact through the sum.
With the Relation Network, $g$ can model relations between pairs of elements much more easily since it receives the pairs as inputs.
This has been successfully used in tasks where relations between set elements are important~\citep{Zambaldi2018, Palm2018, Hu2018a}.
This comes at the cost of $\Theta(n^2)$ time complexity, compared to the $\Theta(n)$ complexity of Deep Sets.

\paragraph{PointNet}
PointNet \citep{Qi2017} is a model designed for 3d point clouds.
There are two main differences to what we have seen earlier:
the permutation-invariant function is not a sum, but a maximum, and there are multiple stages of encoding.

\begin{align}
    h_1(\mX) &= p_1\big(\max_{\vx \in \mX} g_1(\vx)\big) \\
    h_2(\mX) &= p_2\big(\max_{\vx \in \mX} g_2(\text{Concat}[\vx, h_1(\mX)])\big)
\end{align}

Here, the set is first encoded to a feature vector using an elementwise maximum in $h_1$.
This feature vector is concatenated to each element of the set again, and this new set is once again encoded.
This two-stage process lets global information about the set (obtained with $h_1$) be shared with the individual elements, which can then affect the representation in $h_2$.
Concatenation of a fixed feature vector to every element of the set is permutation-equivariant, since it is simply a function applied to every element.
$p_1, p_2, g_1,$ and $g_2$ are once again MLPs with trainable weights.

\paragraph{Attention}
Soft attention attempts to model the human ability to focus on one thing out of many things.
While not a full set encoder by itself, this is used as a building block in not only set encoders, but also in image and text encoders.
The typical soft attention model~\citep{Mnih2014,Ba2015} takes in a set and additional contextual information as input, and ``attends'' to part of the set that is relevant to that context.
This will come in use in \autoref{cha:vqa}, where the context is the question about the set.

\begin{align}
    h(\mX, \vz) = \sum_{\vx \in \mX} g(\vx, \vz) \vx
\end{align}

$g(\vx, \vz)$ is a neural network that outputs a scalar -- which makes this a weighted sum -- and is also often normalised with a softmax function so that it is a weighted average of the elements $\vx$.
The specific weightings depend on the context vector $\vz$.

Elements where $g$ has a high value have a greater influence on the result than where $g$ is low.
By giving certain elements higher weightings than others, the model can ``focus'' on the relevant elements of the set.

A modification of this idea called self-attention~\citep{Vaswani2017} has recently found great success in language modeling.
In essence, \emph{every} $\vx$ is used as $\vz$ to attend on the $\mX$ set (this is where the ``self'' comes from), rather than using a single $\vz$ from a separate model.
This model is related to Relation Networks, since all pairs of $\vx, \vy \in \mX$ are modeled by $g$ again. 

The example set encoders we have covered here demonstrate how compositions of permutation-invariant and permutation-equivariant functions are used in practice.

\subsection{Pooling bottleneck}\label{background sec:bottleneck}
We mentioned earlier that the typical permutation-invariant primitives have no learnable parameters.
Operations like sum and max are very simple, but have to compress a (potentially very large) set into a feature vector \emph{in a single step}.
This heavy compression can discard a lot of information that could be useful for the downstream task~\citep{Murphy2019,Qi2017}.

This is what motivates us to develop \emph{learned}, permutation-invariant approaches in \autoref{cha:perm-optim} and \autoref{cha:fspool}.
We show that by having a more sophisticated, learned model for this step, results are frequently improved.
Due to the nature of composability of these set function properties, replacing the sum or max pooling in an existing set encoder with our approaches is straightforward.

\section{Set decoders}\label{sec:background-decoders}

Set decoders turn a feature vector into a set and are used when the desired output is set-structured.
There are a variety of such tasks, ranging from object detection (predicting the set of objects in an image) to molecule generation (predicting the set of atoms and the set of bonds connecting them).
This problem turns out to be much less studied than the set encoder case.

There are two new aspects to consider: how can we make this set prediction with a neural network, and how can we compute a loss between sets to use as the training objective?

\subsection{Set losses}\label{background sec:setloss}
Regardless of how the set is predicted, we have to compute the loss between the predicted set and a ground-truth set in order to train the model.
The problem here is that both predicted set as well as ground-truth set are in an arbitrary order.
Na\"ively computing a pairwise loss -- such as a mean squared error -- does not work, since there is no guarantee that the elements are in the same order in the matrix representation.

\begin{myexample}
    Let us say that the prediction that a model has made is:

    \begin{equation}
        \left[
            \begin{array}{ll}
                1 & 4 \\
                2 & 5 \\
                3 & 6 \\
            \end{array}
        \right]
    \end{equation}

    but the set label was stored as:

    \begin{equation}
        \left[
            \begin{array}{ll}
                4 & 1 \\
                5 & 2 \\
                6 & 3 \\
            \end{array}
        \right].
    \end{equation}

    The model made the correct prediction of $\{[1,2,3]^\T, [4,5,6]^\T\}$, but a mean squared error would incorrectly think that the prediction was wrong.
    The order of the elements in the target set is arbitrary, so it is impossible for the model to guess the order.
    By pairing up the first column in the prediction with the second column in the label and vice versa, this issue is resolved.
\end{myexample}

Since the loss should not be affected by the order of either set, we need something that is permutation-invariant in both its arguments.
One such loss is the \emph{Chamfer loss}, which matches up every element of a predicted set $\mhY = [\vhy_1, \vhy_2, \ldots]$ to the closest element in the target set $\mY = [\vy_1, \vy_2, \ldots]$ and vice versa.
A normal pairwise loss -- like a mean squared error -- is used to measure this closeness.

\begin{equation}\label{background eq:chamfer}
    \Lcha(\mhY, \mY) = \sum_i \min_j \norm{\vhy_i - \vy_j }^2 + \sum_j \min_i \norm{\vhy_i - \vy_j }^2
\end{equation}

\begin{figure}
    \centering
    \centerfloat{\resizebox{\textwidth}{!}{%
    \begin{tikzpicture}[x=2.5cm,y=2.5cm, scale=1.4]
        \node at (0.5, 1.35) {\footnotesize Chamfer loss};
        \node (true0) [draw=Oranges-4-4, circle, fill=Oranges-4-3, inner sep=0.7mm] at (0, 0) {};
        \node (true1) [draw=Oranges-4-4, circle, fill=Oranges-4-3, inner sep=0.7mm] at (0, 1) {};
        \node (true2) [draw=Oranges-4-4, circle, fill=Oranges-4-3, inner sep=0.7mm] at (1, 1) {};
        \node (true3) [draw=Oranges-4-4, circle, fill=Oranges-4-3, inner sep=0.7mm] at (1, 0) {};
        \node (pred0) [draw=Blues-4-4!50!black, circle, fill=Blues-4-4, inner sep=0.7mm] at (0.4, 0.1) {};
        \node (pred1) [draw=Blues-4-4!50!black, circle, fill=Blues-4-4, inner sep=0.7mm] at (0.38, -0.3) {};
        \node (pred2) [draw=Blues-4-4!50!black, circle, fill=Blues-4-4, inner sep=0.7mm] at (0.8, 1.15) {};
        \node (pred3) [draw=Blues-4-4!50!black, circle, fill=Blues-4-4, inner sep=0.7mm] at (-0.1, 0.75) {};
        
        \draw [sedge, color=Blues-4-4, bend left=12] (pred0) to (true0);
        \draw [sedge, color=Blues-4-4] (pred1) to (true0);
        \draw [sedge, color=Blues-4-4, bend left=12] (pred2) to (true2);
        \draw [sedge, color=Blues-4-4, bend left=12] (pred3) to (true1);
        \draw [sedge, color=Oranges-4-3, bend right=12] (pred0) to (true0);
        \draw [sedge, color=Oranges-4-3] (pred0) to (true3);
        \draw [sedge, color=Oranges-4-3, bend right=12] (pred2) to (true2);
        \draw [sedge, color=Oranges-4-3, bend right=12] (pred3) to (true1);
    \end{tikzpicture}
    \hspace{2cm}
    \begin{tikzpicture}[x=2.5cm,y=2.5cm, scale=1.4]
        \node at (0.5, 1.35) {\footnotesize Hungarian loss};
        \node (true0) [draw=Oranges-4-4, circle, fill=Oranges-4-3, inner sep=0.7mm] at (0, 0) {};
        \node (true1) [draw=Oranges-4-4, circle, fill=Oranges-4-3, inner sep=0.7mm] at (0, 1) {};
        \node (true2) [draw=Oranges-4-4, circle, fill=Oranges-4-3, inner sep=0.7mm] at (1, 1) {};
        \node (true3) [draw=Oranges-4-4, circle, fill=Oranges-4-3, inner sep=0.7mm] at (1, 0) {};
        \node (pred0) [draw=Blues-4-4!50!black, circle, fill=Blues-4-4, inner sep=0.7mm] at (0.4, 0.1) {};
        \node (pred1) [draw=Blues-4-4!50!black, circle, fill=Blues-4-4, inner sep=0.7mm] at (0.38, -0.3) {};
        \node (pred2) [draw=Blues-4-4!50!black, circle, fill=Blues-4-4, inner sep=0.7mm] at (0.8, 1.15) {};
        \node (pred3) [draw=Blues-4-4!50!black, circle, fill=Blues-4-4, inner sep=0.7mm] at (-0.1, 0.75) {};
        
        \draw [sedge, color=black] (pred0) to (true3);
        \draw [sedge, color=black] (pred1) to (true0);
        \draw [sedge, color=black] (pred2) to (true2);
        \draw [sedge, color=black] (pred3) to (true1);
    \end{tikzpicture}
    }}
    \caption[Visualisations of set losses]{Example of the assignments of the Chamfer loss (left) and Hungarian loss (right).
    Orange points mark the target set, blue points mark the predicted set.
    The arrows show which direction the loss pulls a predicted point.
    Each point of one colour (e.g. orange) is matched up with the closest point of the other colour (e.g. blue).
    Blue arrows denote assignments from the first term in \autoref{background eq:chamfer}, orange arrows denote assignments from the second term.}
    \label{background fig:loss}
\end{figure}
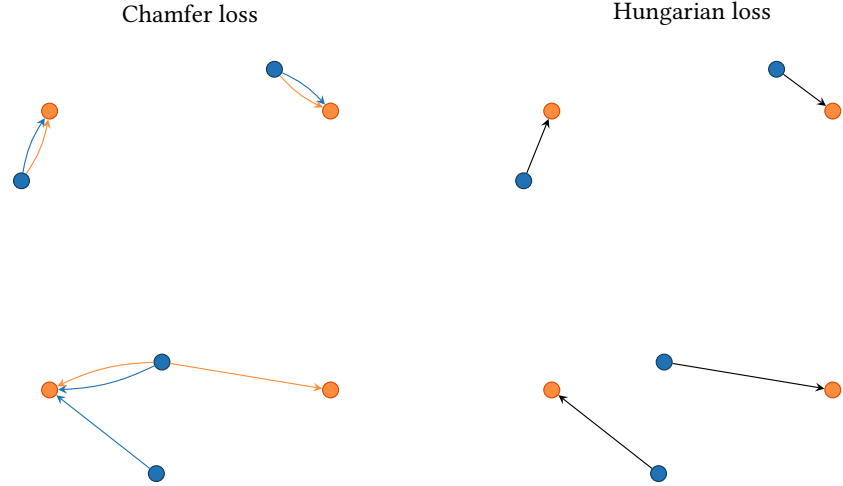

The nested sums and minimisations ensure the desired permutation invariance in both arguments.
One issue is that one element in one set can be matched up with multiple elements in the other set.
This means that there can be issues with different set sizes and duplicates: the loss between $[1, 1, 2]$ and $[1, 2, 2, 2]$ is 0, even though the multisets that are represented are clearly different.
Conceptually, the Chamfer loss treats the inputs as proper sets, not the multisets that we usually work with.

A more sophisticated loss that does not have this problem involves the linear assignment problem with the pairwise losses as assignment costs:

\begin{equation}
    \Lhun(\mhY, \mY) = \min_{\pi \in \Pi} \sum_i \norm{\vhy_i - \vy_{\pi(i)}}^2
\end{equation}

where $\Pi$ is the space of all permutations.
This minimisation can be solved exactly with the Hungarian algorithm and has the benefit of assigning each element in one set to exactly one element from the other set.
This comes with the trade-off of $\Theta(n^3)$ time complexity and that it is not easily parallelisable.
We visualise the difference between these two losses in \autoref{background fig:loss}, which shows the Hungarian loss matching up elements one-to-one.

\subsection{Predicting sets}
Now that we know how to define the training objective for set prediction models, we need a model that is able to predict sets.
The main approach to predicting sets in the literature is to use an MLP or RNN that simply predicts the matrix representation of the set.
To predict a set of size $n$ with vectors of size $d$, an MLP with $nd$ number of outputs is used.
Alternatively, the initial state of an RNN (such as an LSTM~\citep{Hochreiter1997}) with $d$ hidden units is initialised with the input feature vector and run for $n$ steps to produce the $n \times d$ outputs.
To handle sets with variable sizes, one option is to pad all sets in a dataset with zeros to a fixed size and concatenate an additional feature to each set element, indicating whether that element is a padding element.
These two approaches are essentially the only general ways to predict sets in the literature without making domain-specific assumptions about the sets. 

\begin{sidenote}{Aside: Object detection}
    For some specialised tasks like object detection, there are existing approaches like Faster R-CNN~\citep{Ren2015} that are not based on sets, but a combination of heuristics.
    These are specifically tuned to the image domain input and have certain drawbacks that a set-based approach avoids.
    A set-based approach has the advantage of being fully end-to-end differentiable, which potentially improves the quality of the predictions compared to a multi-stage object detector like Faster R-CNN.
    Traditional object detectors also require post-processing of the outputs outside of the training algorithm with techniques like non-maximum suppression, which is somewhat inelegant.

    The goal of this thesis is to explore approaches which work generically for sets, without requiring input domain-specific changes.
    Therefore, we limit ourselves to feature vectors as input when predicting sets.
    Structured data like image feature maps can always be turned into a feature vector, but not necessarily vice versa.
\end{sidenote}

These methods are somewhat unsatisfying, since they treat the set as if it were just a list, with the only difference to predicting a list being the use of the set loss.
The MLP and RNN outputs are \emph{ordered}, but they are used to predict sets, which are \emph{unordered}.
We indeed find in \autoref{fspool sec:problem} that this can cause a \emph{responsibility problem}.
The mismatch between ordered MLP or RNN outputs and unordered sets results in discontinuity issues which hinders learning.
The study in \autoref{cha:fspool} is in part about this problem and our solution to it in the auto-encoding setting.
Then, we build on this work to develop a set prediction method in \autoref{cha:dspn} without the responsibility problem that is no longer limited to the auto-encoder setting.
We show that our method can perform tasks like end-to-end object detection in a purely set-based way.

\section{Other sets in machine learning}\label{sec:background-other}
In this section, we discuss some related topics on sets in machine learning.
This puts our work on sets into a broader context.

\subsection{Pooling in CNNs}
Global average pooling, which is commonly applied at the end of CNNs such as ResNets~\citep{He2016}, average the feature vectors across all spatial positions in the CNN feature map.
Essentially, global average pooling treats the different spatial positions as if they were a set.
The same applies to global max pooling in order to summarise a feature map into a feature vector, which has been used by~\citet{Perez2018}.
This can lead to surprising consequences where small patches of the input image can be shuffled around and processed independently (just like a set) without much loss of performance~\citep{Brendel2018}.

On a smaller scale, this applies to the non-global average and max pooling as well.
For max pooling with a kernel size of $2\times2$, the four input feature vectors are treated as if they were a set.

Since these pooling methods are simply permutation-invariant functions, we can substitute them with any other permutation-invariant function.
Our methods in \autoref{cha:perm-optim} and \autoref{cha:fspool} could therefore be used in CNNs.
However, it is unclear whether they would make any significant difference with the much greater importance of convolutions over poolings in CNNs.

\subsection{Per-element prediction}
In some tasks, we have a set as input and want to predict something about each element of the set, rather than the set as a whole.
In other words, these are set-to-set problems.
While each element could be independently predicted, there is some prior belief that there is  additional information to be gained from considering the other elements in the set at the same time.
An example of this is detecting an outlier in a set of inputs~\citep{Lee2019}:
determining an outlier requires finding commonalities between the non-outliers, so it is sensible to make a prediction about all elements in the set at once rather than one-by-one.

Unlike in set encoders, we want the output in these models to be equivariant rather than invariant.
If $h([\va, \vb, \vc]) = [\va', \vb', \vc']$, then changing the input to $h([\vb, \va, \vc])$ should give us $[\vb', \va', \vc']$.
Because the order of the output is always the same as the input, there is no responsibility problem and no need for the assignment-based set losses.
This makes it an easier problem than something like image-to-set.
We will make use of this benefit in \autoref{cha:fspool}.

Taking a step back, supervised learning with a \emph{dataset} is itself a set problem.
The inputs in the dataset are fed into a model -- the machine learning algorithm -- which predicts an output for each element in the set (set-to-set).
There is usually a loss over the dataset, which is permutation-invariant by averaging or summing across the individual losses.
From this, computing things like $\partial \text{ loss} / \partial \text{ parameters}$ is permutation-invariant too.
Training algorithms that do not use the whole dataset at once (like SGD) are made invariant in expectation by randomly sampling from the dataset.
Using a permutation of the dataset that is easier to learn with is known as curriculum learning~\citep{Bengio2009}.

\subsection{Multi-labeling}
Multi-labeling tasks can be seen as set prediction problems, where the goal is predict a set of labels.
These labels come from a fixed, finite set of all possible labels.
The main difference to our set prediction setup is that because there are only a finite number of possible labels, we can pick a fixed arbitrary ordering for these labels just like in classification or word embeddings.
An n-hot vector encoding (a vector with a 0 in a dimension if the label is not present and a 1 when the label is present) therefore suffices for multi-label classification, though more set-oriented methods have also been studied~\citep{Rezatofighi2017, Belanger2016}.

In contrast, the sets we are working with have vectors in $\R^d$ as elements.
With infinitely many possible elements, there is no order we can enforce on the set elements a priori, so the strategy for multi-labeling does not work.
Choosing one anyway results discontinuities in the labels, similar to the responsibility problem that we discuss in \autoref{fspool sec:problem}.
This is why we need something like our model in \autoref{cha:dspn} to predict these more complex sets with elements in $\R^d$.

\subsection{Clustering}
Lastly, clustering can be seen as a set prediction task.
The goal in clustering is to find a partitioning of the input set where ``similar'' points are grouped into the same partition.
So, the output is a \emph{set of sets}, where the inner sets contain the elements of the input.
This is usually done in an unsupervised setting, while our method in \autoref{cha:dspn} works in the supervised learning setting.
Our approach to set prediction could be used to train a \emph{supervised} clustering algorithm like \citet{Finley2005}.

\chapter{Motivation: Counting in visual question answering}\label{cha:vqa}
\chaptermark{Counting in VQA}
In this chapter, we will motivate our work on sets with the visual question answering task.
We will develop a method for counting sets of object proposals that have associated bounding box information.
These sets can be obtained from traditional object detectors.

When we worked on this problem, our focus was not on sets in general, but specifically the set of object proposals.
We later realised the fundamental importance of properly using the given set representation, which motivated us to study sets more generally in later chapters.

These contributions have been published as~\citep{Zhang2018Counting} in the International Conference on Learning Representations (ICLR) 2018 and have been presented at the Visual Question Answering Challenge workshop, hosted at the Conference on Computer Vision and Pattern Recognition (CVPR) 2018.

\begin{figure}
    \centering
    \fontsize{10}{10}\selectfont
    \centerfloat{\resizebox{\figurewidth}{!}{%
    \begin{tikzpicture}[x=1.5cm, y=1.5cm]
        \node (a) {\begin{tikzpicture}[x=1.5cm, y=1.5cm]\pic {a};\end{tikzpicture}};
        \node [right = 0.5cm of a] (A) {\begin{tikzpicture}[x=1.5cm, y=1.5cm]\pic {Acolor};\end{tikzpicture}};
        \node [right = 1 cm of A, yshift=1mm] (C2) {\begin{tikzpicture}[x=1.5cm, y=1.5cm]\pic {C2};\end{tikzpicture}};
        \draw [sedge] ([xshift=2mm]a.east) -- (A);
        \draw [sedge, decoration={snake, post length=0.5mm, amplitude=0.5mm}, decorate] (A) -- (C2);
    \end{tikzpicture}
    }}
    \caption[Overview of counting module]{
        Simplified example about counting the number of cats.
        The light-coloured cat is detected twice and results in a duplicate proposal.
        This shows the conversion from the attention weights $\va$ to a graph representation $\mA$ and the eventual goal of this module with exactly one proposal per true object.
        There are 4 proposals (vertices) capturing 3 underlying objects (groups in dotted lines).
        There are 3 relevant proposals (black with weight 1) and 1 irrelevant proposal (white with weight 0).
        Red edges mark \emph{intra}-object edges between duplicate proposals and blue edges mark the main \emph{inter}-object duplicate edges.
        In graph form, the object groups, colouring of edges, and shading of vertices serve illustration purposes only; the model does not have these access to these directly.
    }
    \label{count fig:overview}
\end{figure}
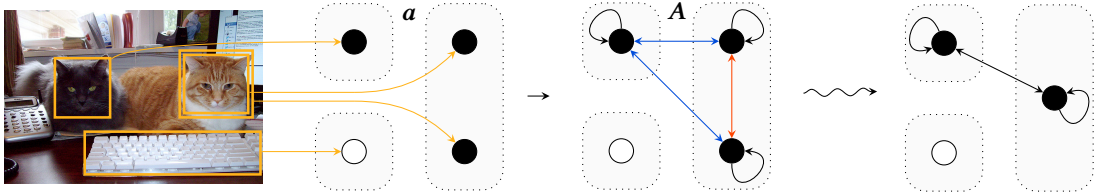

\section{Introduction}\label{count sec:intro}
Consider the problem of counting how many cats there are in \autoref{count fig:overview}.
Solving this involves several rough steps: understanding what instances of that type can look like, finding them in the image, and adding them up.
This is a common task in visual question answering (VQA) -- answering questions about images -- and is rated as among the tasks requiring the lowest human age to be able to answer~\citep{Antol2015}.
However, current models for VQA on natural images struggle to answer \emph{any} counting questions successfully outside of dataset biases~\citep{Jabri2016}.

One reason for this is the presence of a fundamental problem with counting in the widely-used soft attention mechanisms (\autoref{count sec:problems}) that we identify.
Another reason is that unlike standard counting tasks, there is no ground truth labelling of where the objects to count are.
Coupled with the fact that models need to be able to count a large variety of objects and that, ideally, performance on non-counting questions should not be compromised, the task of counting in VQA seems very challenging.

To make this task easier, we can use object proposals -- a set of pairs of a bounding box and object features -- from object detection networks as input instead of learning from pixels directly.
In any moderately complex scene, this runs into the issue of double-counting overlapping object proposals.
This is a problem present in many natural images, which leads to inaccurate counting in real-world scenarios.

Our main contribution is a differentiable neural network module that tackles this problem and consequently can learn to count (\autoref{count sec:counting}).
Used alongside an attention mechanism, this module avoids a fundamental limitation of soft attention while producing strong counting features.
As the input is a set, we build our model to be permutation-invariant to the order of the object proposals.
We then provide experimental evidence of the effectiveness of this module (\autoref{count sec:experiments}).
On a toy dataset, we demonstrate that this module enables robust counting in a variety of scenarios.
On the number category of the VQA v2 Open-Ended dataset~\citep{Goyal2017}, a relatively simple baseline model using the counting module outperforms all previous models -- including large ensembles of state-of-the-art methods -- without degrading performance on other categories.

\section{Related work}\label{count sec:related}
Usually, greedy non-maximum suppression (NMS) is used to eliminate duplicate bounding boxes.
The main problem with using it as part of a model is that its gradient is piecewise constant.
Various differentiable variants such as by \citet{Azadi2017}, \citet{Hosang2017}, and \citet{Henderson2017} exist.
The main difference is that, since we are interested in counting, our module does not need to make discrete decisions about which bounding boxes to keep; it outputs counting features, not a smaller set of bounding boxes.
Our module is also easily integrated into standard VQA models that utilise soft attention without any need for other network architecture changes and can be used without using true bounding boxes for supervision.

On the VQA v2 dataset~\citep{Goyal2017} that we apply our method on, only few advances on counting questions have been made.
The main improvement in accuracy is due to the use of object proposals in the visual processing pipeline, proposed by \citet{Anderson2018}.
Their object proposal network is trained with classes in singular and plural forms, for example ``tree'' versus ``trees'', which only allows primitive counting information to be present in the object features after region-of-interest pooling.
Our approach differs in the way that instead of relying on counting features being present in the input, we create counting features using information present in the attention map over object proposals.
This has the benefit of being able to count anything that the attention mechanism can discriminate instead of only objects that belong to the predetermined set of classes that had plural forms.

Using these object proposals, \citet{Trott2018} train a sequential counting mechanism with a reinforcement learning loss on the counting question subsets of VQA v2 and Visual Genome.
They achieve a small increase in accuracy and can obtain an interpretable set of objects that their model counted, but it is unclear whether their method can be integrated into traditional VQA models due to their loss not applying to non-counting questions.
Since they evaluate on their own dataset, their results can not be easily compared to existing results in VQA.

Methods such as by \citet{Santoro2017} and \citet{Perez2018} can count on the synthetic CLEVR VQA dataset~\citep{Johnson2017} successfully without bounding boxes and supervision of where the objects to count are.
They also use more training data ($\sim$250,000 counting questions in the CLEVR training set versus $\sim$50,000 counting questions in the VQA v2 training set), much simpler objects, and synthetic question structures.

More traditional approaches based on \citet{Lempitsky2010} learn to produce a target density map, from which a count is computed by integrating over it.
In this setting, \citet{Cohen2017} make use of overlaps of convolutional receptive fields to improve counting performance.
\citet{Chattopadhyay2017} use an approach that divides the image into smaller non-overlapping chunks, each of which is counted individually and combined together at the end.
In both of these contexts, the convolutional receptive fields or chunks can be seen as sets of bounding boxes with a fixed structure in their positioning.
Note that while \citet{Chattopadhyay2017} evaluate their models on a small subset of counting questions in VQA, major differences in training setup make their results not comparable to our work.

\section{Problems with soft attention}\label{count sec:problems}

\begin{figure}
    \centering
    \centerfloat{\resizebox{\textwidth}{!}{%
        \begin{tikzpicture}
            \node {\begin{tikzpicture}\pic {probs};\end{tikzpicture}};
        \end{tikzpicture}
    }}
    \caption[Problem with counting using soft attention]{Example demonstrating the problem with using soft attention for counting. The resulting two feature vectors are the same for the two images, even though the images show a different number of stars.}
    \label{count fig:stars}
\end{figure}
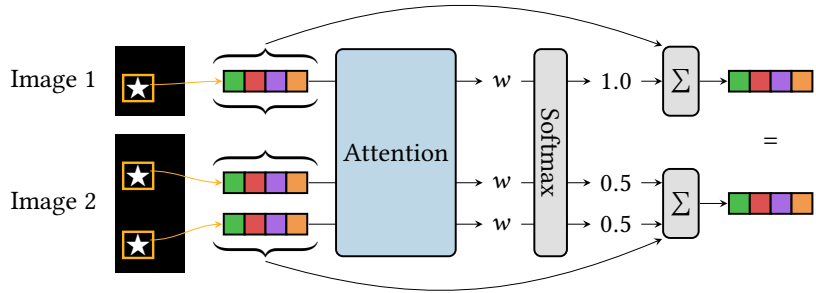

The main message in this section is that using the feature vectors obtained after the attention mechanism is not enough to be able to count;
the attention maps themselves should be used, which is what we do in our counting module.

Models in VQA have consistently benefited from the use of soft attention~\citep{Mnih2014, Bahdanau2015} on the image, commonly implemented with a shallow convolutional network.
It learns to output a weight for the feature vector at each spatial position in the feature map, which is first normalised and then used for performing a weighted sum over the spatial positions to produce a single feature vector.
However, soft spatial attention severely limits the ability for a model to count.

Consider the task of counting the number of stars for two images (\autoref{count fig:stars}): an image showing a single star on a clean background and an image that consists of two side-by-side copies of the first image.
What we will describe applies to both spatial feature maps and sets of object proposals as input, but we focus on the latter case for simplicity.
With an object detection network, we detect one star in the first image and two stars in the second image, producing the same feature vector for all three detections.
The attention mechanism then assigns all three instances of the same star the same weight.

The usual normalisation used for the attention weights is the softmax function, which normalises the weights to sum to 1.
Herein lies the problem:
the star in the first image receives a normalised weight of 1, but the two stars in the second image now each receive a weight of 0.5.
After the weighted sum, we are effectively averaging the two stars in the second image back to a single star.
As a consequence, the feature vector obtained after the weighted sum is exactly the same between the two images and we have lost all information about a possible count from the attention map.
Any method that normalises the weights to sum to 1 suffers from this issue.

Multiple glimpses~\citep{Larochelle2010} -- sets of attention weights that the attention mechanism outputs -- or several steps of attention~\citep{Yang2016, Lu2016} do not circumvent this problem.
Each glimpse or step can not separate out an object each, since the attention weight given to one feature vector does not depend on the other feature vectors to be attended over.
Hard attention~\citep{Ba2015, Mnih2014} and structured attention~\citep{Kim2017} may be possible solutions to this, though no significant improvement in counting ability has been found for the latter so far~\citep{Zhu2017}.
\citet{Ren2017} circumvent the problem by limiting attention to only work within one bounding box at a time, remotely similar to our approach of using object proposal features.

Without normalisation of weights to sum to one, the scale of the output features depends on the number of objects detected.
In an image with 10 stars, the output feature vector is scaled up by 10.
Since deep neural networks are typically very scale-sensitive -- the scale of weight initialisations and activations is generally considered quite important~\citep{Mishkin2016} -- and the classifier would have to learn that joint scaling of all features is somehow related to count, this approach is not reasonable for counting objects.
This is shown in \citet{Teney2017} where they provide evidence that sigmoid normalisation not only degrades accuracy on non-number questions slightly, but also does not help with counting.

\section{Counting module}\label{count sec:counting}
\sectionmark{Model}

In this section, we describe a differentiable mechanism for counting from attention weights, while also dealing with the problem of overlapping object proposals to reduce double-counting of objects.
This involves some nontrivial details to produce counts that are as accurate as possible.
The main idea is illustrated in \autoref{count fig:overview} with the two main steps shown in \autoref{count fig:intra} and \autoref{count fig:inter}.
The use of this module allows a model to count while still being able to exploit the benefits of soft attention.

Our key idea for dealing with overlapping object proposals is to turn these object proposals into a graph that is based on how they overlap.
We then remove and scale edges in a specific way such that an estimate of the number of underlying objects is recovered.

Our general strategy is to primarily design the module for the unrealistic extreme cases of perfect attention maps and bounding boxes that are either fully overlapping or fully distinct.
By introducing some parameters and only using differentiable operations, we give the ability for the module to interpolate between the correct behaviours for these extreme cases to handle the more realistic cases.
These parameters are responsible for handling variations in attention weights and partial bounding box overlaps in a manner suitable for a given dataset.

\subsection{Piecewise linear activation}
To introduce these parameters, we use several piecewise linear functions $f_1, \ldots, f_8$ as activation functions, approximating arbitrary functions with domain and range [0, 1].
We show how these functions look after training in \autoref{count fig:trained-plins} (\autopageref{count fig:trained-plins}).
The shapes of these functions are learned to handle the specific nonlinear interactions necessary for dealing with overlapping proposals.
Through their parametrisation we enforce that $f_k(0) = 0$, $f_k(1) = 1$, and that they are monotonically increasing.
The first two properties are required so that the extreme cases that we explicitly handle are left unchanged.
In those cases, $f_k$ is only applied to values of 0 or 1, so the activation functions can be safely ignored for understanding how the module handles them.
By enforcing monotonicity, we can make sure that, for example, an increased value in an attention map should never result in the prediction of the count to decrease.

Intuitively, the interval $[0, 1]$ is split into $d$ equal size intervals.
Each contains a line segment that is connected to the neighbouring line segments at the boundaries of the intervals.
These line segments form the shape of the activation function.

For each function $f_k$, there are $d$ weights $w_{k1}, \ldots, w_{kd}$, where the weight $w_{ki}$ is the gradient for the interval $[\frac{i-1}{d}, \frac{i}{d})$.
We arbitrarily fix $d$ to be 16 in this chapter, observing no significant difference when changing it to 8 and 32 in preliminary experiments.
All $w_{ki}$ are enforced to be non-negative by always using the absolute value of them, which yields the monotonicity property.
Dividing the weights by $\sum_m^d |w_{km}|$ yields the property that $f(1) = 1$.
The function can be written as

\begin{equation}
    f_k(x) = \sum_{i=1}^d \max(0, 1 - | d x - i |) \frac{\sum_{j=1}^i |w_{kj}|}{\sum_{m=1}^d |w_{km}|}
\end{equation}

In essence, the $\max$ term selects the two nearest boundary values of an interval, which are normalised cumulative sums over the $w_k$ weights, and linearly interpolates between the two.
This approach is similar to the subgradient approach by \citet{Jaderberg2015} to make sampling from indices differentiable.
All $w_{ki}$ are initialised to 1, which makes the functions linear on initialisation.
When applying $f_k(\vx)$ to a vector-valued input $\vx$, it is assumed to be applied elementwise.
By caching the normalised cumulative sum $ \sum_j^i |w_{kj}| / \sum_m^d |w_{km}|$, this function has linear time complexity with respect to $d$ and is efficiently implementable on GPUs.

Extensions to this are possible through Deep Lattice Networks~\citep{You2017}, which preserve monotonicity across several nonlinear neural network layers.
They would allow $\mA$ and $\mD$ to be combined in more sophisticated ways beyond an elementwise product, possibly improving counting performance as long as the property of the range lying within $[0, 1]$ is still enforced in some way.

\subsection{Input}
Given a set of features from object proposals, an attention mechanism produces a weight for each proposal based on the question.
The counting module takes as input the $n$ largest attention weights $\va = [a_1, \ldots, a_n]^\T$ and their corresponding bounding boxes $\vb = [b_1, \ldots, b_n]^\T$.
We assume that the weights lie in the interval $[0, 1]$, which can easily be achieved by applying a logistic function on the attention map.

In the extreme cases that we explicitly handle, we assume that the attention mechanism assigns a value of 1 to $a_i$ whenever the $i$th proposal contains a relevant object and a value of 0 whenever it does not.
This is in line with what usual soft attention mechanisms learn, as they produce higher weights for relevant inputs.
We also assume that either two object proposals fully overlap (in which case they must be showing the same object and thus receive the same attention weight) or that they are fully distinct (in which case they show different objects).
Keep in mind that while we make these assumptions to make reasoning about the behaviour easier, the learned parameters in the activation functions are intended to handle the more realistic scenarios when the assumptions do not apply.

Instead of partially overlapping proposals, the problem now becomes the handling of \emph{exact duplicate} proposals of underlying objects in a differentiable manner.

\subsection{Deduplication}
We start by changing the vector of attention weights $\va$ into a graph representation in which bounding boxes can be utilised more easily.
Hence, we compute the outer product of the attention weights to obtain an attention matrix.

\begin{equation}
    \mA = \va\va^\T
\end{equation}

$\mA \in \R^{n \times n}$ can be interpreted as an adjacency matrix for a weighted directed graph.
In this graph, the $i$th vertex represents the object proposal associated with $a_i$ and the edge between any pair of vertices $(i, j)$ has weight $a_i a_j$.
In the extreme case where $a_i$ is virtually $0$ or $1$, products are equivalent to logical AND operators.
It follows that the subgraph containing only the vertices satisfying $a_i = 1$ is a complete digraph with self-loops.

In this representation, our objective is to eliminate edges in such a way that, conceptually, the underlying true objects -- instead of proposals thereof -- are the vertices of that complete subgraph.
In order to then turn that graph into a count, recall that the number of edges $|E|$ in a complete digraph with self-loops relates to the number of vertices $|V|$ through $|E| = |V|^2$.
$|E|$ can be computed by summing over the entries in an adjacency matrix and $|V|$ is then the count.
Notice that when $|E|$ is set to the sum over $\mA$, $|E| = \sum_{ij} \va_i \va_j = (\sum_{i} \va_i)^2$, which implies $|V| = \sum_i \va_i$.
This convenient property implies that when all proposals are fully distinct, the module can output the same as simply summing over the original attention weights by default.

There are two types of duplicate edges to eliminate to achieve our objective: \emph{intra}-object edges and \emph{inter}-object edges.

\subsubsection{Intra-object edges}

First, we eliminate intra-object edges between duplicate proposals of a single underlying object.

To compare two bounding boxes, we use the usual intersection-over-union (IoU) metric.
We define the distance matrix $\mD \in \mathbb{R}^{n \times n}$ to be

\begin{equation}
    D_{ij} = 1 - \text{IoU}(b_i, b_j)
\end{equation}

$\mD$ can also be interpreted as an adjacency matrix.
It represents a graph that has edges everywhere except when the two bounding boxes that an edge connects would overlap.

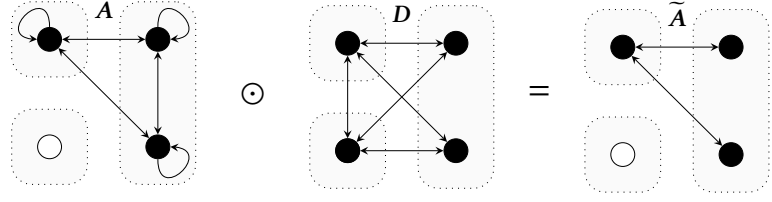
\begin{figure}
    \centering
    \fontsize{10}{10}\selectfont
    \centerfloat{\scalebox{0.95}{
    \begin{tikzpicture}[x=1.5cm, y=1.5cm]
        \clip (-1, -1) rectangle (6.25, 1);
        \node (A) {\begin{tikzpicture}[x=1.5cm, y=1.5cm]\pic {A};\end{tikzpicture}};
        \node [right = of A] (O) {\begin{tikzpicture}[x=1.5cm, y=1.5cm]\pic {O};\end{tikzpicture}};
        \node [right = of O] (B) {\begin{tikzpicture}[x=1.5cm, y=1.5cm]\pic {B};\end{tikzpicture}};
        \node at ($(A)!0.5!(O)$) {\Large $\odot$};
        \node at ($(O)!0.5!(B)$) {\Large $=$};
    \end{tikzpicture}
    }}
    \caption[Intra-object edge removal]{
        Removal of intra-object edges by masking the edges of the attention matrix $\mA$ with the distance matrix $\mD$.
        The black vertices now form a graph without self-loops.
        The self-loops need to be added back in later.
    }
    \label{count fig:intra}
\end{figure}

Intra-object edges are removed by elementwise multiplying ($\odot$) the distance matrix with the attention matrix (\autoref{count fig:intra}).

\begin{equation}
    \AO = f_1 (\mA) \odot f_2 (\mD)
\end{equation}

$\AO$ no longer has self-loops, so we need to add them back in at a later point to still satisfy $|E| = |V|^2$.
Notice that we start making use of the activation functions mentioned earlier to handle intermediate values in the interval $(0, 1)$ for both $\mA$ and $\mD$.
They regulate the influence of attention weights that are not close to 0 or 1 and the influence of partial overlaps.

\subsubsection{Inter-object edges}

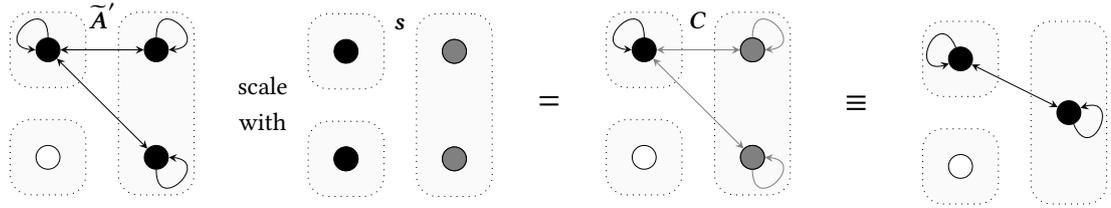
\begin{figure}
    \centering
    \fontsize{10}{10}\selectfont
    \centerfloat{\scalebox{0.95}{
        \begin{tikzpicture}[x=1.5cm, y=1.5cm]
        \clip (-1, -1) rectangle (9.4, 1);
        \node (B) {\begin{tikzpicture}[x=1.5cm, y=1.5cm]\pic {Bprime};\end{tikzpicture}};
        \node [right = of B] (s) {\begin{tikzpicture}[x=1.5cm, y=1.5cm]\pic {s};\end{tikzpicture}};
        \node [right = of s] (C) {\begin{tikzpicture}[x=1.5cm, y=1.5cm]\pic {C};\end{tikzpicture}};
        \node [right = of C] (C2) {\begin{tikzpicture}[x=1.5cm, y=1.5cm]\pic {C2};\end{tikzpicture}};
        \draw [draw=no] (B) -- (s) node [midway, above] {scale} node[midway, below] {with};
        \node at ($(s)!0.5!(C)$) {\Large $=$};
        \node at ($(C)!0.5!(C2)$) {\Large $\equiv$};
    \end{tikzpicture}
    }}
    \caption[Inter-object edge removal]{
        Removal of duplicate inter-object edges by computing a scaling factor for each vertex and scaling $\AO '$ accordingly.
        $\AO '$ is $\AO$ with self-loops already added back in.
        The scaling factor for one vertex is computed by counting how many vertices have outgoing edges to the same set of vertices; all edges of the two proposals on the right are scaled by $0.5$.
        This can be seen as averaging proposals within each object and is equivalent to removing duplicate proposals altogether under a sum.
    }
    \label{count fig:inter}
\end{figure}

Second, we eliminate inter-object edges between duplicate proposals of different underlying objects.

The main idea (depicted in \autoref{count fig:inter}) is to count the number of proposals associated to each individual object, then scale down the weight of their associated edges by that number.
If there are two proposals of a single object, the edges involving those proposals should be scaled by 0.5.
In essence, this averages over the proposals within each underlying object because we only use the sum over the edge weights to compute the count at the end.
Conceptually, this reduces multiple proposals of an object down to one as desired.
Since we do not know how many proposals belong to an object, we have to estimate this.
We do this by using the fact that proposals of the same object are similar.

Keep in mind that $\AO$ has no self-loops nor edges between proposals of the same object.
As a consequence, two nonzero rows in $\AO$ are the same if and only if the proposals are the same.
If the two rows differ in at least one entry, then one proposal overlaps a proposal that the other proposal does not overlap, so they must be different proposals.
This means for comparing rows, we need a similarity function that satisfies the criteria of taking the value $1$ when they differ in no places and $0$ if they differ in at least one place.
We define a differentiable similarity between proposals $i$ and $j$ as

\begin{equation}
    \text{Sim}_{ij} =  f_3(1 - |a_i - a_j|) \prod_k f_3(1 - | X_{ik} - X_{jk} |)
\end{equation}
where $\mX = f_4(\mA) \odot f_5(\mD)$ is the same as $\AO$ except with different activation functions.
The $\prod$ term compares the rows of proposals $i$ and $j$.
Using this term instead of $f_4(1 - D_{ij})$ was more robust to inaccurate bounding boxes in initial experiments.

Note that the $f_3(1 - |a_i - a_j|)$ term handles the edge case when there is only one proposal to count.
Since $\mX$ does not have self-loops, $\mX$ contains only zeros in that case, which causes the row corresponding to $a_i = 1$ to be incorrectly similar to the rows where $a_{j \neq i} = 0$.
By comparing the attention weights through that term as well, this issue is avoided.

Now that we can check how similar two proposals are, we count the number of times any row is the same as any other row and compute a scaling factor $s_i$ for each vertex $i$.

\begin{equation}
    s_i =  1 / \sum_j \text{Sim}_{ij}
\end{equation}

The time complexity of computing $\vs = [s_1, \ldots, s_n]^\T$ is $\Theta(n^3)$ as there are $n^2$ pairs of rows and $\Theta(n)$ operations to compute the similarity of any pair of rows.

Since these scaling factors apply to each vertex, we have to expand $\vs$ into a matrix using the outer product in order to scale both incoming and outgoing edges of each vertex.
We can also add self-loops back in, which need to be scaled by $\vs$ as well.
Then, the count matrix $\mC$ is

\begin{equation}
    \mC = \m \AO \odot \vs \vs^\T + \text{diag}(\vs \odot f_1(\va \odot \va))
\end{equation}

where diag$(\cdot)$ expands a vector into a diagonal matrix with the vector on the diagonal.

The scaling of self-loops involves a non-obvious detail.
Recall that the diagonal that was removed when going from $\mA$ to $\AO$ contains the entries $f_1(\va \odot \va)$.
Notice however that we are scaling this diagonal by $\vs$ and not $\vs \odot \vs$.
This is because the number of inter-object edges scales quadratically with respect to the number of proposals per object, but the number of self-loops only scales linearly.

\subsection{Output}
Under a sum, $\mC$ is now equivalent to a complete graph with self-loops that involves all relevant objects instead of relevant proposals as originally desired.

To turn $\mC$ into a count $c$, we set $|E| = \sum_{i,j} C_{ij}$ as mentioned and
\begin{equation}
    c = |V| = \sqrt{|E|}
\end{equation}

We verified experimentally that when our extreme case assumptions hold, $c$ is always an integer and equal to the correct count, regardless of the number of duplicate object proposals.

To avoid issues with scale when the number of objects is large, we turn this single feature into several classes, one for each possible number.
Since we only used the object proposals with the largest $n$ weights, the predicted count $c$ can be at most $n$.
We define the output $\vo = [o_0, o_1, \ldots, o_n]^\T$ to be
\begin{equation}\label{count eq:output}
    o_i = \max(0, 1 - |c - i|)
\end{equation}

This results in a vector that is 1 at the index of the count and 0 everywhere else when $c$ is exactly an integer, and a linear interpolation between the two corresponding one-hot vectors when the count falls in-between two integers.

\subsection{Output confidence}
Finally, we might consider a prediction made from values of $\va$ and $\mD$ that are either close to $0$ or close to $1$ to be more reliable -- we explicitly handle these after all -- than when many values are close to $0.5$.
To incorporate this idea, we scale $\vo$ by a confidence value in the interval $[0, 1]$.

We define $p_{\va}$ and $p_{\mD}$ to be the average distances to $0.5$.
The choice of $0.5$ is not important, because the module can learn to change it by changing where $f_6(x) = 0.5$ and $f_7(x) = 0.5$.
\begin{align}
    p_{\va} &= \frac{1}{n} \sum_{i} |f_6(a_{i}) - 0.5|\\
    p_{\mD} &= \frac{1}{n^2} \sum_{i, j} |f_7(D_{ij}) - 0.5|
\end{align}

Then, the output of the module with confidence scaling is
\begin{equation}
    \widetilde{\vo} =  f_8(p_{\va} + p_{\mD}) \cdot \vo
\end{equation}

In summary, we only used differentiable operations to deduplicate object proposals and obtain a feature vector that represents the predicted count.
This allows easy integration into any model with soft attention, enabling a model to count from an attention map.
Each step that we applied is either permutation-equivariant or permutation-invariant, which makes our whole model permutation-invariant.

\section{Experiments}\label{count sec:experiments}
We provide the source code to reproduce our experiments at \url{https://github.com/Cyanogenoid/vqa-counting}.

\subsection{Toy task}\label{count sec:toy}
\begin{figure}
    \centering
    \centerfloat{
        \includegraphics[width=0.46\figurewidth, trim={0 4.5mm 0 4.5mm}, clip]{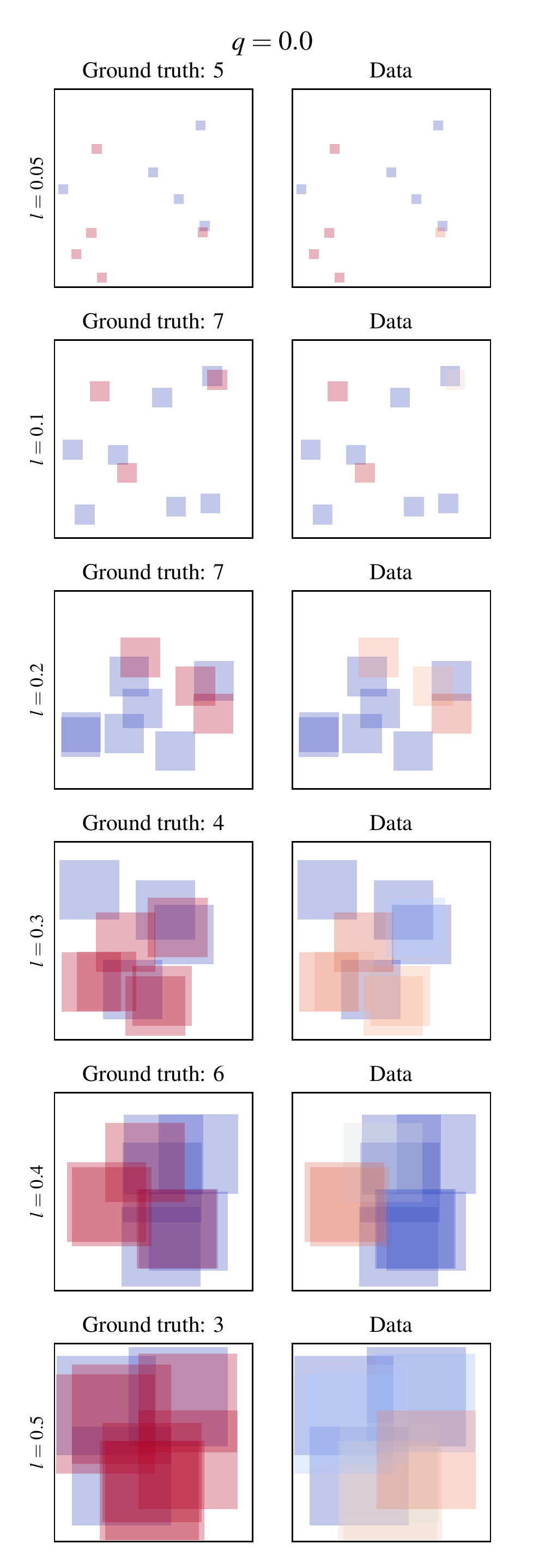}
        \hspace{1cm}
        \includegraphics[width=0.46\figurewidth, trim={0 4.5mm 0 4.5mm}, clip]{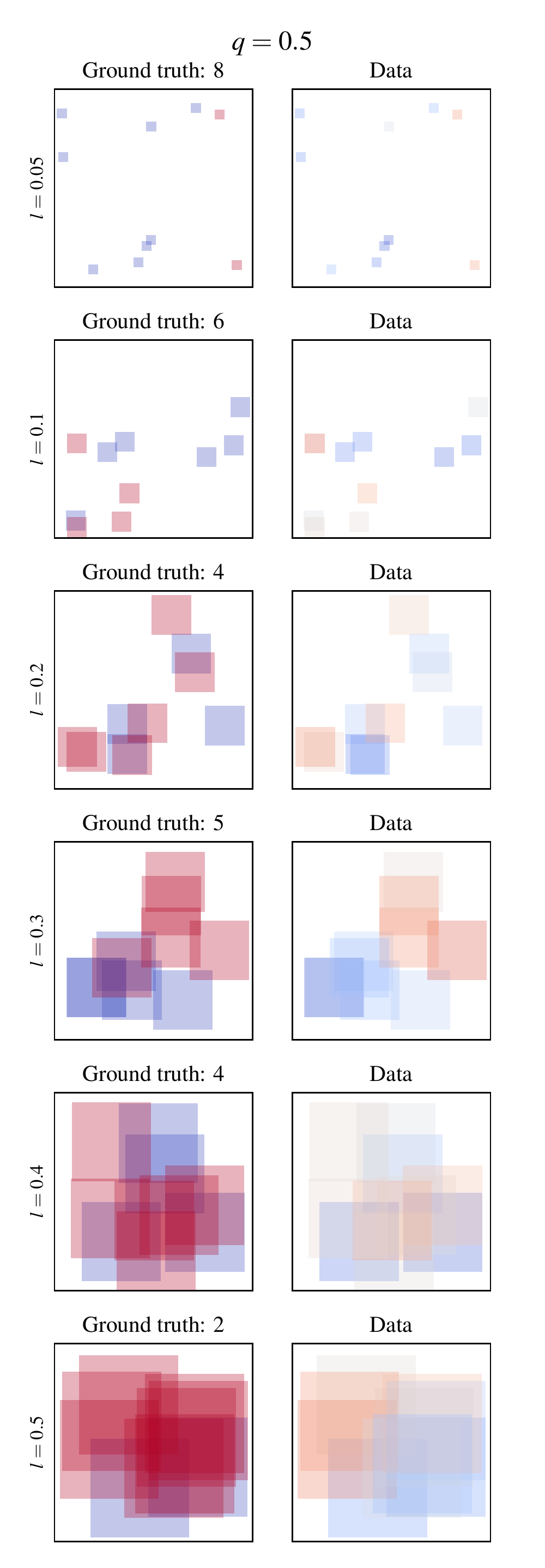}
    }
    \vspace{-3mm}
    \caption[Samples from the toy counting dataset]{
        Example toy dataset data for varying bounding box side lengths $l$ and noise $q$.
        The ground truth column shows bounding boxes of randomly placed true objects (blue) and of irrelevant objects (red).
        The data column visualises the samples that are actually used as input (dark blues represent weights close to 1, dark reds represent weights close to 0, lighter colours represent weights closer to 0.5).
        The weight of the $i$th bounding box $b_i$ is defined as
        $a_i = (1 - q) \text{ score} + qz$
        where the score is the maximum overlap of $b_i$ with any true bounding box or 0 if there are no true bounding boxes and $z$ is drawn from $U(0, 1)$.
        Note how this turns red bounding boxes that overlap a lot with a blue bounding box in the ground truth column into a blue bounding box in the data column, which simulates the duplicate proposal that we have to deal with.
        Best viewed in colour.
    }
    \label{count fig:toy}
\end{figure}

First, we design a simple toy task to evaluate counting ability.
This dataset is intended to only evaluate the performance of counting; thus, we skip any processing steps that are not directly related such as the processing of an input image.
Samples from this dataset are given in \autoref{count fig:toy}.

The classification task is to predict an integer count $\hat{c}$ of true objects, uniformly drawn from 0 to 10 inclusive, from a set of bounding boxes and the associated attention weights.
10 square bounding boxes with side length $l \in (0, 1]$ are placed in a square image with unit side length.
The x and y coordinates of their top left corners are uniformly drawn from $U(0, 1 - l)$ so that the boxes do not extend beyond the image border.
$l$ is used to control the overlapping of bounding boxes: a larger $l$ leads to the fixed number of objects to be more tightly packed, increasing the chance of overlaps.
$\hat{c}$ number of these boxes are randomly chosen to be true bounding boxes.
The \emph{score} of a bounding box is the maximum IoU overlap of it with any true bounding box.
Then, the attention weight is a linear interpolation between the score and a noise value drawn from $U(0, 1)$, with $q \in [0, 1]$ controlling this trade-off.
$q$ is the attention noise parameter: when $q$ is 0, there is no noise and when $q$ is 1, there is no signal.
Increasing $q$ also indirectly simulates imprecise placements of bounding boxes in real datasets.

We compare the counting module against a simple baseline that simply sums the attention weights and turns the sum into a feature vector with \autoref{count eq:output}.
Both models are followed by a linear projection to the classes 0 to 10 inclusive and a softmax activation.
They are trained with cross-entropy loss for 1000 iterations using Adam~\citep{Kingma2015} with a learning rate of 0.01 and a batch size of 1024.

\begin{figure}
    \centering
    \centerfloat{\includegraphics[width=\figurewidth]{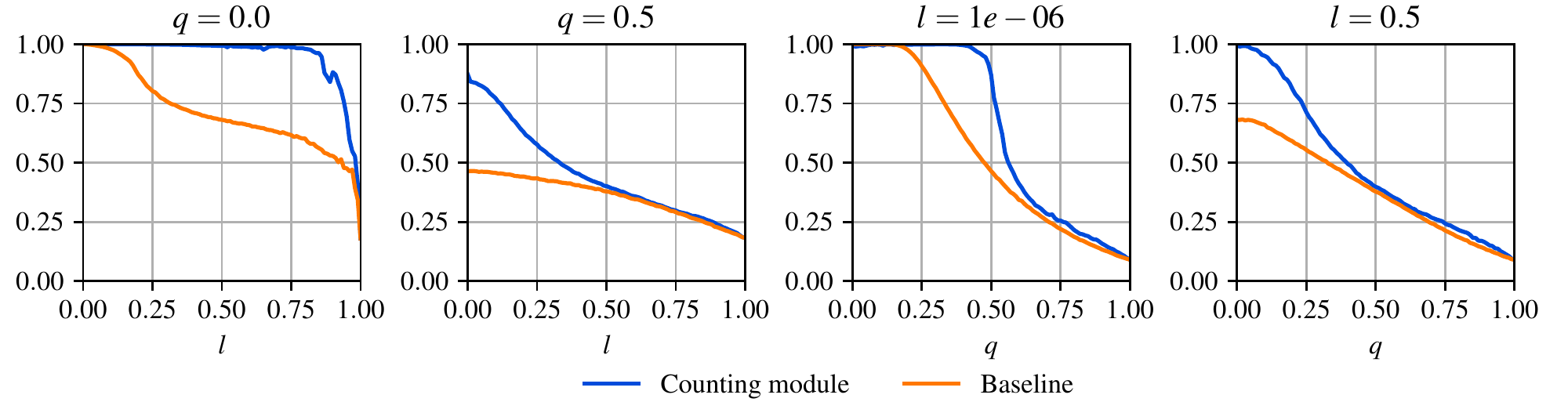}}
    \caption[Accuracies on toy dataset for varying dataset parameters]{Accuracies on the toy task as side length $l$ and noise $q$ are varied in 0.01 step sizes. Our counting module is in blue, the baseline is in orange.}
    \label{count fig:toy-acc}
\end{figure}

\subsubsection{Results}
The results of varying $l$ while keeping $q$ fixed at various values and vice versa are shown in \autoref{count fig:toy-acc}.
Regardless of $l$ and $q$, the counting module performs better than the baseline in most cases, often significantly so.
Particularly when the noise is low, the module can deal with high values for $l$ very successfully, showing that it accomplishes the goal of increased robustness to overlapping proposals.
The module also handles moderate noise levels decently as long as the overlaps are limited.
The performance when both $l$ and $q$ are high is closely matched by the baseline, likely due to the high difficulty of those parametrisations leaving little information to extract in the first place.

\begin{figure}
    \centering
    \centerfloat{
    \includegraphics[width=0.5\figurewidth]{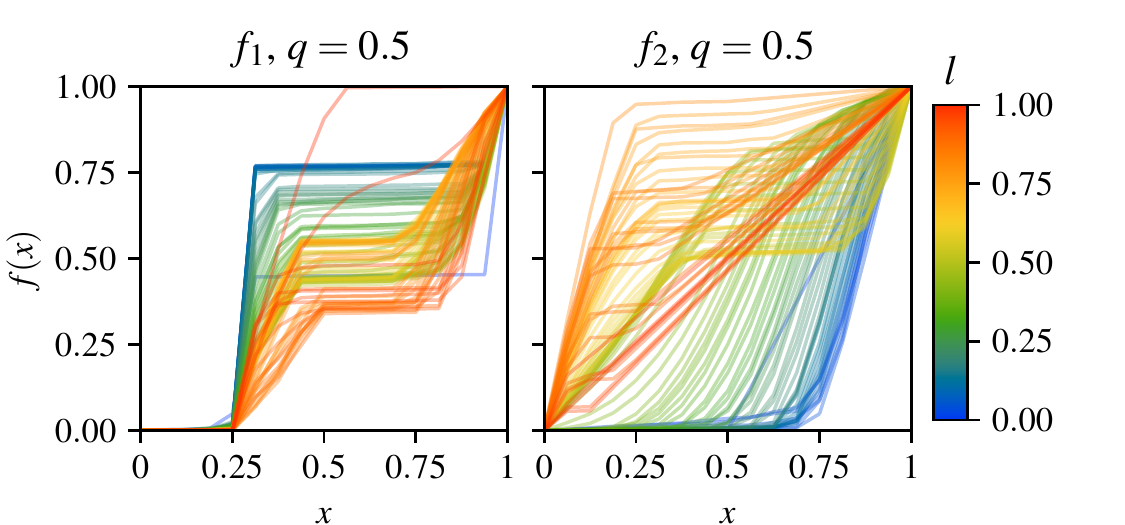}
    \includegraphics[width=0.5\figurewidth]{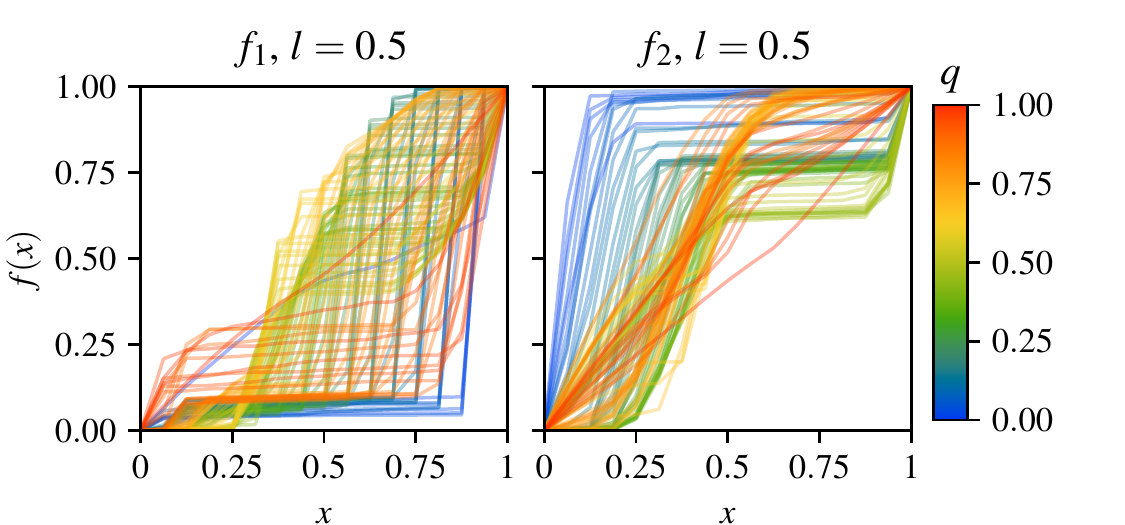}
    }
    \caption[Shapes of trained activation functions on toy dataset]{
        Shapes of trained activation functions $f_1$ (attention weights) and $f_2$ (bounding box distances) for varying bounding box side lengths (left) or the noise (right) in the dataset, varied in 0.01 step sizes.
        Best viewed in colour.
    }
    \label{count fig:trained-plins}
\end{figure}

\begin{figure}
    \centering
    \centerfloat{
    \includegraphics[width=\figurewidth, trim={1cm 0 0 0}, clip]{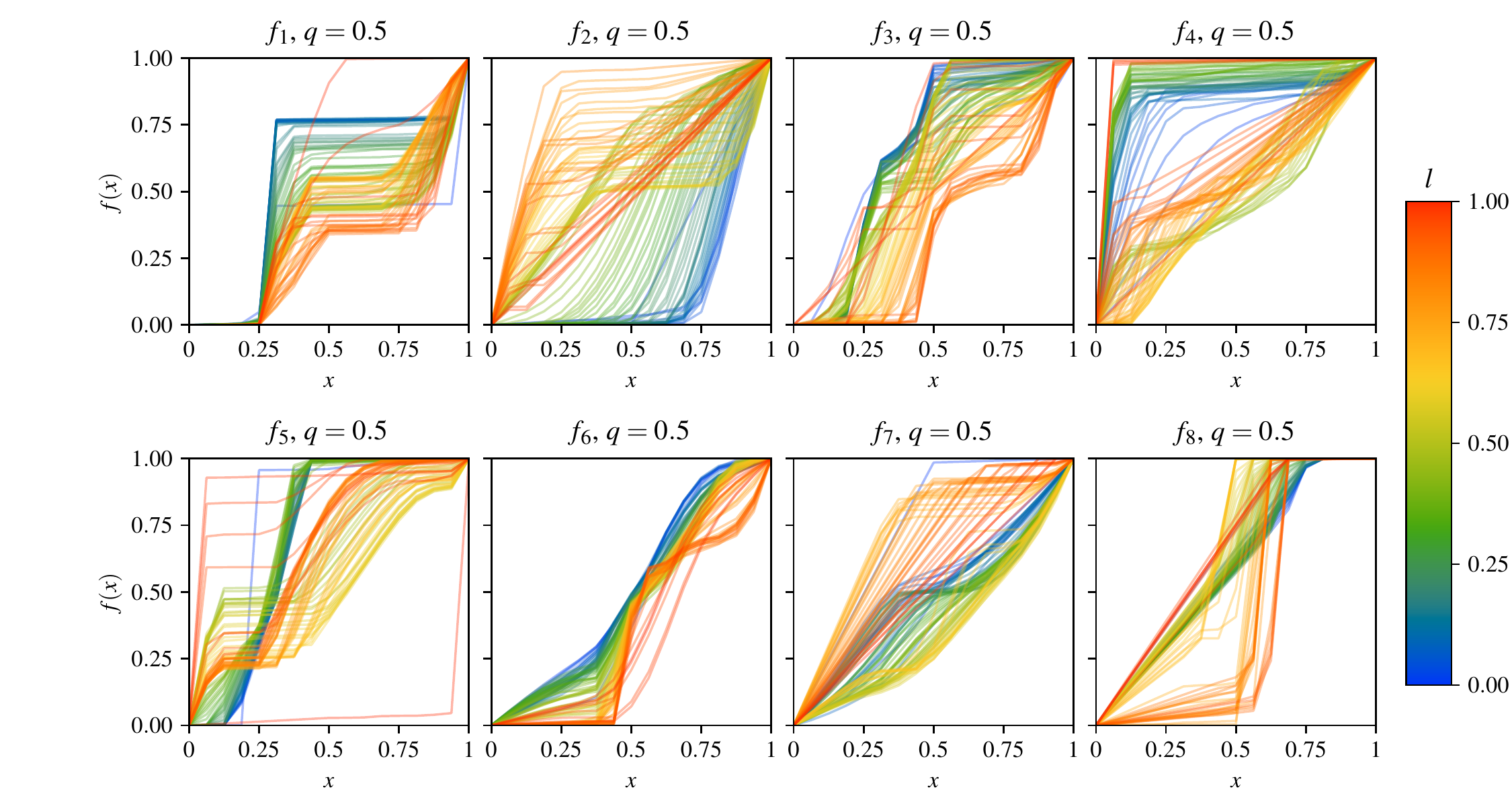}
    }
    \caption[Shapes of trained activation functions for varying noise on toy dataset]{Shape of activation functions as $l$ is varied for $q=0.5$ on the toy dataset.
    Each line shows the shape of the activation function when $l$ is set to the value associated to its colour.
    Best viewed in colour.}
    \label{count fig:rainbows1}
\end{figure}

\begin{figure}
    \centering
    \centerfloat{
    \includegraphics[width=\figurewidth, trim={1cm 0 0 0}, clip]{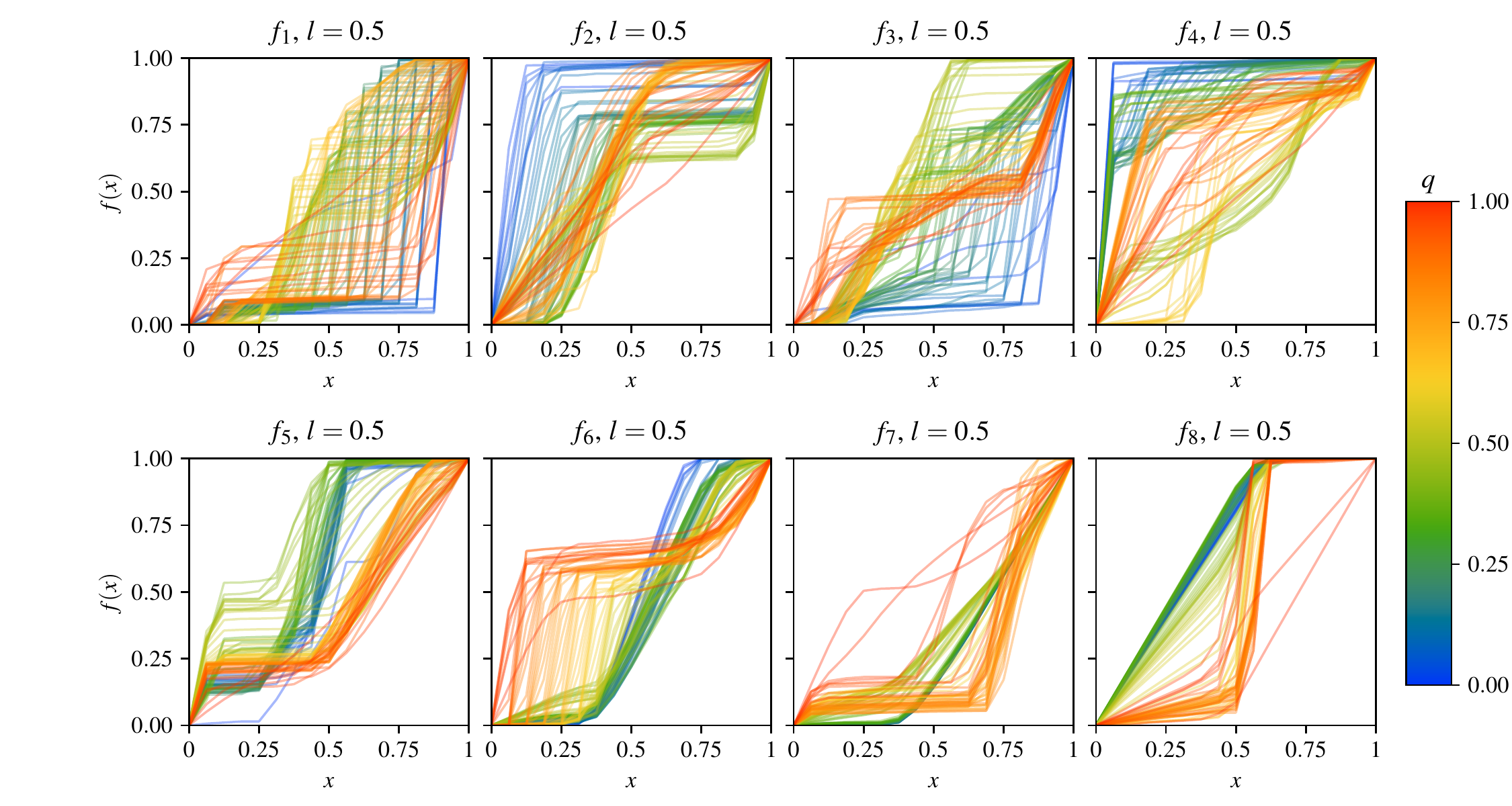}
    }
    \caption[Shapes of trained activation functions for varying box sizes on toy dataset]{Shape of activation functions as $q$ is varied for $l=0.5$ on the toy dataset.
    Each line shows the shape of the activation function when $q$ is set to the value associated to its colour.
    Best viewed in colour.}
    \label{count fig:rainbows2}
\end{figure}

We can also look at the shape of the activation functions themselves, shown in \autoref{count fig:trained-plins} (full version: \autoref{count fig:rainbows1} and \autoref{count fig:rainbows2}), to understand how the behaviour changes with varying dataset parameters.
For simplicity, we limit our description to the two easiest-to-interpret functions in \autoref{count fig:trained-plins}: $f_1$ for the attention weights and $f_2$ for the bounding box distances.

When increasing the side length, the height of the ``step'' in $f_1$ decreases to compensate for the generally greater degree of overlapping bounding boxes.
A similar effect is seen with $f_2$: it varies over requiring a high pairwise distance when $l$ is low -- when partial overlaps are most likely spurious -- and judging small distances enough for proposals to be considered different when $l$ is high.
At the highest values for $l$, there is little signal in the overlaps left since everything overlaps with everything, which explains why $f_2$ returns to its default linear initialisation for those parameters.

When varying the amount of noise, without noise $f_1$ resembles a step function where the step starts close to $x=1$ and takes a value of close to 1 after the step.
Since a true proposal will always have a weight of 1 when there is no noise, anything below this can be safely zeroed out.
With increasing noise, this step moves away from 1 for both $x$ and $f_1(x)$, capturing the uncertainty when a bounding box belongs to a true object.
With lower $q$, $f_2$ considers a pair of proposals to be distinct for lower distances, whereas with higher $q$, $f_2$ follows a more sigmoidal shape.
This can be explained by the model taking the increased uncertainty of the precise bounding box placements into account by requiring higher distances for proposals to be considered completely different.

\subsection{VQA}\label{count sec:vqa}
VQA v2~\citep{Goyal2017} is the updated version of VQA v1~\citep{Antol2015} where greater care has been taken to reduce dataset biases through balanced pairs:
for each question, a pair of images is identified where the answer to that question differs.
The standard accuracy metric on this dataset accounts for disagreements in human answers by averaging $\min(\frac{1}{3} \text{\emph{ agreeing}}, 1)$ over all 10-choose-9 subsets of human answers, where \emph{agreeing} is the number of human answers that agree with the given answer.
This can be shown to be equal to $\min(0.3 \text{\emph{ agreeing}}, 1)$ without averaging.

\newcommand{\agree}{\#}
\begin{sidenote}{Aside: Simplifying the evaluation metric}
    We managed to simplify the official metric of averaging over the 10-choose-9 subsets as follows.

    There are two cases for the one answer to be discarded:

    \begin{enumerate}
        \item the discarded answer \emph{is not} the predicted answer \\$\implies$ accuracy stays the same
        \item the discarded answer \emph{is} the predicted answer \\$\implies$ we have to subtract 1 from the number of agreeing answers.
    \end{enumerate}

    We use $\agree$ to denote the number of human answers agreeing with the prediction.
    There are $10 - \agree$ of case 1 and $\agree$ of case 2, therefore the accuracy is:

    \begin{equation}
        0.1 \cdot \left((10 - \agree) \min\left(\frac{\agree}{3}, 1\right) + 
                         \agree \min\left(\frac{\agree - 1}{3}, 1\right)\right)
    \end{equation}

    We know that when $\agree = 0$, the accuracy is 0, and when $\agree \geq 4$, the accuracy is 1.
    In the remaining cases $1 \leq \agree \leq 3$, we know that $\frac{\agree - 1}{3} < \frac{\agree}{3} \leq 1$, so all the $\min$s can be removed.
    This allows us to move the $\agree$ outside:

    \begin{equation}
        0.1 \cdot \frac{1}{3} \cdot \agree \left((10 - \agree) + (\agree - 1) \right)
    \end{equation}

    which simplifies to $0.3 \agree$.
    Lastly, we can combine all the cases together to get $\min(0.3 \agree, 1)$ as accuracy metric.
\end{sidenote}

\begin{figure}
    \centering
    \fontsize{10}{10}\selectfont
    
    \centerfloat{\resizebox{\figurewidth}{!}{%
        \begin{tikzpicture}[x=1.5cm, y=1.5cm]
            \pic{count model};
        \end{tikzpicture}
    }}
    \caption[Network architecture for VQA]{
        Schematic view of a model using our counting module.
        The modifications made to the baseline model when including the counting module are marked in red.
        Blue blocks mark modules with trainable parameters, grey blocks mark modules without trainable parameters.
        White {\fontfamily{cmr}\textcircled{w}} mark linear layers, either linear projections or convolutions with a spatial size of 1 depending on the context.
        Dropout with drop probability 0.5 is applied before the GRU and every {\fontfamily{cmr}\textcircled{w}}, except before the {\fontfamily{cmr}\textcircled{w}} after the counting module.
        \Large $\diamond$ \normalsize stands for the fusion function we define in \autoref{app sec:vqa}, BN stands for batch normalisation, $\sigma$ stands for a logistic, and Embedding is a word embedding that has been fed through a $\tanh$ function.
        The two glimpses of the attention mechanism are represented with the two lines exiting the {\fontfamily{cmr}\textcircled{w}}.
        Note that one of the two glimpses is shared with the counting module.
    }
    \label{count fig:model}
\end{figure}
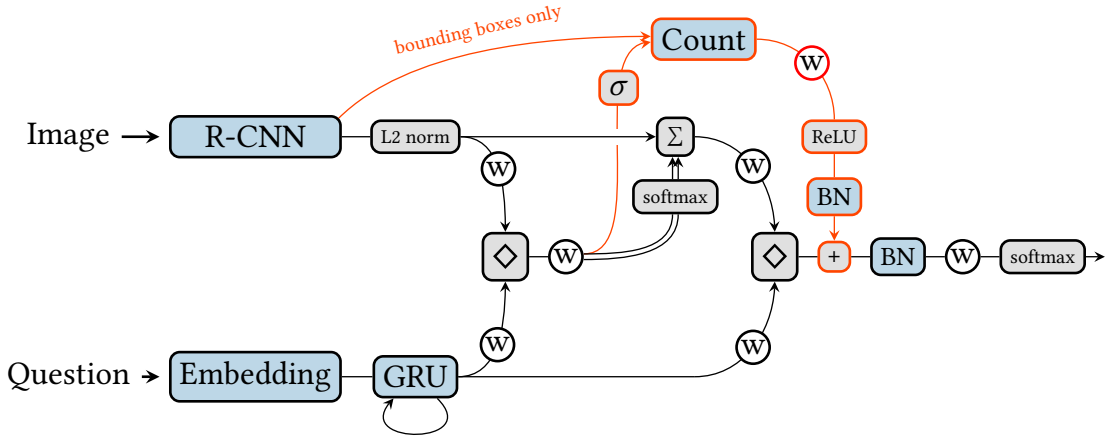

\paragraph{Model}\label{count sec:arch}
Our baseline model is based on the work of \citet{Kazemi2017}, which outperformed most previous VQA models on the VQA v1 dataset with a simple baseline architecture.
We adapt the model to the VQA v2 dataset and make various tweaks that improve validation accuracy slightly, which we describe in full in \autoref{app sec:vqa}.
The architecture is illustrated in \autoref{count fig:model}.

We have not performed any tuning of this baseline to maximise the performance difference between it and the baseline with counting module.
To augment this model with the counting module, we extract the attention weights of the first attention glimpse (there are two in the baseline) before softmax normalisation, and feed them into the counting module after applying a logistic function.
Since object proposal features from \citet{Anderson2018} vary from 10 to 100 per image, a natural choice for the number of top-$n$ proposals to use is 10.
The output of the module is linearly projected into the same space as the hidden layer of the classifier, followed by ReLU activation, batch normalisation, and addition with the features in the hidden layer.

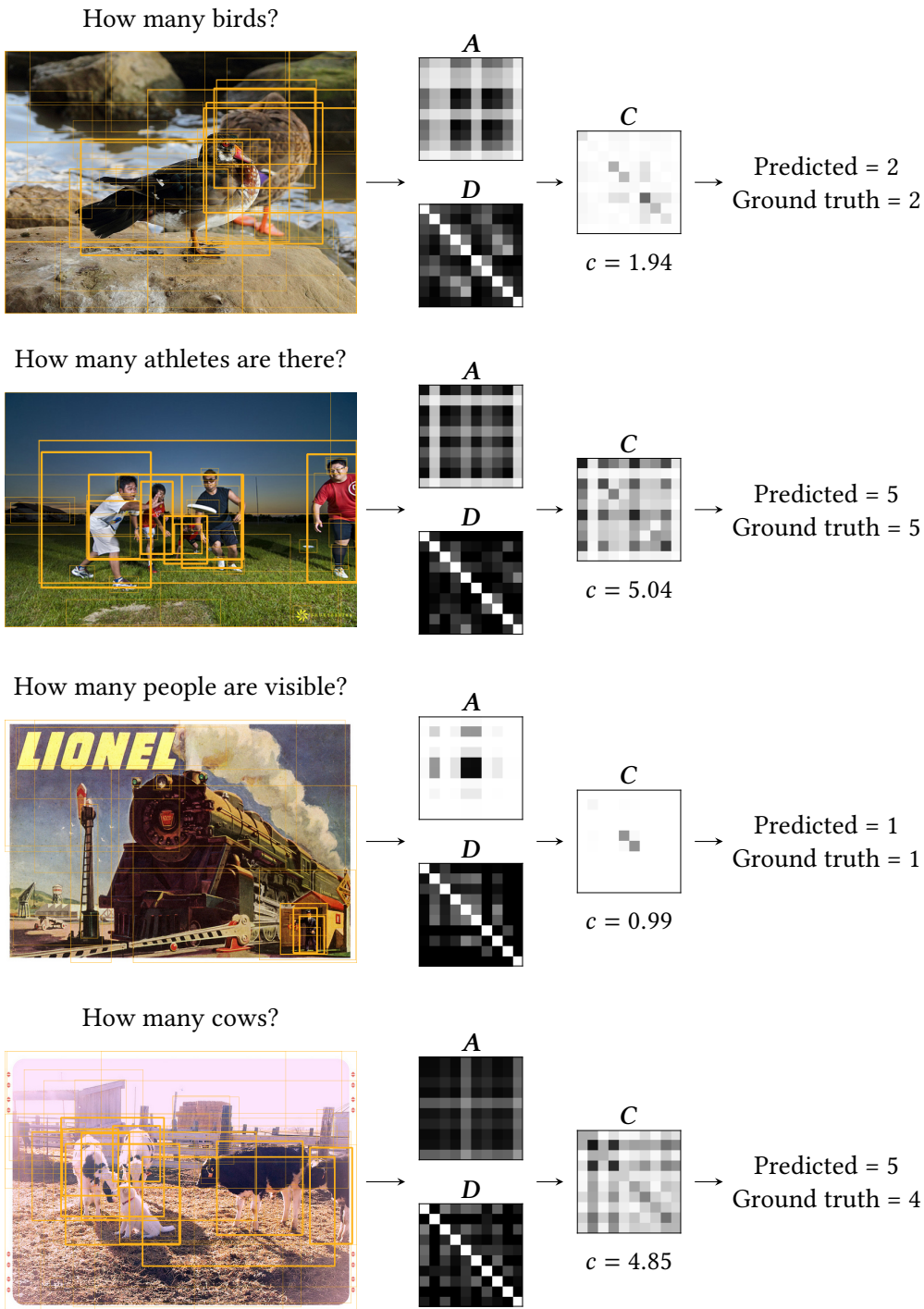
\begin{figure}
    \centering
    
    \centerfloat{{
        \begin{tikzpicture}[x=1.5cm, y=1.5cm]
        \node (img) at (0, 0) {\includegraphics[width=5cm]{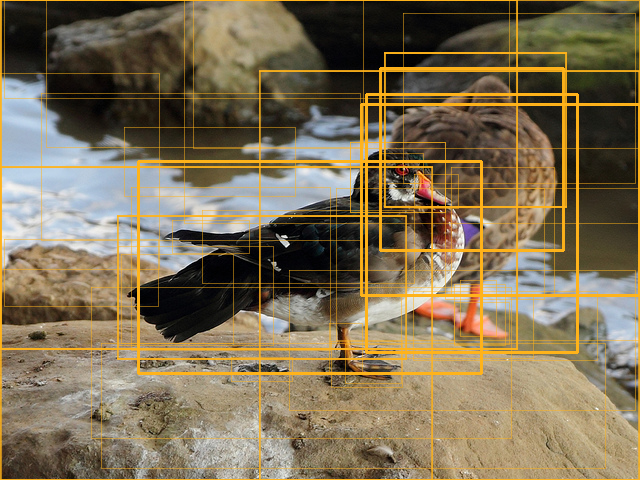}};
        \node (q) [above = 0.0 cm of img] {How many birds?};
        \node (a) at (2.75, 0.7) {\includegraphics[width=2cm]{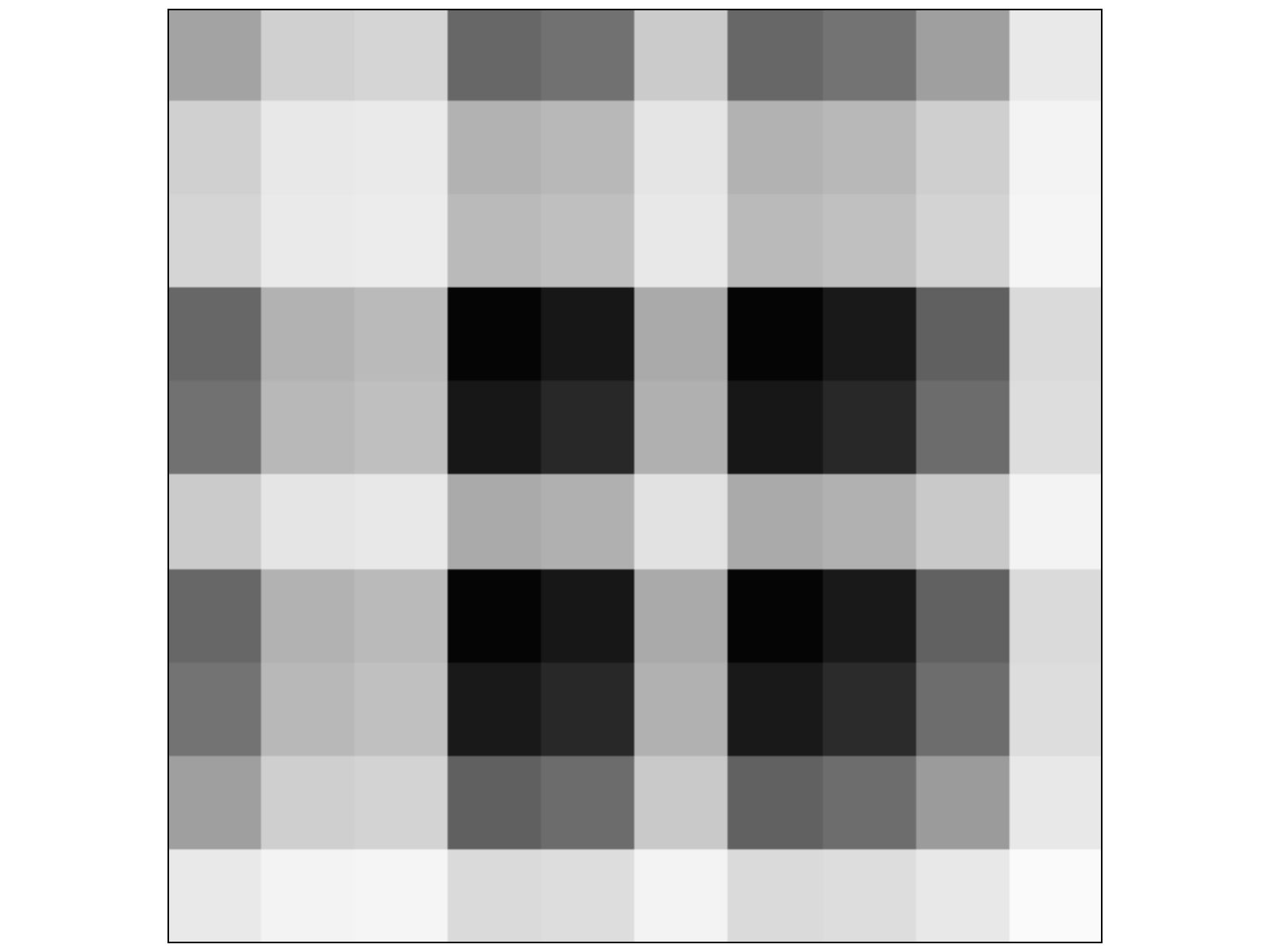}};
        \node (d) at (2.75, -0.7) {\includegraphics[width=2cm]{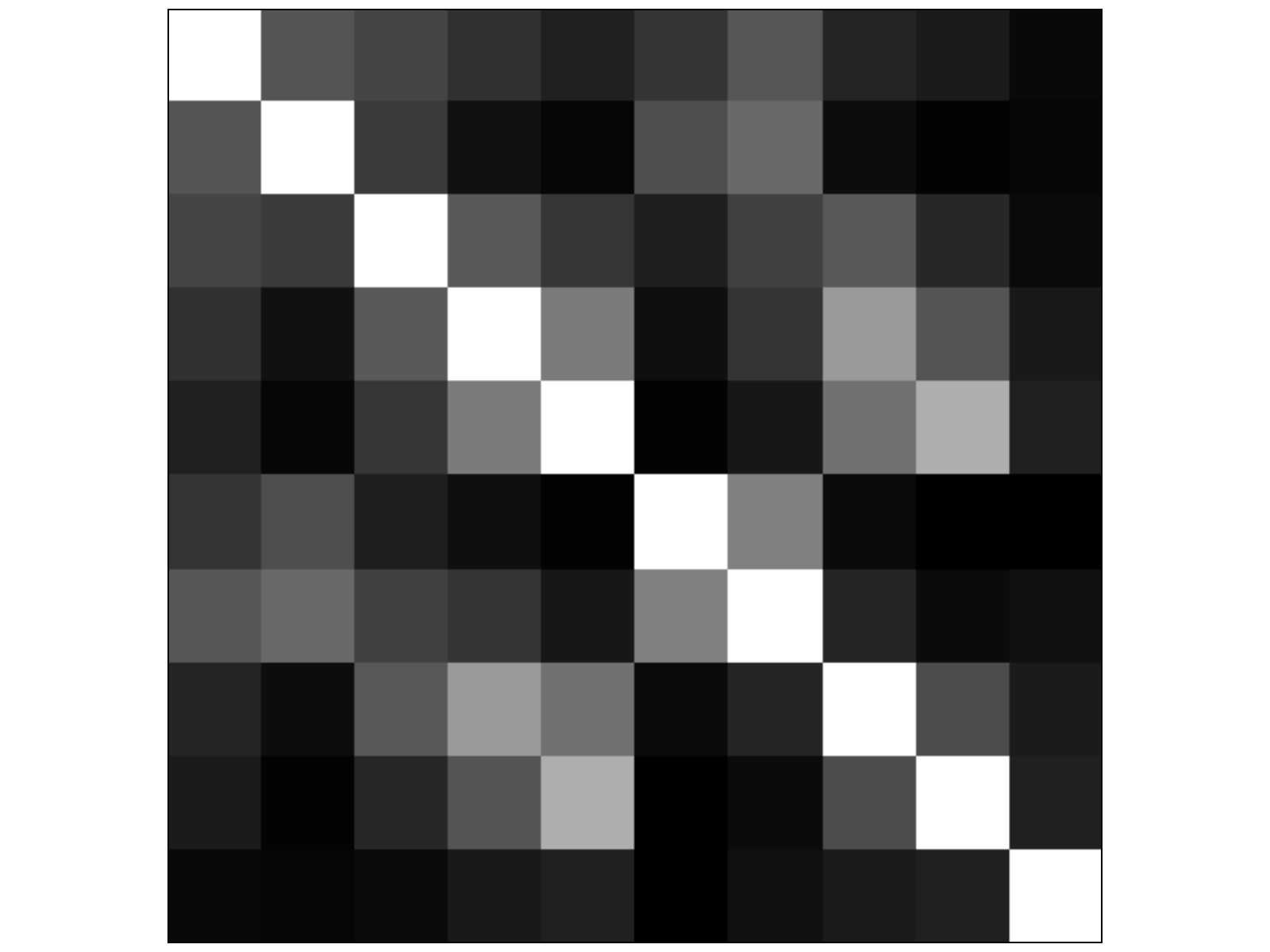}};
        \node (c) at (4.25, 0) {\includegraphics[width=2cm]{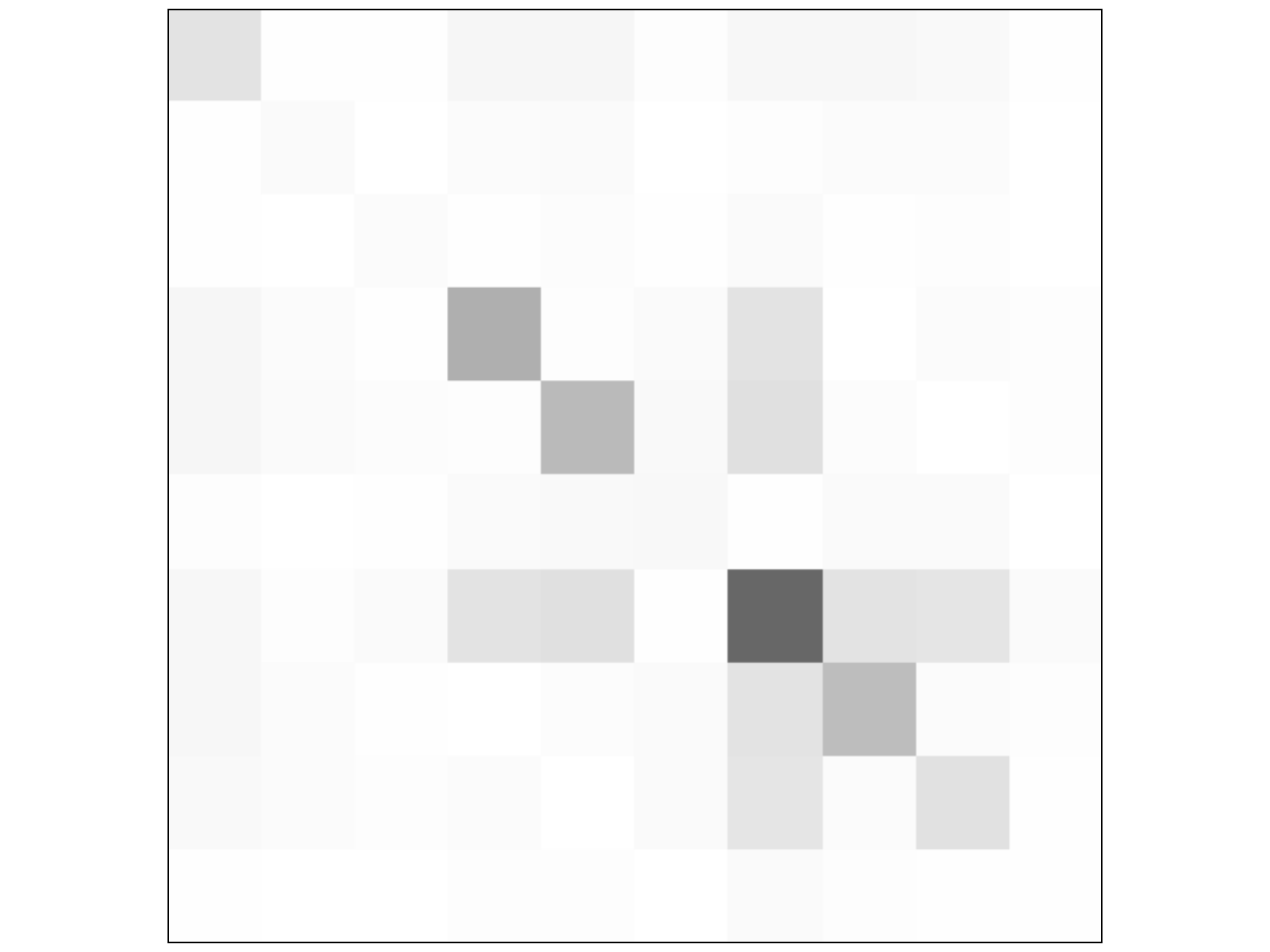}};
        \node [below = 0.0 cm of c] {$c = 1.94$};
        \node (r) [right = 0.2 cm of c, align=center] {Predicted = 2 \\ Ground truth = 2};
        \node [above = -0.2 cm of a] {$\mA$};
        \node [above = -0.2 cm of d] {$\mD$};
        \node [above = -0.2 cm of c] {$\mC$};
        \draw [sedge] (img) -- ([xshift=0.2cm]a.west |- img);
        \draw [sedge] ([xshift=-0.2cm]a.east |- img) -- ([xshift=0.2cm]c.west);
        \draw [sedge] ([xshift=-0.2cm]c.east) -- (r);
        \node at (0, -1.4) {};
    \end{tikzpicture}}}
    
    \centerfloat{{
        \begin{tikzpicture}[x=1.5cm, y=1.5cm]
        \node (img) at (0, 0) {\includegraphics[width=5cm]{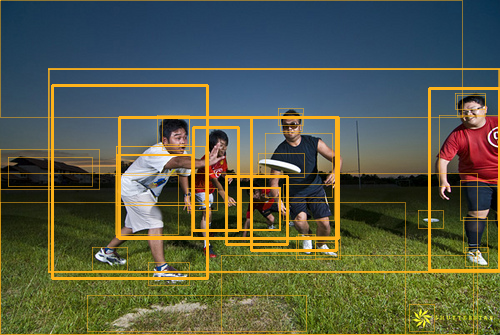}};
        \node (q) [above = 0.0 cm of img] {How many athletes are there?};
        \node (a) at (2.75, 0.7) {\includegraphics[width=2cm]{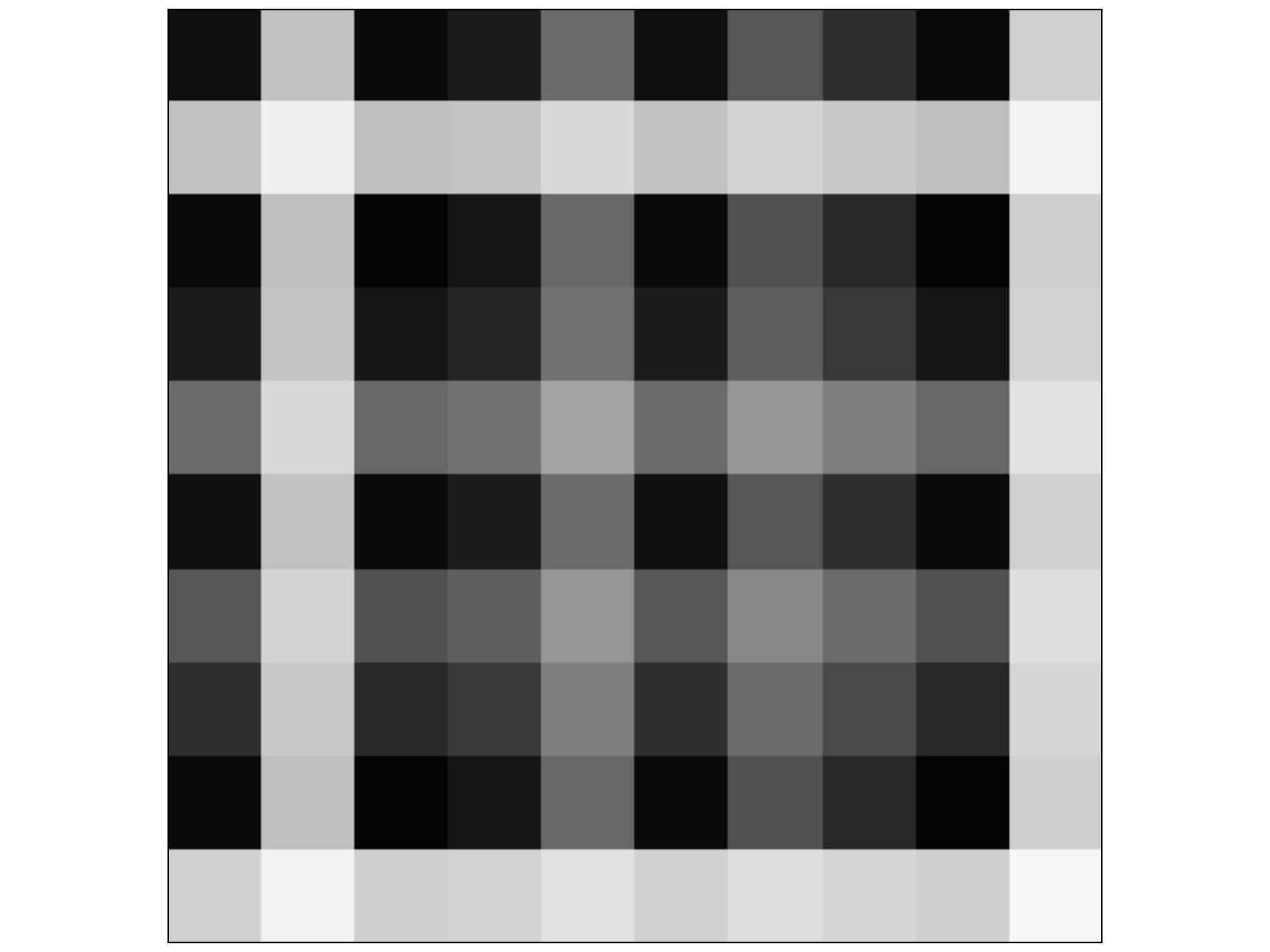}};
        \node (d) at (2.75, -0.7) {\includegraphics[width=2cm]{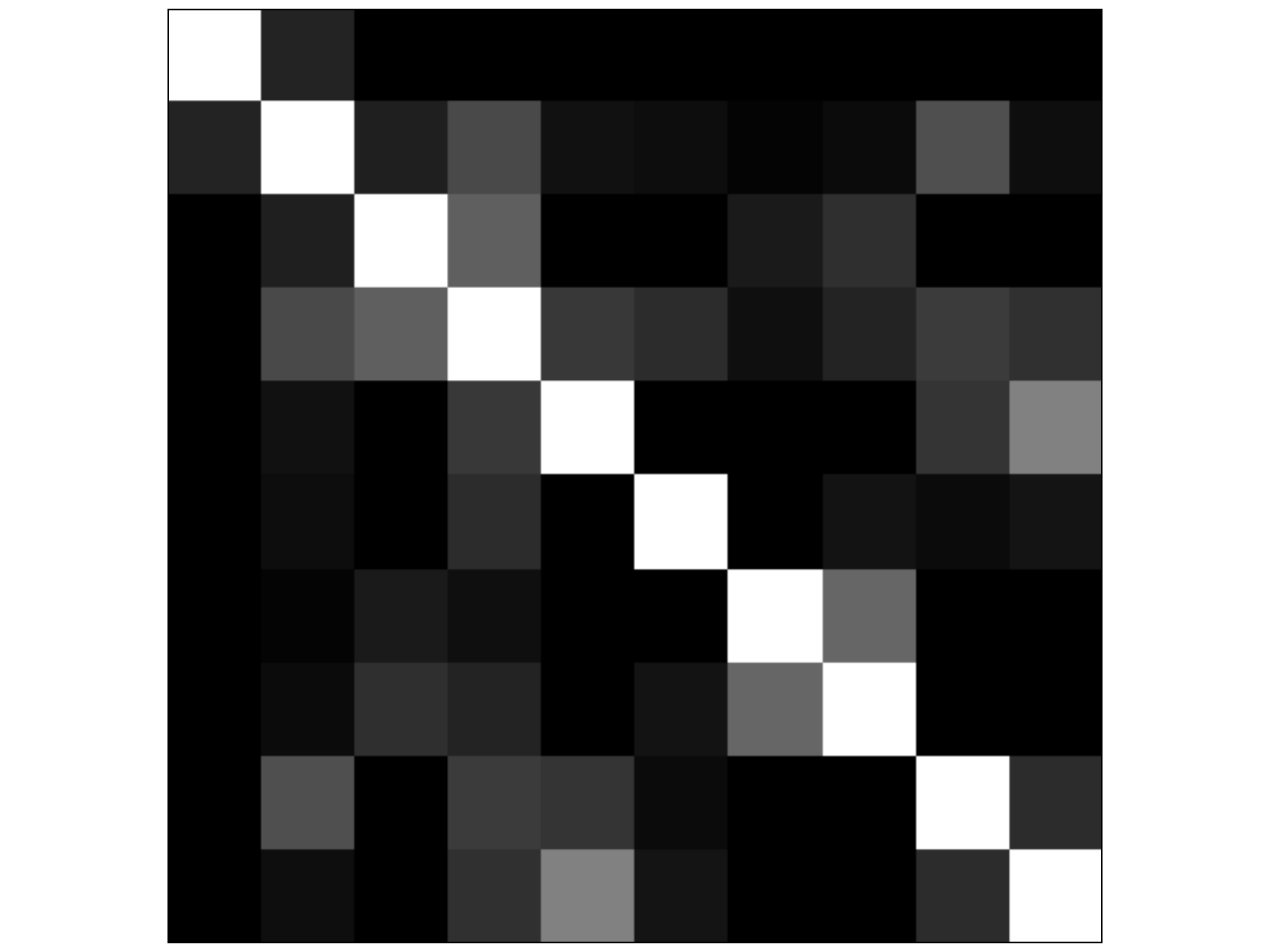}};
        \node (c) at (4.25, 0) {\includegraphics[width=2cm]{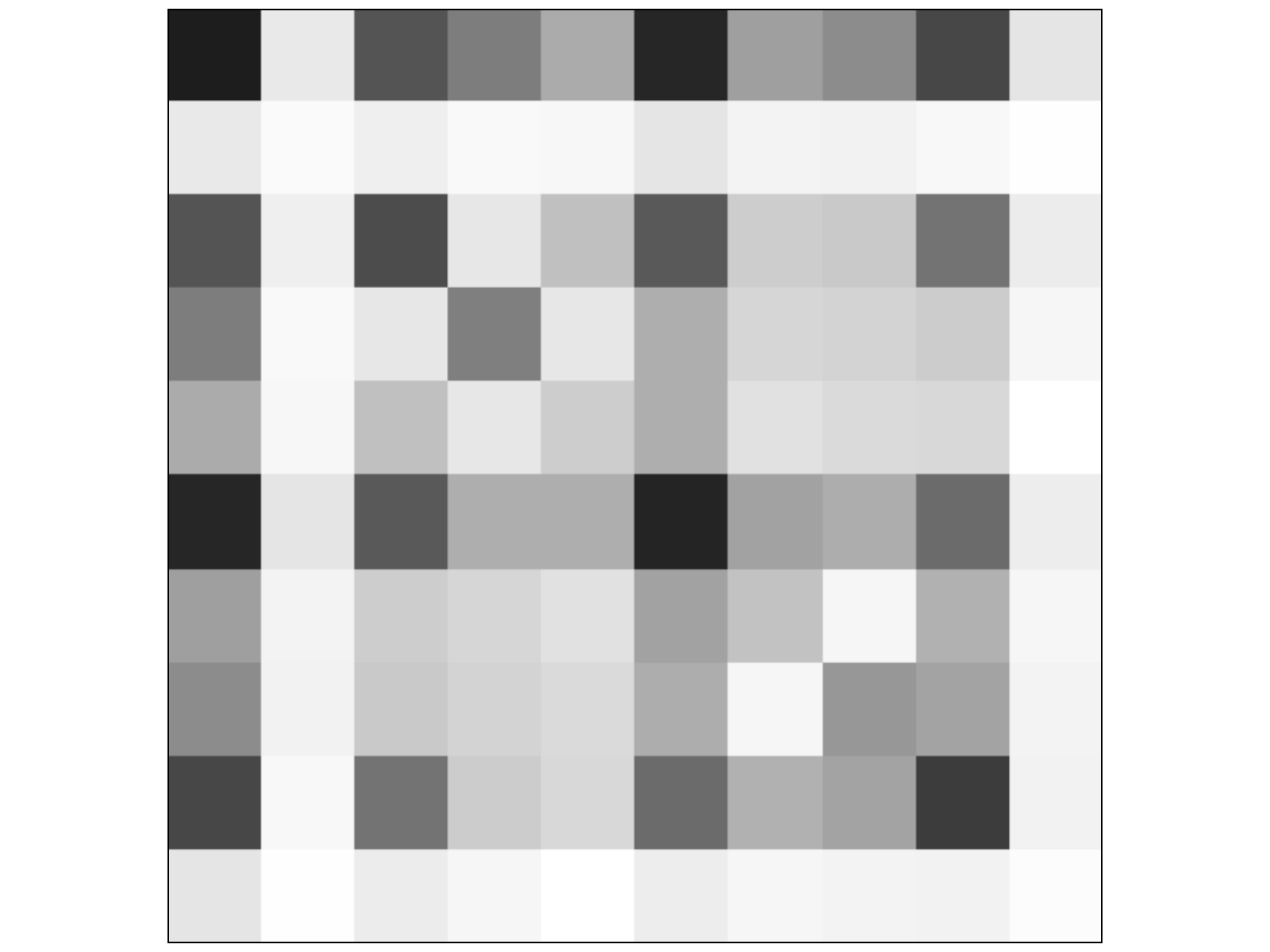}};
        \node [below = 0.0 cm of c] {$c = 5.04$};
        \node (r) [right = 0.2 cm of c, align=center] {Predicted = 5 \\ Ground truth = 5};
        \node [above = -0.2 cm of a] {$\mA$};
        \node [above = -0.2 cm of d] {$\mD$};
        \node [above = -0.2 cm of c] {$\mC$};
        \draw [sedge] (img) -- ([xshift=0.2cm]a.west |- img);
        \draw [sedge] ([xshift=-0.2cm]a.east |- img) -- ([xshift=0.2cm]c.west);
        \draw [sedge] ([xshift=-0.2cm]c.east) -- (r);
        \node at (0, -1.4) {};
    \end{tikzpicture}}}
    
    \centerfloat{{
        \begin{tikzpicture}[x=1.5cm, y=1.5cm]
        \node (img) at (0, 0) {\includegraphics[width=5cm]{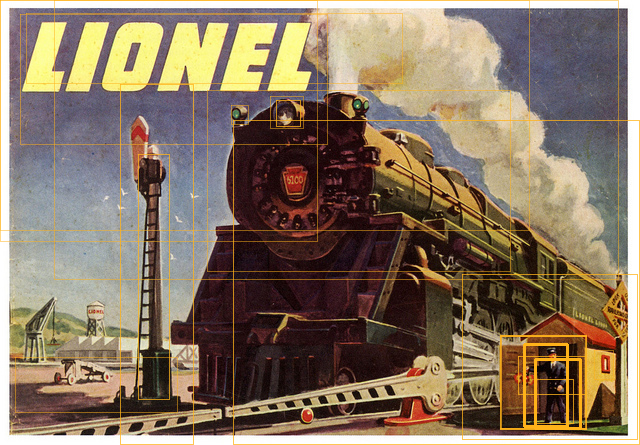}};
        \node (q) [above = 0.0 cm of img] {How many people are visible?};
        \node (a) at (2.75, 0.7) {\includegraphics[width=2cm]{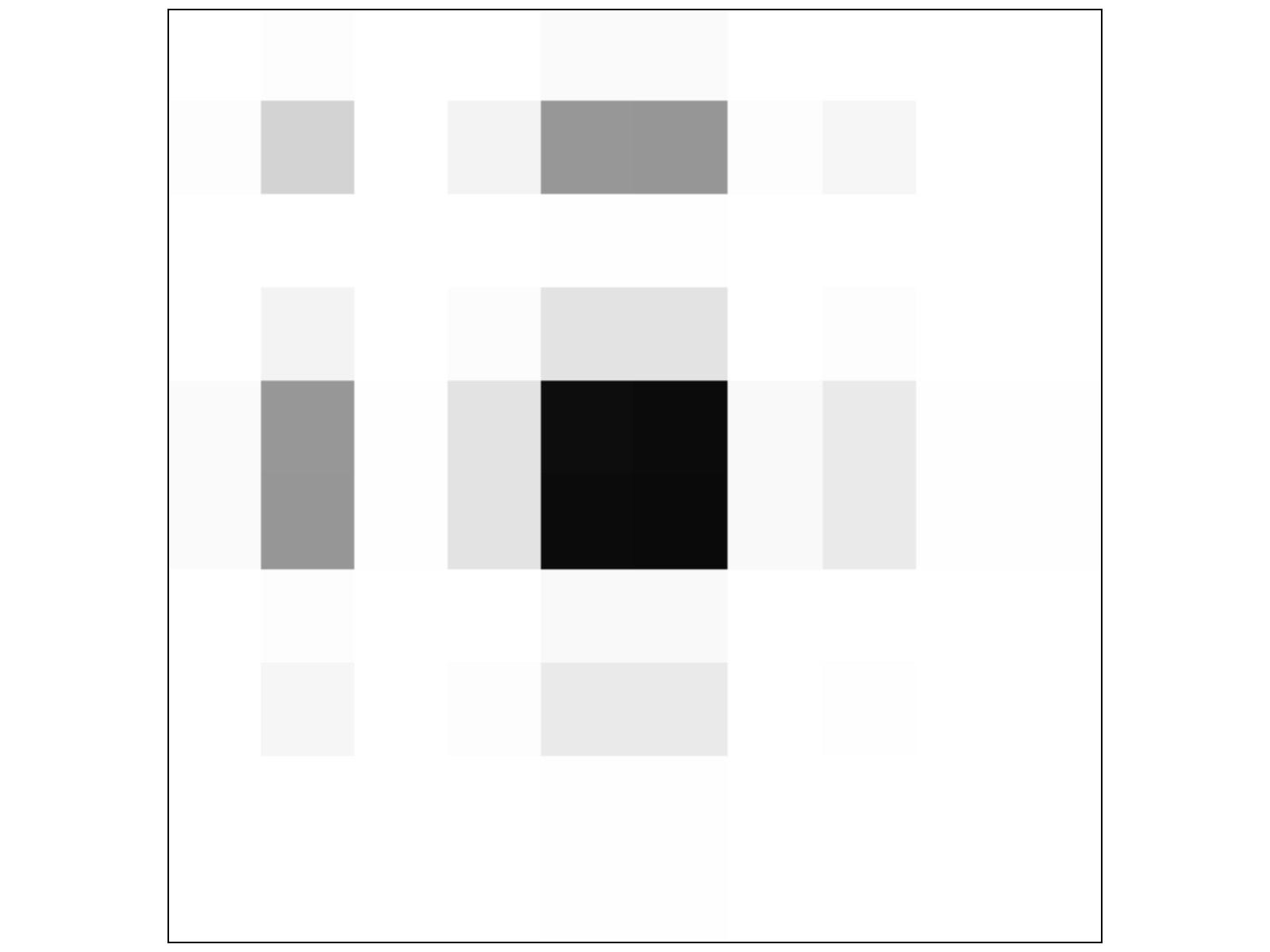}};
        \node (d) at (2.75, -0.7) {\includegraphics[width=2cm]{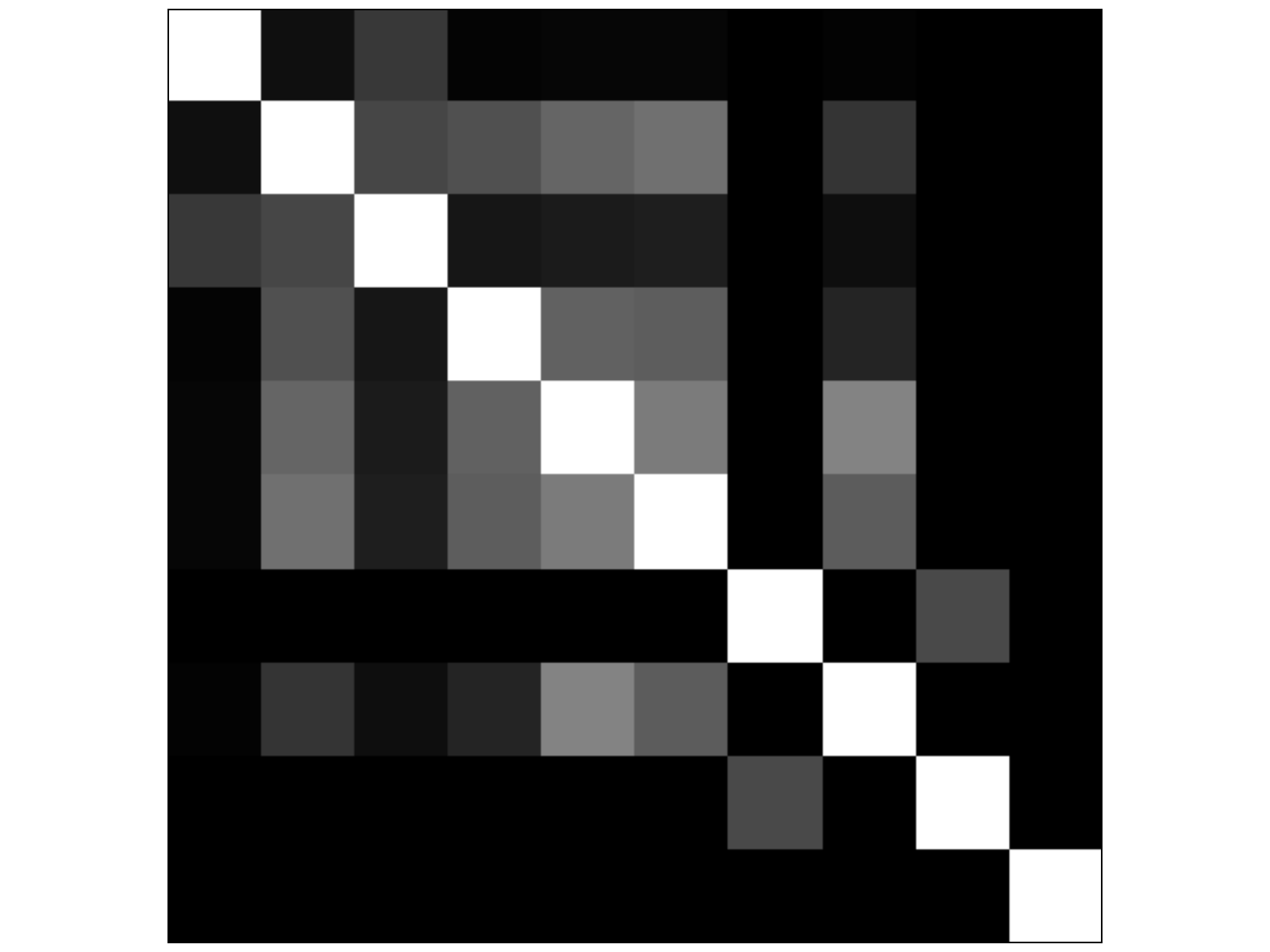}};
        \node (c) at (4.25, 0) {\includegraphics[width=2cm]{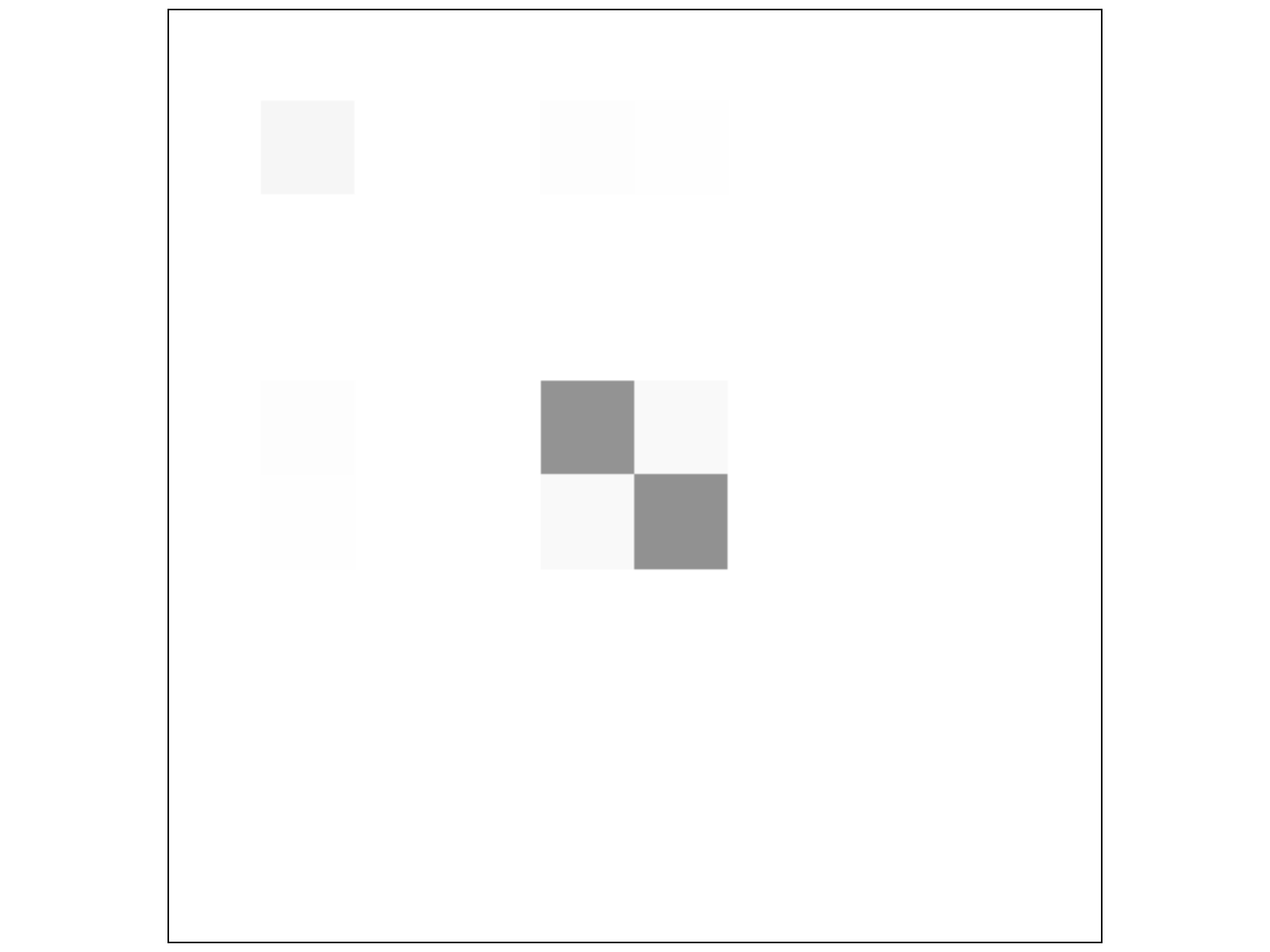}};
        \node [below = 0.0 cm of c] {$c = 0.99$};
        \node (r) [right = 0.2 cm of c, align=center] {Predicted = 1 \\ Ground truth = 1};
        \node [above = -0.2 cm of a] {$\mA$};
        \node [above = -0.2 cm of d] {$\mD$};
        \node [above = -0.2 cm of c] {$\mC$};
        \draw [sedge] (img) -- ([xshift=0.2cm]a.west |- img);
        \draw [sedge] ([xshift=-0.2cm]a.east |- img) -- ([xshift=0.2cm]c.west);
        \draw [sedge] ([xshift=-0.2cm]c.east) -- (r);
        \node at (0, -1.4) {};
    \end{tikzpicture}}}
    
    \centerfloat{{
        \begin{tikzpicture}[x=1.5cm, y=1.5cm]
        \node (img) at (0, 0) {\includegraphics[width=5cm]{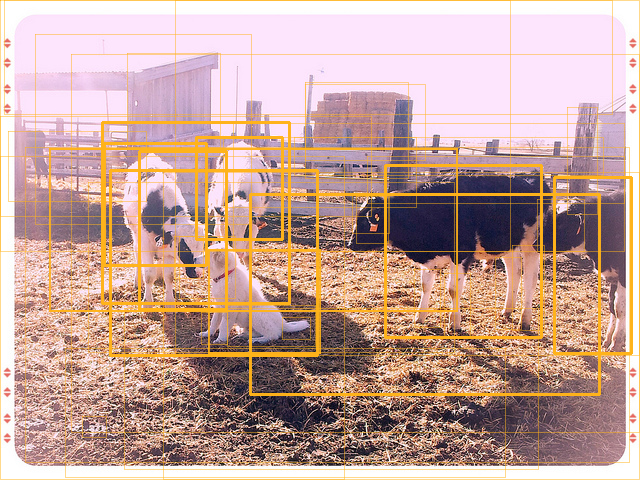}};
        \node (q) [above = 0.0 cm of img] {How many cows?};
        \node (a) at (2.75, 0.7) {\includegraphics[width=2cm]{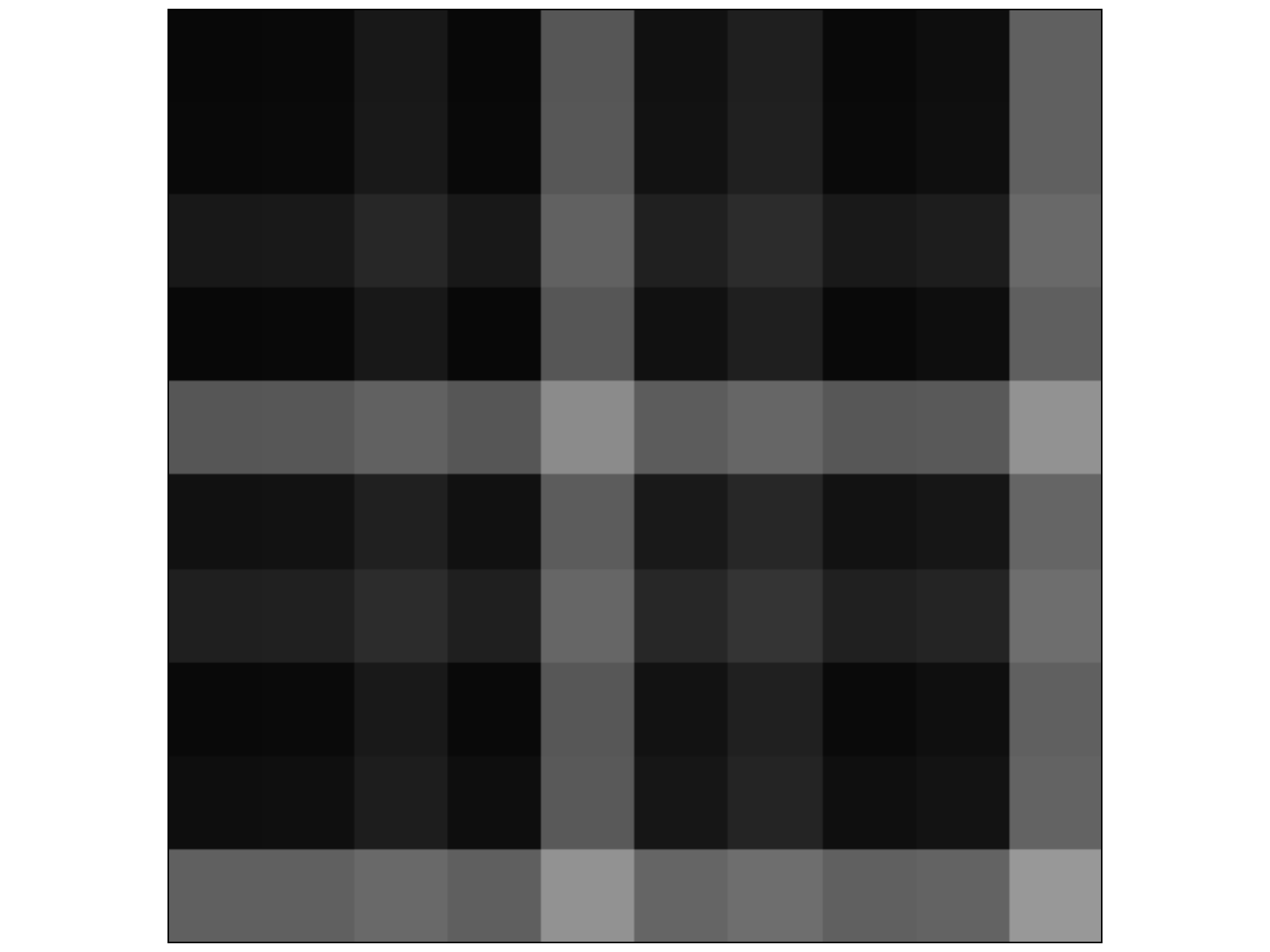}};
        \node (d) at (2.75, -0.7) {\includegraphics[width=2cm]{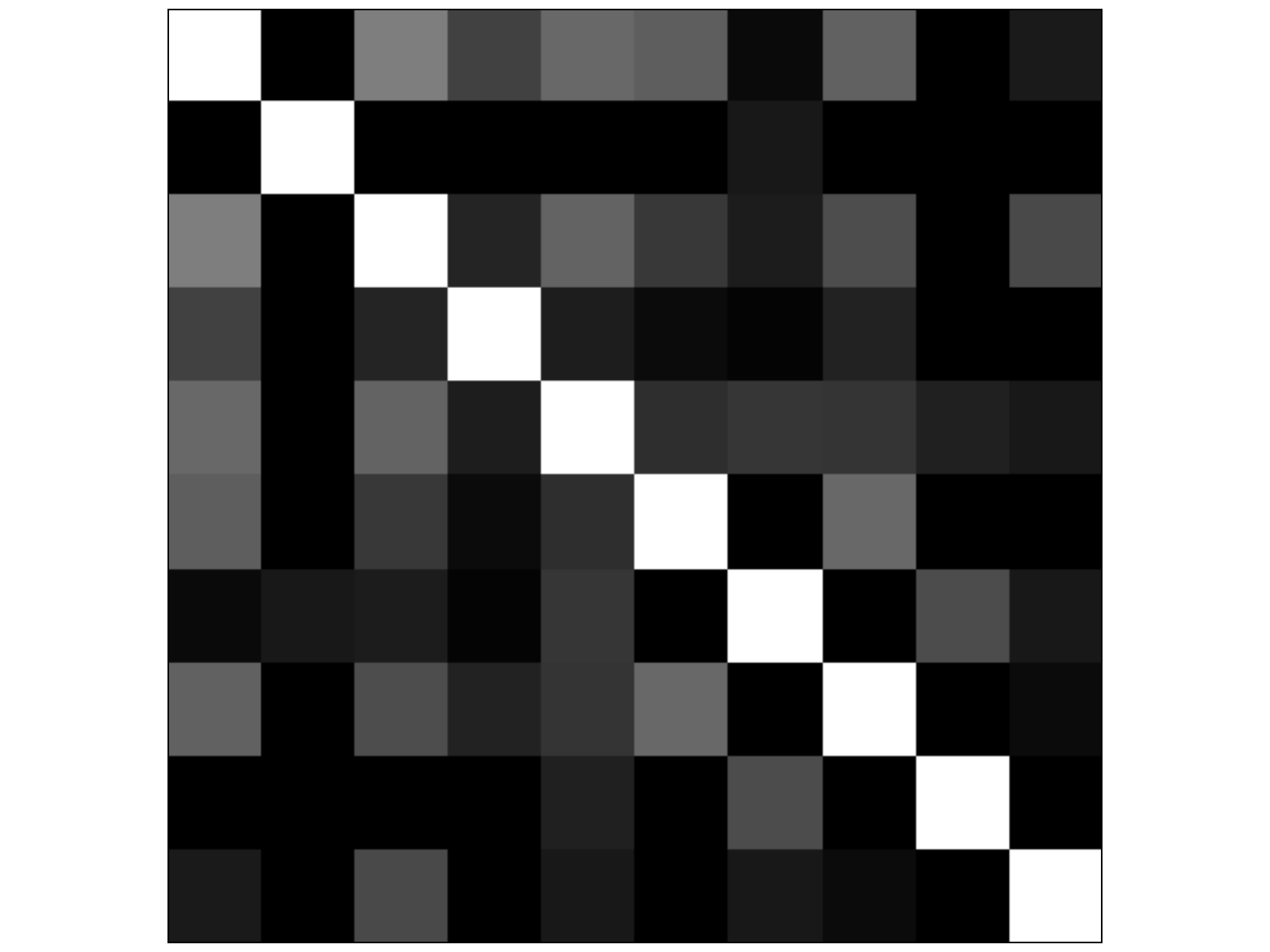}};
        \node (c) at (4.25, 0) {\includegraphics[width=2cm]{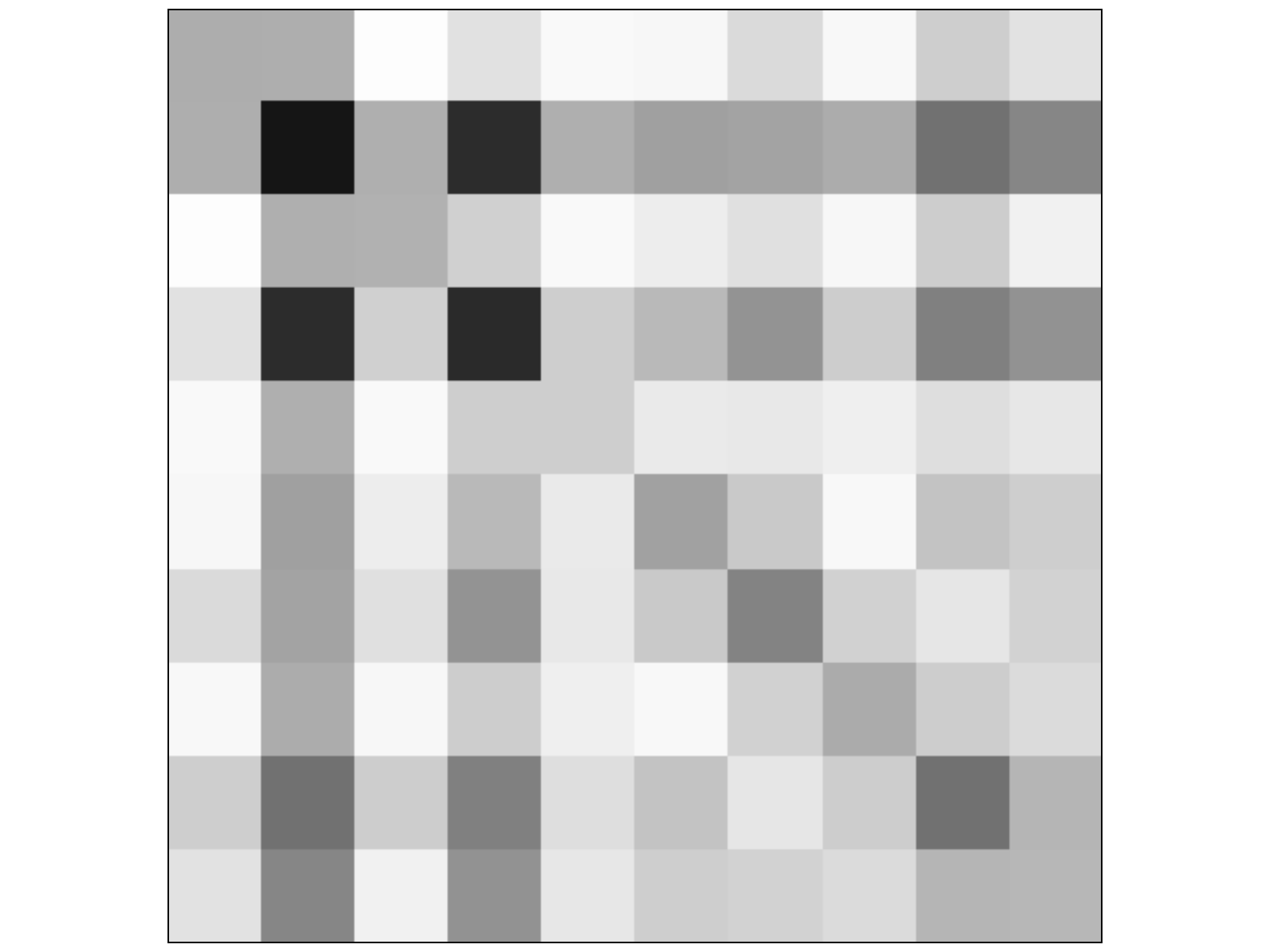}};
        \node [below = 0.0 cm of c] {$c = 4.85$};
        \node (r) [right = 0.2 cm of c, align=center] {Predicted = 5 \\ Ground truth = 4};
        \node [above = -0.2 cm of a] {$\mA$};
        \node [above = -0.2 cm of d] {$\mD$};
        \node [above = -0.2 cm of c] {$\mC$};
        \draw [sedge] (img) -- ([xshift=0.2cm]a.west |- img);
        \draw [sedge] ([xshift=-0.2cm]a.east |- img) -- ([xshift=0.2cm]c.west);
        \draw [sedge] ([xshift=-0.2cm]c.east) -- (r);
    \end{tikzpicture}}}
    \caption[Example inputs and model activations]{
        Selection of validation images with overlaid bounding boxes, values of the attention matrix $\mA$, distance matrix $\mD$, and the resulting count matrix $\mC$.
        White entries represent values close to 1, black entries represent values close to 0.
        The count $c$ is the usual square root of the sum over the elements of $\mC$.
        Notice how particularly in the third example, $\mA$ clearly contains more rows/columns with high activations than there are actual objects (a sign of overlapping bounding boxes) and the counting module successfully removes intra- and inter-object edges to arrive at the correct prediction regardless.
        The prediction is not necessarily -- though often is -- the rounded value of $c$.
    }
    \label{count fig:examples}
\end{figure}

\paragraph{Results}
\autoref{count tab:vqa-test} shows the results on the official VQA v2 leader board.
The baseline with our module has a significantly higher accuracy on number questions without compromising accuracy on other categories compared to the baseline result.
Despite our single-model baseline being substantially worse than the state-of-the-art, by simply adding the counting module we outperform even the 8-model ensemble in \citet{Zhou2017} on the number category.
We expect further improvements in number accuracy when incorporating their techniques to improve the quality of attention weights, especially since the current state-of-the-art models suffer from the problems with counting that we mention in \autoref{count sec:problems}.
Some qualitative examples of inputs and activations within the counting module are shown in \autoref{count fig:examples}.

Since the writing of this chapter, further VQA challenges have been run.
The winners of the 2018 challenge were \citet{Kim2018}, who used our counting model to improve the counting results in their model.
They achieved a 54.04\% accuracy on the number category of the VQA v2 test set, which is a 2.65\% improvement over our results.
Without our counting module, they obtained 50.66\% accuracy in the number category, so a significant part of their improved number results are due to our model.
In other words, they independently confirmed that a better attention model (which they developed) leads to even better counting results.

\begin{table}
    \setlength\tabcolsep{4.2pt}
    \caption[VQA v2 test results]{
        Results on VQA v2 of the top models along with our results.
        Entries marked with (Ens.) are ensembles of models.
        At the time of writing, our model with the counting module places third among all entries.
        All models listed here use object proposal features and are trained on the training and validation sets.
        The top-performing ensemble models use additional pre-trained word embeddings, which we do not use.
    }
    \label{count tab:vqa-test}
    \centering
    \centerfloat{\begin{tabular}{l c c c c c c c c}
        \toprule
        & \multicolumn{4}{c}{VQA v2 test-dev} & \multicolumn{4}{c}{VQA v2 test}\\
        \cmidrule(l{4pt}){2-5} \cmidrule(l{4pt}){6-9}
        Model & Yes/No & \textbf{Number} & Other & All & Yes/No & \textbf{Number} & Other & All \\
        \hline
        \citet{Teney2017} & 81.82 & 44.21 & 56.05 & 65.32 & 82.20 & 43.90 & 56.26 & 65.67 \\
        \citet{Teney2017} (Ens.)& 86.08 & 48.99 & 60.80 & 69.87 & 86.60 & 48.64 & 61.15 & 70.34 \\
        \citet{Zhou2017} & 84.27 & 49.56 & 59.89 & 68.76 & -- & -- & -- & -- \\
        \citet{Zhou2017} (Ens.) & -- & -- & -- & -- & 86.65 & 51.13 & 61.75 & 70.92 \\
        \hline
        Baseline & 82.98 & 46.88 & 58.99 & 67.50 & 83.21 & 46.60 & 59.20 & 67.78 \\
        + counting& 83.14 & \textbf{51.62} & 58.97 & 68.09 & 83.56 & \textbf{51.39} & 59.11 & 68.41 \\
        \bottomrule
    \end{tabular}}
\end{table}

\begin{table}
    \setlength\tabcolsep{7pt}
    \caption[VQA v2 validation results]{
        Results on the VQA v2 validation set with models trained only on the training set.
        Reported are the mean accuracies and sample standard deviations ($\pm$) over 4 random initialisations.
    }
    \label{count tab:vqa-val}
    \centering
    \centerfloat{\begin{tabular}{l c c c c c c}
        \toprule
        & \multicolumn{3}{c}{VQA accuracy} & \multicolumn{3}{c}{Balanced pair accuracy} \\
        \cmidrule(l{4pt}){2-4} \cmidrule(l{4pt}){5-7}
        Model & Number & \textbf{Count} & All & Number & \textbf{Count} & All \\
        \hline
        Baseline & 44.83\tiny$\pm$0.2 & 51.69\tiny$\pm$0.2 & 64.80\tiny$\pm$0.0 & 17.34\tiny$\pm$0.2 & 20.02\tiny$\pm$0.2 & 36.44\tiny$\pm$0.1 \\
        + NMS & 44.60\tiny$\pm$0.1 & 51.41\tiny$\pm$0.1 & 64.80\tiny$\pm$0.1 & 17.06\tiny$\pm$0.1 & 19.72\tiny$\pm$0.1 & 36.44\tiny$\pm$0.2 \\
        + counting& 49.36\tiny$\pm$0.1 & \textbf{57.03}\tiny$\pm$0.0 & 65.42\tiny$\pm$0.1 & 23.10\tiny$\pm$0.2 & \textbf{26.63}\tiny$\pm$0.2 & 37.19\tiny$\pm$0.1 \\
        \bottomrule
    \end{tabular}}
\end{table}

We also evaluate our models on the validation set of VQA v2, shown in \autoref{count tab:vqa-val}.
This allows us to consider only the counting questions within number questions, since number questions include questions such as "what time is it?" as well.
We treat any question starting with the words "how many" as a counting question.
As we expect, the benefit of using the counting module on the counting question subset is higher than on number questions in general.
Additionally, we try an approach where we simply replace the counting module with NMS, using the average of the attention glimpses as scoring, and one-hot encoding the number of proposals left.
The NMS-based approach, using an IoU threshold of 0.5 and no score thresholding based on validation set performance, does not improve on the baseline, which suggests that the piecewise gradient of NMS is a major problem for learning to count in VQA and that conversely, there is a substantial benefit to being able to differentiate through the counting module.

Additionally, we can evaluate the accuracy over balanced pairs as proposed by \citet{Teney2017}: the ratio of balanced pairs on which the VQA accuracy for both questions is 1.0.
This is a much more difficult metric, since it requires the model to find the subtle details between images instead of being able to rely on question biases in the dataset.
First, notice how all balanced pair accuracies are greatly reduced compared to their respective VQA accuracy.
More importantly, the absolute accuracy improvement of the counting module is still fully present with the more challenging metric, which is further evidence that the module can properly count rather than simply fitting better to dataset biases.

\begin{figure}
    \centering
    \centerfloat{\resizebox{\figurewidth}{!}{%
        \includegraphics[width=1.1\linewidth, trim={1cm 0 0 0},clip]{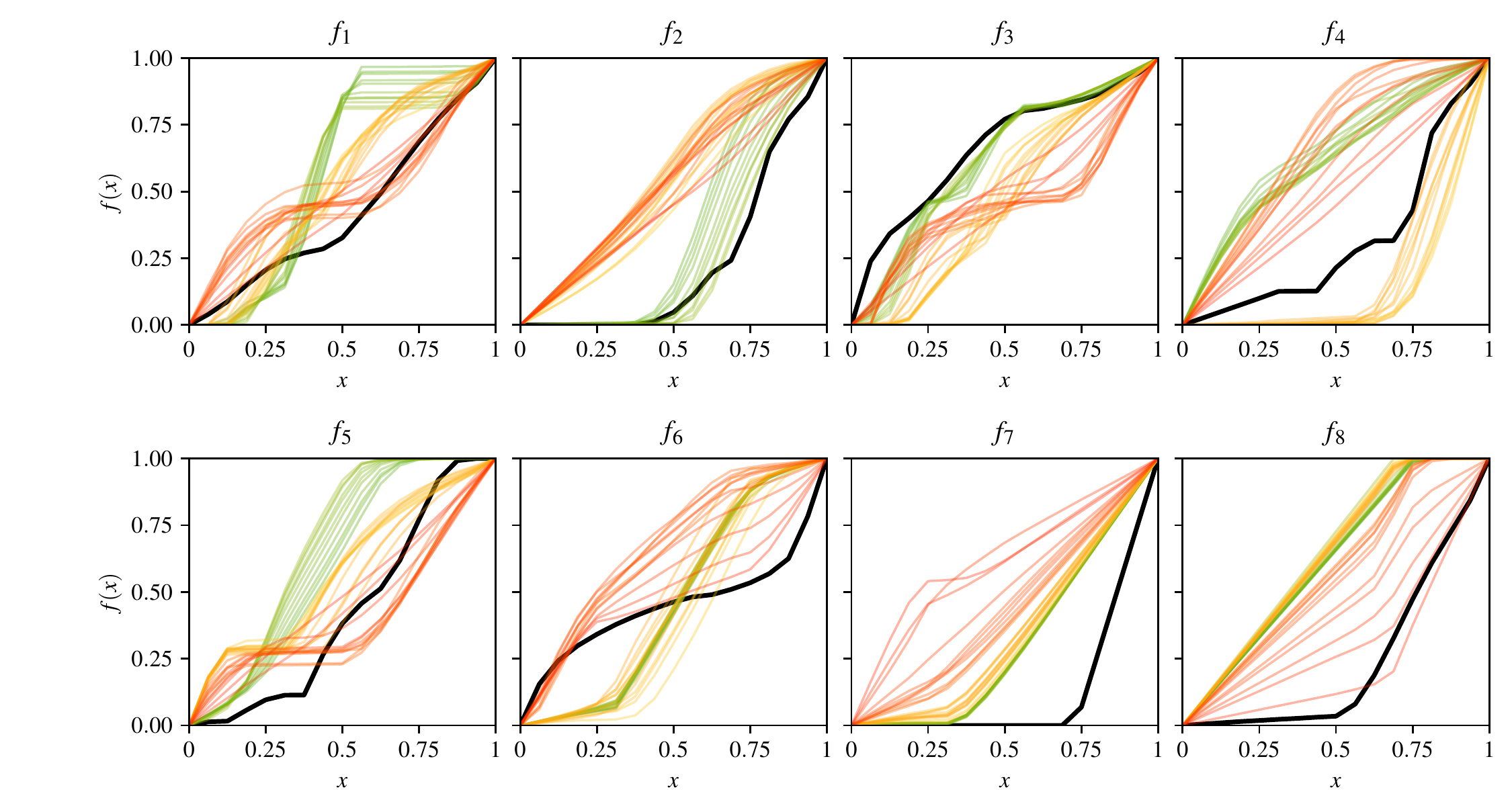}
    }}
    \caption[Shape of trained activation functions on VQA v2]{
        Shape of activation functions for a model trained on the train and validation sets of VQA v2 (thick black), compared against the shapes when parametrising the toy dataset with $q$ around 0.4 (green), 0.7 (orange), or 1.0 (red) with fixed $l=0.2$.
    Best viewed in colour.
    }
    \label{count fig:trained}
\end{figure}

When looking at the activation functions of the trained model, shown in \autoref{count fig:trained}, we find that some characteristics of them are shared with high-noise parametrisations of the toy dataset.
This suggests that the current attention mechanisms and object proposal network are still very inaccurate, which explains the perhaps small-seeming increase in counting performance.
This provides further evidence that the balanced pair accuracy is maybe a more reflective measure of how well current VQA models perform than the overall VQA accuracies of over 70\% of the current top models.

\section{Conclusion}\label{count sec:conclusion}
After understanding why VQA models struggle to count, we designed a counting module that alleviates this problem through differentiable bounding box deduplication.
The module can readily be used alongside any future improvements in VQA models, as long as they still use soft attention as all current top models on VQA v2 do.
It has uses outside of VQA as well: for many counting tasks, it can allow an object-proposal-based approach to work without ground-truth objects available as long as there is a -- possibly learned -- per-proposal scoring (for example using a classification score) and a notion of how dissimilar a pair of proposals are.
Since each step in the module has a clear purpose and interpretation, the learned weights of the activation functions are also interpretable.
The design of the counting module is an example showing how by encoding inductive biases into a deep learning model, challenging problems such as counting of arbitrary objects can be approached when only relatively little supervisory information is available.

For future research, it should be kept in mind that VQA v2 requires a versatile skill set that current models do not have.
To make progress on this dataset, we advocate focusing on \emph{understanding} of what the current shortcomings of models are and finding ways to mitigate them.

\chapter{Set encoder: Permutation-optimisation}\label{cha:perm-optim}
\chaptermark{Permutation-optimisation}
After having motivated the usefulness of sets through the VQA benchmark task, we move on to developing a method for extracting information from sets in a more general setting.
Based on the idea that lists are easier to work with than sets, we aim to let the model \emph{learn} how to turn a set into a list.

These contributions have been published as~\citep{Zhang2019PermOptim} in the International Conference on Learning Representations (ICLR) 2019.

\section{Introduction}\label{perm sec:intro}
Consider a task where each input sample is a \emph{set} of feature vectors with each feature vector describing an object in an image (for example: $\{$person, table, cat$\}$).
Because there is no a priori ordering of these objects, it is important that the model is invariant to the order that the elements appear in the set.
However, this puts restrictions on what can be learned efficiently.
As we covered in \autoref{sec:background-encoders}, the typical approach is to compose elementwise operations with permutation-invariant reduction operations, such as summing~\citep{Zaheer2017} or taking the maximum~\citep{Qi2017} over the whole set.
Since the reduction operator compresses a set of any size down to a single descriptor, this can be a significant bottleneck in what information about the set can be represented efficiently.

We take an alternative approach based on an idea explored in \citet{Vinyals2015}, where they find that some permutations of sets allow for easier learning on a task than others.
They do this by ordering the set elements in some predetermined way and feeding the resulting sequence into a recurrent neural network.
For instance, it makes sense that if the task is to output the top-n numbers from a set of numbers, it is useful if the input is already sorted in descending order before being fed into an RNN.
This approach leverages the representational capabilities of traditional sequential models such as LSTMs, but requires some prior knowledge of what order might be useful.

Our idea is to learn such a permutation purely from data \emph{without requiring a priori knowledge} (\autoref{perm sec:model}).
The key aspect is to turn a set into a sequence in a way that is both permutation-invariant, as well as differentiable so that it is learnable.
Our main contribution is a Permutation-optimisation (PO) module that satisfies these requirements:
it optimises a permutation in the forward pass of a neural network using pairwise comparisons.
By feeding the resulting sequence into a traditional model such as an LSTM, we can learn a flexible, permutation-invariant representation of the set while avoiding the bottleneck that a simple reduction operator would introduce.
Techniques used in our model may also be applicable to other set problems where permutation-invariance is desired, building on the literature of approaches to dealing with permutation-invariance (\autoref{perm sec:related}).

In four different experiments, we show improvements over existing methods (\autoref{perm sec:experiments}).
The former two tasks measure the ability to learn a particular permutation as target: number sorting and image mosaics.
We achieve state-of-the-art performance with our model, which shows that our method is suitable for representing permutations in general.
The latter two tasks test whether a model can learn to solve a task that requires it to come up with a suitable permutation implicitly: classification from image mosaics and visual question answering.
We provide no supervision of what the permutation should be;
the model has to learn by itself what permutation is most useful for the task at hand.
Here, our model also beats the existing models and we improve the performance of a state-of-the-art model in visual question answering (VQA) with it.
This shows that our PO module is able to learn good permutation-invariant representations of sets using our approach.

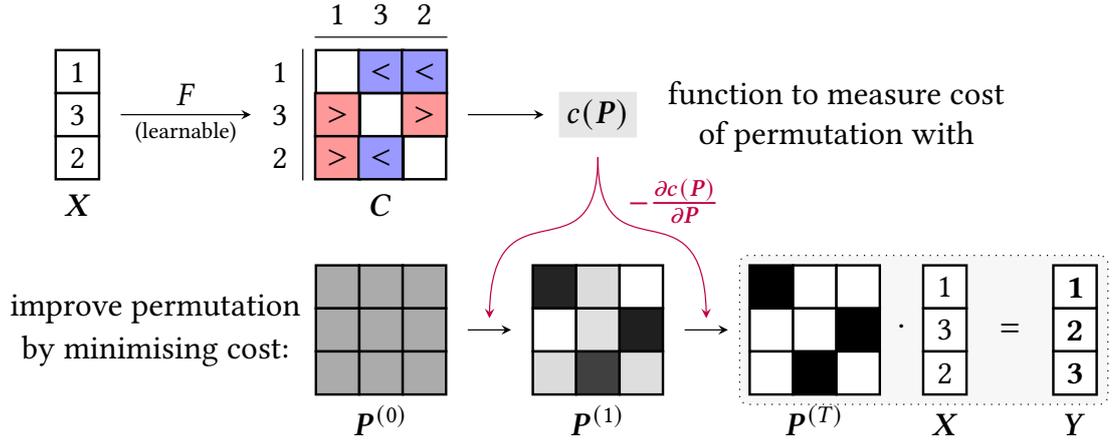
\begin{figure}
    \centering
    \centerfloat{\resizebox{\figurewidth}{!}{%
        \begin{tikzpicture}
        \node {\begin{tikzpicture}\pic {overview};\end{tikzpicture}};
        \end{tikzpicture}
    }}
    \caption[Overview of Permutation-optimisation module]{Overview of Permutation-optimisation module.
        In the ordering cost $\mC$, elements of $\mX$ are compared to each other (blue represents a negative value, red represents a positive value).
        Gradients are applied to unnormalised permutations $\widetilde{\mP}^{(t)}$, which are normalised to proper permutations $\mP^{(t)}$.
    }
    \label{perm fig:overview}
\end{figure}

\sectionmark{Model}
\section{Permutation-optimisation module}\label{perm sec:model}
\sectionmark{Model}

\begin{tcolorbox}[blanker, float=t, oversize=2cm]
    \begin{algorithm}[H]
        \caption{Forward pass of permutation-optimisation algorithm}
        \label{perm alg:pseudo}
        \begin{algorithmic}[1]
            \State \textbf{Input}: $\mX \in \R^{N \times M}$ with $\vx_i$ as rows in arbitrary order

            \State \textbf{Learnable parameters}: weights that parametrise $F$, step size $\eta$

            \State

            \State $C_{ij} \gets \text{normed}(F(\vx_i, \vx_j))$
            \Comment{ordering costs (\autoref{perm eq:ordering})}

            \State initialise $\widetilde{\mP}$
            \Comment{uniform or linear assignment init (\autoref{perm eq:pinit})}

            \For{$\text{t} \gets 1, T$}
                \State $\mP \gets S(\widetilde{\mP})$
                \Comment{normalise assignment (\autoref{perm eq:sinkhorn}}

                \State $\mG \gets \partial c(\mP) / \partial \mP$
                \Comment{compute gradient of assignment (\autoref{perm eq:grad})}

                \State $\widetilde{\mP} \gets \widetilde{\mP} - \eta \mG$
                \Comment{gradient descent step on assignment (\autoref{perm eq:update})}
            \EndFor

            \State $\mP \gets S(\widetilde{\mP})$

            \State $\mY \gets \mP \mX$
            \Comment{permute rows of $\mX$ to obtain output $\mY$}
        \end{algorithmic}
    \end{algorithm}
\end{tcolorbox}

We will now describe a differentiable, and thus learnable model to turn an \emph{unordered} set $\{\vx_i\}_N$ with feature vectors as elements into an \emph{ordered} sequence of these feature vectors.
An overview of the algorithm is shown in \autoref{perm fig:overview} and pseudo-code is shown in \autoref{perm alg:pseudo}.
The input set is represented as a matrix $\mX = [\vx_1, \ldots, \vx_N]^\T$ with the feature vectors $\vx_i$ as rows in some arbitrary order.
Note that this representation of sets is transposed compared to \autoref{cha:background} ($\R^{n \times d}$ instead of $\R^{d \times n}$) to ease the exposition in this chapter.
In the algorithm, it is important to not rely on the arbitrary order so that $\mX$ is correctly treated as a set.
The goal is then to learn a permutation matrix $\mP$ such that when permuting the rows of the input through $\mY = \mP \mX$, the output is ordered correctly according to the task at hand.
When an entry $P_{ik}$ takes the value $1$, it can be understood as assigning the $i$th element to the $k$th position in the output.

Our main idea is to first relate pairs of elements through an \emph{ordering cost}, parametrised with a neural network.
This pairwise cost tells us whether an element $i$ should preferably be placed before or after element $j$ in the output sequence.
Using this, we can define a \emph{total cost} that measures how good a given permutation is (\autoref{perm subsec:total}).
The second idea is to optimise this total cost in each forward pass of the module (\autoref{perm subsec:optim}).
By minimising the total cost of a permutation, we improve the quality of a permutation with respect to the current ordering costs.
Crucially, the ordering cost function -- and thus also the total cost function -- is learned.
In doing so, the module is able to learn how to generate a permutation as is desired.

In order for this to work, it is important that the optimisation process itself is differentiable so that the ordering cost is learnable.
Because permutations are inherently discrete objects, a continuous relaxation of permutations is necessary.
For optimisation, we perform gradient descent on the total cost for a fixed number of steps and unroll the iteration, similar to how recurrent neural networks are unrolled to perform backpropagation-through-time.
Because the \emph{inner gradient} (total cost differentiated with respect to permutation) is itself differentiable with respect to the ordering cost, the whole model is kept differentiable and we can train it with a standard supervised learning loss.\

Note that as long as the ordering cost is computed appropriately (\autoref{perm subsec:order}), all operations used turn out to be permutation-equivariant or invariant.
Thus, we have a model that respects the symmetries of sets while producing an output without those symmetries: a sequence.
This can be naturally extended to outputs where the target is not a sequence, but grids and lattices (\autoref{perm subsec:lattice}).

\subsection{Total cost function}\label{perm subsec:total}
The total cost function measures the quality of a given permutation and should be lower for better permutations.
Because this is the function that will be optimised, it is important to understand what it expresses precisely.

The main ingredient for the total cost of a permutation is the pairwise ordering cost (details in \autoref{perm subsec:order}).
By computing it for all pairs, we obtain a cost matrix $\mC$ where the entry $C_{ij}$ represents the ordering cost between $i$ and $j$:
the cost of placing element $i$ anywhere before $j$ in the output sequence.
An important constraint that we put on $\mC$ is that $C_{ij} = - C_{ji}$.
In other words, if one ordering of $i$ and $j$ is ``good'' (negative cost), then the opposite ordering obtained by swapping them is ``bad'' (positive cost).
Additionally, this constraint means that $C_{ii} = 0$.
This makes sure that two very similar feature vectors in the input will be similarly ordered in the output because their pairwise cost goes to $0$.

In this chapter we use a straightforward definition of the total cost function: a sum of the ordering costs over all pairs of elements $i$ and $j$.
When considering the pair $i$ and $j$, if the permutation maps $i$ to be before $j$ in the output sequence, this cost is simply $C_{ij}$.
Vice versa, if the permutation maps $i$ to be after $j$ in the output sequence, the cost has to be flipped to $C_{ji}$.
To express this idea, we define the total cost $c \colon \mathbb{R}^{N \times N} \mapsto \mathbb{R}$ of a permutation $\mP$ as:

\begin{equation}
    c(\mP) = \sum_{ij} C_{ij} \sum_{k} P_{ik} \left(\sum_{k' > k} P_{jk'} - \sum_{k' < k} P_{jk'}\right)
\end{equation}

This can be understood as follows:
If the permutation assigns element $i$ to position $u$ (so $P_{iu} = 1$) and element $j$ to position $v$ (so $P_{jv} = 1$), the sums over $k$ and $k'$ simplify to $1$ when $v > u$ and $-1$ when $v < u$;
permutation matrices are binary and only have one $1$ in any row and column, so all other terms in the sums are $0$.
That means that the term for each $i$ and $j$ becomes $C_{ij}$ when $v > u$ and $- C_{ij} = C_{ji}$ when $v < u$, which matches what we described previously.

\subsection{Optimisation problem}\label{perm subsec:optim}
Now that we can compute the total cost of a permutation, we want to optimise this cost with respect to a permutation.
After including the constraints to enforce that $\mP$ is a valid permutation matrix, we obtain the following optimisation problem:

\begin{equation}
    \begin{aligned}
        & \minimize_{\mP}
        & & c(\mP) \\
        & \text{subject to}
          & & \forall i,k \colon P_{ik} \in \{0, 1\}, \\
          & & & \forall i \colon \sum_k P_{ik} = 1, \sum_k P_{ki} = 1\\
    \end{aligned}
\end{equation}

Optimisation over $\mP$ directly is difficult due to the discrete and combinatorial nature of permutations.
To make optimisation feasible, a common relaxation is to replace the constraint that $P_{ik} \in \{0, 1\}$ with $P_{ik} \in [0, 1]$ \citep{Fogel2013}.
With this change, the feasible set for $\mP$ expands to the set of doubly-stochastic matrices, known as the Birkhoff or assignment polytope.
Rather than hard permutations, we now have soft assignments of elements to positions, analogous to the latent assignments when fitting a mixture of Gaussians model using Expectation-Maximisation.

Note that we do not need to change our total cost function after this relaxation.
Instead of discretely flipping the sign of $C_{ij}$ depending on whether element $i$ comes before $j$ or not, the sums over $k$ and $k'$ give us a weight for each $C_{ij}$ that is based on how strongly $i$ and $j$ are assigned to positions.
This weight is positive when $i$ is on average assigned to earlier positions than $j$ and negative vice versa.

In order to perform optimisation of the cost under our constraints, we reparametrise $\mP$ with the Sinkhorn operator $S$ from \citet{Adams2011} so that the constraints are always satisfied.

We found this to lead to better solutions than projected gradient descent in initial experiments.
After first exponentiating all entries of a matrix, $S$ repeatedly normalises all rows, then all columns of the matrix to sum to 1, which converges to a doubly-stochastic matrix in the limit.

\begin{equation}
    \mP = S(\widetilde{\mP})
\end{equation}

This ensures that $\mP$ is always approximately a doubly-stochastic matrix.
$\widetilde{\mP}$ can be thought of as the unnormalised permutation while $\mP$ is the normalised permutation.
By changing our optimisation to minimise $\widetilde{\mP}$ instead of $\mP$ directly, all constraints are always satisfied and we can simplify the optimisation problem to $\min_{\widetilde{\mP}} c(\mP)$ without any constraints.

\begin{sidenote}[float]{Aside: Sinkhorn operator}
    The Sinkhorn operator $S$ as defined in \citet{Adams2011} is:

    \begin{align}
        \mathcal{T}_r(\mX)_{ij} &= X_{ij} / \sum_k X_{ik} \\
        \mathcal{T}_c(\mX)_{ij} &= X_{ij} / \sum_k X_{kj} \\
        S^{(0)}(\mX) &= \exp(\mX) \\
        S^{(l+1)}(\mX) &= \mathcal{T}_c\bigg(\mathcal{T}_r\left(S^{(l)}(\mX)\right)\bigg) \\
        S(\mX) &= S^{(L)}(\mX)\label{perm eq:sinkhorn}
    \end{align}

    $\mathcal{T}_r$ normalises each row, $\mathcal{T}_c$ normalises each column of a square matrix $\mX$ to sum to one.
    This formulation is different from the normal Sinkhorn operator by \citet{Sinkhorn1964} by exponentiating all entries first and running for a fixed number of steps $L$ instead of for steps approaching infinity.
    \citet{Mena2018} include a temperature parameter on the exponentiation, which acts analogously to temperature in the softmax function.
    In this chapter, we fix $L$ to 4.
\end{sidenote}

It is now straightforward to optimise $\widetilde{\mP}$ with standard gradient descent.
First, we compute the gradient:

\begin{align}
    \frac{\partial c(\mP)}{\partial {P}_{pq}} &= 2 \sum_{j} C_{pj} \left(\sum_{k' > q} P_{jk'} - \sum_{k' < q} P_{jk'}\right) \label{perm eq:grad} \\
    \frac{\partial c(\mP)}{\partial \widetilde{P}_{pq}} &= \frac{\partial \mP}{\partial \widetilde{P}_{pq}} \cdot \frac{\partial c(\mP)}{\partial \mP} \label{perm eq:truegrad}
\end{align}

From \autoref{perm eq:grad}, it becomes clear that this gradient is itself differentiable with respect to the ordering cost $C_{ij}$, which allows it to be learned.
In practice, both ${\partial c(\mP)} / {\partial \widetilde{\mP}}$ as well as $\partial [\partial c(\mP) / \partial \widetilde{\mP}] / \partial \mC$ can be computed with automatic differentiation.
However, some implementations of automatic differentiation require the computation of $c(\mP)$ which we do not use.
In this case, implementing ${\partial c(\mP)} / {\partial \widetilde{\mP}}$ explicitly can be more efficient.
Also notice that if we define $B_{jq} = \sum_{k' > q} P_{jk'} - \sum_{k' < q} P_{jk'}$, \autoref{perm eq:grad} is just the matrix multiplication $\mC \mB$ and is thus efficiently computable.

For the optimisation, $\mP$ has to be initialised in a permutation-equivariant way to preserve permutation-invariance of the algorithm.
In this chapter, we consider a uniform initialisation so that all $P_{ik} = 1 / N$ (\newterm{PO-U} model, left) and an initialisation that linearly assigns~\citep{Mena2018} each element to each position (\newterm{PO-LA} model, right).

\begin{equation}
    \widetilde{P}_{ik}^{(0)} = 0 \quad \mathrel{\text{or}} \quad \widetilde{P}_{ik}^{(0)} = \vw_k \vx_i
    \label{perm eq:pinit}
\end{equation}

where $\vw_k$ is a different weight vector for each position $k$.
Then, we perform gradient descent for a fixed number of steps $T$.
The iterative update using the gradient and a (learnable) step size $\eta$ converges to the optimised permutation $\mP^{(T)}$:

\begin{equation}
    \widetilde{\mP}^{(t + 1)} = \widetilde{\mP}^{(t)} - \eta \ \frac{\partial c(\mP^{(t)})}{\partial \mP^{(t)}}\label{perm eq:update}
\end{equation}

One peculiarity of this is that we update $\widetilde{\mP}$ with the gradient of the normalised permutation $\mP$, not of the unnormalised permutation $\widetilde{\mP}$ as normal.
In other words, we do gradient descent on $\widetilde{\mP}$ but in \autoref{perm eq:truegrad} we set $\partial P_{uv} / \partial \widetilde{P}_{pq} = 1$ when $u = p, v = q$, and 0 everywhere else.
We found that this results in significantly better permutations experimentally; we believe that this is because $\partial \mP / \partial \widetilde{\mP}$ vanishes too quickly from the Sinkhorn normalisation, which biases $\mP$ away from good permutation matrices wherein all entries are close to 0 and 1.
We justify this in more detail in \autoref{perm app:sinkgrad}.

The runtime of this algorithm is dominated by the computation of gradients of $c(\mP)$, which involves a matrix multiplication of two $N \times N$ matrices.
In total, the time complexity of this algorithm is $T$ times the complexity of this matrix multiplication, which is $\Theta(N^3)$ in practice.
We found that typically, small values for $T$ such as $4$ are enough to get good permutations.

\subsection{Ordering cost function}\label{perm subsec:order}
The ordering cost $C_{ij}$ is used in the total cost and tells us what the pairwise cost for placing $i$ before $j$ should be.
The key property to enforce is that the function $F$ that produces the entries of $\mC$ is anti-symmetric ($F(\vx_i, \vx_j) = -F(\vx_j, \vx_i)$).
A simple way to achieve this is to define $F$ as:

\begin{equation}
    F(\vx_i, \vx_j) = f(\vx_i, \vx_j) - f(\vx_j, \vx_i)
\end{equation}

We can then use a small neural network for $f$ to obtain a learnable $F$ that is always anti-symmetric.

Lastly, $\mC$ is normalised to have unit Frobenius norm.
This results in simply scaling the total cost obtained, but it also decouples the scale of the outputs of $F$ from the step size parameter $\eta$ to make optimisation more stable at inference time.
$\mC$ is then defined as:

\begin{align}
    \widetilde{C}_{ij} &= F(\vx_i, \vx_j) \\
    C_{ij} &= \widetilde{C}_{ij} / \Vert \widetilde{\mC} \Vert_F \label{perm eq:ordering}
\end{align}

\subsection{Extending permutations to lattices}\label{perm subsec:lattice}
In some tasks, it may be natural to permute the set into a lattice structure instead of a sequence.
For example, if it is known that the set contains parts of an image, it makes sense to arrange these parts back to an image by using a regular grid.
We can straightforwardly adapt our model to this by considering each row and column of the target grid as an individual permutation problem.
The total cost of an assignment to a grid is the sum of the total costs over all individual rows and columns of the grid.
The gradient of this new cost is then the sum of the gradients of these individual problems.
This results in a model that considers both row-wise and column-wise pairwise relations when permuting a set of inputs into a grid structure, and more generally, into a lattice structure.

\subsection{Justification for alternative update}\label{perm app:sinkgrad}
We mentioned previously that we perform gradient descent with the post-Sinkhorn gradients on the pre-Sinkhorn matrix.
Here, we justify why this is a reasonable thing to do.

First, the gradient of $S(\mX)$ is:
\begin{align}
    \frac{\partial S(\mX)}{\partial \mX} &=
    \frac{\partial \exp(\mX)}{\partial \mX}
    \frac{\partial \mathcal{T}_r\big(\exp(\mX)\big)}{\partial \exp(\mX)}
    \frac{\partial \mathcal{T}_c\Big(\mathcal{T}_r\big(\exp(\mX)\big)\Big)}{\partial \mathcal{T}_r\big(\exp(\mX)\big)}
    \cdots \label{perm eq:sinkgrad}
    \\
    \frac{\partial \mathcal{T}_r(\mX)_{uv}}{\partial X_{ij}} &= \bbone_{u=i} \frac{\bbone_{v=j}\sum_k X_{uk} - X_{uv}}{(\sum_k X_{uk})^2} \\
    \frac{\partial \mathcal{T}_c(\mX)_{uv}}{\partial X_{ij}} &= \bbone_{v=j} \frac{\bbone_{u=i}\sum_k X_{kv} - X_{uv}}{(\sum_k X_{kv})^2}
\end{align}

where $\bbone$ is the indicator function that returns 1 if the condition is true and 0 otherwise.

We compared the entropy of the permutation matrices obtained with and without using the ``proper'' gradient with $\partial S(\widetilde{\mP}) / \partial \widetilde{\mP}$ as term in it and found that our version has a significantly lower entropy.
To understand this, it is enough to focus on the first two terms in \autoref{perm eq:sinkgrad}, which is essentially the gradient of a softmax function applied row-wise to $\mP$.

Let $\vx$ be a row in $\mP$ and $s_i$ be the $i$th entry in the softmax function applied to $\vx$.
Then, the gradient is:
\begin{equation}
    \frac{\partial s_i}{\partial x_j} = s_i (\bbone_{i=j} - s_j)
\end{equation}

Since this is a product of entries in a probability distribution, the gradient vanishes quickly as we move towards a proper permutation matrix (all entries very close to 0 or 1).
By using our alternative update and thus removing this term from our gradient, we can avoid the vanishing gradient problem.

Gradient descent is not efficient when the gradient vanishes towards the optimum and the optimum -- in our case a permutation matrix with exact ones and zeros as entries -- is infinitely far away.
Since we prefer to use a small number of steps in our algorithm for efficiency, we want to reach a good solution as quickly as possible.
This justifies effectively ignoring the step size that the gradient suggests and simply taking a step in a similar direction as the gradient in order to be able to saturate the Sinkhorn normalisation sufficiently, thus obtaining a doubly stochastic matrix that is closer to a proper permutation matrix in the end.

\subsection{Quadratic programming formulation}
We can write our total cost function as a quadratic program in the standard $\vx^\T \mQ \vx$ form with linear constraints. (We leave out the constraints here as they are not particularly interesting.)
This connects our work to the extensive quadratic programming literature.

First, we can define $\mO \in \R^{N \times N}$ as:

\begin{equation}
    O_{k k'} =
    \begin{cases}
        -1 & \text{if } k > k' \\
        0 & \text{if } k = k' \\
        1 & \text{if } k < k' \\
    \end{cases}
\end{equation}

and with it, $\mQ \in \R^{N^2 \times N^2}$ as:

\begin{equation}
    Q_{(ik)(jk')} = C_{ij} O_{kk'}
\end{equation}

Then we can write the cost function as:

\begin{align}
    c(\mP) &= \sum_{ij} C_{ij} \sum_{k k'} P_{ik} P_{jk'} O_{kk'} \\
        &= \sum_{ik} \sum_{jk'} P_{ik} (C_{ij} O_{kk'}) P_{jk'} \\
        &= \sum_{(ik)} \sum_{(jk')} P_{(ik)} Q_{(ik)(jk')} P_{(jk')}\\
        &= \vp^\T \mQ \vp
\end{align}

where there is some bijection between a pair of indices $(i, k)$ and the index $l$, and $\vp$ is a flattened version of $\mP$ with $p_l = P_{ik}$.
$\mQ$ is indefinite because the total cost can be negative:
a uniform initialisation for $\mP$ has a cost of 0, better permutations have negative cost, worse permutations have positive cost.
Thus, the problem is non-convex and the problem is possibly NP-hard.
Also, since we have flattened $\mP$ into $\vp$, the number of optimisation variables is quadratic in the set size $N$.
Even if this were a convex quadratic program, methods such as OptNet~\citep{Amos2017} have cubic time complexity in the number of optimisation variables, which makes it $O(N^6)$ for our case.

\section{Related work}\label{perm sec:related}
The most relevant work to ours is the inspiring study by \citet{Mena2018}, where they discuss the reparametrisation that we use and propose a model that can also learn permutations implicitly in principle.
Their model uses a simple elementwise linear map from each of the $N$ elements of the set to the $N$ positions, normalised by the Sinkhorn operator.
This can be understood as classifying each element individually into one of the $N$ classes corresponding to positions, then normalising the predictions so that each class only occurs once within this set.
However, processing the elements individually means that their model does not take relations between elements into account properly; elements are placed in absolute positions, not relative to other elements.
Our model differs from theirs by considering pairwise relations when creating the permutation.
By basing the cost function on pairwise comparisons, it is able to order elements such that local relations in the output are taken into account.
We believe that this is important for learning from permutations implicitly, because networks such as CNNs and RNNs rely on local ordering more than absolute positioning of elements.
It also allows our model to process variable-sized sets, which their model is not able to do.

Our work is closely related to the set function literature, where the main constraint is invariance to ordering of the set.
While it is always possible to simply train using as many permutations of a set as possible, using a model that is naturally permutation-invariant increases learning and generalisation capabilities through the correct inductive bias in the model.
There are some similarities with relation networks~\citep{Santoro2017} in considering all pairwise relations between elements as in our pairwise ordering function.
However, they sum over all non-linearly transformed pairs, which can lead to the bottleneck we mention in \autoref{perm sec:intro}.
Meanwhile, by using an RNN on the output of our model, our approach can encode a richer class of functions:
it can still learn to simply sum the inputs, but it can also learn more complex functions where the learned order between elements is taken into account.
The concurrent work by~\citet{Murphy2019} discusses various approximations of averaging the output of a neural network over all possible permutations, with our method falling under their categorisation of a learned canonical input ordering.
Our model is also relevant to neural networks operating on graphs such as graph convolutional networks~\citep{Kipf2017}.
Typically, a set function is applied to the set of neighbours for each node, with which the state of the node is updated.
Our module combined with an RNN is thus an alternative set function to perform this state update with.

\citet{Noroozi2016} and \citet{Cruz2017} show that it is possible to use permutation learning for representation learning in a self-supervised setting.
The model in \citet{Cruz2017} is very similar to \citet{Mena2018}, including use of a Sinkhorn operator, but they perform significantly more processing on images with a large CNN (AlexNet) beforehand with the main goal of learning good representations for that CNN.
We instead focus on using the permuted set itself for representation learning in a supervised setting.

We are not the first to explore the usefulness of using optimisation in the forward pass of a neural network (for example, \citet{Stoyanov2011,Domke2012,Belanger2017}).
However, we believe that we are the first to show the potential of optimisation for processing sets because -- with an appropriate cost function -- it is able to preserve permutation-invariance.
In OptNet~\citep{Amos2017}, exact solutions to convex quadratic programs are found in a differentiable way through various techniques.
Unfortunately, our quadratic program is non-convex, which makes finding an optimal solution possibly NP-hard~\citep{Pardalos1991}.
We thus fall back to the simpler approach of gradient descent on the reparametrised problem to obtain a non-optimal, but reasonable solution.

Note that our work differs from learning to rank approaches such as \citet{Burges2005} and \citet{Severyn2015}, as there the end goal is the permutation itself.
This usually requires supervision on what the target permutation should be, producing a permutation with hard assignments at the end.
We require our model to produce soft assignments so that it is easily differentiable, since the main goal is not the permutation itself, but processing it further to form a representation of the set being permuted.
This means that other approaches that produce hard assignments such as Ptr-Net~\citep{Vinyals2015a} are also unsuitable for implicitly learning permutations, although using a variational approximation through \citet{Mena2018} to obtain a differentiable permutation with hard assignments is a promising direction to explore for the future.
Due to the lack of differentiability, existing literature on solving minimum feedback arc set problems~\citep{Charon2007} can not be easily used for set representation learning either.




\section{Experiments}\label{perm sec:experiments}
Throughout the text, we will refer to our model with uniform assignment as PO-U, with linear assignment initialisation as PO-LA, and the model from \citet{Mena2018} as LinAssign.
We perform a qualitative analysis of what comparisons are learned in \autoref{perm app:analysis}.
Precise experimental details can be found in \autoref{perm app:details} and our implementation for all experiments is available at \url{https://github.com/Cyanogenoid/perm-optim} for full reproducibility.

An interesting aspect we observed throughout all experiments is how the learned step size $\eta$ changes during training.
At the start of training, it decreases from its initial value of 1, thus reducing the influence of the permutation mechanism.
Then, $\eta$ starts rising again, usually ending up at a value above 1 at the end of training.
This can be explained by the ordering cost being very inaccurate at the start of training, since it has not been trained yet.
Through training, the ordering cost improves and it becomes more beneficial for the influence of the PO module on the permutation to increase.

\subsection{Sorting numbers}\label{perm sec:numbersort}

\begin{figure}
    \centering
    \fontsize{10}{10}\selectfont
    \centerfloat{\resizebox{\textwidth}{!}{%
        \begin{tikzpicture}
            \node {\begin{tikzpicture}\pic {arch sort};\end{tikzpicture}};
        \end{tikzpicture}
    }}
    \caption[Network architecture for number sorting]{Network architecture for number sorting. The MLP in $F$ computes $f$ and is shared across pairs of set elements. The PO block performs the optimisation with the given costs and permutes the input set.}
    \label{perm fig:arch-sort}
\end{figure}
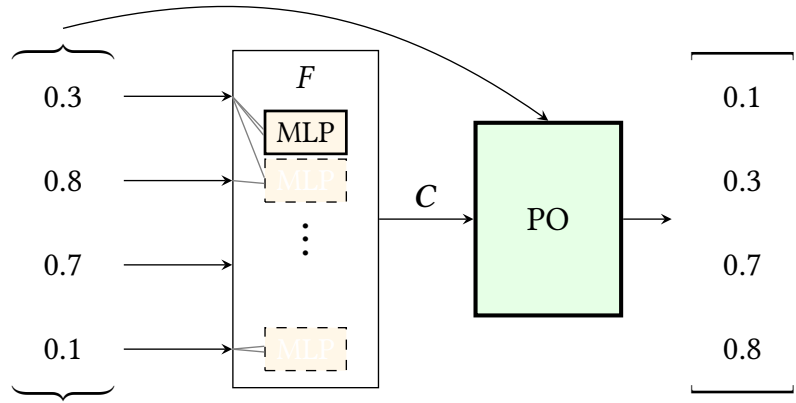

We start with the toy task of turning a set of random unsorted numbers into a sorted list.
For this problem, we train with fixed-size sets of numbers drawn uniformly from the interval $[0, 1]$ and evaluate on different intervals to determine generalisation ability (for example: $[0, 1], [0, 1000], [1000, 1001]$).
We use the correctly ordered sequence as training target and minimise the mean squared error.
Following \citet{Mena2018}, during evaluation we use the Hungarian algorithm for solving a linear assignment problem with $- \mP$ as the assignment costs.
This is done to obtain a permutation with hard assignments from our soft permutation.
We show our model architecture in \autoref{perm fig:arch-sort}.

\paragraph{Results}
Our PO-U model is able to sort all sizes of sets that we tried -- $5$ to $1024$ numbers -- perfectly, including generalising to all the different evaluation intervals without any mistakes.
This is in contrast to all existing end-to-end learning-based approaches such as \citet{Mena2018}, which starts to make mistakes on $[0, 1]$ at 120 numbers and no longer generalises to sets drawn from $[1000, 1001]$ at 80 numbers.
\citet{Vinyals2015} already starts making mistakes on $5$ numbers.
Our stark improvement over existing results is evidence that the inductive biases due to the learned pairwise comparisons in our model are suitable for learning permutations, at least for this particular toy problem.
In \autoref{perm sec:analysis-sorting}, we investigate what it learns that allows it to generalise this well.

\subsection{Re-assembling image mosaics}\label{perm sec:mosaic explicit}

\begin{figure}
    \centering
    \centerfloat{\resizebox{\figurewidth}{!}{%
        \begin{tikzpicture}
        \node {\begin{tikzpicture}\pic {arch classify};\end{tikzpicture}};
        \end{tikzpicture}
    }}
    \caption[Network architecture for image mosaic tasks]{Network architecture for image mosaic tasks. The small CNN and the MLP in $F$ is shared across set elements and pairs of set elements respectively. The PO block performs the optimisation with the given row and column costs and permutes the input set. The ResNet-18 network at the end is only present in the implicit permutation setting.}
    \label{perm fig:arch-classify}
\end{figure}
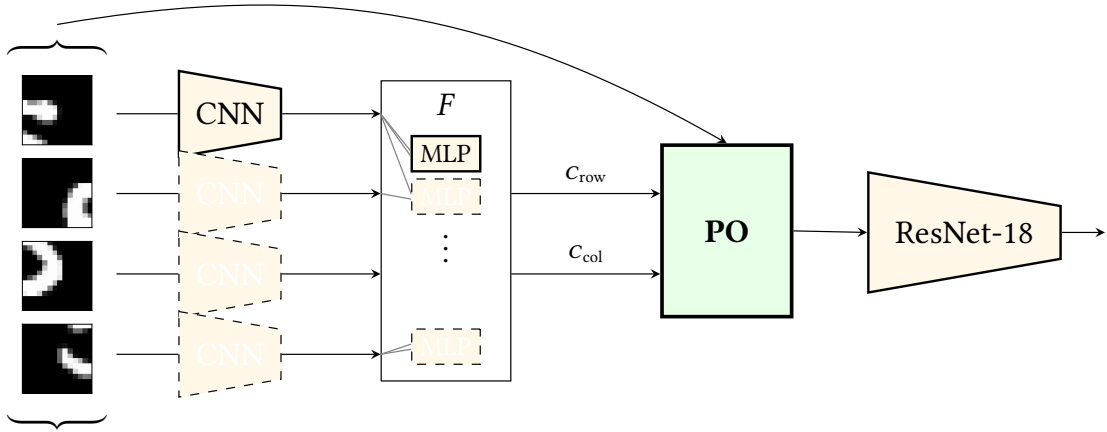

As second task, we consider a problem where the model is given images that are split into $n \times n$ equal-size tiles and the goal is to re-arrange this set of tiles back into the original image.
We take these images from either MNIST, CIFAR10, or a version of ImageNet with images resized down to $64 \times 64$ pixels.
For this task, we use the alternative cost function described in \autoref{perm subsec:lattice} to arrange the tiles into a grid rather than a sequence;
this lets our model take relations within rows and columns into account.
Again, we minimise the mean squared error to the correctly permuted image and use the Hungarian algorithm during evaluation, matching the experimental setup in \citet{Mena2018}.
Due to the lack of reference implementation of their model for this experiment, we use our own implementation of their model, which we verified to reproduce their MNIST results closely.
Unlike them, we decide to not arbitrarily upscale MNIST images to get improved results for all models.
We show our model architecture in \autoref{perm fig:arch-classify}.

\begin{table}
    \setlength\tabcolsep{4pt}
    \centering
    \caption[MSE of explicit image mosaic reconstruction]{Mean squared error of image mosaic reconstruction for different datasets and number of tiles an image is split into. Lower is better. LinAssign* is the model by \citet{Mena2018}, LinAssign is our reproduction of their model, PO-U and PO-LA are our models with uniform and linear assignment initialisation respectively.}
    \label{perm tab:mosaic}
    \centerfloat{\begin{tabular}{c c c c c c c c c c c c c}
\toprule
& \multicolumn{4}{c}{MNIST} & \multicolumn{4}{c}{CIFAR10} & \multicolumn{4}{c}{ImageNet $64 \times 64$} \\
\cmidrule(l{4pt}){2-5} \cmidrule(l{4pt}){6-9} \cmidrule(l{4pt}){10-13}
Model & \small $2 \times 2$ & \small $3 \times 3$ & \small $4 \times 4$ & \small $5 \times 5$ & \small $2 \times 2$ & \small $3 \times 3$ & \small $4 \times 4$ & \small $5 \times 5$ & \small $2 \times 2$ & \small $3 \times 3$ & \small $4 \times 4$ & \small $5 \times 5$ \\
\midrule
\textit{LinAssign*} & \textit{0.00} & \textit{0.00} & \textit{0.26} & \textit{0.18} & \textit{--} & \textit{--} & \textit{--} & \textit{--} & \textit{0.22} & \textit{0.31} & \textit{--} & \textit{--} \\
LinAssign & 0.00 & 0.00 & 0.33 & 0.08 & 0.37 & 0.49 & 1.34 & 1.12 & 0.60 & 1.10 & 1.33 & 1.44 \\
PO-U & \textbf{0.00} & 0.02 & 0.46 & 0.45 & \textbf{0.11} & 0.44 & 1.23 & 1.26 & \textbf{0.14} & 0.69 & 1.20 & \textbf{1.31} \\
PO-LA & 0.00 & \textbf{0.00} & \textbf{0.07} & \textbf{0.01} & 0.18 & \textbf{0.16} & \textbf{1.07} & \textbf{0.70} & 0.16 & \textbf{0.62} & \textbf{1.13} & 1.32 \\
\bottomrule
    \end{tabular}}
\end{table}

\begin{table}
    \setlength\tabcolsep{4pt}
    \centering
    \caption[Accuracy with explicit image mosaic reconstruction]{Accuracy of image mosaic reconstruction. Higher is better. A permutation is considered correct if all tiles are placed correctly. Because of indistinguishable tiles at higher tile counts (for example multiple completely blank tiles on MNIST) it becomes very unlikely to guess the correct ground-truth matching at higher tile counts. LinAssign* results come from \citet{Mena2018}.}
    \label{perm tab:mosaic2}

    \centerfloat{\begin{tabular}{c c c c c c c c c c c c c}
        \toprule
& \multicolumn{4}{c}{MNIST} & \multicolumn{4}{c}{CIFAR10} & \multicolumn{4}{c}{ImageNet $64 \times 64$} \\
\cmidrule(l{4pt}){2-5} \cmidrule(l{4pt}){6-9} \cmidrule(l{4pt}){10-13}
Model & \small $2 \times 2$ & \small $3 \times 3$ & \small $4 \times 4$ & \small $5 \times 5$ & \small $2 \times 2$ & \small $3 \times 3$ & \small $4 \times 4$ & \small $5 \times 5$ & \small $2 \times 2$ & \small $3 \times 3$ & \small $4 \times 4$ & \small $5 \times 5$ \\
\midrule
\textit{LinAssign*} & \textit{100} & \textit{72} & \textit{3} & \textit{0} & \textit{--} & \textit{--} & \textit{--} & \textit{--} & \textit{81} & \textit{47} & \textit{0} & \textit{0} \\
LinAssign & 99.7 & 66.9 & 0.8 & 0.0 & 68.8 & 30.3 & 0.0 & 0.0 & 47.7 & 4.0 & 0.0 & \textbf{0.0} \\
PO-U & \textbf{100.0} & 65.9 & 0.2 & 0.0 & 86.2 & 34.4 & 0.0 & 0.0 & \textbf{85.9} & 19.2 & 0.1 & 0.0 \\
PO-LA & 99.9 & \textbf{73.1} & \textbf{1.8} & \textbf{0.0} & \textbf{87.3} & \textbf{66.0} & \textbf{0.6} & \textbf{0.3} & 84.2 & \textbf{28.6} & \textbf{0.1} & 0.0 \\
\bottomrule
    \end{tabular}}
\end{table}

\paragraph{Results}
The mean squared errors for the different image datasets and different number of tiles an image is split into are shown in \autoref{perm tab:mosaic}.
The corresponding accuracies when training a classifier on these reconstructions are shown in  \autoref{perm tab:mosaic2}.
First, notice that in essentially all cases, our model with linear assignment initialisation (PO-LA) performs best, often significantly so.
On the two more complex datasets CIFAR10 and ImageNet, this is followed by our PO-U model, then the LinAssign model.
We analyse what types of comparisons PO-U learns in \autoref{perm sec:analysis-mosaic}.

On MNIST, LinAssign performs better than PO-U on higher tile counts because images are always centred on the object of interest.
That means that many tiles only contain the background and end up completely blank;
these tiles can be more easily assigned to the borders of the image by the LinAssign model than our PO-U model because the absolute position is much more important than the relative positioning to other tiles.
This also points towards an issue for these cases in our cost function:
because two tiles that have the same contents are treated the same by our model, it is unable to place one blank tile on one side of the image and another blank tile on the opposite side, as this would require treating the two tiles differently.
This issue with backgrounds is also present on CIFAR10 to a lesser extent:
notice how for the $3 \times 3$ case, the error of PO-U is much closer to LinAssign on CIFAR10 than on ImageNet, where PO-U is much better comparatively.
This shows that the PO-U model is more suitable for more complex images when relative positioning matters more.
PO-LA is able to combine the best of both methods.

\begin{figure}
    \centering
    \centerfloat{\resizebox{\figurewidth}{!}{%
    \includegraphics[width=\textwidth]{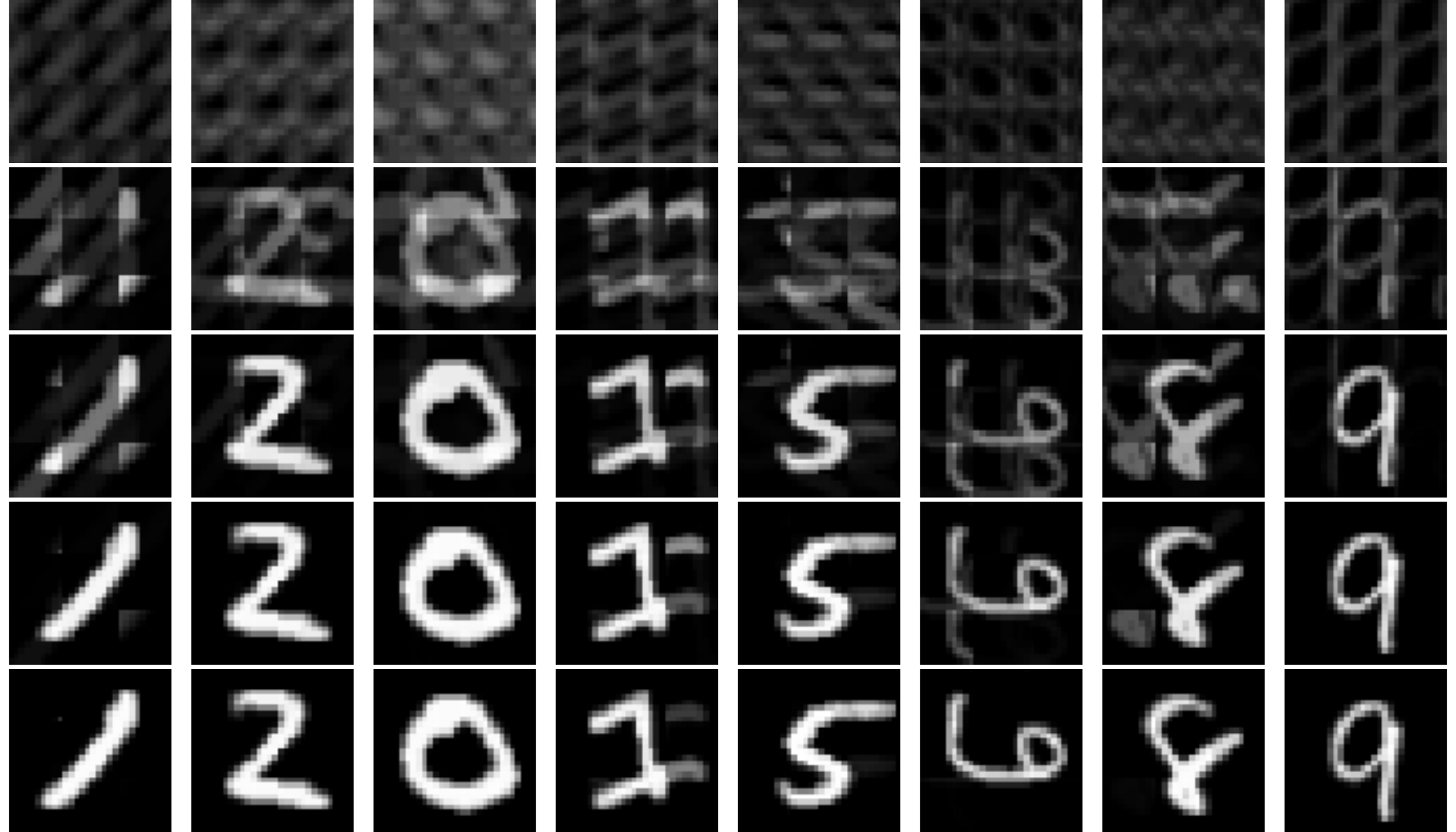}
    }}
    \caption[Example explicit reconstructions on MNIST $3 \times 3$]{Example reconstructions of PO-U as they are being optimised on MNIST $3 \times 3$ with explicit supervision. These examples have not been cherry-picked.}
    \label{perm fig:mosaic-mnist}
\end{figure}

\begin{figure}
    \centering
    \centerfloat{\resizebox{\figurewidth}{!}{%
    \includegraphics[width=\textwidth]{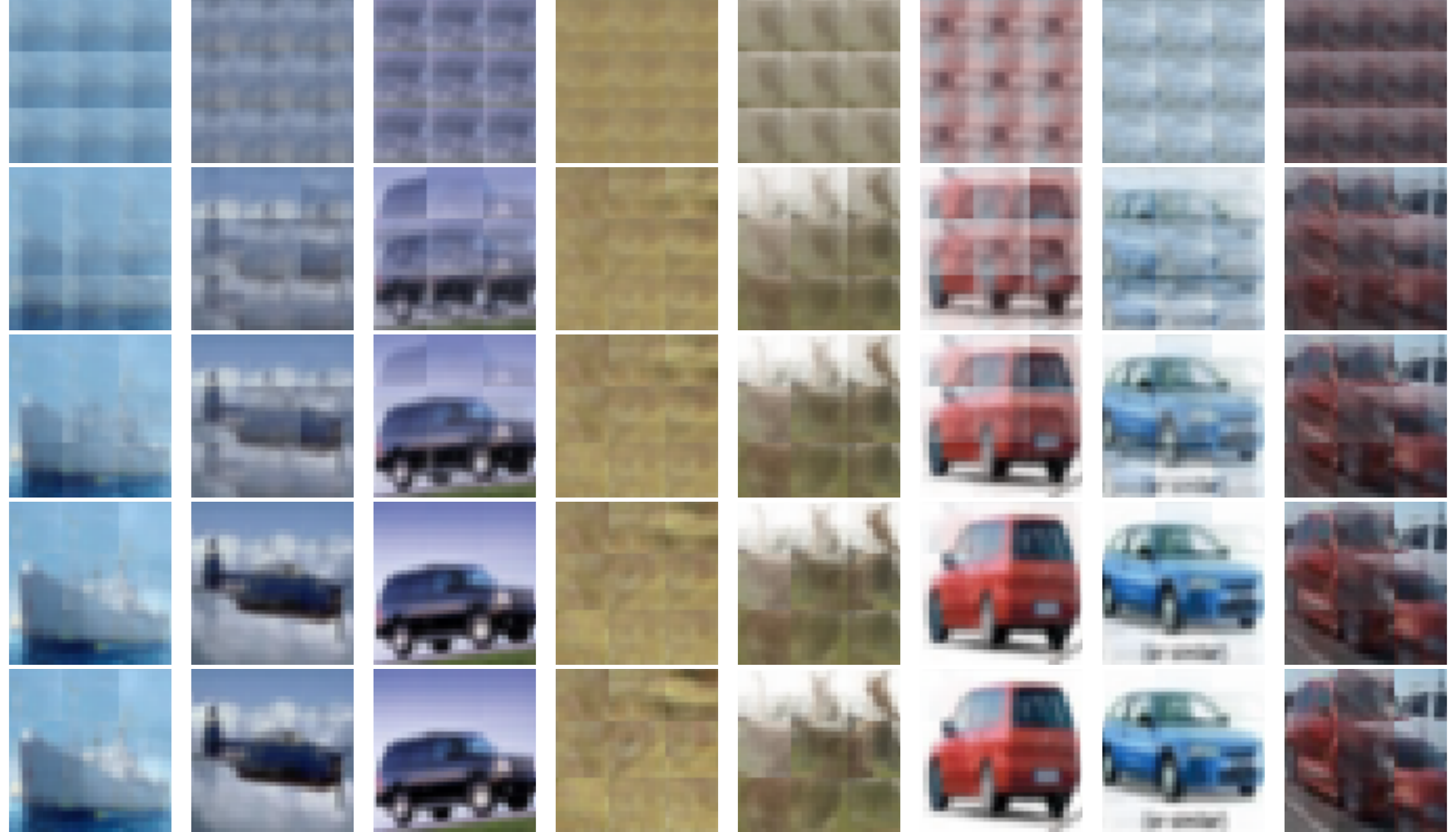}
    }}
    \caption[Example explicit reconstructions on CIFAR10 $3 \times 3$]{Example reconstructions of PO-U as they are being optimised on CIFAR10 $3 \times 3$ with explicit supervision. These examples have not been cherry-picked.}
    \label{perm fig:mosaic-cifar10}
\end{figure}

\begin{figure}
    \centering
    \centerfloat{\resizebox{\figurewidth}{!}{%
    \includegraphics[width=\textwidth]{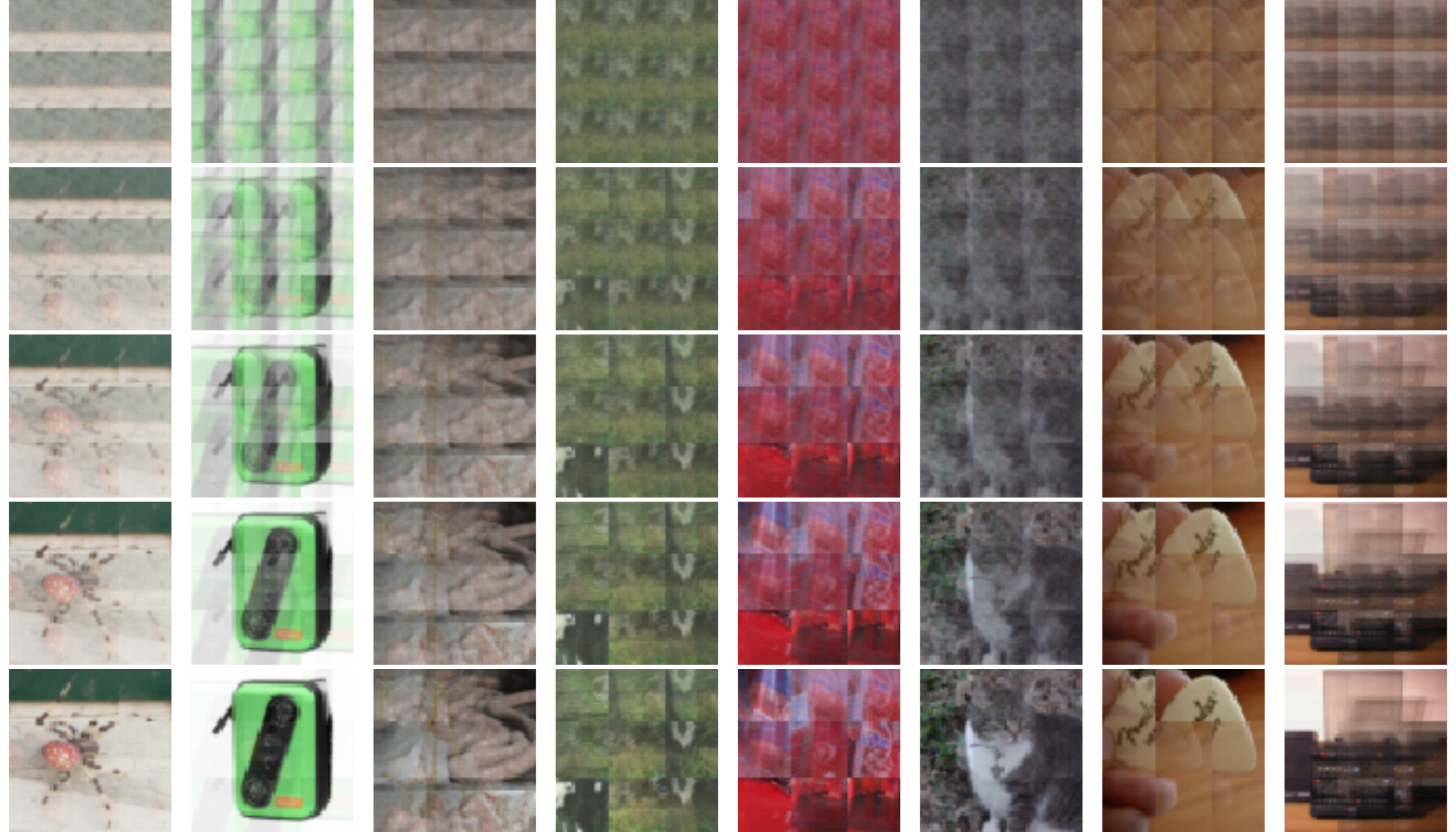}
    }}
    \caption[Example explicit reconstructions on ImageNet $3 \times 3$]{Example reconstructions of PO-U as they are being optimised on ImageNet $3 \times 3$ with explicit supervision. These examples have not been cherry-picked.}
    \label{perm fig:mosaic-imagenet}
\end{figure}

In \autoref{perm fig:mosaic-mnist}, \autoref{perm fig:mosaic-cifar10}, and \autoref{perm fig:mosaic-imagenet}, we show some example reconstructions that have been learnt by our PO-U model.
Starting from a uniform assignment at the top, the figures show reconstructions as a permutation is being optimised.
Generally, it is able to reconstruct most images fairly well.


\subsection{Implicit permutations through classification}\label{perm sec:mosaic implicit}
We now turn to tasks where the goal is not producing the permutation itself, but learning a suitable permutation for a different task.
For these tasks, we do not provide explicit supervision on what the permutation should be; an appropriate permutation is learned implicitly while learning to solve another task.

As the dataset, we use a straightforward modification of the image mosaic task.
The image tiles are assigned to positions on a grid as before, which are then concatenated into a full image.
This image is fed into a standard image classifier (ResNet-18~\citep{He2016}) which is trained with the usual cross-entropy loss to classify the image.
The idea is that the network has to learn \emph{some} permutation of the image tiles so that the classifier can classify it accurately.
This is not necessarily the permutation that restores the original image faithfully.

One issue with this set-up we observed is that with big tiles, it is easy for a CNN to ignore the artefacts on the tile boundaries, which means that simply permuting the tiles randomly gets to almost the same test accuracy as using the original image.
To prevent the network from avoiding to solve the task, we first pre-train the CNN on the original dataset without permuting the image tiles.
Once it is fully trained, we freeze the weights of this CNN and train only the permutation mechanism.

\begin{table}
    \setlength\tabcolsep{4pt}
    \centering
    \caption[Accuracy with implicit reconstructions]{
        Accuracy of classification from implicitly-learned image reconstructions through permutations. \textit{max} shows the accuracy of the pre-trained model on the original images, \textit{min} shows the accuracy of the pre-trained model on images with randomly permuted tiles.}
    \label{perm tab:classify}
    \centerfloat{\begin{tabular}{c c c c c c c c c c c c c}
\toprule
& \multicolumn{4}{c}{MNIST} & \multicolumn{4}{c}{CIFAR10} & \multicolumn{4}{c}{ImageNet $64 \times 64$} \\
\cmidrule(l{4pt}){2-5} \cmidrule(l{4pt}){6-9} \cmidrule(l{4pt}){10-13}
Model & \small $2 \times 2$ & \small $3 \times 3$ & \small $4 \times 4$ & \small $5 \times 5$ & \small $2 \times 2$ & \small $3 \times 3$ & \small $4 \times 4$ & \small $5 \times 5$ & \small $2 \times 2$ & \small $3 \times 3$ & \small $4 \times 4$ & \small $5 \times 5$ \\
\midrule
\emph{max} & \textit{99.5} & \textit{99.5} & \textit{99.5} & \textit{99.3} & \textit{81.0} & \textit{81.0} & \textit{81.9} & \textit{80.2} & \textit{31.2} & \textit{33.4} & \textit{31.2} & \textit{33.5} \\
\emph{min} & \textit{36.6} & \textit{22.5} & \textit{17.1} & \textit{14.6} & \textit{36.5} & \textit{26.4} & \textit{22.9} & \textit{18.0} & \textit{11.4} & \textit{7.8} & \textit{3.5} & \textit{2.7} \\
LinAssign & \textbf{99.4} & 99.2 & 86.0 & 84.2 & 64.6 & 33.8 & 33.4 & \textbf{32.5} & 13.1 & 5.8 & 5.3 & 3.3 \\
PO-U & 99.3 & 98.7 & 67.9 & 69.2 & 70.8 & \textbf{41.6} & 33.3 & 29.7 & \textbf{24.6} & \textbf{12.1} & \textbf{7.3} & \textbf{5.1} \\
PO-LA & 99.3 & \textbf{99.4} & \textbf{93.3} & \textbf{89.8} & \textbf{71.6} & 40.7 & \textbf{34.2} & 32.3 & 23.4 & 10.9 & 6.3 & 4.4 \\
\bottomrule
    \end{tabular}}
\end{table}

\begin{table}

    \setlength\tabcolsep{4pt}
    \centering
    \caption[MSE of implicit reconstructions]{Mean squared error of implicitly-learned reconstruction. Lower is better.}
    \label{perm tab:classify2}
    \centerfloat{\begin{tabular}{c c c c c c c c c c c c c}

        \toprule
& \multicolumn{4}{c}{MNIST} & \multicolumn{4}{c}{CIFAR10} & \multicolumn{4}{c}{ImageNet $64 \times 64$} \\
\cmidrule(l{4pt}){2-5} \cmidrule(l{4pt}){6-9} \cmidrule(l{4pt}){10-13}
Model & \small $2 \times 2$ & \small $3 \times 3$ & \small $4 \times 4$ & \small $5 \times 5$ & \small $2 \times 2$ & \small $3 \times 3$ & \small $4 \times 4$ & \small $5 \times 5$ & \small $2 \times 2$ & \small $3 \times 3$ & \small $4 \times 4$ & \small $5 \times 5$ \\
\midrule
\emph{max} & \textit{0.00} & \textit{0.00} & \textit{0.00} & \textit{0.00} & \textit{0.00} & \textit{0.00} & \textit{0.00} & \textit{0.00} & \textit{0.00} & \textit{0.00} & \textit{0.00} & \textit{0.00} \\
\emph{min} & \textit{1.59} & \textit{1.63} & \textit{1.91} & \textit{1.72} & \textit{1.76} & \textit{1.91} & \textit{2.11} & \textit{2.01} & \textit{1.48} & \textit{1.72} & \textit{1.79} & \textit{1.83} \\
LinAssign & 0.02 & 0.05 & 0.73 & 1.11 & 0.56 & 1.29 & 1.53 & 1.62 & 0.88 & 1.33 & 1.32 & \textbf{1.39} \\
PO-U & 0.02 & 0.15 & 1.38 & 1.24 & 0.33 & 1.03 & \textbf{1.44} & \textbf{1.34} & 0.44 & 1.16 & \textbf{1.28} & 1.47 \\
PO-LA & \textbf{0.01} & \textbf{0.03} & \textbf{0.50} & \textbf{0.93} & \textbf{0.28} & \textbf{1.00} & 1.47 & 1.55 & \textbf{0.41} & \textbf{1.15} & 1.41 & 1.43 \\
\bottomrule

    \end{tabular}}

\end{table}

\paragraph{Results}
We show the classification results in \autoref{perm tab:classify} and the corresponding MSE reconstruction losses in \autoref{perm tab:classify2}.
Note that the models are not trained to minimise MSE loss.
Generally, a similar trend to the image mosaic task with explicit supervision can be seen.
Our PO-LA model usually performs best, although for ImageNet PO-U is consistently better.
This is evidence that for more complex images, the benefits of linear assignment decrease (and can actually detract from the task in the case of ImageNet) and the importance of the optimisation process in our model increases.
With higher number of tiles on MNIST, even though PO-U does not perform well, PO-LA is clearly superior to only using LinAssign.
This is again due to the fully black tiles not being able to be sorted well by the cost function with uniform initialisation.

\begin{figure}
    \centering
    \centerfloat{\resizebox{\figurewidth}{!}{%
    \includegraphics[width=\figurewidth]{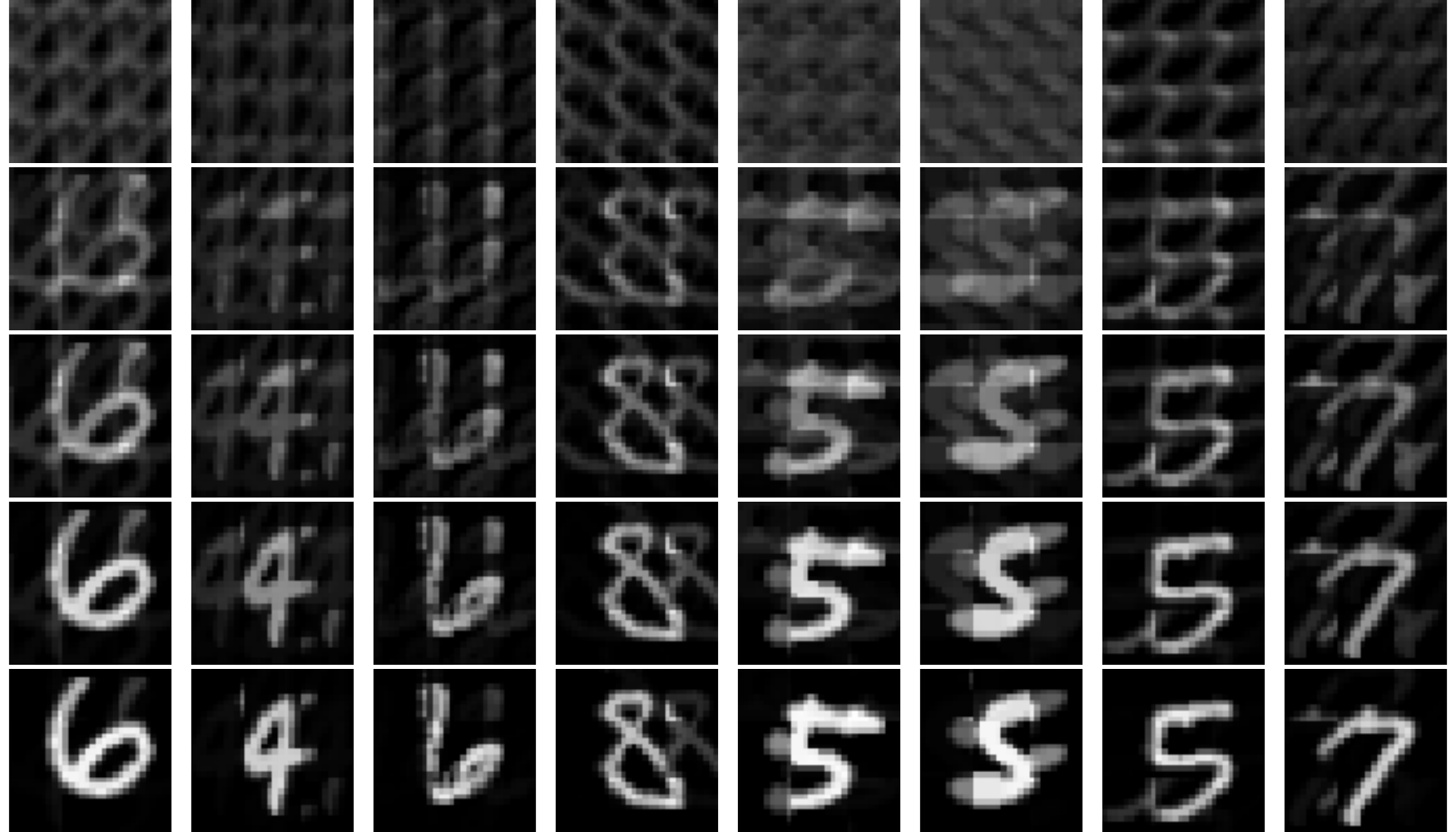}
    }}
    \caption[Example implicit reconstructions on MNIST]{Example reconstructions of PO-U as they are being optimised on MNIST $3 \times 3$ with implicit supervision. These examples have not been cherry-picked.}
    \label{perm fig:classify-mnist}
\end{figure}
\begin{figure}
    \centering
    \centerfloat{\resizebox{\figurewidth}{!}{%
    \includegraphics[width=\figurewidth]{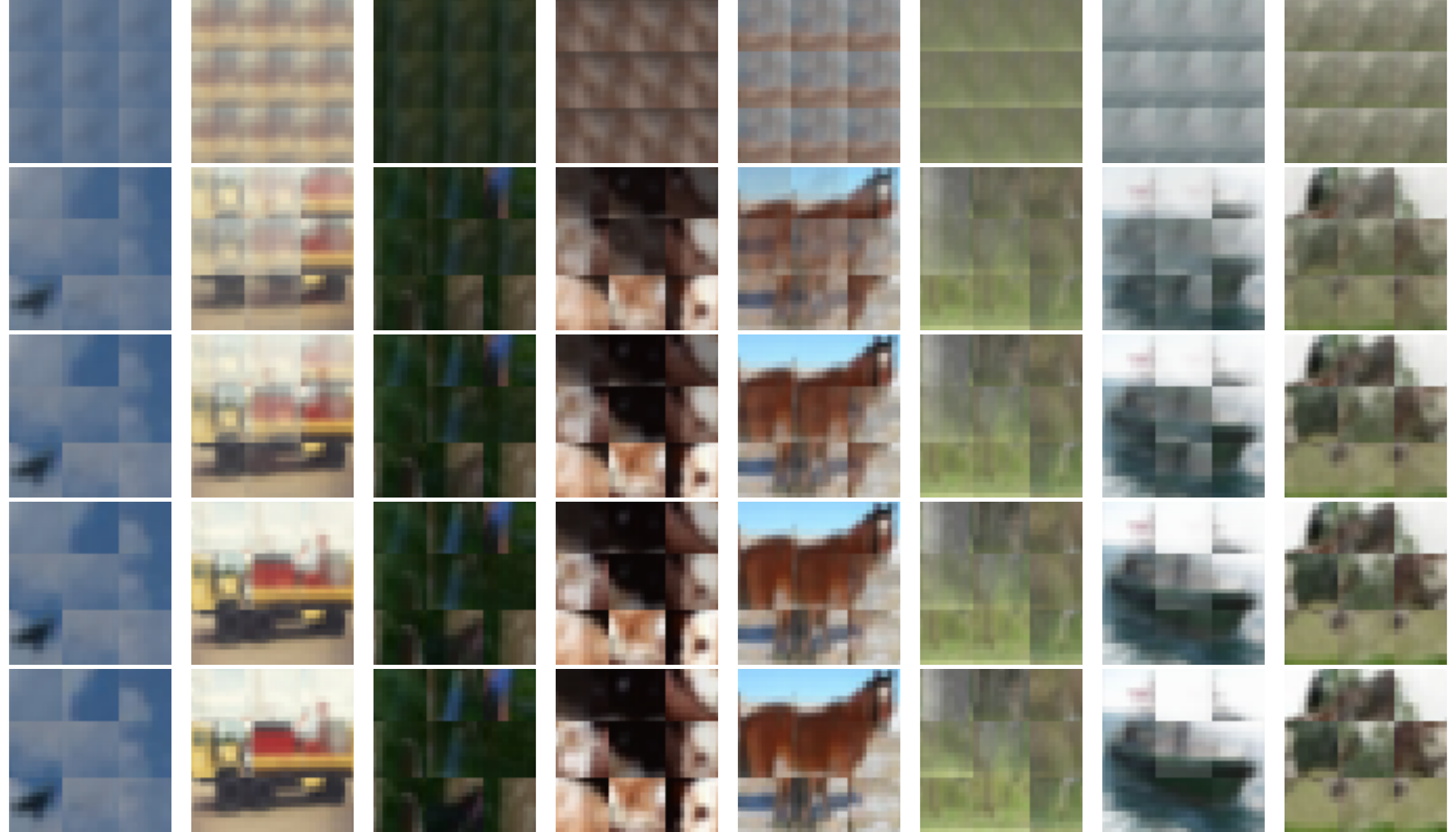}
    }}
    \caption[Example implicit reconstructions on CIFAR10 $3 \times 3$]{Example reconstructions of PO-U as they are being optimised on CIFAR10 $3 \times 3$ with implicit supervision. These examples have not been cherry-picked.}
    \label{perm fig:classify-cifar10}
\end{figure}
\begin{figure}
    \centering
    \centerfloat{\resizebox{\figurewidth}{!}{%
    \includegraphics[width=\figurewidth]{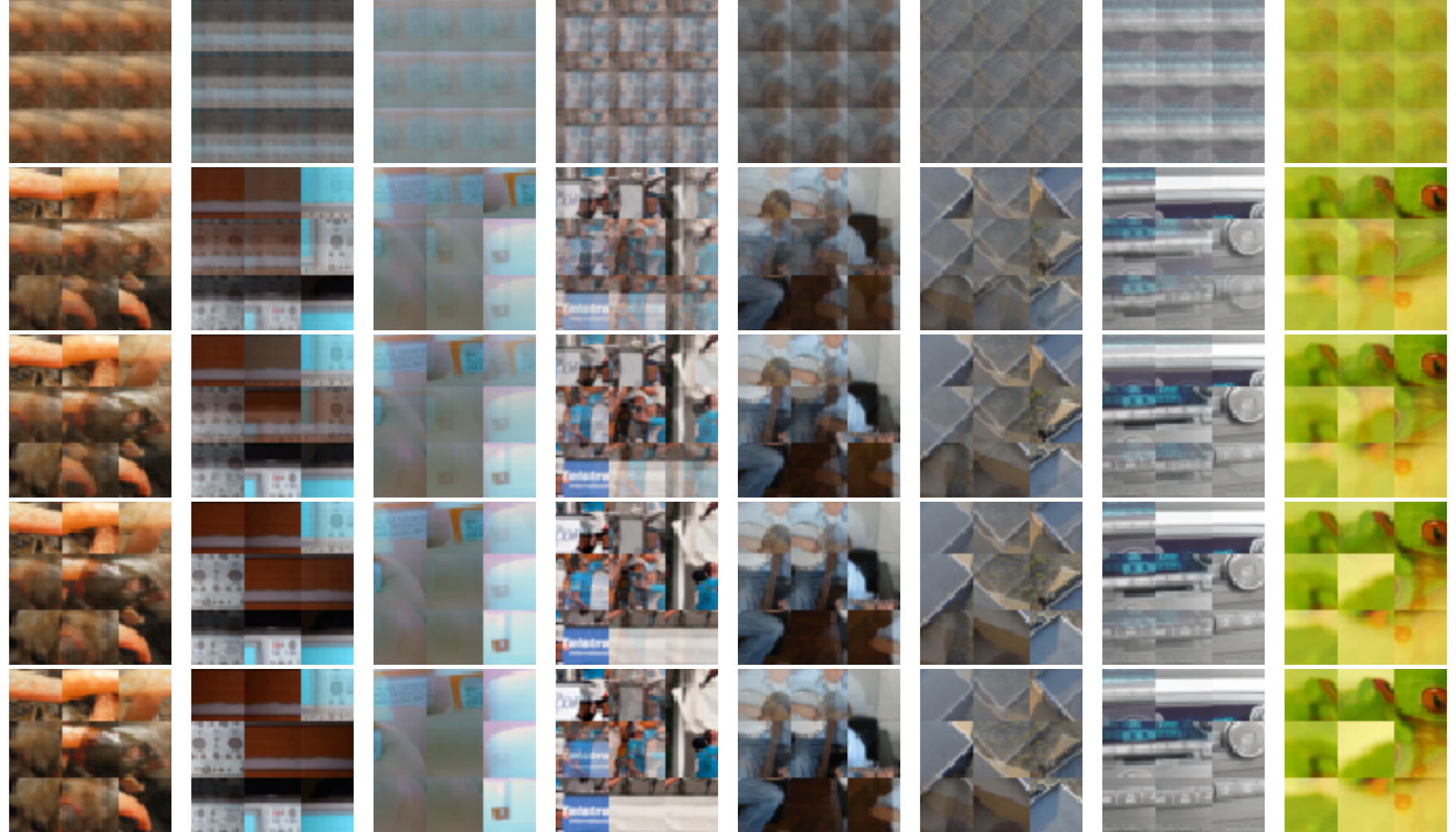}
    }}
    \caption[Example implicit reconstructions on $3 \times 3$]{Example reconstructions of PO-U as they are being optimised on ImageNet $3 \times 3$ with implicit supervision. These examples have not been cherry-picked.}
    \label{perm fig:classify-imagenet}
\end{figure}
\begin{figure}
    \centering
    \centerfloat{\resizebox{\figurewidth}{!}{%
    \includegraphics[width=\figurewidth]{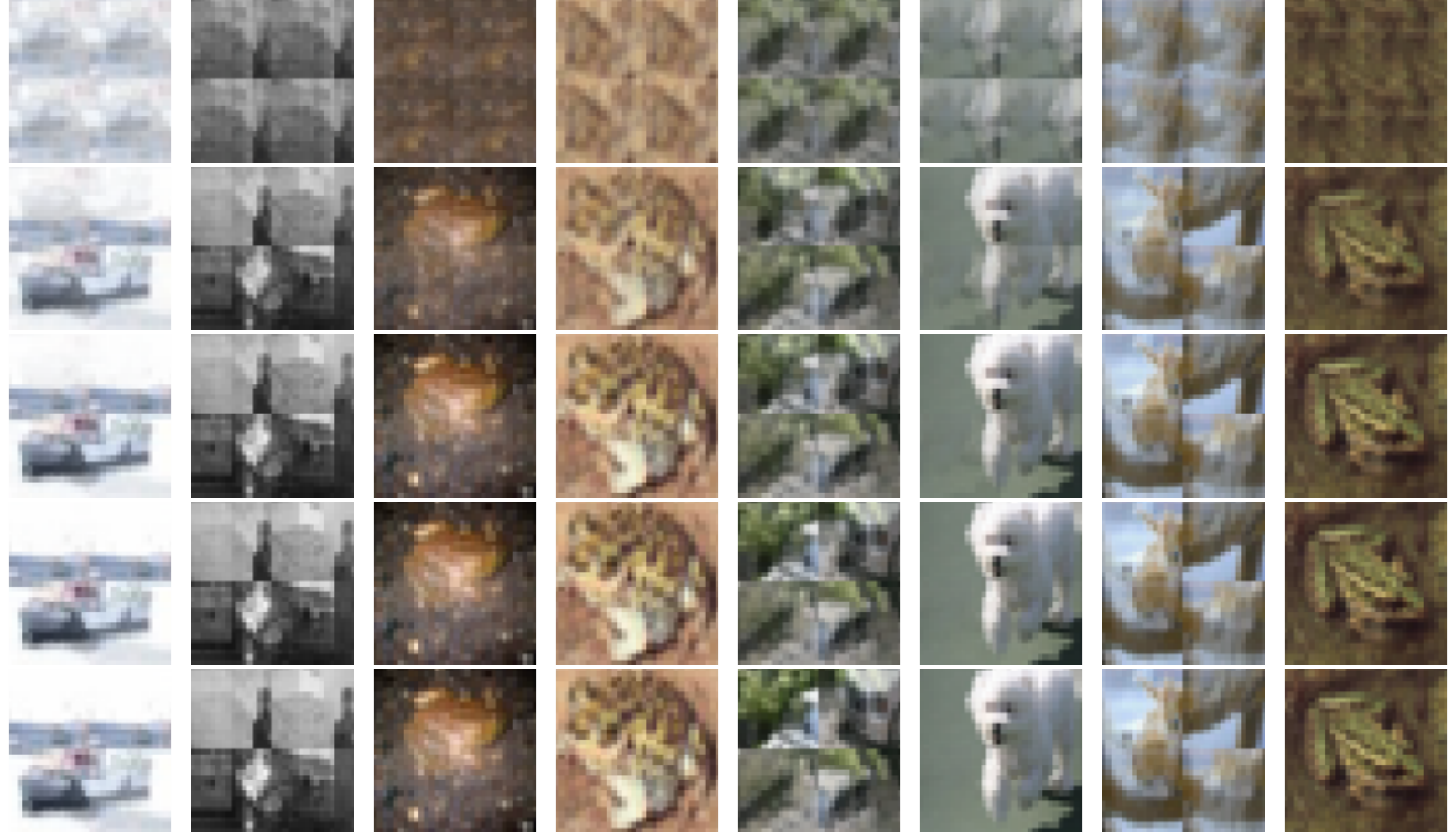}
    }}
    \caption[Example implicit reconstructions on CIFAR10 $2 \times 2$]{Example reconstructions of PO-U as they are being optimised on CIFAR10 $2 \times 2$ with implicit supervision. These examples have not been cherry-picked.}
    \label{perm fig:classify-cifar10-2}
\end{figure}
\begin{figure}
    \centering
    \centerfloat{\resizebox{\figurewidth}{!}{%
    \includegraphics[width=\figurewidth]{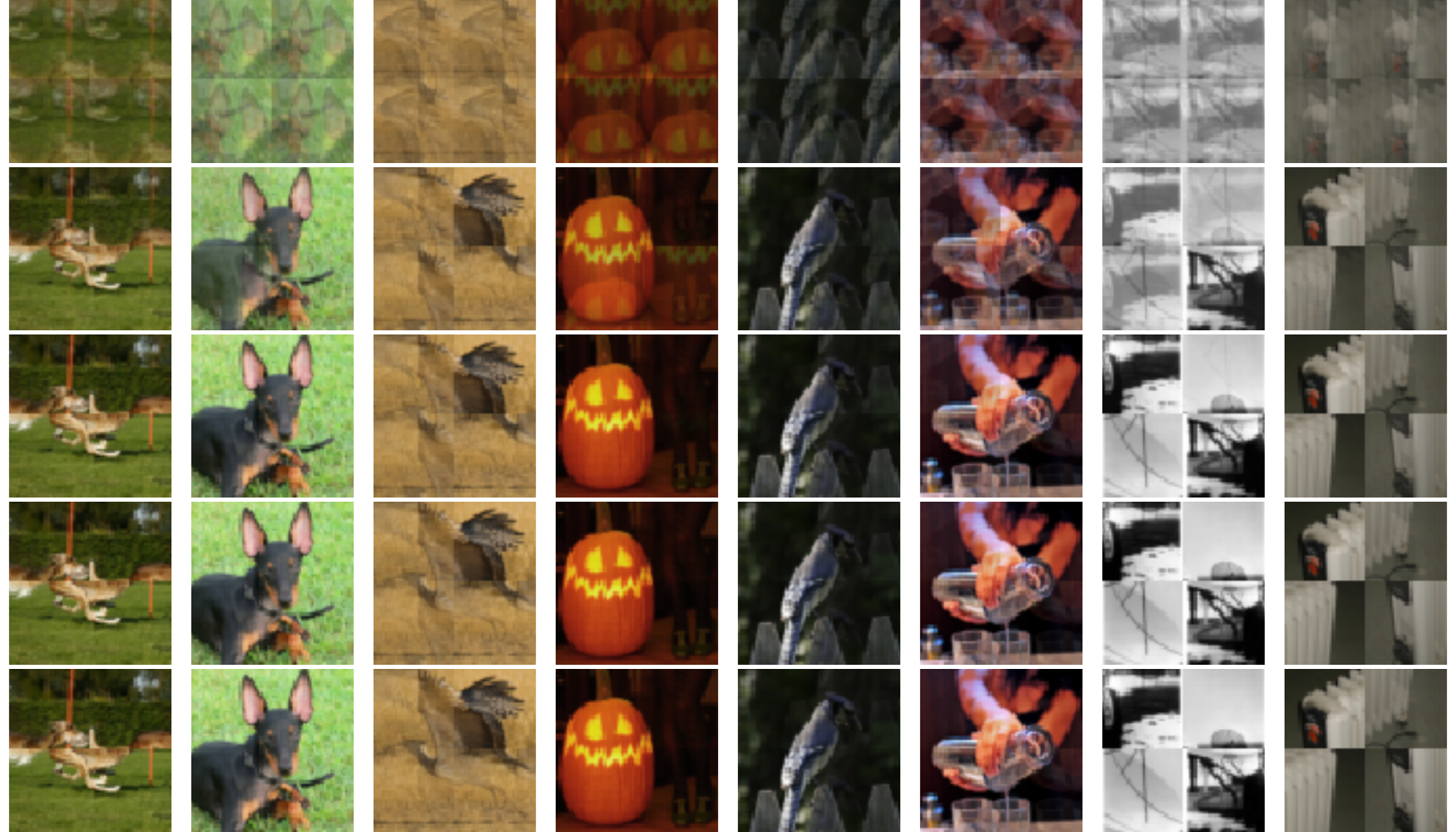}
    }}
    \caption[Example implicit reconstructions on $2 \times 2$]{Example reconstructions of PO-U as they are being optimised on ImageNet $2 \times 2$ with implicit supervision. These examples have not been cherry-picked.}
    \label{perm fig:classify-imagenet-2}
\end{figure}

In \autoref{perm fig:classify-mnist}, \autoref{perm fig:classify-cifar10}, and \autoref{perm fig:classify-imagenet}, we show some example reconstructions that have been learnt by our PO-U model on $3 \times 3$ versions of the image datasets.
Because the quality of implicit CIFAR10 and ImageNet reconstructions are relatively poor, we also include \autoref{perm fig:classify-cifar10-2}, and \autoref{perm fig:classify-imagenet-2} on $2 \times 2$ versions.
Starting from a uniform assignment at the top, the figures show reconstructions as a permutation is being optimised.
The reconstructions here are clearly noisier than before due the supervision only being implicit.
This is evidence that while our method is superior to existing methods in terms of reconstruction error and accuracy of the classification, there is still plenty of room for improvement to allow for better implicitly learned permutations.
Keep in mind that it is not necessary for the permutation to produce the original image exactly, as long as the CNN can consistently recognise what the permutation method has learned.
Our models tend to naturally learn reconstructions that are more similar to the original image than the LinAssign model.

\subsection{Visual question answering}\label{perm sec:vqa}
As the last task, we consider the much more complex problem of visual question answering (VQA): answering questions about images.
We use the VQA v2 dataset~\citep{Antol2015,Goyal2017}, which in total contains around 1 million questions about 200,000 images from MS-COCO with 6.5 million human-provided answers available for training.
We use bottom-up attention features~\citep{Anderson2018} as representation for objects in the image, which for each image gives us a \emph{set} (size varying from 10 to 100 per image) of bounding boxes and the associated feature vector that encodes the contents of the bounding box.
These object proposals have no natural ordering a priori.

\begin{figure}
    \centering
    \centerfloat{\resizebox{\figurewidth}{!}{%
    \begin{tikzpicture}
        \node {\begin{tikzpicture}\pic {arch ban};\end{tikzpicture}};
    \end{tikzpicture}
    }}
    \caption[Network architecture for visual question answering]{Network architecture for visual question answering task using BAN with 8 glimpses (2 shown) as baseline. We add a shared PO-U module with an LSTM to process its output for each glimpse. The outputs of the BAN attention and the LSTM are added into the hidden state of the BAN network.}
    \label{perm fig:arch-ban}
\end{figure}
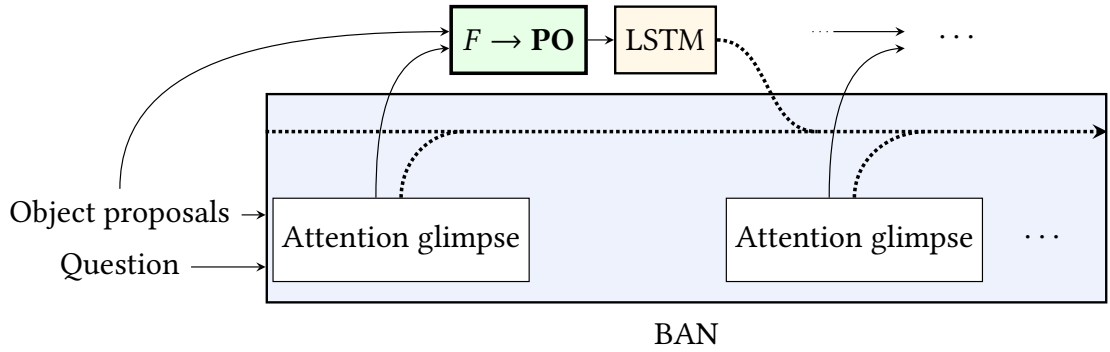

We use the state-of-the-art BAN model~\citep{Kim2018} as baseline and perform a straightforward modification to it to incorporate our module (see \autoref{perm fig:arch-ban}).
For each element in the set of object proposals, we concatenate the bounding box coordinates, features, and the attention value that the baseline model generates.
Our model learns to permute this set into a sequence, which is fed into an LSTM.
We take the last cell state of the LSTM to be the representation of the set, which is fed back into the baseline model.
This is done for each of the eight attention glimpses in the BAN model.
We include another baseline model (BAN + LSTM) that skips the permutation learning, directly processing the set with the LSTM.

\begin{table}
    \setlength\tabcolsep{7pt}
    \centering
    \caption[Accuracy on VQA v2 validation set]{Accuracy on VQA v2 validation set, mean of 10 runs. The sample standard deviation over these runs is shown after the $\pm$ symbol. Overall includes the other three question categories.}

    \label{perm tab:vqa}
    \centerfloat{\begin{tabular}{c c @{\hskip 0.6cm} c c c}
\toprule
Model & Overall & Yes/No & Number & Other \\
\midrule
BAN & 65.96 $\pm$ 0.16 & 83.34 $\pm$ 0.09 & 49.24 $\pm$ 0.56 & 57.17 $\pm$ 0.14\\
BAN + LSTM & 66.06 $\pm$ 0.13 & 83.29 $\pm$ 0.13 & 49.64 $\pm$ 0.37 & 57.30 $\pm$ 0.13\\
BAN + PO-U & \textbf{66.33} $\pm$ 0.09 & \textbf{83.50} $\pm$ 0.10 & \textbf{50.42} $\pm$ 0.46 & \textbf{57.48} $\pm$ 0.10 \\
\bottomrule
    \end{tabular}}
\end{table}

Our results on the validation set of VQA v2 are shown in \autoref{perm tab:vqa}.
We improve on the overall performance of the state-of-the-art model by 0.37\% -- a significant improvement for this dataset -- with 0.27\% of this improvement coming from the learned permutation.
This shows that there is a substantial benefit to learning an appropriate permutation through our model in order to learn better set representations.
Our model significantly improves on the number category, despite the inclusion of our counting module from \autoref{cha:vqa} specifically targeted at number questions in the baseline.
This is evidence that the representation learned through the permutation is non-trivial.
Note that the improvement through our model is not simply due to increased model size and computation: \citet{Kim2018} found that significantly increasing BAN model size, increasing computation time similar in scale to including our model, does not yield any further gains.

\section{Analysis of learned comparisons}\label{perm app:analysis}

\subsection{Number sorting}\label{perm sec:analysis-sorting}
\begin{figure}
    \centering
    \centerfloat{\resizebox{\textwidth}{!}{%
    \includegraphics[width=0.415\linewidth]{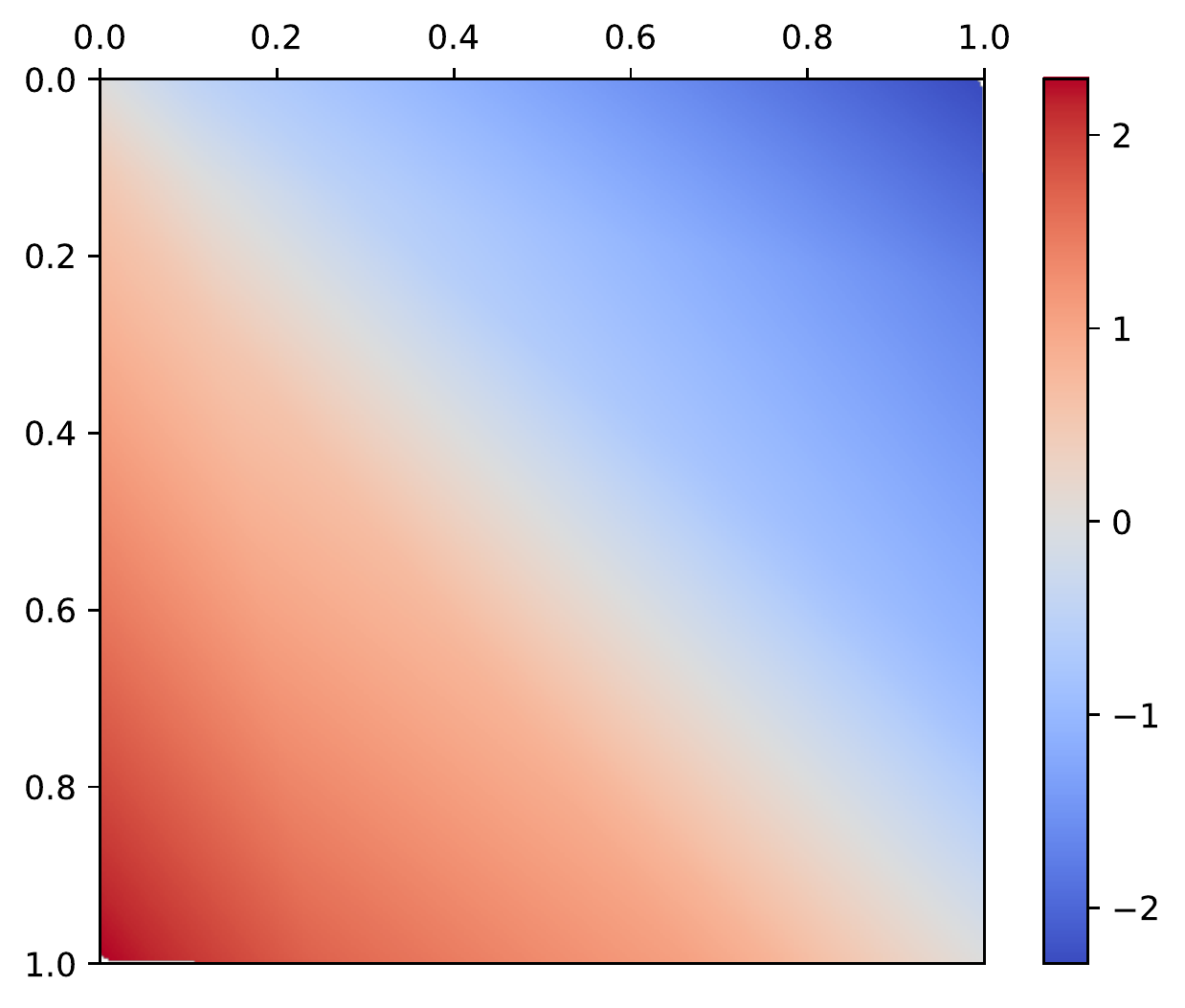}
    \includegraphics[width=0.45\linewidth]{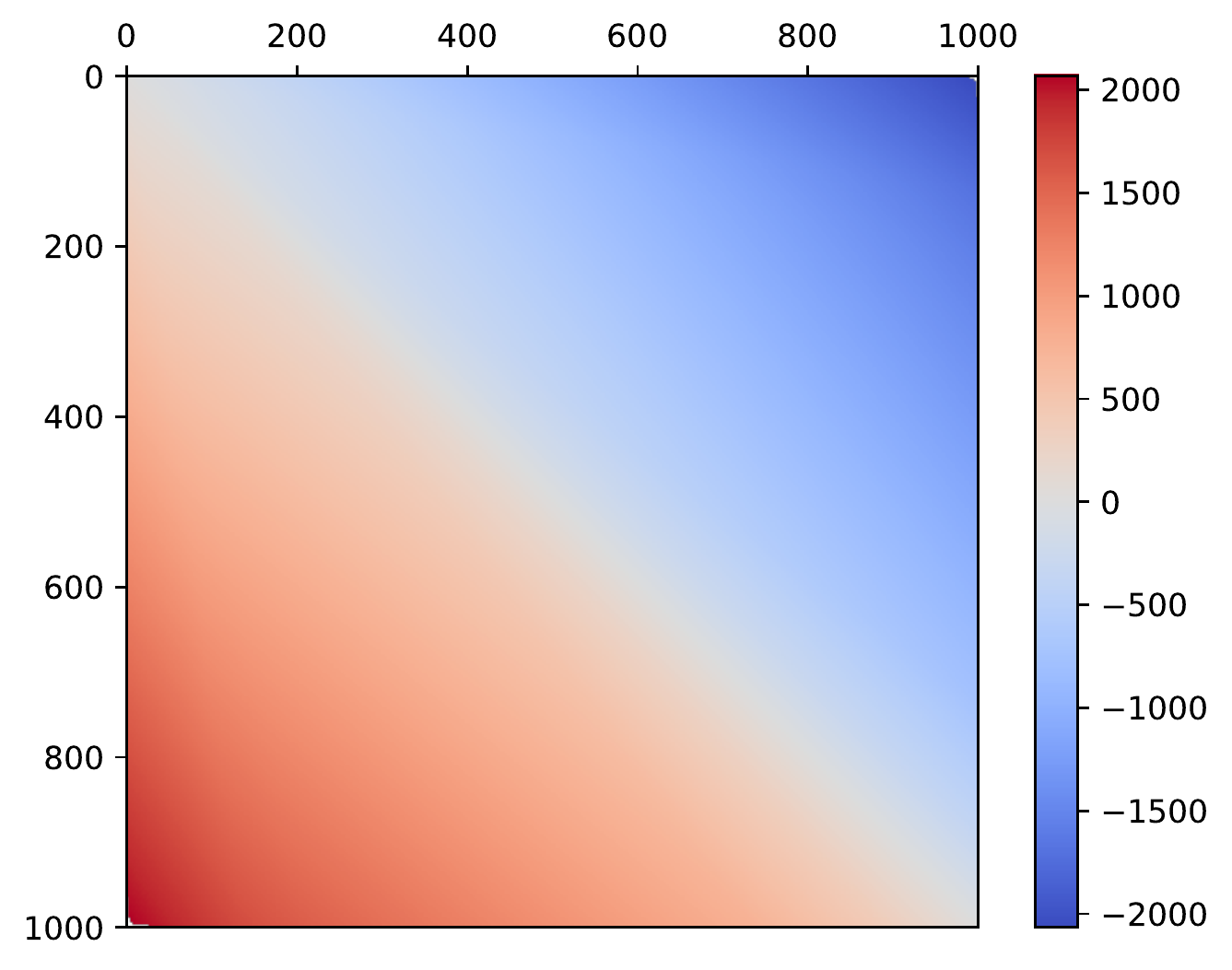}
    }}
    \caption[Outputs of $F$ for different pairs of numbers as input]{Outputs of $F$ for different pairs of numbers as input. Red indicates that the number on the left should be ordered after the number at the top, blue indicates the opposite. Evaluation intervals are $[0, 1]$ (left) and $[0, 1000]$ (right).}
    \label{perm fig:F}
\end{figure}

First, we investigate what comparison function $F$ is learned for the number sorting task.
We start with plotting the outputs of $F$ for different pairs of inputs in \autoref{perm fig:F}.
From this, we can see that it learns a sensible comparison function where it outputs a negative number when the first argument is lower than the second, and a positive number vice versa.

\begin{figure}
    \centering
    \centerfloat{\resizebox{\textwidth}{!}{%
    \includegraphics[width=0.425\linewidth]{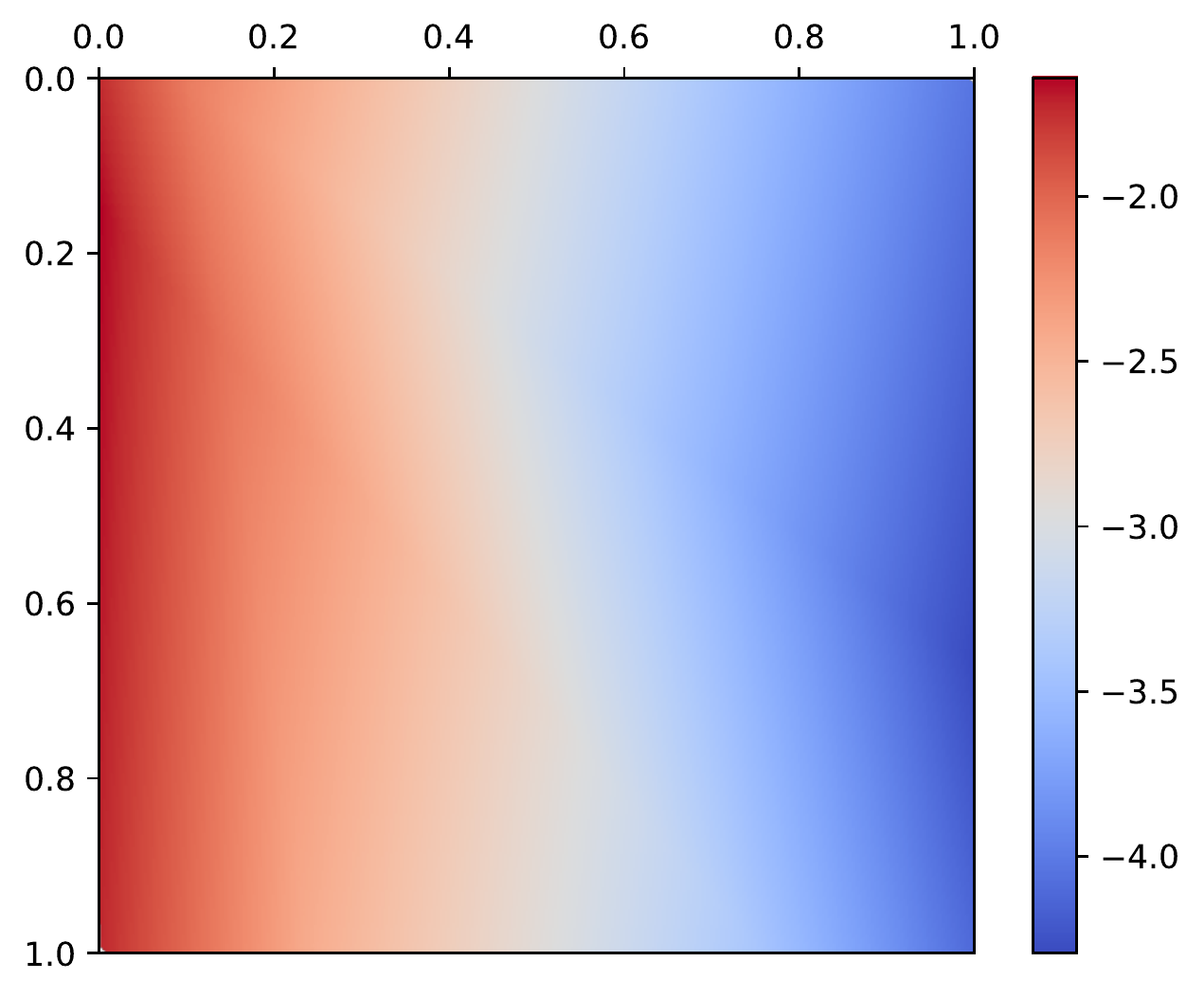}
    \includegraphics[width=0.45\linewidth]{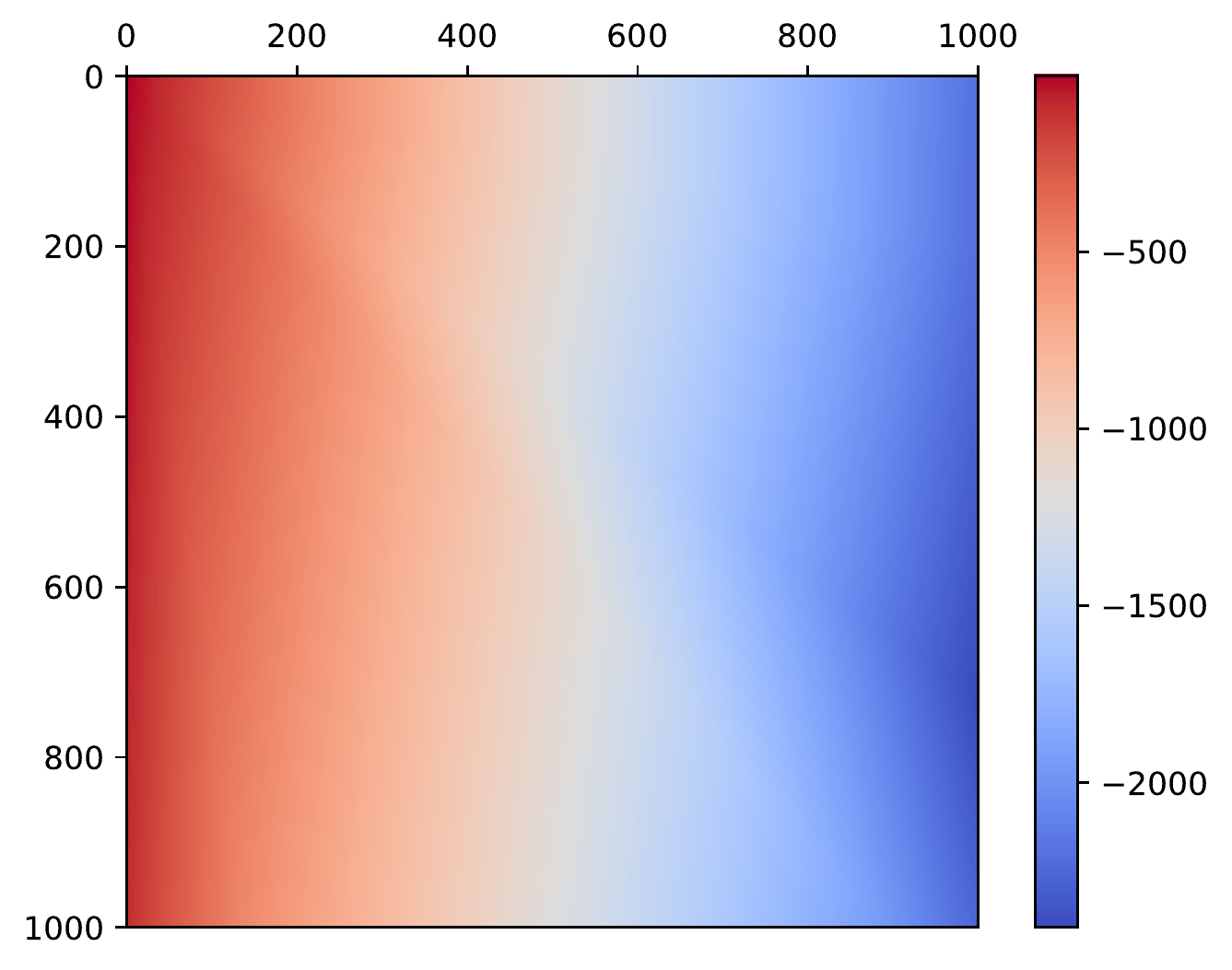}
    }}
    \caption[Outputs of $f$ for different pairs of numbers as input]{Outputs of $f$ for different pairs of numbers as input. Evaluation intervals are $[0, 1]$ (left) and $[0, 1000]$ (right).}
    \label{perm fig:f}
\end{figure}

The easiest way to achieve this is to learn $f(x_i, x_j) = x_i$, which results in $F(x_i, x_j) = x_i - x_j$.
By plotting the outputs of the learned $f$ in \autoref{perm fig:f} we can see that something close to this has indeed been learned.
The learned $f$ mostly depends on the second argument and is a scaled and shifted version of it.
It has not learned to completely ignore the first argument, but the deviations from it are small enough that the cost function of the permutation is able to compensate for it.
We can see that there is a faint grey diagonal area going from $(0, 0)$ to $(1, 1)$ and to $(1000, 1000)$, which could be an artefact from $F$ having small gradients due to its skew-symmetry when two numbers are close to each other.

\subsection{Image mosaics}\label{perm sec:analysis-mosaic}
Next, we investigate the behaviour of $F$ on the image mosaic task.
Since our model uses the outputs of $F$ in the optimisation process, we find it easier to interpret $F$ over $f$ in the subsequent analysis.

\subsubsection{Outputs}
\begin{figure}
    \centering
    \centerfloat{\resizebox{\textwidth}{!}{%
        \begin{tikzpicture}
            \node at (-0.6, 3.5) {\includegraphics[width=0.24\linewidth]{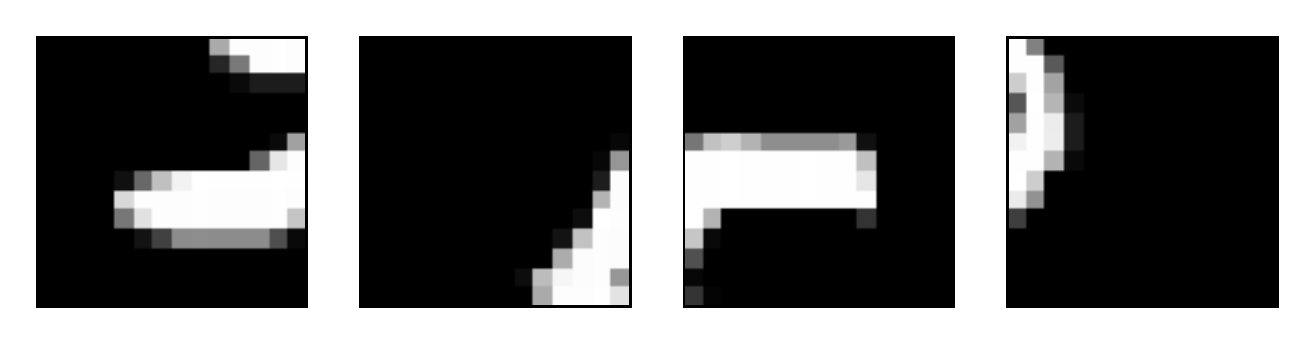}};
            \node at (-4, 0) {\includegraphics[width=0.068\linewidth]{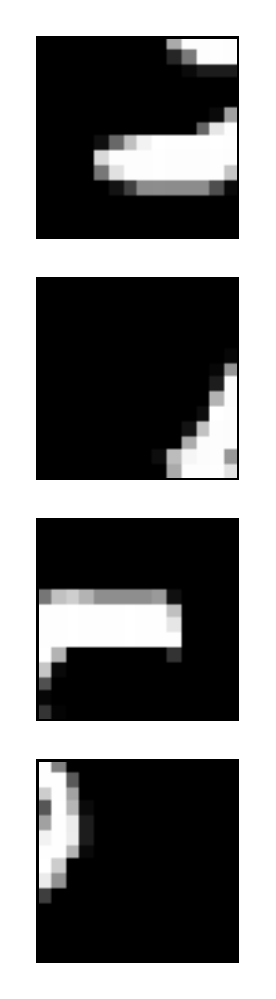}};
            \node at (0, 0) {\includegraphics[width=0.3\linewidth]{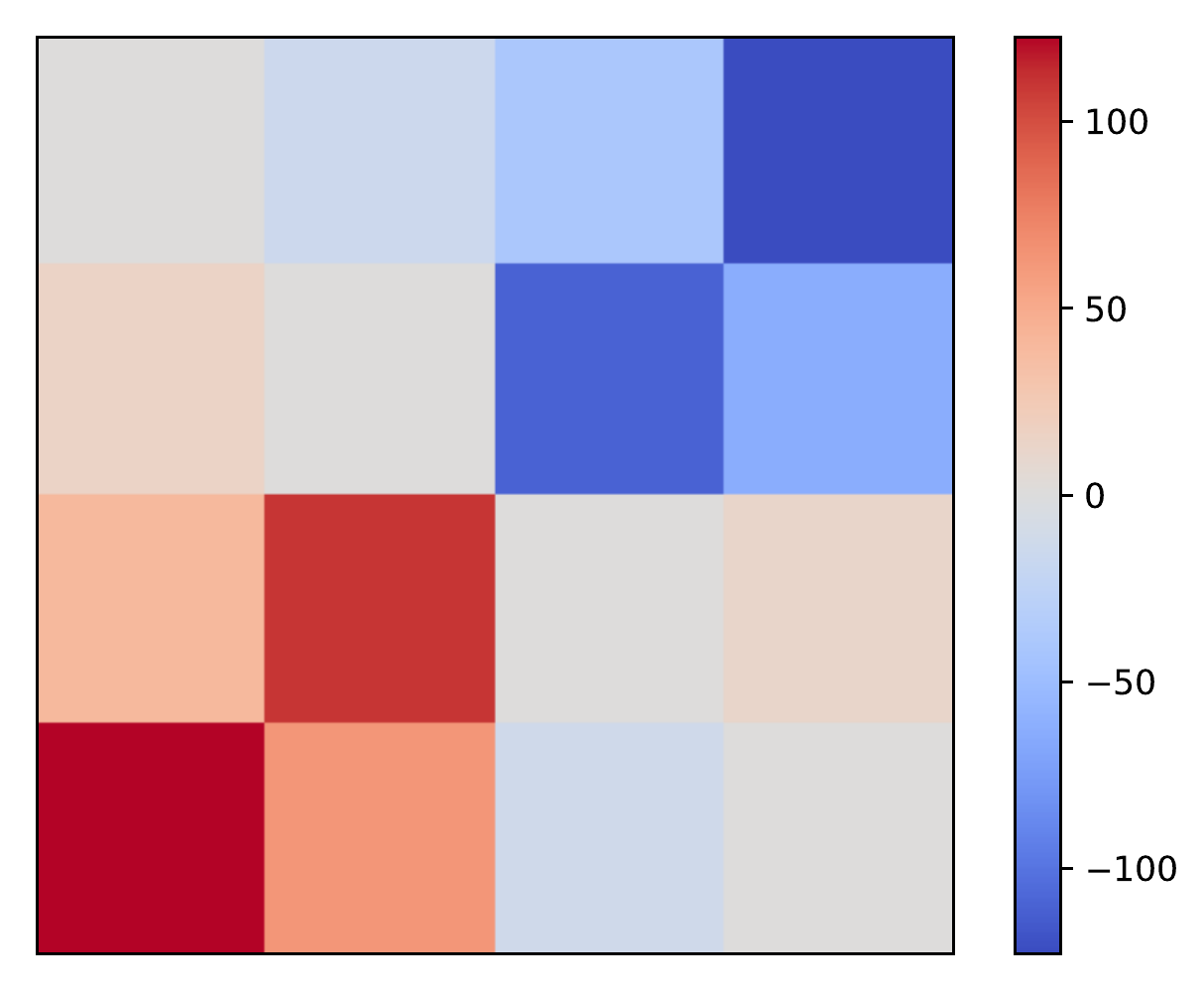}};
        \end{tikzpicture}
        \begin{tikzpicture}
            \node at (-0.6, 3.5) {\includegraphics[width=0.24\linewidth]{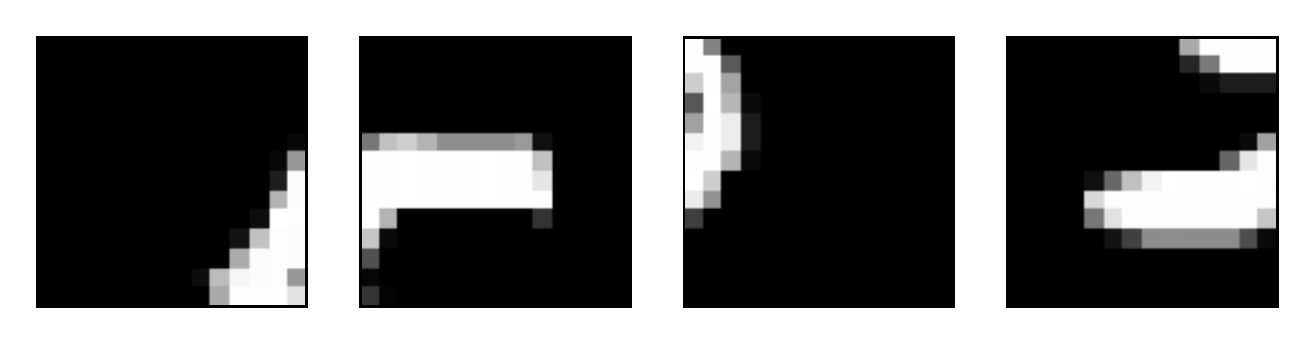}};
            \node at (-4, 0) {\includegraphics[width=0.068\linewidth]{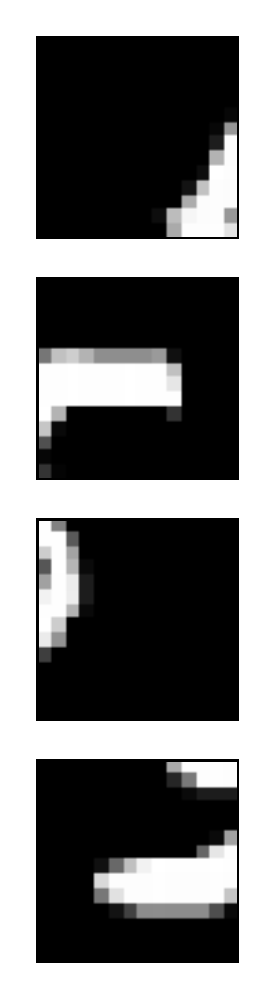}};
            \node at (0, 0) {\includegraphics[width=0.3\linewidth]{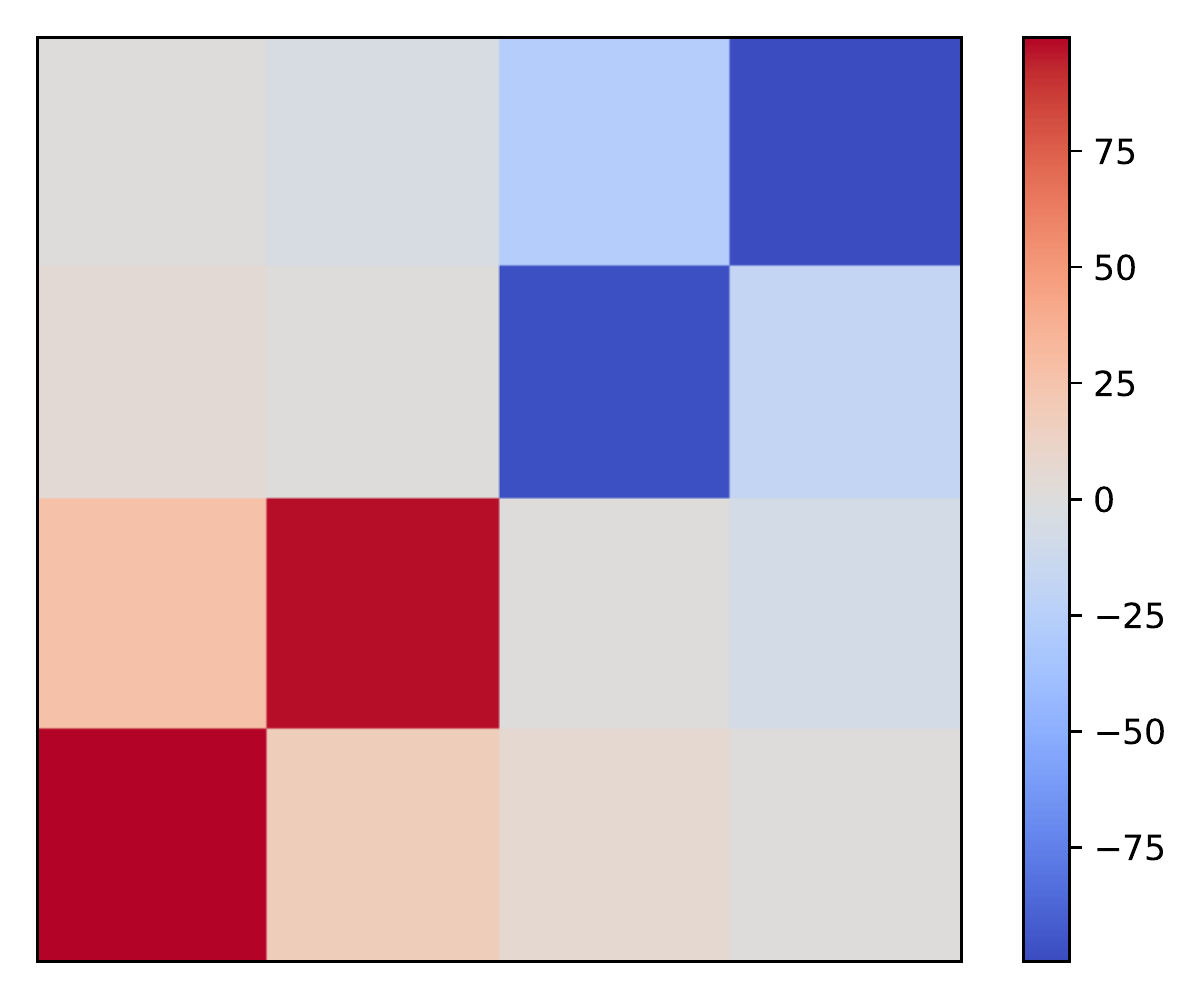}};
        \end{tikzpicture}
    }}
    \caption[Outputs of $F_1$ and $F_2$ for pairs of tiles from an image in MNIST]{Outputs of $F_1$ (left half, row comparisons) and $F_2$ (right half, column comparisons) for pairs of tiles from an image in MNIST. For $F_1$, the tiles are sorted left-to-right if only $F_1$ was used as cost. For $F_2$, the tiles are sorted top-to-bottom if only $F_2$ was used as cost. Blue indicates that the tile to the left of this entry should be ordered left of the tile at the top for $F_1$, the tile on the left should be ordered above the tile at the top for $F_2$. The opposite applies for red. The saturation of the colour indicates how strong this ordering is.}
    \label{perm fig:costs}
\end{figure}

We start by looking at the output of $F_1$ (costs for left-to-right ordering) and $F_2$ (costs for top-to-bottom ordering) for MNIST $2 \times 2$, shown in \autoref{perm fig:costs}.
First, there is a clear entry in each row and column of both $F_1$ and $F_2$ that has the highest absolute cost (high colour saturation) whenever the corresponding tiles fit together correctly.
This shows that it successfully learned to be confident what order two tiles should be in when they fit together.
From the two 2-by-2 blocks of red and blue on the anti-diagonal, we can also see that it has learned that for the per-row comparisons ($F_1$), the tiles that should go into the left column should generally compare to less than (i.e. should be permuted to be to the left of) the tiles that go to the right.
Similarly, for the per-column comparisons ($F_2$) tiles that should be at the top compare to less than tiles that should be at the bottom.
Lastly, $F_1$ has a low absolute cost when comparing two tiles that belong in the same column.
These are the entries in the matrix at the coordinates (1, 2), (2, 1), (4, 3), and (3, 4).
This makes sense, as $F_1$ is concerned with whether one tile should be to the left or right of another, so tiles that belong in the same column should not have a preference either way.
A similar thing applies to $F_2$ for tiles that belong in the same row.

\subsubsection{Sensitivity to positions}
\begin{figure}
    \centering
    \centerfloat{\resizebox{\figurewidth}{!}{%
    \includegraphics[width=0.27\figurewidth]{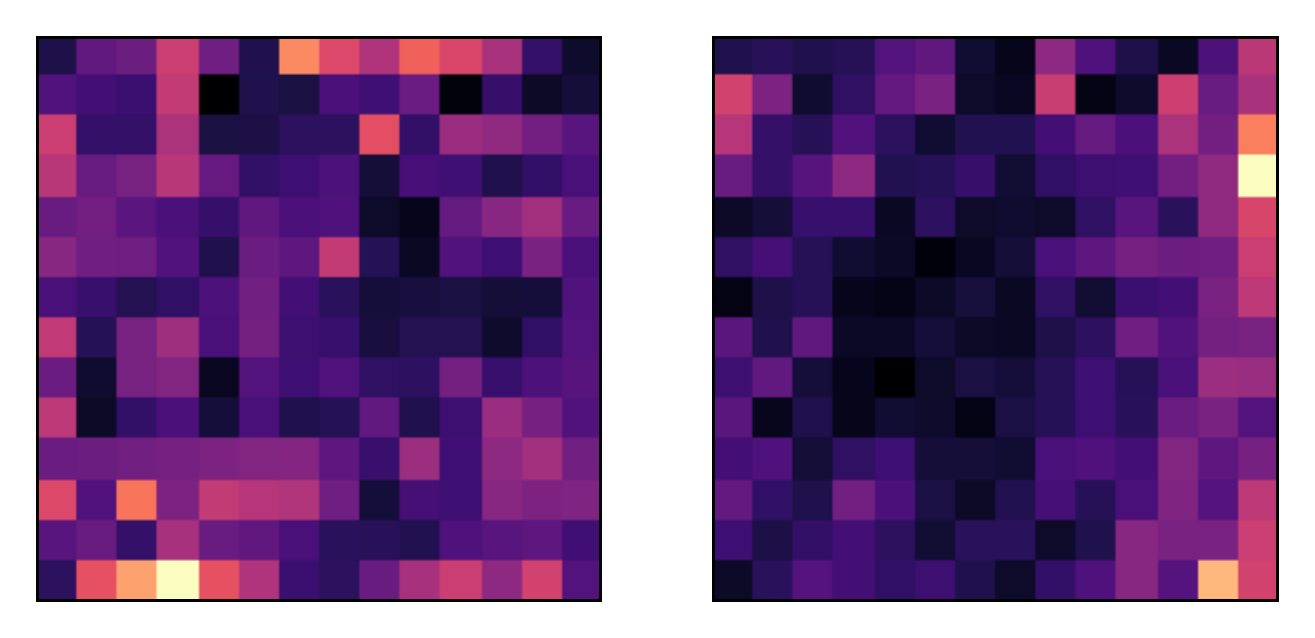}
    \hspace{5mm}
    \includegraphics[width=0.27\figurewidth]{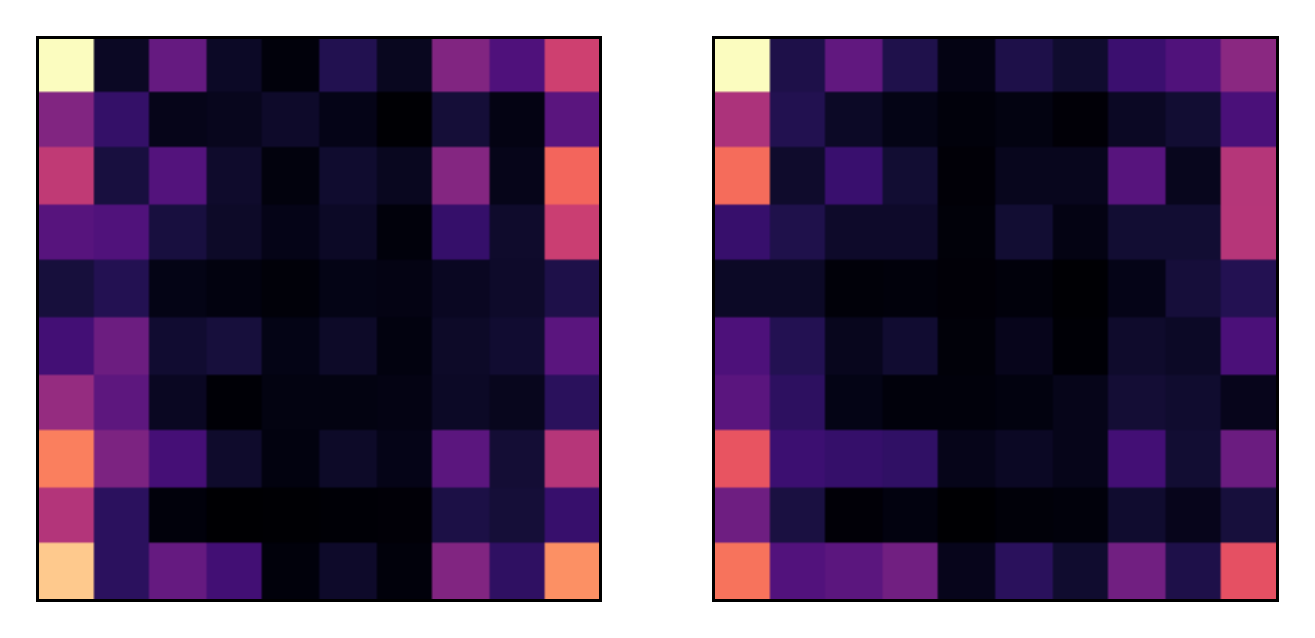}
    \hspace{5mm}
    \includegraphics[width=0.27\figurewidth]{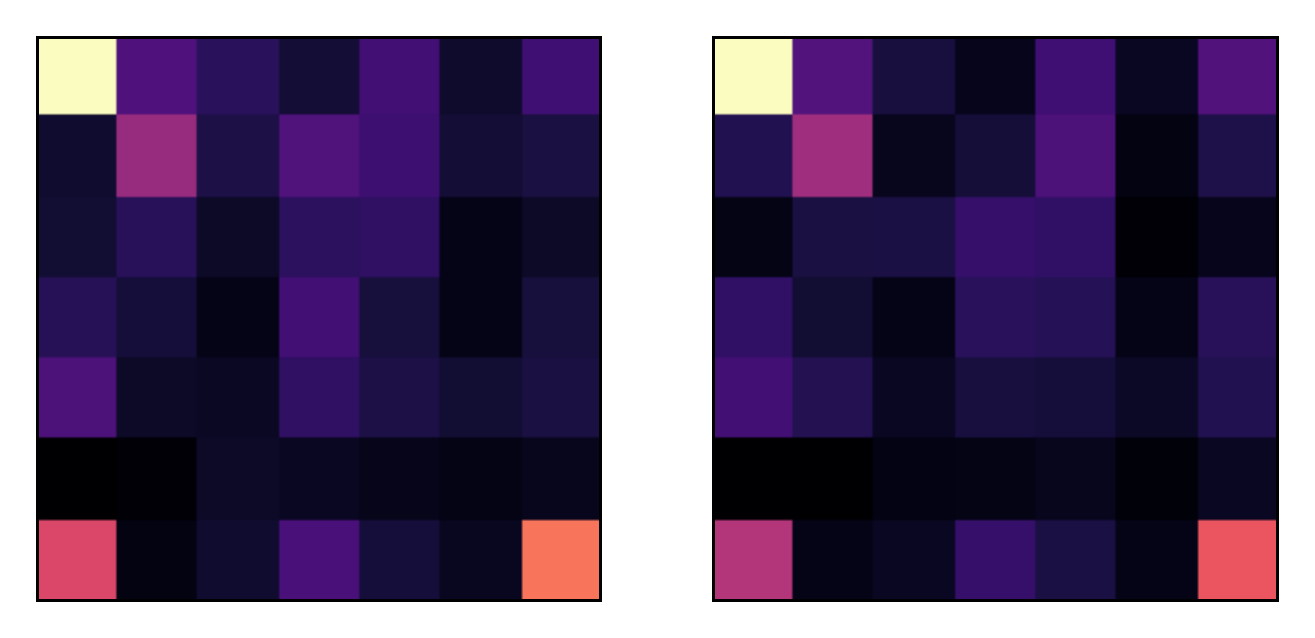}
    }}
    \caption[Sensitivity to positions within a tile for MNIST]{Sensitivity to positions within a tile for MNIST $2 \times 2$ (left), $3 \times 3$ (middle), and $4 \times 4$ (right). The left plot of each pair shows $F_1$, the right plot shows $F_2$.}
    \label{perm fig:sensitivity-mnist}
\end{figure}

\begin{figure}
    \centering
    \centerfloat{\resizebox{\figurewidth}{!}{%
    \includegraphics[width=0.27\figurewidth]{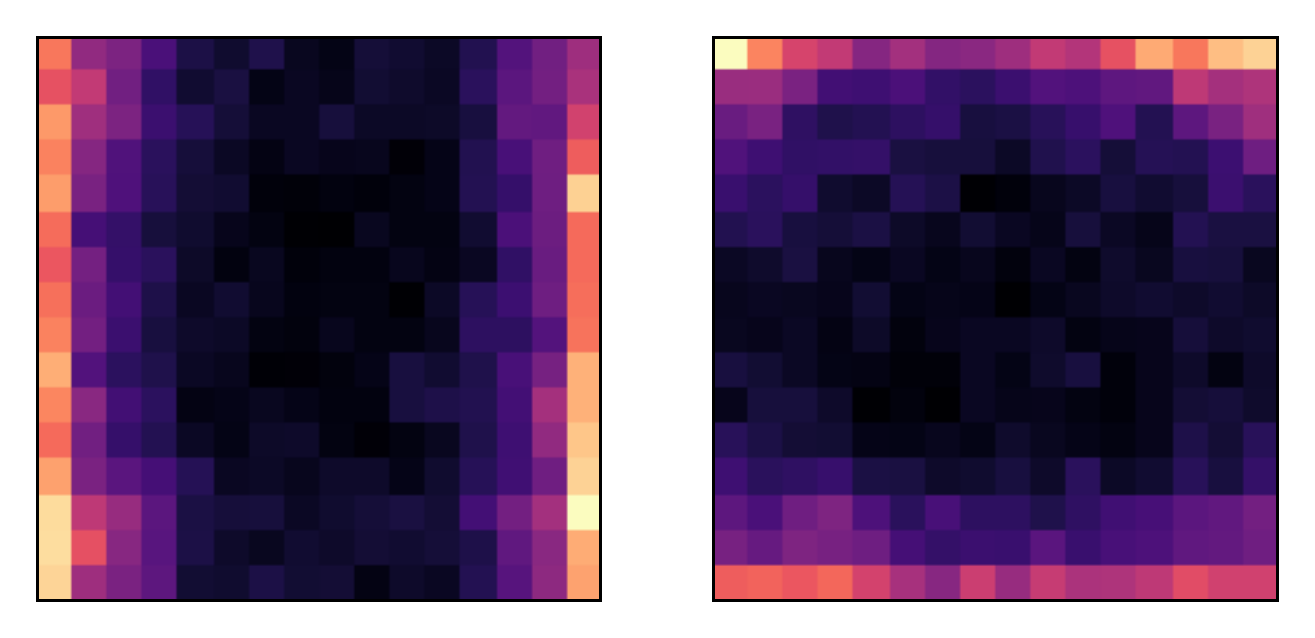}
    \hspace{5mm}
    \includegraphics[width=0.27\figurewidth]{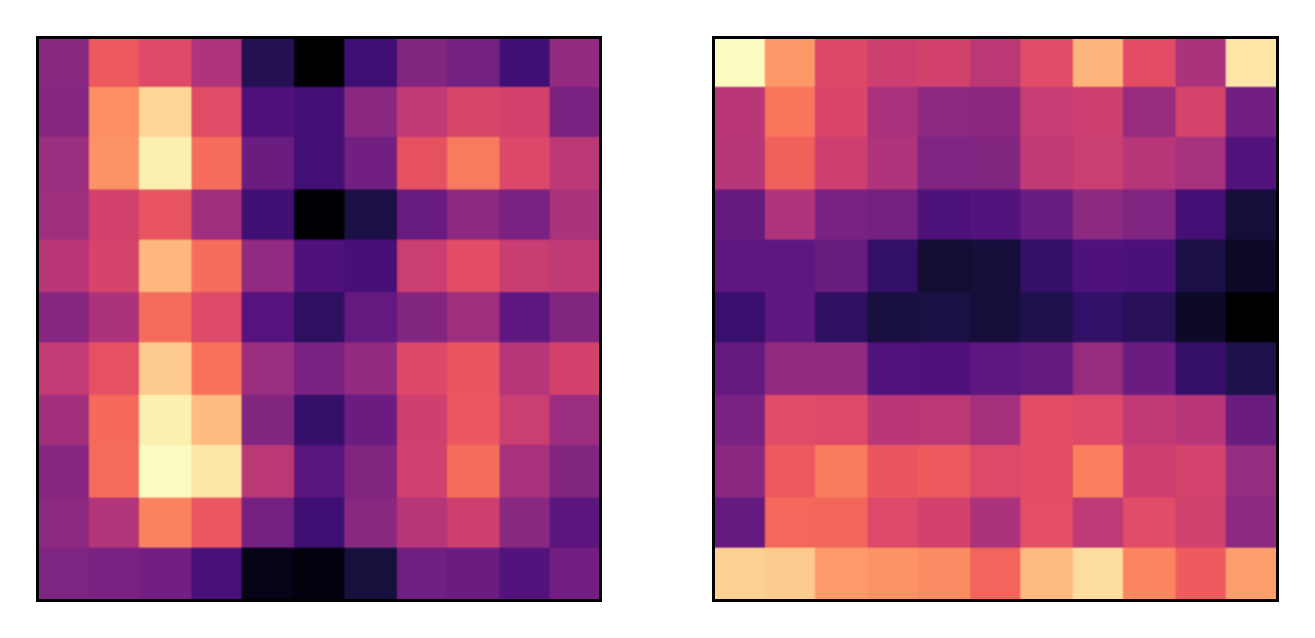}
    \hspace{5mm}
    \includegraphics[width=0.27\figurewidth]{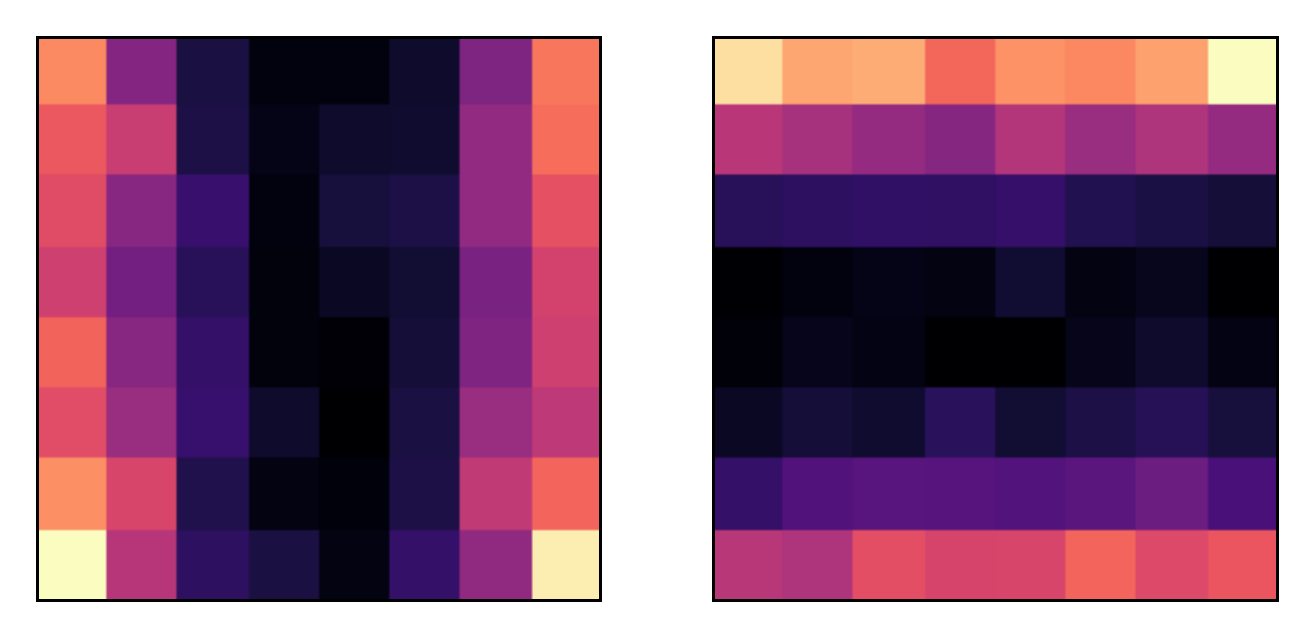}
    }}
    \caption[Sensitivity to positions within a tile for CIFAR10]{Sensitivity to positions within a tile of the comparisons for CIFAR10 $2 \times 2$ (left), $3 \times 3$ (middle), and $4 \times 4$ (right). The left plot of each pair shows $F_1$, the right plot shows $F_2$.}
    \label{perm fig:sensitivity-cifar10}
\end{figure}

Next, we investigate what positions within the tiles $F_1$ and $F_2$ are most sensitive to.
This illustrates what areas of the tiles are usually important for making comparisons.
We do this by computing the gradients of the absolute values of $F$ with respect to the input tiles and averaging over many inputs.
For MNIST $2 \times 2$ (\autoref{perm fig:sensitivity-mnist}, left), it learns no particular spatial pattern for $F_1$ and puts slightly more focus away from the centre of the tile for $F_2$.
As we will see later, it learns something that is very content-dependent rather than spatially-dependent.
With increasing numbers of tiles on MNIST, it tends to focus more on edges, and especially on corners.
For the CIFAR10 dataset (\autoref{perm fig:sensitivity-cifar10}), there is a much clearer distinction between left-right comparisons for $F_1$ and top-bottom comparisons for $F_2$.
For the $2 \times 2$ and $4 \times 4$ settings, it relies heavily on the pixels on the left and right borders for left-to-right comparisons, and top and bottom edges for top-to-bottom comparisons.
Interestingly, $F_1$ in the $3 \times 3$ setting (middle pair) on CIFAR10 focuses on the left and right halves of the tiles, but specifically avoids the borders.
A similar thing applies to $F_2$, where a greater significance is given to pixels closer to the middle of the image rather than only focusing on the edges.
This suggests that it learns to not only match up edges as with the other tile numbers, but also uses the content within the tile to do more sophisticated content-based comparisons.

\begin{figure}
    \centering
    \centerfloat{\resizebox{\figurewidth}{!}{%
    \includegraphics[width=0.495\figurewidth]{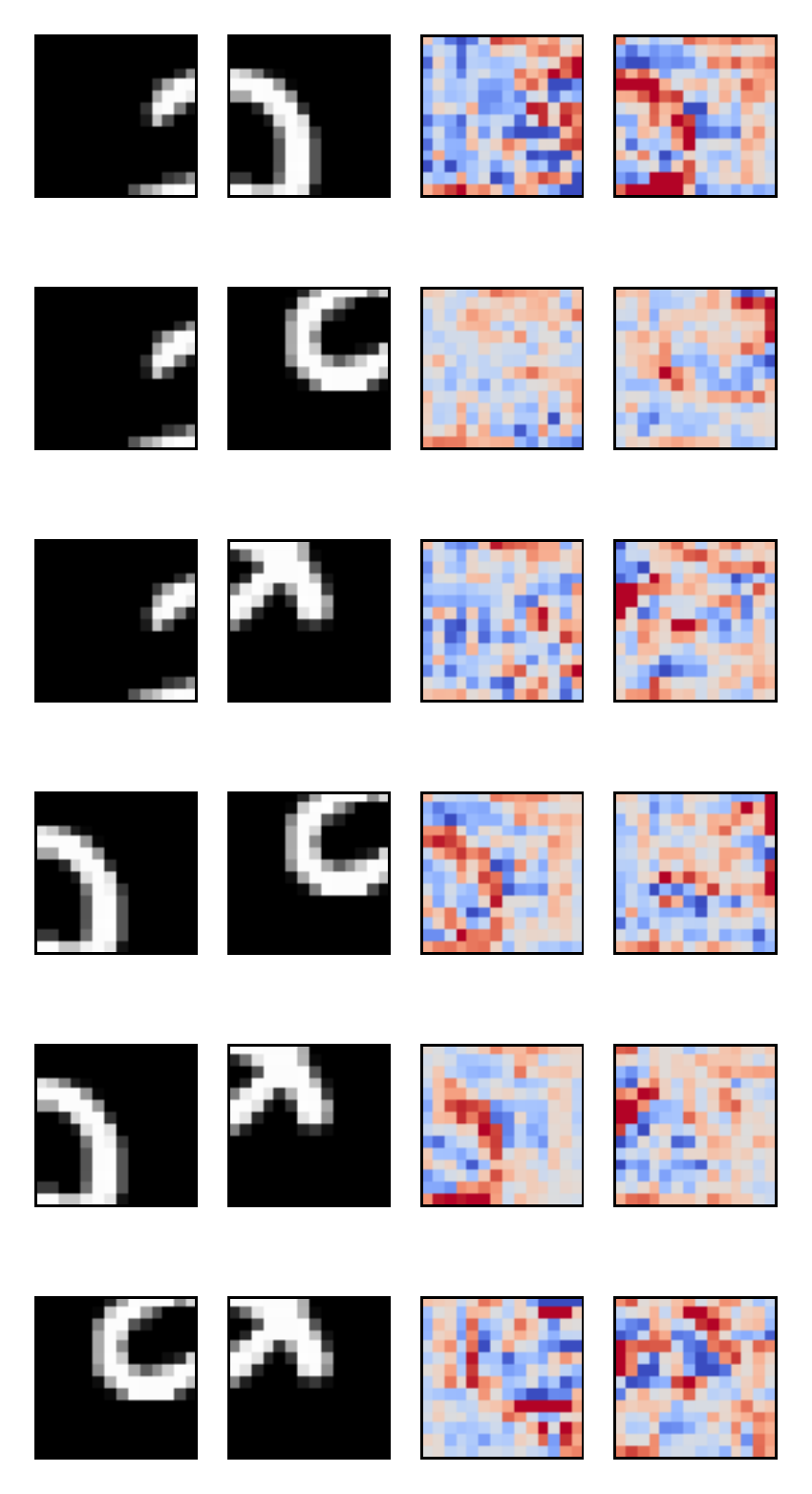}
    \hfill
    \includegraphics[width=0.495\figurewidth]{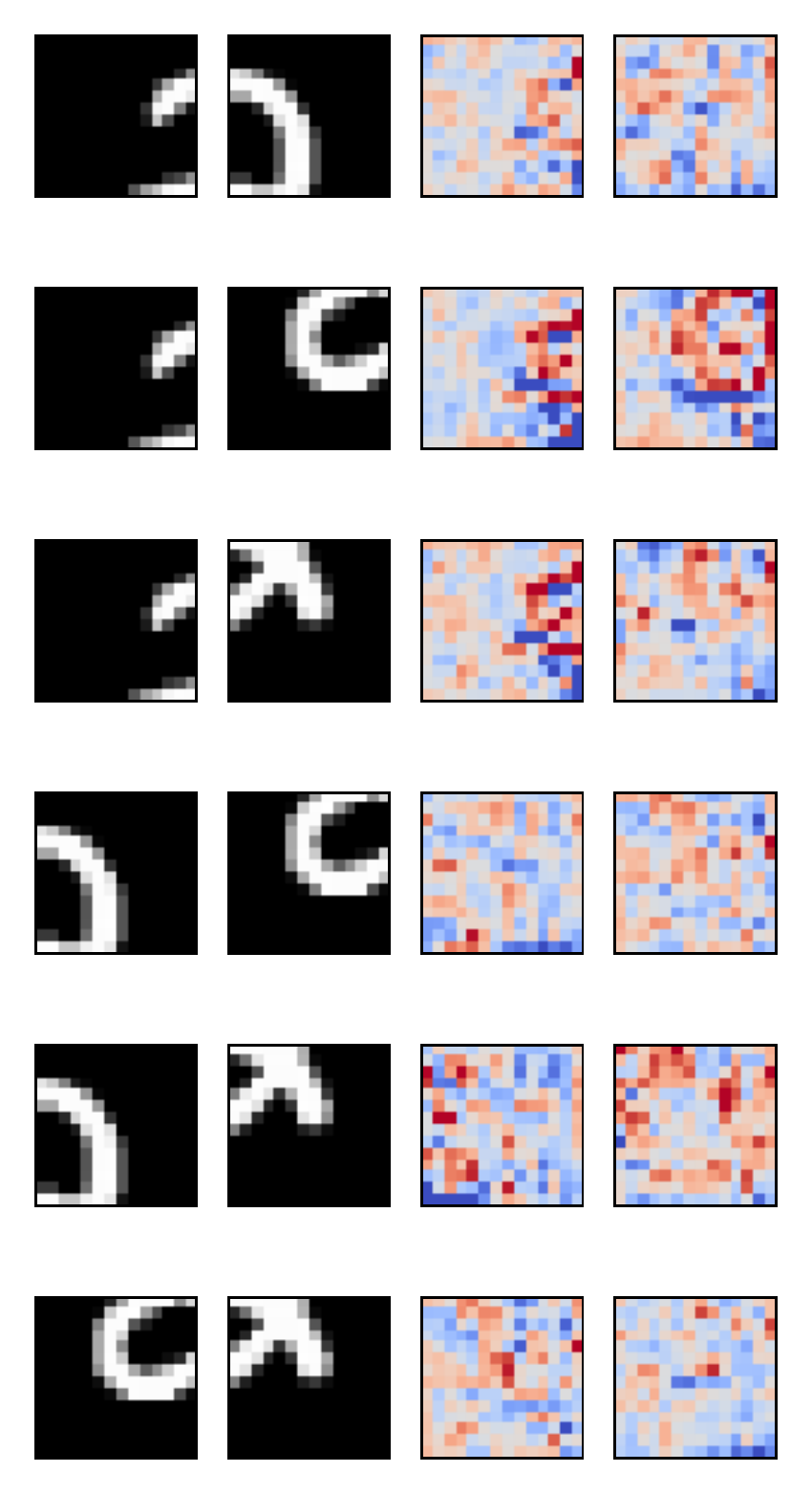}
    }}
    \caption[Gradient maps of pairs of tiles from MNIST]{Gradient maps of pairs of tiles from MNIST for $F_1$ (left half) and $F_2$ (right half). Each group of four consists of: tile 1, tile 2, gradient of $F(t_1, t_2)$ with respect to tile 1, gradient of $F(t_2, t_1)$ with respect to tile 2.}
    \label{perm fig:gradients-mnist}
\end{figure}

\begin{figure}
    \centering
    \centerfloat{\resizebox{\figurewidth}{!}{%
    \includegraphics[width=0.495\figurewidth]{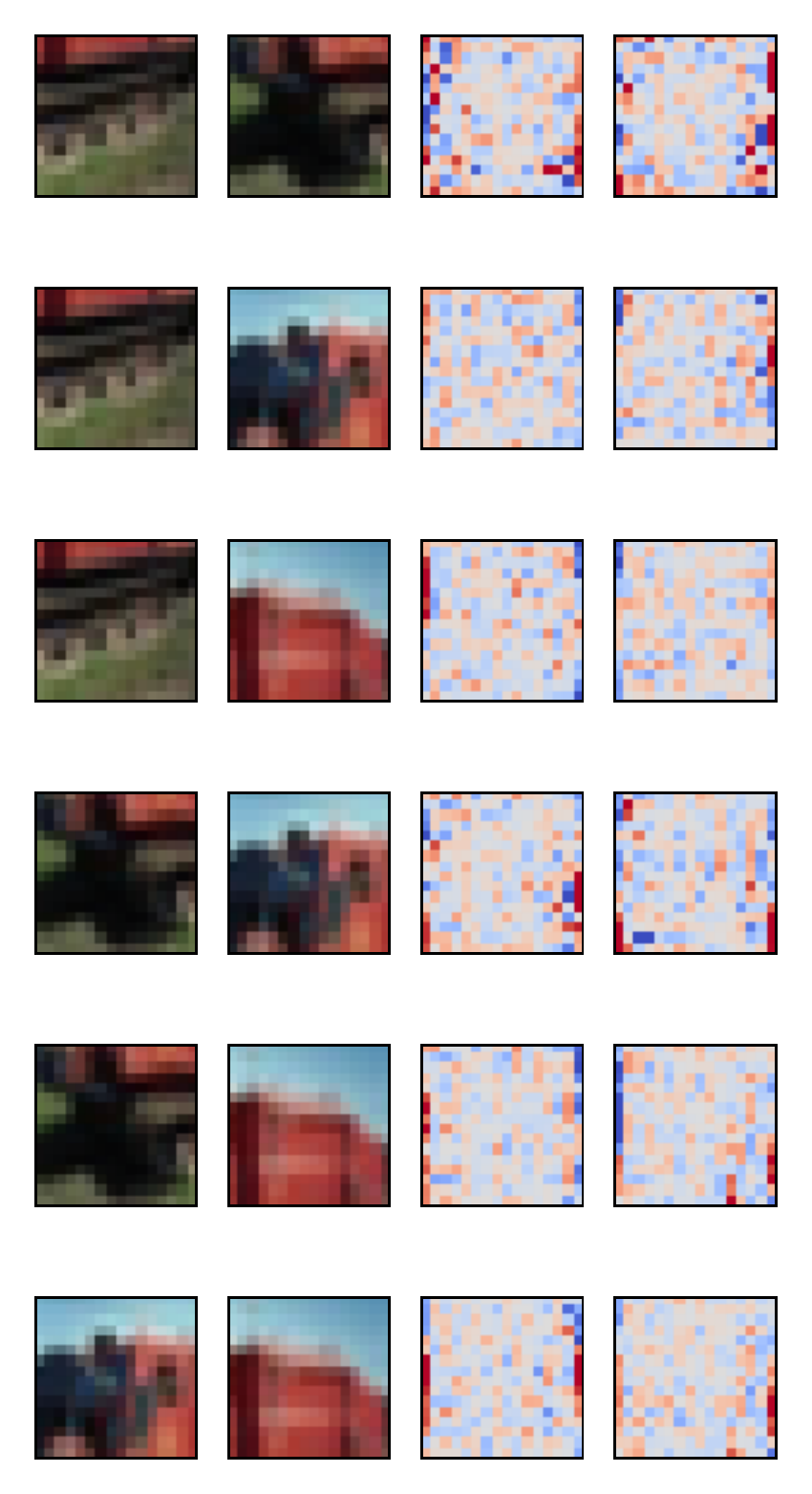}
    \hfill
    \includegraphics[width=0.495\figurewidth]{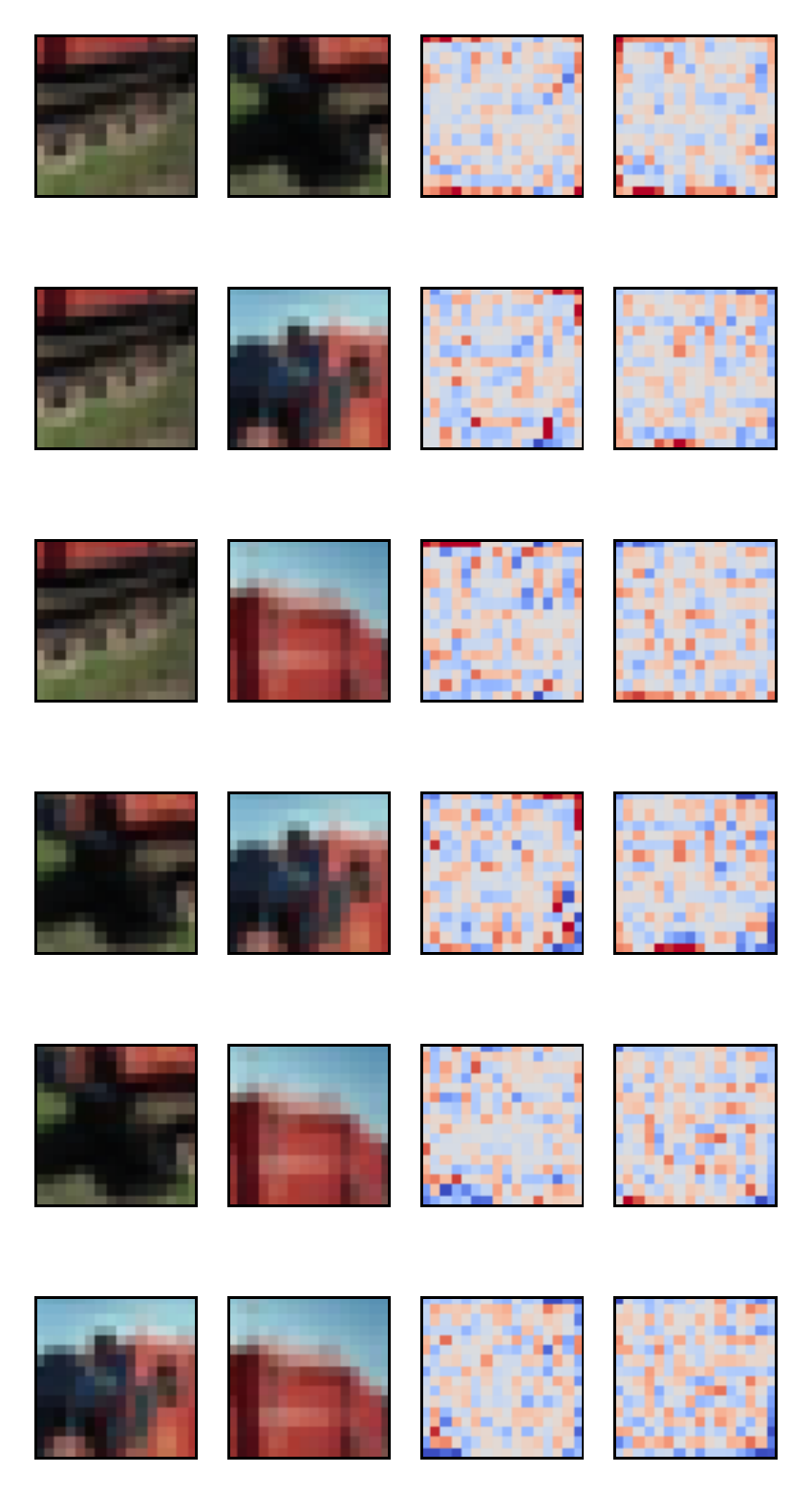}
    }}
    \caption[Gradient maps of pairs of tiles from CIFAR10]{Gradient maps of pairs of tiles from CIFAR10 for $F_1$ (left half) and $F_2$ (right half). Each group of four consists of: tile 1, tile 2, gradient of $F(t_1, t_2)$ with respect to tile 1, gradient of $F(t_2, t_1)$ with respect to tile 2.}
    \label{perm fig:gradients-cifar10}
\end{figure}

\subsubsection{Per-tile gradients}
Lastly, we can look at the gradients of $F$ with respect to the input tiles for specific pairs of tiles, shown in \autoref{perm fig:gradients-mnist} and \autoref{perm fig:gradients-cifar10}.
This gives us a better insight into what changes to the input tiles would affect the cost of the comparison the most.
These figures can be understood as follows: for each pair of tiles, we have the corresponding two gradient maps next to them.
Brightening the pixels for the blue entries in these gradient maps would order the corresponding tile more strongly towards the left for $F_1$ and towards the top for $F_2$.
The opposite applies to brightening the pixels with red entries.
Vice versa, darkening pixels with blue entries orders the tile more strongly towards the right for $F_1$ and the bottom for $F_2$.
More saturated colours in the gradient maps correspond to greater effects on the cost when changing those pixels.

We start with gradients on the tiles for an input showing the digit 2 on MNIST $2 \times 2$ in \autoref{perm fig:gradients-mnist}.
We focus on the first row, left side, which shows a particular pair of tiles from this image and their gradients of $F_1$ (left-to-right ordering), and we share some of our observations here:

\begin{itemize}
    \item The gradients of the second tile show that to encourage the permutation to place it to the right of the first tile, it is best to increase the brightness of the curve in tile 2 that is already white (red entries in tile 2 gradient map) and decrease the black pixels around it (blue entries).
    This means that it recognised that this type of curve is important in determining that it should be placed to the right, perhaps because it matches up with the start of the curve from tile 1.
    We can imagine the curve in the gradient map of tile 2 roughly forming part of a 7 rather than a 2 as well, so it is not necessarily looking for the curve of a 2 specifically.
    \item In the gradient map of the first tile, we can see that to encourage it to be placed to the left of tile 2, increasing the blue entries would form a curve that would make the first tile look like part of an 8 rather than a 2, completing the other half of the curve from tile 2.
    This means that it has learned that to match something with the shape in tile 2, a loop that completes it is best, but the partial loop that we have in tile 1 satisfies part of this too.
    \item Notice how the gradient of tile 1 changes quite a bit when going from row 1 to row 3, where it is paired up with different tiles.
    This suggests that the comparison has learned something about the specific comparison between tiles being made, rather than learning a general trend of where the tile should go.
    The latter is what a linear assignment model is limited to doing because it does not model pairwise interactions.
    \item In the third row, we can see that even though the two tiles do not match up, there is a red blob on the left side of the tile 2 gradient map.
    This blob would connect to the top part of the line in tile 1, so it makes sense that making the two tiles match up more on the border would encourage tile 2 to be ordered to the right of tile 1.
\end{itemize}

Similar observations apply to the right half of \autoref{perm fig:gradients-mnist}, such as row 5, where tile 1 (which should go above tile 2) should have its pixels in the bottom left increased and tile 2 should have its pixels in the top left increased in order for tile 1 to be ordered before (i.e. above) tile 2 more strongly.

On CIFAR10 $2 \times 2$ in \autoref{perm fig:gradients-cifar10}, it is enough to focus on the borders of the tiles.
Here, it is striking how specifically it tries to match edge colours between tiles.
For example, consider the blue sky in the left half ($F_1$), row 6.
To order tile 1 to the left of tile 2, we should change tile 1 to have brighter sky and darker red on the right border, and also darken the black on the left border so that it matches up less well with the right border of tile 2, where more of the bright sky is visible.
For tile 2, the gradient shows that it should also match up more on the left border, and have increase the amount of bright pixels, i.e. sky, on the right border, again so that it matches up less well with the left border of tile 1 if they were to be ordered the opposing way.

\section{Discussion}\label{perm sec:discussion}
In this chapter, we proposed our Permutation-optimisation module for learning permutations of sets using an optimisation-based approach.
In various experiments, we verified the merit of our approach for learning permutations and, from them, set representations.
We think that the optimisation-based approach to processing sets is currently underappreciated and hope that the techniques and results in this chapter will inspire new algorithms for processing sets in a permutation-invariant manner.
Of course, there is plenty of work to be done.
For example, we have only explored one possible function for the total cost; different functions capturing different properties may be used.
The main drawback of our approach is the cubic time complexity in the set size compared to the quadratic complexity of \citet{Mena2018}, which limits our model to tasks where the number of elements is relatively small.
While this is acceptable on the real-world dataset that we used -- VQA with up to 100 object proposals per image -- with only a 30\% increase in computation time, our method does not scale to the much larger set sizes encountered in domains such as point cloud classification.
Improvements in the optimisation algorithm may improve this situation, perhaps through a divide-and-conquer approach.

We believe that going beyond tensors as basic data structures is important for enabling higher-level reasoning.
As a fundamental mathematical object, sets are a natural step forward from tensors for modelling unordered collections.
The property of permutation invariance lends itself to greater abstraction by allowing data that has no obvious ordering to be processed, and we took a step towards this by learning an ordering that existing neural networks are able to take advantage of.

\chapter{Set auto-encoder: Featurewise sort pooling}\label{cha:fspool}
\chaptermark{Featurewise sort pooling}

In this chapter, we will present a multitude of contributions to the set neural network literature.
We identify the responsibility problem in existing set prediction models.
We then develop a set encoder that is simpler and faster than our model in \autoref{cha:perm-optim}.
It is based on a similar idea of turning the set into an ordered representation;
we use numerical sorting instead of trying to learn the ordering.
To avoid the responsibility problem, we develop a set decoder that is paired with our set encoder.

These contributions have been presented as~\citep{Zhang2019FSPool} at the Sets \& Partitions workshop, hosted at the Neural Information Processing Systems (NeurIPS) 2019 conference, and have been submitted to the International Conference on Learning Representations (ICLR) 2020.

\section{Introduction}\label{fspool sec:intro}
Consider the following task: you have a dataset wherein each data point is a \emph{set} of 2-d points that form the vertices of a regular polygon, and the goal is to learn an auto-encoder on this dataset.
The only variable is the rotation of this polygon around the origin, with the number of points, size, and centre of it fixed.
Because the inputs and outputs are sets, this problem has some unique challenges.

\paragraph{Encoder:}
This turns the set of points into a latent space.
The order of the elements in the set is irrelevant, so the feature vector the encoder produces should be invariant to permutations of the elements in the set.
While there has been recent progress on learning such functions~\citep{Zaheer2017, Qi2017}, they compress a set of any size down to a single feature vector in one step.
This can be a significant bottleneck in what these functions can represent efficiently, particularly when relations between elements of the set need to be modeled~\citep{Murphy2019, Zhang2019PermOptim}.

\paragraph{Decoder:}
This turns the latent space back into a set.
The elements in the target set have an arbitrary order, so a standard reconstruction loss cannot be used na\"ively -- the decoder would have to somehow output the elements in the same arbitrary order.
Methods like those in \citet{Achlioptas2018} therefore use an assignment mechanism to match up elements (\autoref{fspool sec:background}), after which a usual reconstruction loss can be computed.
Surprisingly, their model is still unable to solve the polygon reconstruction task with close-to-zero reconstruction error, despite the apparent simplicity of the dataset.


In this chapter, we introduce a set pooling method for neural networks that addresses both the encoding bottleneck issue and the decoding failure issue.
We make the following contributions:

\begin{enumerate}
    \item
        We identify the \emph{responsibility problem} (\autoref{fspool sec:problem}).
        This is a fundamental issue with existing set prediction models that has not been considered in the literature before, explaining why these models struggle to model even the simple polygon dataset.

    \item
    We introduce \textsc{FSPool}:
    a differentiable, sorting-based pooling method for variable-size sets (\autoref{fspool sec:fsort}).
    By using our pooling in the encoder of a set auto-encoder and \emph{inverting the sorting} in the decoder, we can train it with the usual MSE loss for reconstruction \emph{without} the need for an assignment-based loss.
    This avoids the responsibility problem.

    \item
        We show that our auto-encoder can learn polygon reconstructions with close-to-zero error, which is not possible with existing set auto-encoders (\autoref{fspool sec:polygons}).
        This benefit transfers over to a set version of MNIST, where the quality of reconstruction and learned representation is improved (\autoref{fspool sec:mnist}).
        In further classification experiments on CLEVR (\autoref{fspool sec:clevr}) and several graph classification datasets (\autoref{fspool sec:graph}), using FSPool in a set encoder improves over many non-trivial baselines.
\end{enumerate}

\section{Background}\label{fspool sec:background}
The problem with predicting sets is that the output order of the elements is arbitrary, so computing an elementwise mean squared error does not make sense; there is no guarantee that the elements in the target set happen to be in the same order as they were generated.
The existing solution around this problem is an assignment-based loss, which assigns each predicted element to its ``closest'' neighbour in the target set first, after which a traditional pairwise loss can be computed.

We have a predicted set $\mhY$ with feature vectors as elements and a ground-truth set $\mY$, and we want to measure how different the two sets are.
These sets can be represented as matrices with the feature vectors placed in the columns in some arbitrary order, so $\mhY = [\vhy_1, \ldots, \vhy_n]$ and $\mY = [\vy_1, \ldots, \vy_n]$ with $n$ as the set size (columns) and $d$ as the number of features per element (rows).
In this work, we assume that these two sets have the same size.
The usual way to produce $\mhY$ is with a multi-layer perceptron (MLP) that has $d \times n$ outputs.

\paragraph{Hungarian loss}
One way to do this assignment is to find a linear assignment that minimises the total loss, which can be solved with the Hungarian algorithm in $O(n^3)$ time.
With $\Pi$ as the space of all $n$-length permutations:

\begin{equation}
    \mathcal{L}_H(\mhY, \mY) = \min_{\pi \in \Pi} \sum_i^n \norm{ \vhy_i - \vy_{\pi(i)} }^2
\end{equation}

\paragraph{Chamfer loss}
Alternatively, we can assign each element directly to the closest element in the target set.
To ensure that all points in the target set are covered, a term is added to the loss wherein each element in the target set is also assigned to the closest element in the predicted set.
This has $O(n^2)$ time complexity and can be run efficiently on GPUs.

\begin{equation}
    \mathcal{L}_{C}(\mhY, \mY) = \sum_i \min_{j} \norm{ \vhy_i - \vy_j }^2 + \sum_j \min_i \norm{ \vhy_i - \vy_j }^2
\end{equation}

Both of these losses are examples of permutation-invariant functions:
the loss is the same regardless of how the columns of $\mY$ and $\mhY$ are permuted.

\section{Responsibility problem}\label{fspool sec:problem}

\begin{figure}
    \centering
    \centerfloat{\resizebox{\linewidth}{!}{%
        \begin{tikzpicture}
            \node {\begin{tikzpicture}\pic {symmetry};\end{tikzpicture}};
        \end{tikzpicture}
    }}
    \caption[Visualisation of responsibility problem]{Discontinuity (red arrow) when rotating the set of points. The coloured points denote which output of the network is responsible for which point. In the top path, the set rotated by $90\degree$ is the same set (exactly the same shape before and after rotation) and encodes to the same feature vector, so the output responsibility (colouring) must be the same too. In this example, after $30\degree$ and a further small clockwise rotation by $\epsilon$, the point that each output pair is responsible for has to suddenly change.}
    \label{fspool fig:symmetry}
\end{figure}
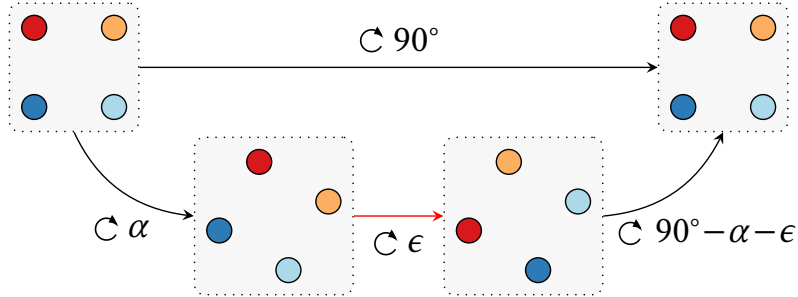

It turns out that standard neural networks struggle with modeling symmetries that arise because there are $n!$ different list representations of the same set, which we highlight here with an example.
Suppose we want to train an auto-encoder on our polygon dataset and have a square (so a set of 4 points with the x-y coordinates as features) with some arbitrary initial rotation (see \autoref{fspool fig:symmetry}).
Each pair in the 8 outputs of the MLP decoder is \emph{responsible} for producing one of the points in this square (\autoref{fspool fig:ae}).
We mark each such pair with a different colour in the figure.

\begin{figure}
    \centering
    \centerfloat{\resizebox{1.15\linewidth}{!}{%
        \sffamily
        \begin{tikzpicture}
            \pic at (-3, 0) {square2};
            \begin{scope}[on background layer]
                \node (s1) [object, fit=(a) (b) (c) (d)] {};
            \end{scope}

            \node (enc) [trapezium, draw, rotate=-90, inner ysep=3.5mm, trapezium angle=70, fill=Oranges-4-1] at (3.5, 0) {\scriptsize Encoder};
            \node [box, scale=0.5, fill=Blues-4-3] at (5.5, 0.5) {};
            \node (fv) [box, scale=0.5, fill=Oranges-4-3] at (5.5, 0) {};
            \node [box, scale=0.5, fill=Purples-4-3] at (5.5, -0.5) {};
            \node (dec) [trapezium, draw, rotate=90, inner ysep=3.5mm, trapezium angle=70, fill=Oranges-4-1] at (7.5, 0) {\scriptsize ~~MLP~~};

            \node [color=Purples-4-4!80!black] (i4) at (0,  1.5) {(-1, -1)};
            \node [color=Purples-4-4!80!black] (i3) at (0,  0.5) {(~1, -1)};
            \node [color=Purples-4-4!80!black] (i2) at (0, -0.5) {(~1, ~1)};
            \node [color=Purples-4-4!80!black] (i1) at (0, -1.5) {(-1, ~1)};

            \node [color=RdYlBu-4-1] (o4) at (11,  1.5) {(-1, ~1)};
            \node [color=RdYlBu-4-2] (o3) at (11,  0.5) {(~1, ~1)};
            \node [color=RdYlBu-4-3] (o2) at (11, -0.5) {(~1, -1)};
            \node [color=RdYlBu-4-4] (o1) at (11, -1.5) {(-1, -1)};

            \pic at (14, 0) {square};
            \begin{scope}[on background layer]
                \node (s2) [object, fit=(a) (b) (c) (d)] {};
            \end{scope}

            \draw [sedge] (i1) -- (enc.south |- i1);
            \draw [sedge] (i2) -- (enc.south |- i2);
            \draw [sedge] (i3) -- (enc.south |- i3);
            \draw [sedge] (i4) -- (enc.south |- i4);
            \draw [sedge] (enc) -- (fv);
            \draw [sedge] (fv) -- (dec);
            \draw [sedge] (dec.south |- o1) -- (o1);
            \draw [sedge] (dec.south |- o2) -- (o2);
            \draw [sedge] (dec.south |- o3) -- (o3);
            \draw [sedge] (dec.south |- o4) -- (o4);
        \end{tikzpicture}
    }}
    \caption[Set auto-encoder for demonstrating the responsibility problem]{Each pair of outputs of the auto-encoder is \emph{responsible} for one of the points of the set. This responsibility is marked with the different colours of the outputs.}
    \label{fspool fig:ae}
\end{figure}
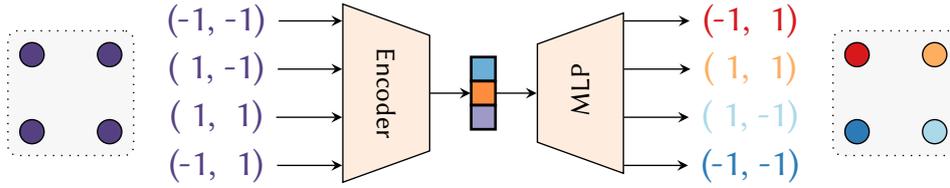

If we rotate the square (top left in figure) by 90 degrees (top right in figure), we simply permute the elements within the set.
They are the same set, so they also encode to the same latent representation and decode to the same list representation.
This means that each output is still responsible for producing the point at the same position after the rotation, i.e. the dark red output is still responsible for the top left point, the light red output is responsible for the top right point, etc.
However, this also means that at some point during that 90 degree rotation (bottom path in figure), \emph{there must exist a discontinuous jump} (red arrow in figure) in how the outputs are assigned.
We know that the 90 degree rotation must start and end with the top left point being produced by the dark red output.
Thus, we know that there is a rotation where all the outputs must simultaneously change which point they are responsible for, so that completing the rotation results in the top left point being produced by the dark red output.
\emph{Even though we change the set continuously, the list representation (MLP or RNN outputs) must change discontinuously.}

This is a challenge for neural networks to learn, since they can typically only model functions without discontinuous jumps.
As we increase the number of vertices in the polygon (number of set elements), it must learn an increasing frequency of situations where all the outputs must discontinuously change at once, which becomes very difficult to model. Our experiment in \autoref{fspool sec:polygons} confirms this.

This example highlights a more general issue: whenever there are at least two set elements that can be smoothly interchanged, these discontinuities arise.
For example, the set of bounding boxes in object detection can be interchanged in much the same way as the points of our square here.
An MLP or RNN that tries to generate these (like in \citet{Rezatofighi2018, Stewart2016}) must handle which of its outputs is responsible for what element in a discontinuous way.
Note that traditional object detectors like Faster R-CNN do not have this responsibility problem, because they do not treat object detection as a proper set prediction task with their anchor-based approach.
When the set elements come from a finite domain (often a set of labels) and not $\R^d$, it does not make sense to interpolate set elements.
Thus, the responsibility problem does not apply to methods for such problems, for example \citet{Welleck2018, Rezatofighi2017}.

\subsection{Formal statement}\label{app:responsibility}
The following theorem is a more formal treatment of the responsibility problem resulting in discontinuities.

\newcommand{\setxa}{
    \begin{smallmatrix}
        -\cos(\theta)\\
        -\sin(\theta)
    \end{smallmatrix}
}
\newcommand{\setxb}{
    \begin{smallmatrix}
        \cos(\theta)\\
        \sin(\theta)
    \end{smallmatrix}
}
\newcommand{\setxz}{
    \begin{smallmatrix}
        -1 & 1 \\
        0 & 0 \\
    \end{smallmatrix}
}

\begin{thm}
    For any set function $f: \mathcal{S}_n^d \to \R^{d \times n}$ ($d \geq 2$, $n \geq 2$, $\mathcal{S}_n^d$ is the set of all sets of size $n$ with elements in $\R^d$) from a set of points $\mS = \{ \vx_1$, $\vx_2$, $\ldots$, $\vx_n \}$ to a list representation of that set $\mL = [\vx_{\sigma(1)}, \vx_{\sigma(2)}, \ldots, \vx_{\sigma(n)} ]$ with some fixed permutation $\sigma \in \Pi$, there will be a discontinuity in $f$: there exists an $\varepsilon > 0$ such that for all $\delta > 0$, there exist two sets $\mS_1$ and $\mS_2$ where:

\begin{equation}
    d_s(\mS_1, \mS_2) < \delta \quad \text{and} \quad d_l(f(\mS_1), f(\mS_2)) \geq \varepsilon.
\end{equation}

$d_s$ is a measure of the distance between two sets (e.g. Chamfer loss) and $d_l$ is the sum of Euclidean distances ($d_l(\mA, \mB) = \sum_j \norm{\va_j - \vb_j}_2$).

\end{thm}

\begin{figure}[b]
    \centering
    \centerfloat{
    \includegraphics[width=0.5\textwidth]{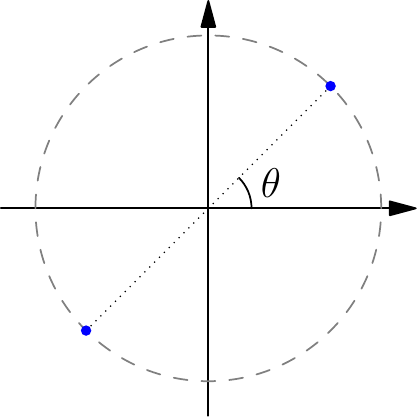}
    }
    \caption{Example of the set containing two points.}
    \label{fig:cmap}
\end{figure}

\begin{proof}
We prove the theorem by considering mappings from a set of two points in two dimensions.
For larger sets or sets with more dimensions, we can isolate two points and two dimensions and ignore the remaining points and dimensions.

Let us consider the set of two points $\mS(\theta) = \left\{ \left[\setxa\right], \left[\setxb\right] \right\}$ (see \autoref{fig:cmap}).
This is mapped to a list $\mL(\theta) = f(\mS(\theta))$.
Without loss of generality, we can assume that our list representation for $\theta = 0$ is
$\mL(0)
= \left[\begin{smallmatrix}
        -\cos(0) & \cos(0) \\
        -\sin(0) & \sin(0) \\
\end{smallmatrix}\right]
= \left[ \setxz \right]$.
Since the order of set elements is irrelevant and $f$ is a (permutation-invariant) set function, $\mS(\pi) = \mS(0)$ and therefore $\mL(\pi) = \mL(0) = \left[ \setxz \right]$.
This implies that for at least one value of $\theta = \theta^*$, there is a change in responsibility such that for $\theta \leq \theta^*$, the list representation will be
$\mL_1(\theta) = \left[ \setxa \setxb \right]$
while for $\theta > \theta^*$, the list representation will be
$\mL_2(\theta) = \left[ \setxb \setxa \right]$
in order to satisfy $\mL(\pi) = \mL(0)$.
For any $\theta$, $d_l(\mL_1(\theta), \mL_2(\theta)) = 4$.

Let $\varepsilon = 3.9$ and $\delta$ be given.
We can find a sufficiently small $\alpha > 0$ so that $d_s(\mS(\theta^*), \mS(\theta^* + \alpha)) < \delta$ and $d_l(\mL(\theta^*), \mL(\theta^* + \alpha)) > \varepsilon$.
\end{proof}


%

\sectionmark{Model}
\section{Featurewise sort pooling}\label{fspool sec:fsort}
\sectionmark{Model}

The main idea behind our pooling method is simple: sorting each feature across the elements of the set and performing a weighted sum.
The numerical sorting ensures the property of permutation-invariance.
The difficulty lies in how to determine the weights for the weighted sum in a way that works for variable-sized sets.

A key insight for auto-encoding is that we can store the permutation that the sorting applies in the encoder and apply the inverse of that permutation in the decoder.
This allows the model to restore the arbitrary order of the set element so that it no longer needs an assignment-based loss for training.
This avoids the problem in \autoref{fspool fig:symmetry}, because rotating the square by 90$\degree$ also permutes the outputs of the network accordingly.
Thus, there is no longer a discontinuity in the outputs during this rotation.
In other words, we make the auto-encoder permutation-\emph{equivariant}: permuting the input set also permutes the neural network's output in the same way.
This also means that $\mL(\pi) \neq \mL(0)$, so the proof no longer applies.

We describe the model for the simplest case of encoding fixed-size sets in \autoref{fspool sec:fixed}, extend it to variable-sized sets in \autoref{fspool sec:variable}, then discuss how to use this in an auto-encoder in \autoref{fspool sec:auto}.

\subsection{Fixed-size sets}\label{fspool sec:fixed}
We are given a set of $n$ feature vectors $\mX = [\vx_1, \ldots, \vx_n]$ where each $\vx_i$ is a column vector of dimension $d$ placed in some arbitrary order in the columns of $\mX \in \R^{d \times n}$.
From this, the goal is to produce a single feature vector in a way that is invariant to permutation of the columns in the matrix.

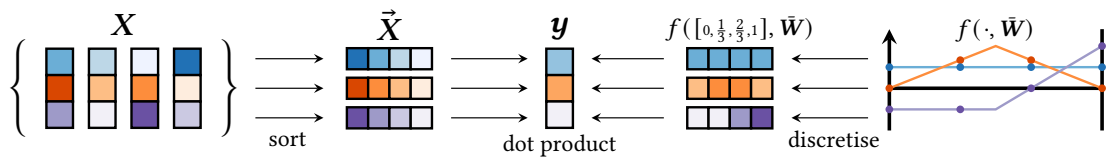
\begin{figure*}[t]
    \centering
    \centerfloat{\resizebox{\figurewidth}{!}{%
        \begin{tikzpicture}
            \node {\begin{tikzpicture}\pic {model};\end{tikzpicture}};
        \end{tikzpicture}
    }}
    \caption[Overview of FSPool model]{Overview of our \textsc{FSPool} model for variable-sized sets.
    In this example, the weights define piecewise linear functions with two pieces.
    The four dots on each line correspond to the positions where $f$ is evaluated for a set of size four.
    }
    \label{fspool fig:model}
\end{figure*}

We first sort each of the $d$ features across the elements of the set by numerically sorting within the rows of $\mX$ to obtain the matrix of sorted features $\mvX$:

\begin{equation}
    \vec{X}_{i,j} = \textsc{Sort}(\mX_{i,:})_j
\end{equation}

where $\mX_{i,:}$ is the $i$th row of $\mX$ and $\textsc{Sort}(\cdot)$ sorts a vector in descending order.
While this may appear strange since the columns of $\mvX$ no longer correspond to individual elements of the set, there are good reasons for this.
A transformation (such as with an MLP) prior to the pooling can ensure that the features being sorted are mostly independent so that little information is lost by treating the features independently.
Also, if we were to sort whole elements by one feature, there would be discontinuities whenever two elements swap order.
This problem is avoided by our featurewise sorting.

Efficient parallel implementations of \textsc{Sort} are available in Deep Learning frameworks such as PyTorch, which uses a bitonic sort ($O(\log^2 n)$ parallel time, $O(n \log^2 n)$ comparisons).
While the permutation that the sorting applies is not differentiable, gradients can still be propagated pathwise according to this permutation in a similar way as for max pooling.

Then, we apply a learnable weight matrix $\mW \in \R^{d \times n}$ to $\mvX$ by elementwise multiplying and summing over the columns (row-wise dot products).

\begin{equation}
    y_i = \sum_j^n W_{i,j} \vec{X}_{i,j} \label{fspool eq:fixed}
\end{equation}

$\vy \in \R^{d}$ is the final pooled representation of $\mvX$.
The weight vector allows different weightings of different ranks and is similar in spirit to the parametric version of the gather step in Gather-Excite~\citep{Hu2018}.
This is a generalisation of both max and sum pooling, since max pooling can be obtained with the weight vector $[1, 0, \ldots, 0]$ and sum pooling can be obtained with the $\mathbf{1}$ vector.
Thus, it is also a maximally powerful pooling method for multisets~\citep{Xu2019} while being potentially more flexible~\citep{Murphy2019} in what it can represent.

\subsection{Variable-size sets}\label{fspool sec:variable}
When the size $n$ of sets can vary, our previous weight matrix can no longer have a fixed number of columns.
To deal with this, we define a \emph{continuous} version of the weight vector in each row: we use a fixed number of weights to parametrise a piecewise linear function $f : [0, 1] \to \R$, also known as calibrator function~\citep{Jaderberg2015}.
For a set of size three, this function would be evaluated at 0, 0.5, and 1 to determine the three weights for the weighted sum.
For a set of size four, it would be evaluated at 0, 1/3, 2/3, and 1.
This decouples the number of columns in the weight matrix from the set size that it processes, which allows it to be used for variable-sized sets.

To parametrise a piecewise linear function $f$, we have a weight vector $\vbw \in \R^{k}$ where $k - 1$ is the number of pieces defined by the $k$ points.
With the ratio $r \in [0, 1]$,

\begin{equation}
    f(r, \vbw) = \sum_{i=1}^k \max\left(0, 1 - | r (k - 1) - (i - 1) | \right) \bar{w}_i
\end{equation}

The $\max(\cdot)$ term selects the two nearest points to $r$ and linearly interpolates them.
For example, if $k = 3$, choosing $r \in [0, 0.5]$ interpolates between the first two points in the weight vector with $(1 - 2r) w_1 + 2r w_2$.

We have a different $\vbw$ for each of the $d$ features and place them in the rows of a weight matrix $\mbW \in \R^{d \times k}$, which no longer depends on $n$.
Using these rows with $f$ to determine the weights:

\begin{equation}
    {y}_i = \sum_{j=1}^n f\left(\frac{j - 1}{n - 1}, \mbW_{i, :}\right) \vec{X}_{i,j} \label{fspool eq:variable}
\end{equation}

$\vy$ is now the pooled representation with a potentially varying set size $n$ as input.
When $n = k$, this reduces back to \autoref{fspool eq:fixed}.
For most experiments, we simply set $k=20$ without tuning it.

\subsection{Auto-encoder}\label{fspool sec:auto}
To create an auto-encoder, we need a decoder that turns the latent space back into a set.
Analogously to image auto-encoders, we want this decoder to roughly perform the operations of the encoder in reverse.
The FSPool in the encoder has two parts: sorting the features, and pooling the features.
Thus, the FSUnpool version should ``unpool'' the features, and ``unsort'' the features.
For the former, we define an unpooling version of \autoref{fspool eq:variable} that distributes information from one feature vector to a variable-size list of feature vectors.
For the latter, the idea is to store the permutation of the sorting from the encoder and use the inverse of it in the decoder to unsort it.
This allows the auto-encoder to restore the original ordering of set elements, which makes it permutation-equivariant.

With $\vyp \in \R^d$ as the vector to be unpooled, we define the unpooling similarly to \autoref{fspool eq:variable} as

\begin{equation}
    \vec{X}'_{i,j} = f\left(\frac{j - 1}{n - 1}, \mbW'_{i, :}\right) y'_{i}
\end{equation}

In the non-autoencoder setting, the lack of differentiability of the permutation is not a problem due to the pathwise differentiability.
However, in the auto-encoder setting we make use of the permutation in the decoder.
While gradients can still be propagated through it, it introduces discontinuities whenever the sorting order in the encoder for a set changes, which we empirically observed to be a problem.
To avoid this issue, we need the permutation that the sort produces to be differentiable.
To achieve this, we use the recently proposed sorting networks~\citep{Grover2019}, which is a continuous relaxation of numerical sorting.
This gives us a differentiable approximation of a permutation matrix $\mP_i \in [0, 1]^{n \times n}, i \in \{1, \ldots, d\}$ for each of the $d$ features, which we can use in the decoder while still keeping the model fully differentiable.
It comes with the trade-off of increased computation costs with $O(n^2)$ time and space complexity, so we only use the relaxed sorting in the auto-encoder setting.
It is possible to decay the temperature of the relaxed sort throughout training to 0, which allows the more efficient traditional sorting algorithm to be used at inference time.

Lastly, we can use the inverse of the permutation from the encoder to restore the original order.

\begin{equation}
    {X'}_{i,j} = (\mvX'_{i, :} \mP_i^\T)_{j}
\end{equation}

where $\mP_i^\T$ permutes the elements of the $i$th row in $\vec{X}'$.

Because the permutation is stored and used in the decoder, this makes our auto-encoder similar to a U-net architecture~\citep{Long2015} since it is possible for the network to skip the small latent space.
Typically we find that this only starts to become a problem when $d$ is too big, in which case it is possible to only use a subset of the $P_i$ in the decoder to counteract this.

\section{Related work}\label{fspool sec:related}
We are proposing a differentiable function that maps a \emph{set} of feature vectors to a single feature vector.
This has been studied in many works such as Deep Sets~\citep{Zaheer2017} and PointNet~\citep{Qi2017}, with universal approximation theorems being proven.
In our notation, the Deep Sets model is $g(\sum_{j} h(\mX_{:, j}))$ where $h: \R^d \to \R^p$ and $g: \R^p \to \R^q$.
Since this is $O(n)$ in the set size $n$, it is clear that while it may be able to approximate any set function, problems that depend on higher-order interactions between different elements of the set will be difficult to model aside from pure memorisation.
This explains the success of relation networks (RN), which simply perform this sum over all \emph{pairs} of elements, and has been extended to higher orders by \citet{Murphy2019}.
Our work proposes an alternative operator to the sum that is intended to allow some relations between elements to be modeled through the sorting, while not incurring as large of a computational cost as the $O(n^2)$ complexity of RNs.

\paragraph{Sorting-based set functions}
The use of sorting has often been considered in the set learning literature due to its natural way of ensuring permutation-invariance.
The typical approach is to sort elements of the set as units rather than our approach of sorting each feature individually.

For example, the similarly-named SortPooling~\citep{Zhang2018} sorts the elements based on one feature of each element.
However, this introduces discontinuities into the optimisation whenever two elements swap positions after the sort.
For variable-sized sets, they simply truncate (which again adds discontinuities) or pad the sorted list to a fixed length and process this with a CNN, treating the sorted vectors as a sequence.
Similarly, \citet{Cangea2018} and \citet{Gao2019} truncate to a fixed-size set by computing a score for each element and keeping elements with the top-k scores.
In contrast, our pooling handles variable set sizes without discontinuities through the featurewise sort and continuous weight space.
\citet{Gao2019} propose a graph auto-encoder where the decoder use the ``inverse'' of what the top-k operator does in the encoder, similar to our approach.
Instead of numerically sorting, \citet{Mena2018} and our Permutation-optimisation model (\autoref{cha:perm-optim}) \emph{learn} an ordering of set elements instead.
Our FSPool model introduced here has the benefit of only $\Theta(n \log n)$ time complexity, compared to Permutation-optimisation with $\Theta(n^3)$, so it is much easier to use with larger sets and thus a greater variety of tasks.

Outside of the set learning literature, rank-based pooling in a convolutional neural network has been used in \citet{Shi2016}, where the rank is turned into a weight.
Sorting within a single feature vector has been used for modeling more powerful functions under a Lipschitz constraint for Wasserstein GANs~\citep{Anil2019} and improved robustness to adversarial examples~\citep{Cisse2017}.


\paragraph{Set prediction}
Assignment-based losses combined with an MLP or similar are a popular choice for various auto-encoding and generative tasks on point clouds~\citep{Fan2017, Yang2018, Achlioptas2018}.
An interesting alternative approach is to perform the set generation sequentially~\citep{Stewart2016,Johnson2017, You2018}.
The difficulty lies in how to turn the set into one or multiple sequences, which these papers try to solve in different ways.

\section{Experiments}\label{fspool sec:experiments}
We start with two auto-encoder experiments, then move to tasks where we replace the pooling in an established model with FSPool.
Full results can be found in the appendices, experimental details can be found in \autoref{fspool app:details}, and we provide our code for reproducibility at \url{https://github.com/Cyanogenoid/fspool}.

\subsection{Rotating polygons}\label{fspool sec:polygons}
\begin{figure}
    \centerfloat{
        \includegraphics[width=0.2\figurewidth]{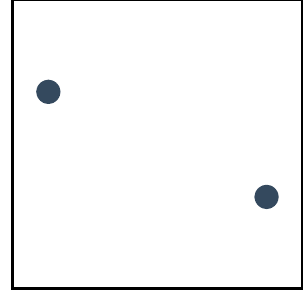}
        \hspace{0.0666\figurewidth}
        \includegraphics[width=0.2\figurewidth]{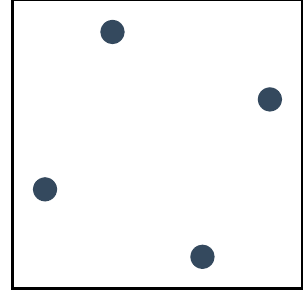}
        \hspace{0.0666\figurewidth}
        \includegraphics[width=0.2\figurewidth]{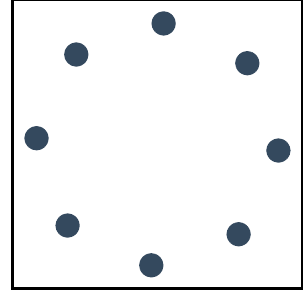}
        \hspace{0.0666\figurewidth}
        \includegraphics[width=0.2\figurewidth]{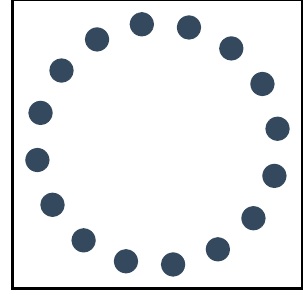}
    }
    \caption[Examples from polygon dataset]{Examples from polygon dataset for $n \in \{2, 4, 8, 16\}$.}
    \label{fspool fig:poly}
\end{figure}

We start with our simple dataset of auto-encoding regular polygons (\autoref{fspool sec:problem}), with each point in a set corresponding to the x-y coordinate of a vertex in that polygon.
This dataset is designed to explicitly test whether the responsibility problem occurs in practice.
We keep the set size the same within a training run and only vary the rotation.
We try this with set sizes of increasing powers of 2.
We show some examples of different set sizes in \autoref{fspool fig:poly}.

\paragraph{Model}
The encoder contains a 2-layer MLP applied to each set element, FSPool, and a 2-layer MLP to produce the latent space.
The decoder contains a 2-layer MLP, FSUnpool, and a 2-layer MLP applied on each set element.
We train this model to minimise the mean squared error.
As baseline, we use a model where the decoder has been replaced with an MLP and train it with either the linear assignment or Chamfer loss (equivalent to AE-EMD and AE-CD models in \citet{Achlioptas2018}).
We also include a random baseline that outputs a polygon with the correct size and centre, but random rotation.

\begin{table}
    \centering
    \caption[MSE on Polygon dataset]{
        Direct mean squared error (in hundredths) on Polygon dataset with different number of points in the set. Lower is better.
    }
    \centerfloat{
    \begin{tabular}{lcccccc}
    \toprule
    Set size & 2 & 4 & 8 & 16 & 32 & 64 \\
    \midrule
    \textsc{FSPool} & \textbf{0.000} & \textbf{0.001} & \textbf{0.000} & \textbf{0.000} & \textbf{0.000} & \textbf{0.0001} \\
    \textsc{Random} & 100.323 & 100.134 & 99.367 & 99.951 & 99.438 & 99.523 \\
    \bottomrule
    \end{tabular}
    }
    \label{fspool tab:graph-params-direct}

    \vspace{5mm}
    \centering
    \caption[Chamfer loss on Polygon dataset]{
        Chamfer loss (in hundredths) on Polygon dataset with different number of points in the set. Lower is better.
    }
    \centerfloat{
    \begin{tabular}{lcccccc}
    \toprule
    Set size & 2 & 4 & 8 & 16 & 32 & 64 \\
    \midrule
    \textsc{FSPool} & \textbf{0.001} & \textbf{0.001} & \textbf{0.001} & \textbf{0.000} & \textbf{0.001} & \textbf{0.002} \\
    \textsc{MLP} + Chamfer & 1.189 & 1.771 & 0.274 & 1.272 & 0.316 & 0.085 \\
    \textsc{MLP} + Hungarian & 1.517 & 0.400 & 0.251 & 1.266 & 0.326 & 0.081 \\
    \textsc{Random} & 72.848 & 19.866 & 5.112 & 1.271 & 0.322 & 0.081 \\
    \bottomrule
    \end{tabular}
    }
    \label{fspool tab:graph-params-chamfer}

    \vspace{5mm}
    \centering
    \caption[Hungarian loss on Polygon dataset]{
        Hungarian loss (in hundredths) on Polygon dataset with different number of points in the set. Lower is better.
    }
    \centerfloat{
    \begin{tabular}{lcccccc}
    \toprule
    Set size & 2 & 4 & 8 & 16 & 32 & 64 \\
    \midrule
    \textsc{FSPool} & \textbf{0.000} & \textbf{0.001} & \textbf{0.000} & \textbf{0.000} & \textbf{0.000} & \textbf{0.001} \\
    \textsc{MLP} + Chamfer & 0.595 & 0.885 & 0.137 & 0.641 & 0.160 & 0.285 \\
    \textsc{MLP} + Hungarian & 0.758 & 0.200 & 0.126 & 0.634 & 0.163 & 0.040 \\
    \textsc{Random} & 36.424 & 9.933 & 2.556 & 0.635 & 0.161 & 0.041\\

    \bottomrule
    \end{tabular}
    }
    \label{fspool tab:graph-params-hungarian}
\end{table}

\paragraph{Results}
First, we verified that if the latent space is always zeroed out, the model with FSPool is unable to train, suggesting that the latent space is being used and is necessary.
In \autoref{fspool tab:graph-params-direct}, \autoref{fspool tab:graph-params-chamfer}, and \autoref{fspool tab:graph-params-hungarian}, we show the results of various model and training loss combinations.
For our training runs with set sizes up to 128, our auto-encoder is able to reconstruct the point set close to perfectly.
Meanwhile, the baseline converges significantly slower with high reconstruction error when the number of points is 8 or fewer and outputs the same set irrespective of input above that, regardless of loss function.
This shows that FSPool with the direct MSE training loss is clearly better than the baseline trained with either Hungarian or Chamfer loss.

Even when significantly increasing the latent size, dimensionality of layers, tweaking the learning rate, and replacing FSPool in the encoder with sum, mean, or max, the baseline trained with the Hungarian or Chamfer loss fails completely at 16 points.
We verified that for 4 points, the baseline shows the discontinuous jump behaviour in the outputs as we predict in \autoref{fspool fig:symmetry}.

This experiment highlights the difficulty of learning this simple dataset with traditional approaches due to the responsibility problem, while our model is able to fit this dataset with ease.

\subsection{Noisy MNIST reconstruction}\label{fspool sec:mnist}
Next, we turn to the harder task of auto-encoding MNIST images -- turned into sets of points -- using a denoising auto-encoder.
Each pixel that is above the mean pixel level is considered to be part of the set with its x-y coordinates as feature, scaled to be within the range of [0, 1].
The set size varies between examples and is 133 on average.
We add Gaussian noise to the points in the set and use the set without noise as training target for the denoising auto-encoder.

\paragraph{Model}
We use exactly the same architecture as on the polygon dataset.
As baseline models, we combine sum/mean/max pooling encoders with MLP/LSTM decoders and train with the Chamfer loss.
This closely corresponds to the AE-CD approach~\citep{Achlioptas2018} with the MLP decoder and the model by \citet{Stewart2016} with the LSTM decoder.


\begin{figure}
    \centering
    \centerfloat{
        \includegraphics[width=\figurewidth, trim={0 3.5mm 0 0},clip]{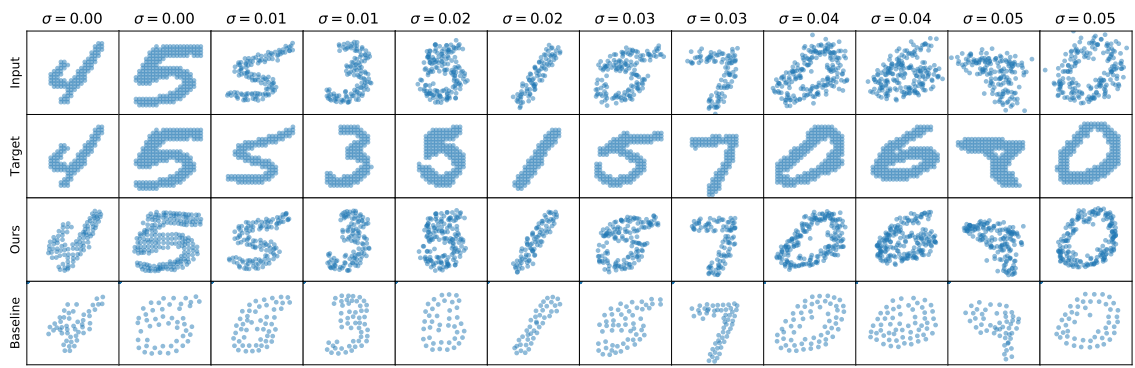}
    }
    \caption[MNIST reconstructions for varying noise]{MNIST as point sets with different amounts of Gaussian noise ($\sigma$) and their reconstructions. The baseline uses sum pooling and an MLP decoder, which had the best quantitative results among the baselines. We used the best network for our model ($0.28 \times 10^4$ average Chamfer loss) and the best network for the baseline model ($0.20 \times 10^4$ average Chamfer loss). The examples are not cherry-picked.}
    \label{fspool fig:mnist}
\end{figure}

\begin{table}
    \setlength{\tabcolsep}{4pt}
    \centering
    \caption[Chamfer losses on MNIST]{Test Chamfer loss (in 1000ths) for MNIST for different input noise levels $\sigma$ over 6 runs. Lower is better.}
    \centerfloat{\begin{tabular}{lcccccc}
    \toprule
    Noise $\sigma$ & 0.00 & 0.01 & 0.02 & 0.03 & 0.04 & 0.05\\
    \midrule
    \textsc{FSPool + FSUnpool} & 0.42\tiny$\pm$0.06 & 0.34\tiny$\pm$0.05 & 0.36\tiny$\pm$0.02 & 0.38\tiny$\pm$0.03 & 0.41\tiny$\pm$0.00 & 0.44\tiny$\pm$0.01 \\
    \textsc{Sum + MLP} & \textbf{0.30}\tiny$\pm$0.04 & \textbf{0.28}\tiny$\pm$0.03 & \textbf{0.28}\tiny$\pm$0.03 & \textbf{0.28}\tiny$\pm$0.03 & \textbf{0.27}\tiny$\pm$0.01 & \textbf{0.31}\tiny$\pm$0.04 \\
    \textsc{Sum + RNN} & 0.76\tiny$\pm$0.46 & 0.58\tiny$\pm$0.06 & 0.57\tiny$\pm$0.09 & 0.54\tiny$\pm$0.11 & 0.64\tiny$\pm$0.13 & 0.78\tiny$\pm$0.39 \\
    \textsc{Max + MLP} & 1.29\tiny$\pm$0.23 & 1.37\tiny$\pm$0.28 & 1.23\tiny$\pm$0.16 & 1.74\tiny$\pm$0.32 & 1.27\tiny$\pm$0.19 & 1.43\tiny$\pm$0.30 \\
    \textsc{Mean + MLP} & 1.41\tiny$\pm$0.12 & 1.22\tiny$\pm$0.18 & 1.33\tiny$\pm$0.29 & 1.25\tiny$\pm$0.09 & 1.31\tiny$\pm$0.15 & 1.49\tiny$\pm$0.31 \\
    \bottomrule
    \end{tabular}}
    \label{fspool tab:mnist1}

    \vspace{12mm}
    \setlength{\tabcolsep}{4pt}
    \centering
    \caption[Chamfer losses on MNIST with mask feature]{Test Chamfer loss (in 1000ths) for MNIST with additional mask features on every element for different input noise levels $\sigma$ over 6 runs. Lower is better.}
    \centerfloat{\begin{tabular}{lcccccc}
    \toprule
    Noise $\sigma$ & 0.00 & 0.01 & 0.02 & 0.03 & 0.04 & 0.05\\
    \midrule
    \textsc{FSPool + FSUnpool} & \textbf{0.28}\tiny$\pm$0.03 & \textbf{0.21}\tiny$\pm$0.01 & \textbf{0.25}\tiny$\pm$0.03 & \textbf{0.25}\tiny$\pm$0.01 & \textbf{0.27}\tiny$\pm$0.01 & \textbf{0.30}\tiny$\pm$0.01 \\
    \textsc{Sum + MLP} & 0.87\tiny$\pm$0.43 & 0.90\tiny$\pm$0.39 & 0.61\tiny$\pm$0.40 & 0.99\tiny$\pm$0.84 & 0.63\tiny$\pm$0.27 & 0.61\tiny$\pm$0.24 \\
    \textsc{Sum + RNN} & 0.58\tiny$\pm$0.13 & 0.69\tiny$\pm$0.16 & 0.60\tiny$\pm$0.18 & 1.35\tiny$\pm$1.53 & 0.73\tiny$\pm$0.12 & 0.63\tiny$\pm$0.13 \\
    \textsc{Max + MLP} & 5.91\tiny$\pm$3.10 & 4.78\tiny$\pm$3.05 & 7.10\tiny$\pm$2.40 & 5.05\tiny$\pm$2.87 & 5.85\tiny$\pm$3.11 & 4.57\tiny$\pm$2.08 \\
    \textsc{Mean + MLP} & 5.92\tiny$\pm$3.10 & 4.55\tiny$\pm$3.17 & 6.84\tiny$\pm$2.84 & 7.11\tiny$\pm$2.39 & 3.04\tiny$\pm$0.78 & 6.34\tiny$\pm$2.53 \\
    \bottomrule
    \end{tabular}}
    \label{fspool tab:mnist1-masked}
\end{table}

\paragraph{Results}
We show the quantitative results for the default MNIST setting in \autoref{fspool tab:mnist1}.
Interestingly, the sum pooling baseline has a lower Chamfer reconstruction error than our model, despite the example outputs in \autoref{fspool fig:mnist} looking clearly worse.
This demonstrates the weakness of the Chamfer loss we mentioned in \autoref{background sec:setloss} when comparing multisets.
Our model avoids this weakness by being trained with a normal MSE loss (with the trade-off of a potentially higher Chamfer loss), which is not possible with the baselines.
This MSE loss ensures that our model always has to predict the correct set size.
The sum pooling baseline has a better test Chamfer loss because it is trained to minimise it, but it is also solving an easier task, since it does not need to distinguish padding from non-padding elements.

The main reason for this difference comes from the shortcoming of the Chamfer loss in distinguishing sets with duplicates or near-duplicates.
For example, the Chamfer loss between $[1, 1.001, 9]$ and $[1, 9, 9.001]$ is close to 0.
Most points in an MNIST set are quite close to many other points and there are many duplicate padding elements, so this problem with the Chamfer loss is certainly present on MNIST.
That is why minimising MSE can lead to different results with higher Chamfer loss than minimising Chamfer loss directly, even though the qualitative results seem worse for the latter.

We can make the comparison between our model and the baselines more similar by forcing the models to predict an additional ``mask feature'' for each set element.
This takes the value 1 when the point is present (non-padding element) and 0 (padding element) when not.
This setting is useful for tasks where the predicted set size matters, as it allows points at the coordinates (0, 0) to be distinguished from padding elements.
The results of this variant are shown in \autoref{fspool tab:mnist1-masked}.
Now, our model is clearly better: even though our auto-encoder minimises an MSE loss, the test Chamfer loss is also much better than all the baselines.
Having to predict this additional mask feature does not affect our model predictions much because our model structure lets our model ``know'' which elements are padding elements, while this is much more challenging for the baselines.

We can also qualitatively compare our FSPool-FSUnpool model against the best baseline, which uses the sum pooling encoder and MLP decoder (\autoref{fspool fig:mnist}).
In general, our model can reconstruct the digits much better than the baseline, which tends to predict too few points even though it always has 342 (the maximum set size) times 2 outputs available.
Occasionally, the baseline also makes big errors such as turning 5s into 8s (first $\sigma=0.01$ example), which we have not observed with our model.


\subsection{Noisy MNIST classification}\label{fspool sec:mnist-class}
Instead of auto-encoding MNIST sets, we can also classify them.
We use the same dataset and replace the set decoder in our model and the baseline with a 2-layer MLP classifier.
We consider three variants: using the trained auto-encoder weights for the encoder and freezing them, not freezing them (finetuning), and training all weights from random initialisation.
This tests how informative the learned representations of the pre-trained auto-encoder and the encoder are.

\paragraph{Results}
We show our results for 1 or 10 epochs and $\sigma = 0.05$ in \autoref{fspool tab:mnist}, for $\sigma = 0.00$ in \autoref{fspool tab:mnist2}, and for 100 epochs and both $\sigma$ values in \autoref{fspool tab:mnist-100}.
These are based on pre-trained models from the default MNIST setting without mask feature.

\begin{table}
    \setlength{\tabcolsep}{4pt}
    \centering
    \caption[Classification accuracies on MNIST with noise for 1 and 10 epochs]{Classification accuracy (mean $\pm$ stdev) on MNIST $\sigma = 0.05$ over 6 runs (different pre-trained networks between runs). Frozen: training with frozen pre-trained auto-encoder weights. Unfrozen: unfrozen auto-encoder weights (fine-tuning). Random init: auto-encoder weights not used.}
    \centerfloat{\begin{tabular}{lcccccc}
    \toprule
    & \multicolumn{3}{c}{1 epoch of training} & \multicolumn{3}{c}{10 epochs of training}\\
    \cmidrule{2-4} \cmidrule(l{4pt}){5-7}
    & Frozen & Unfrozen & Random init & Frozen & Unfrozen & Random init\\
    \midrule
    \textsc{FSPool} & \textbf{82.2}\%\tiny$\pm$2.1 & \textbf{86.9}\%\tiny$\pm$1.3 & \textbf{84.7}\%\tiny$\pm$1.9 & \textbf{84.3}\%\tiny$\pm$ 1.8 & \textbf{91.5}\%\tiny$\pm$0.5 & \textbf{91.9}\%\tiny$\pm$0.5 \\
    \textsc{Sum}    & 76.6\%\tiny$\pm$1.3 & 68.7\%\tiny$\pm$3.5 & 30.3\%\tiny$\pm$5.6 & 79.0\%\tiny$\pm$1.0 & 77.7\%\tiny$\pm$2.3 & 72.7\%\tiny$\pm$3.4 \\
    \textsc{Mean}   & 25.7\%\tiny$\pm$3.6 & 32.2\%\tiny$\pm$10.5 & 30.1\%\tiny$\pm$1.6 & 36.8\%\tiny$\pm$5.0 & 75.0\%\tiny$\pm$2.7 & 73.0\%\tiny$\pm$1.7 \\
    \textsc{Max}   & 73.6\%\tiny$\pm$1.3 & 73.0\%\tiny$\pm$3.5 & 56.1\%\tiny$\pm$5.6 & 77.3\%\tiny$\pm$0.9 & 80.4\%\tiny$\pm$1.8 & 76.9\%\tiny$\pm$1.3 \\
    \bottomrule
    \end{tabular}}
    \label{fspool tab:mnist}

    \vspace{5mm}
    \setlength{\tabcolsep}{4pt}
    \centering
    \caption[Classification accuracies on MNIST without noise for 1 and 10 epochs]{Classification accuracy (mean $\pm$ stdev) on MNIST $\sigma = 0.00$ over 6 runs.}
    \centerfloat{\begin{tabular}{lcccccc}
    \toprule
    & \multicolumn{3}{c}{1 epoch of training} & \multicolumn{3}{c}{10 epochs of training}\\
    \cmidrule{2-4} \cmidrule(l{4pt}){5-7}
    & Frozen & Unfrozen & Random init & Frozen & Unfrozen & Random init\\
    \midrule
    \textsc{FSPool} & \textbf{86.3}\%\tiny$\pm$1.6 & \textbf{92.3}\%\tiny$\pm$1.1 & \textbf{90.5}\%\tiny$\pm$1.2 & \textbf{88.2}\%\tiny$\pm$1.4 & \textbf{96.0}\%\tiny$\pm$0.3 & \textbf{96.1}\%\tiny$\pm$0.3 \\
    \textsc{Sum}    & 82.3\%\tiny$\pm$1.2 & 77.9\%\tiny$\pm$3.4 & 35.3\%\tiny$\pm$8.3 & 85.0\%\tiny$\pm$0.8 & 84.2\%\tiny$\pm$2.5 & 78.4\%\tiny$\pm$3.9 \\
    \textsc{Mean}   & 27.0\%\tiny$\pm$3.3 & 43.5\%\tiny$\pm$7.1 & 31.2\%\tiny$\pm$1.0 & 42.0\%\tiny$\pm$7.7 & 76.7\%\tiny$\pm$2.6 & 77.2\%\tiny$\pm$2.2 \\
    \textsc{Max}   & 82.0\%\tiny$\pm$1.8 & 84.1\%\tiny$\pm$1.4 & 62.9\%\tiny$\pm$3.5 & 86.8\%\tiny$\pm$0.9 & 91.9\%\tiny$\pm$1.3 & 87.7\%\tiny$\pm$1.2 \\
    \bottomrule
    \end{tabular}}
    \label{fspool tab:mnist2}

    \vspace{5mm}
    \setlength{\tabcolsep}{4pt}
    \centering
    \caption[Classification accuracies on MNIST with and without noise for 100 epochs]{Classification accuracy (mean $\pm$ stdev) on MNIST for 100 epochs over 6 runs.}
    \centerfloat{\begin{tabular}{lcccccc}
    \toprule
    & \multicolumn{3}{c}{$\sigma=0.05$, 100 epochs} & \multicolumn{3}{c}{$\sigma=0.00$, 100 epochs}\\
    \cmidrule{2-4} \cmidrule(l{4pt}){5-7}
    & Frozen & Unfrozen & Random init & Frozen & Unfrozen & Random init\\
    \midrule
    \textsc{FSPool} & \textbf{84.9}\%\tiny$\pm$1.7 & \textbf{93.9}\%\tiny$\pm$0.4 & \textbf{94.0}\%\tiny$\pm$0.3 & 88.6\%\tiny$\pm$1.6 & \textbf{97.4}\%\tiny$\pm$0.3 & \textbf{97.5}\%\tiny$\pm$0.3 \\
    \textsc{Sum}    & 79.8\%\tiny$\pm$1.0 & 85.3\%\tiny$\pm$1.1 & 83.1\%\tiny$\pm$1.9 & 85.6\%\tiny$\pm$0.9 & 89.5\%\tiny$\pm$2.5 & 88.3\%\tiny$\pm$1.4 \\
    \textsc{Mean}   & 48.2\%\tiny$\pm$6.9 & 86.5\%\tiny$\pm$0.8 & 84.1\%\tiny$\pm$2.3 & 57.0\%\tiny$\pm$7.7 & 90.3\%\tiny$\pm$1.3 & 91.1\%\tiny$\pm$0.8 \\
    \textsc{Max}   & 78.8\%\tiny$\pm$0.8 & 84.7\%\tiny$\pm$1.0 & 84.6\%\tiny$\pm$0.9 & \textbf{89.2}\%\tiny$\pm$0.8 & 95.3\%\tiny$\pm$0.7 & 95.1\%\tiny$\pm$1.5 \\
    \bottomrule
    \end{tabular}}
    \label{fspool tab:mnist-100}
\end{table}

The FSPool-based models are consistently superior to all the baselines in all training settings.
Even though our model can store information in the permutation that skips the latent space (the model could ``cheat'' the reconstruction by somehow storing everything in the permutation matrix), our latent space contains more information to correctly classify a set, even when the weights are fixed.
Our model with fixed encoder weights already performs better after 1 epoch of training than the baseline models with unfrozen weights after 10 epochs of training.
This shows the benefit of the FSPool-FSUnpool auto-encoder to the representation.
When allowing the encoder weights to change (Unfrozen and Random init), our results again improve significantly over the baselines.

Interestingly, switching the relaxed sort to the unrelaxed sort in our model when using the fixed auto-encoder weights does not hurt accuracy.
Training the FSPool model takes 45 seconds per epoch on a GTX 1080 GPU, only slightly more than the baselines with 37 seconds per epoch.

Note that while~\citet{Qi2017} report an accuracy of $\sim$99\% on a similar set version of MNIST, our model uses noisy sets as input and is much smaller and simpler: we have 3820 parameters, while their model has 1.6 million parameters.
Our model also does not use dropout, batch norm, a branching network architecture, nor a stepped learning rate schedule.
When we try to match their model size, our accuracies for $\sigma=0.00$ increase to $\sim$99\% as well.

\subsection{CLEVR}\label{fspool sec:clevr}
CLEVR~\citep{Johnson2017} is a visual question answering dataset where the task is to classify an answer to a question about an image.
The images show scenes of 3D objects with different attributes, and the task is to answer reasoning questions such as ``what size is the sphere that is left of the green thing''.
Since we are interested in sets, we use this dataset with the ground-truth state description -- the set of objects (maximum size 10) and their attributes -- as input instead of an image of the rendered scene.

\paragraph{Model}
For this dataset, we compare against relation networks (RN) \citep{Santoro2017} -- explicitly modeling all pairwise relations --, Janossy pooling~\citep{Murphy2019}, and regular pooling functions.
While the original RN paper reports a result of 96.4\% for this dataset, we use a tuned implementation by \citet{Messina2018} with 2.6\% better accuracy.
For our model, we modify this to not operate on pairwise relations and replace the existing sum pooling with FSPool.
We use the same hyperparameters for our model as the strong RN baseline without further tuning them.



\begin{table}
    \centering
    \caption[Results on CLEVR]{
        CLEVR results over 10 runs: mean $\pm$ stdev of accuracy after 350 epochs, epochs to reach an accuracy milestone, and wall time required with a 1080 Ti GPU.
        * averages over only 8 runs because 2 runs did not reach 99\%.
        MAC~\citep{Hudson2018} is a model specifically designed for CLEVR and the state-of-the-art for \emph{image inputs} and without program supervision.%
    }
    \begin{tabular}{lccccc}
    \toprule
    {}     &  & \multicolumn{3}{c}{Epochs to reach accuracy} & Time for\\
    \cmidrule{3-5}
    Model  & Accuracy & 98.00\% &    98.50\% &   99.00\%           & 350 epochs\\
    \midrule
    \textsc{FSPool}   &   \textbf{99.27}\%\tiny$\pm$0.18 & \textbf{141}\tiny{$\pm$~5} & \textbf{166}\tiny{$\pm$16} & \textbf{209}\tiny{$\pm$33}  & ~~8.8 h\\
    \textsc{RN}       &   98.98\%\tiny$\pm$0.25 & 144\tiny{$\pm$~6} &     189\tiny{$\pm$29} &    *268\tiny{$\pm$46}~~~  & 15.5 h\\
    \textsc{Janossy} & 97.00\%\tiny$\pm$0.54 & -- & -- & -- & 11.5 h   \\
    \textsc{Sum}      &   99.05\%\tiny$\pm$0.17 & 146\tiny{$\pm$13} &     191\tiny{$\pm$40} &    281\tiny{$\pm$56}  & ~~8.0 h\\
    \textsc{Mean}      &   98.96\%\tiny$\pm$0.27 & 169\tiny{$\pm$~6} &     225\tiny{$\pm$31} &    273\tiny{$\pm$33} & ~~8.0 h\\
    \textsc{Max}      &   96.99\%\tiny$\pm$0.26 & -- &     -- &    -- & ~~8.0 h\\
    \textsc{MAC} & 99.0~\%~~~~~~~& -- & -- & --& --  \\
    \bottomrule
    \end{tabular}
    \label{fspool tab:clevr}
\end{table}

\begin{figure}
    \centering
    \centerfloat{
        \includegraphics[width=\figurewidth]{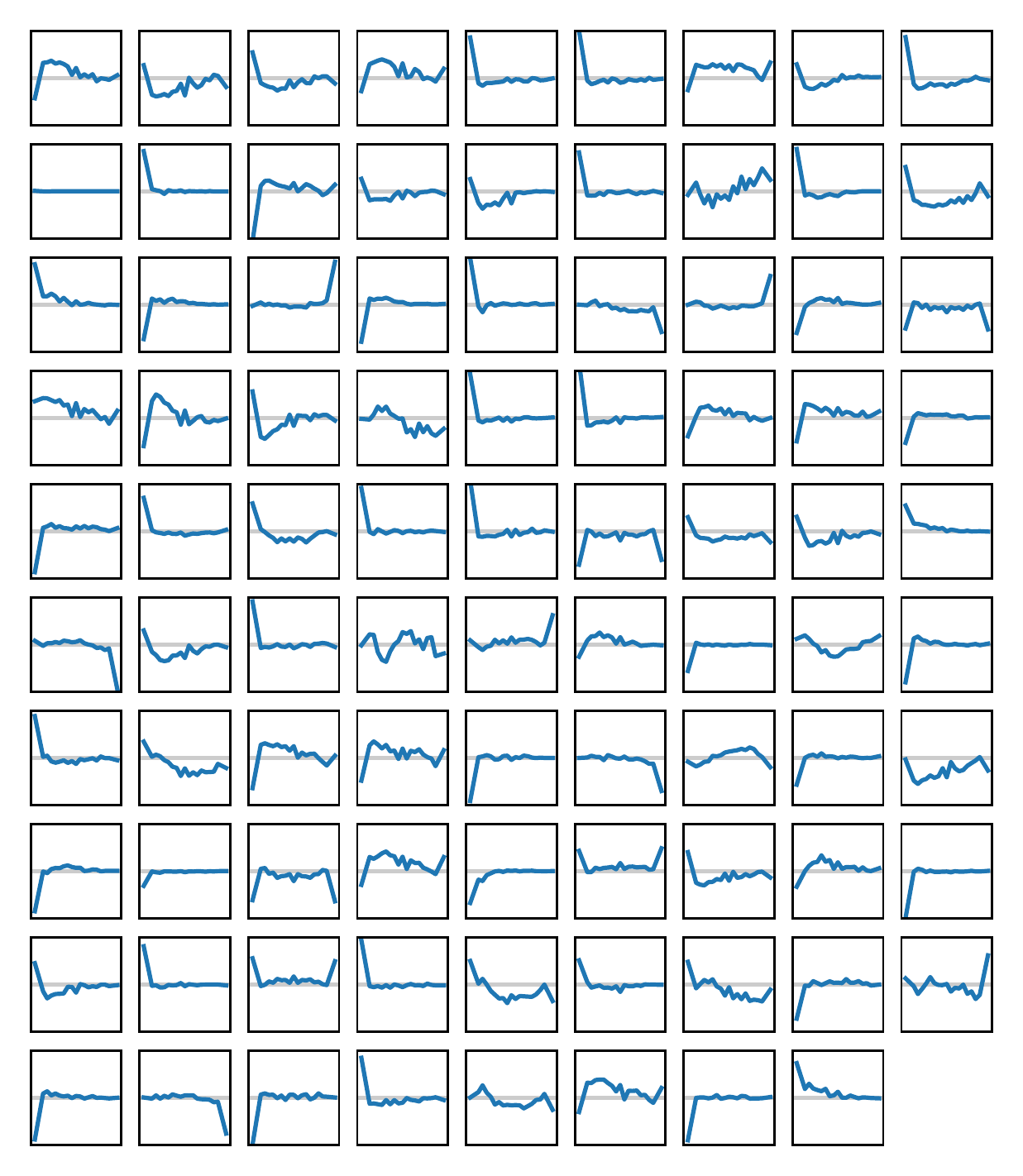}
    }
    \caption[Shapes of learned piecewise linear functions on CLEVR]{Shapes of piecewise linear functions learned by the FSPool model on CLEVR.
        These show $r \in [0, 1]$ on the x-axis and $f(r, \vbw)$ on the y-axis for a particular $\vbw$ of a fully-trained model.
        A common shape among these functions are variants of max pooling: close to 0 weight for most ranks and a large non-zero weight on either the maximum or the minimum value, for example in row 2 column 2.
        There are many functions that simple maximums or sums can \emph{not} easily represent, such as a variant of max pooling with the values slightly below the max receiving a weight of the opposite sign (see row 1 column 1) or the shape in the penultimate row column 5.
        The functions shown here may have a stronger tendency towards 0 values than normal due to the use of weight decay on CLEVR.
    }
    \label{fspool fig:plin}
\end{figure}

\paragraph{Results}
Over 10 runs, \autoref{fspool tab:clevr} shows that our FSPool model reaches the best accuracy and also reaches the listed accuracy milestones in fewer epochs than all baselines.
The difference in accuracy is statistically significant (two-tailed t-tests against sum, mean, RN, all with $p\approx0.01$).
Also, FSPool reaches 99\% accuracy in 5.3 h, while the fastest baseline, mean pooling, reaches the same accuracy in 6.2 h.
Surprisingly, RNs do not provide any benefit here, despite the hyperparameters being explicitly tuned for the RN model.
We show some of the functions $f(\cdot, \mbW)$ that FSPool has learned in \autoref{fspool fig:plin}.
These confirm that FSPool uses more complex functions than just sums or maximums, which allow it to capture more information about the set than other pooling functions.

%

\subsection{Graph classification}\label{fspool sec:graph}
We perform a large number of experiments on various graph classification datasets from the TU repository~\citep{Kersting2016}: 
4 graph datasets from bioinformatics (for example with the graph encoding the structure of a molecule) and 5 datasets from social networks (for example with the graph encoding connectivity between people who worked with each other).
The task is to classify the whole graph into one of multiple classes such as positive or negative drug response.

\paragraph{Model}
We use the state-of-the-art graph neural network GIN~\citep{Xu2019} as baseline.
This involves a series of graph convolutions (which includes aggregation of features from each node's set of neighbours into the node), a readout (which aggregates the set of all nodes into one feature vector), and a classification with an MLP.
We replace the usual sum or mean pooling readout with FSPool $k=5$ for our model.
We repeat 10-fold cross-validation on each dataset 10 times and use the same hyperparameter ranges as \citet{Xu2019} for our model and the GIN baseline.

\paragraph{Experimental setup}
The datasets and node features used are the same as in GIN; we did not cherry-pick them.
Because the social network datasets are purely structural without node features, a constant 1 feature is used on the RDT datasets and the one-hot-encoded node degree is used on the other social network datasets.
The hyperparameter sweep is done based on best validation accuracy for each fold in the cross-validation individually and over the same combinations as specified in GIN.

Note that in GIN, hyperparameters are selected based on best \emph{test} accuracy.
This is a problem, because they consider the number of epochs a hyperparameter when accuracies tend to significantly vary between individual epochs. 
For example, our average result on the PROTEINS dataset would change from 73.8\% to 77.1\%  if we were to select based on best test accuracy, which would be better than their 76.2\%.

While we initially also used $k=20$ in FSPool for this experiment, we found that $k=5$ was consistently an improvement.
The $k=20$ model was still better than the baseline on average by a smaller margin.

\begin{table}
    \centering
    \caption[Graph classification results]{Cross-validation classification results (\%) on various commonly-used graph classification datasets, with the mean cross-validation accuracy averaged over 10 repeats and sample standard deviations ($\pm$).
        Hyperparameters of entries marked with * are known to be selected based on test accuracy instead of validation accuracy, so results are likely not comparable to other existing approaches that were (hopefully) selected based on validation accuracy.
        Our results were selected based on validation accuracy.
    }
    \label{fspool tab:graph}
    \setlength{\tabcolsep}{5pt}
    \centerfloat{\begin{tabular}{lccccc}
    \toprule
    \textit{Social Network} & IMDB-B & IMDB-M & RDT-B & RDT-M5K & COLLAB \\
    \midrule
    Num. graphs & 1000 & 1500 & 2000 & 5000 & 5000 \\
    Num. classes & 2 & 3 & 2 & 5 & 3 \\
    Avg. nodes & 19.8 & 13.0 & 429.6 & 508.5 & 74.5 \\
    Max. nodes & 136 & 89 & 3063 & 2012 & 492 \\
    \midrule
    \textsc{DCNN} \citep{Atwood2016} & 49.1 & 33.5 & -- & -- & 52.1 \\
    \textsc{Patchy-san} \citep{Niepert2016} & 71.0 \tiny$\pm$2.3 & 45.2 \tiny$\pm$2.8 & 86.3 \tiny$\pm$1.6 & 49.1 \tiny$\pm$0.7 & 72.6 \tiny$\pm$2.2 \\
    \textsc{SortPool} \citep{Zhang2018} & 70.0 \tiny$\pm$0.9 & 47.8 \tiny$\pm$0.9 & -- & -- & 73.8 \tiny$\pm$0.5 \\
    \textsc{DiffPool} \citep{Ying2018} & -- & -- & -- & -- & 75.5 \\
    \textsc{WL*} \citep{Xu2019} & 73.8 & 50.9 & 81.0 & 52.5 & 78.9 \\
    \textsc{GIN-Base*} \citep{Xu2019} & 75.1 & 52.3 & 92.4 & 57.5 & 80.2 \\
    \midrule
    \textsc{GIN-FSPool}     & \textbf{72.1} \tiny$\pm$2.0 & \textbf{49.9} \tiny$\pm$1.7 & \textbf{89.1} \tiny$\pm$1.2 & \textbf{51.8} \tiny$\pm$0.9 & 80.0 \tiny$\pm$0.4 \\
    - \textit{epochs} & 95 \tiny$\pm$70 & \textbf{27} \tiny$\pm$23 & \textbf{124} \tiny$\pm$64 & \textbf{66} \tiny$\pm$31 & \textbf{124} \tiny$\pm$56 \\
    \textsc{GIN-Base} & 71.3 \tiny$\pm$1.2 & 48.8 \tiny$\pm$1.7 & 84.8 \tiny$\pm$1.7 & 48.1 \tiny$\pm$2.0 & \textbf{80.3} \tiny$\pm$0.4 \\
    - \textit{epochs} & \textbf{83} \tiny$\pm$73 & 57 \tiny$\pm$59 & 156 \tiny$\pm$58 & 211 \tiny$\pm$27 & 204 \tiny$\pm$26 \\
    \bottomrule
    \end{tabular}}

    \vspace{5mm}

    \begin{tabular}{lcccc}
    \toprule
    \textit{Bioinformatics} & MUTAG & PROTEINS & PTC & NCI1 \\
    \midrule
    Num. graphs & 188 & 1113 & 344 & 4110 \\
    Num. classes & 2 & 2 & 2 & 2 \\
    Avg. nodes & 17.9 & 39.1 & 25.5 & 29.8 \\
    Max. nodes & 28 & 620 & 109 & 111 \\
    \midrule
    \textsc{PK} \citep{Neumann2016} & 76.0 \tiny$\pm$2.7 & 73.7 \tiny$\pm$0.7 & 59.5 \tiny$\pm$2.4 & 82.5 \tiny$\pm$0.5 \\
    \textsc{DCNN} \citep{Atwood2016} & 67.0 & 61.3 & 56.6 & 62.6 \\
    \textsc{Patchy-san} \citep{Niepert2016} & 92.6 \tiny$\pm$4.2 & 75.9 \tiny$\pm$2.8 & 60.0 \tiny$\pm$4.8 & 78.6 \tiny$\pm$1.9\\
    \textsc{SortPool} \citep{Zhang2018} & 85.8 \tiny$\pm$1.7 & 75.5 \tiny$\pm$0.9 & 58.6 \tiny$\pm$2.5 & 74.4 \tiny$\pm$0.5 \\
    \textsc{DiffPool} \citep{Ying2018} & -- & 76.3 & -- & -- \\
    \textsc{WL} \citep{Shervashidze2011} & 84.1 \tiny$\pm$1.9 & 74.7 \tiny$\pm$0.5 & 58.0 \tiny$\pm$2.5 & 85.5 \tiny$\pm$0.5 \\
    \textsc{WL*} \citep{Xu2019} & 90.4 & 75.0 & 59.9 & 86.0 \\
    \textsc{GIN-Base*} \citep{Xu2019} &  89.4 & 76.2 & 64.6 & 82.7 \\
    \midrule
    \textsc{GIN-FSPool}     & \textbf{85.9} \tiny$\pm$2.4 & \textbf{73.8} \tiny$\pm$0.9 & 59.3 \tiny$\pm$1.8 & 79.2 \tiny$\pm$0.6 \\
    - \textit{epochs} & 299 \tiny$\pm$91 & \textbf{69} \tiny$\pm$23 & 214 \tiny$\pm$110 & \textbf{361} \tiny$\pm$54 \\
    \textsc{GIN-Base} & 85.0 \tiny$\pm$1.5 & 73.2 \tiny$\pm$1.2 & \textbf{59.9} \tiny$\pm$2.4 & \textbf{79.4} \tiny$\pm$0.6 \\
    - \textit{epochs} & \textbf{244} \tiny$\pm$95 & 160 \tiny$\pm$123 & \textbf{202} \tiny$\pm$100 & 412 \tiny$\pm$55 \\
    \bottomrule
    \end{tabular}
\end{table}

\paragraph{Results}
We show the results in \autoref{fspool tab:graph}.
On 6 out of 9 datasets, GIN-FSPool achieves better test accuracy than GIN-Base.
On a different 6 datasets, it converges to the best validation accuracy faster.

The most significant improvements are on the two RDT datasets.
Interestingly, these are the two datasets where the number of nodes to be pooled is by far the largest with an average of 400+ nodes per graph, compared to the next largest COLLAB with an average of only 75 nodes.
This is perhaps evidence that FSPool is helping to avoid the bottleneck problem of pooling a large set of feature vectors to a single feature vector.
Mean and sum pooling are mostly able to keep up with FSPool on the smaller set sizes, but are bottlenecked for the RDT datasets with the much larger set sizes.

A Wilcoxon signed-rank test shows that the difference in accuracy to the standard GIN has $p \approx 0.07$ ($W = 7$) and the difference in convergence speed has $p \approx 0.11$ ($W = 9$).
Keep in mind that just because the results have $p > 0.05$, it does not mean that the results are invalid.

We emphasise that the main comparison to be made is between the GIN-Sum and the GIN-FSPool model in the last four rows, since that is the only comparison where the only factor of difference is the pooling method.
When comparing against other models, the network architecture, training hyperparameters, and evaluation methodology can differ significantly.

Keep in mind that while GIN-Base looks much worse than the original GIN-Base*, the difference is that our implementation has hyperparameters properly selected by validation accuracy, while GIN-Base* selected them by test accuracy.
If we were to select based on test accuracy, our implementation frequently outperforms their results.
Also, they only performed a single run of 10-fold cross-validation, while our results are averaged over 10 repeats of 10-fold cross validation.
%



\section{Discussion}\label{fspool sec:discussion}
In this chapter, we identified the responsibility problem with existing approaches for predicting sets and introduced FSPool, which provides a way around this issue in auto-encoders.
In experiments on two datasets of point clouds, we showed that this results in much better reconstructions.
We believe that this is an important step towards set prediction tasks with more complex set elements.
However, because our decoder uses information from the encoder, it is not easily possible to turn it into a generative set model, which is the main limitation of our approach.
Still, we find that using the auto-encoder to obtain better representations and pre-trained weights can be beneficial by itself.
We will use our insights about the responsibility problem to create a model without the auto-encoder limitation in \autoref{cha:dspn}.

In classification experiments, we also showed that simply replacing the pooling function in an existing model with FSPool can give us better results and faster convergence.
We showed that FSPool consistently learns better set representations at a relatively small computational cost, leading to improved results in the downstream task.
Our model thus has immediate applications in various types of set models that have traditionally used sum or max pooling.
For example, we will use FSPool in the next chapter (replacing the pooling in a Relation Network) to improve the quality of the learned representations.
It would be useful to theoretically characterise what types of relations are more easily expressed by FSPool through an analysis like in~\citet{Murphy2019}.
This may result in further insights into how to learn better set representations efficiently.




\chapter{Set decoder: Deep set prediction networks}\label{cha:dspn}
\chaptermark{Deep set prediction networks}

In the previous chapter, we found that all existing approaches for predicting sets suffer from the responsibility problem.
The solution we provided there only covered the auto-encoder setting.
In this chapter, we will develop a solution without this restriction: a model that can predict sets without the responsibility problem in normal supervised learning tasks.

These contributions have been published as~\citep{Zhang2019DSPN} in Advances in Neural Information Processing Systems 32 (NeurIPS), 2019, and have been presented at the Sets \& Partitions workshop, hosted at the Neural Information Processing Systems (NeurIPS) 2019 conference.

\section{Introduction}
You are given a rotation angle and your task is to draw the four corner points of a square that is rotated by that amount.
This is a structured prediction task where the output is a \emph{set}, since there is no inherent ordering to the four points.
Such sets are a natural representation for many kinds of data, ranging from the set of points in a point cloud, to the set of objects in an image (object detection), to the set of nodes in a molecular graph (molecular generation).
Yet, existing machine learning models often struggle to solve even the simple square prediction task as we showed in \autoref{fspool sec:problem}.

The main difficulty in predicting sets comes from the ability to permute the elements in a set freely, which means that there are $n!$ equally good solutions for a set of size $n$.
Models that do not take this set structure into account properly (such as MLPs or RNNs) result in discontinuities, which is the reason why they struggle to solve simple toy set prediction tasks.
We quickly review the background on what the problem is in \autoref{dspn sec:background}.


How can we build a model that properly respects the set structure of the problem so that we can predict sets without running into discontinuity issues?
In this chapter, we aim to address this question. 
Concretely, we contribute the following:

\begin{enumerate}
    \item We propose a model (\autoref{dspn sec:model}, \autoref{alg:dspn}) that can predict a set from a feature vector (vector-to-set) while properly taking the structure of sets into account.
        We explain what properties we make use of that enables this.
        Our model uses backpropagation through a set encoder to decode a set and works for variable-size sets.
        The model is applicable to a wide variety of set prediction tasks since it only requires a feature vector as input.
    \item We evaluate our model on several set prediction datasets (\autoref{dspn sec:experiments}).
        First, we demonstrate that the auto-encoder version of our model is sound on a set version of MNIST.
        Next, we use the CLEVR dataset to show that this works for general set prediction tasks.
        We predict the set of bounding boxes of objects in an image and we predict the set of object attributes in an image, both from a single feature vector.
        Our model is a completely different approach to usual anchor-based object detectors because we pose the task as a set prediction problem, which does not need complicated post-processing techniques such as non-maximum suppression.
\end{enumerate}

\section{Background}\label{dspn sec:background}
\paragraph{Representation}
We are interested in sets of feature vectors with the feature vector describing properties of the element, for example the 2d position of a point in a point cloud.
A set of size $n$ wherein each feature vector has dimensionality $d$ is represented as a matrix $\mY \in \R^{n \times d}$ with the elements as rows in an arbitrary order, $\mY = [\vy_1, \ldots, \vy_n]^\T$.
Note that this representation of sets is transposed compared to \autoref{cha:background} ($\R^{n \times d}$ instead of $\R^{d \times n}$) to ease the exposition in this chapter.
To properly treat this as a set, it is important to only apply operations with certain properties to it~\citep{Zaheer2017}: \emph{permutation-invariance} or \emph{permutation-equivariance}.
In other words, operations on sets should not rely on the arbitrary ordering of the elements.

Set encoders (which turn such sets into feature vectors) are usually built by composing permutation-equivariant operations with a permutation-invariant operation at the end.
A simple example is the Deep Sets model by \citet{Zaheer2017}: $f(\mY) = \sum_i g(\vy_i)$ where $g$ is a neural network.
Because $g$ is applied to every element individually, it does not rely on the arbitrary order of the elements.
We can think of this as turning the set $\{\vy_i\}_{i=1}^n$ into $\{g(\vy_i)\}_{i=1}^n$.
This is permutation-equivariant because changing the order of elements in the input set affects the output set in a predictable way.
Next, the set is summed to produce a single feature vector.
Since summing is commutative, the output is the same regardless of what order the elements are in.
In other words, summing is permutation-invariant.
This gives us an encoder that produces the same feature vector regardless of the arbitrary order the set elements were stored in.

\paragraph{Loss}
In set prediction tasks, we need to compute a loss between a predicted set $\mhY = [\vhy_1, \ldots, \vhy_n]^\T$ and the target set $\mY$.
The main problem is that the elements of each set are in an arbitrary order, so we cannot simply compute a pointwise distance.
The usual solution to this is an assignment mechanism that matches up elements from one set to the other set.
This gives us a loss function that is permutation-invariant in both its arguments.

One such loss is the $O(n^2)$ Chamfer loss, which matches up every element of $\mhY$ to the closest element in $\mY$ and vice versa:

\begin{equation}
    \Lcha(\mhY, \mY) = \sum_i \min_j \norm{ \vhy_i - \vy_j }^2 + \sum_j \min_i \norm{ \vhy_i - \vy_j }^2
\end{equation}

Note that this does not work well for multisets: the loss between $[\va, \va, \vb]$, $[\va, \vb, \vb]$ is 0.
A more sophisticated loss that does not have this problem involves the linear assignment problem with the pairwise losses as assignment costs:

\begin{equation}
    \Lhun(\mhY, \mY) = \min_{\pi \in \Pi} \sum_i \norm{ \vhy_i - \vy_{\pi(i)} }^2
\end{equation}

where $\Pi$ is the space of permutations, which can be solved with the Hungarian algorithm in $O(n^3)$ time.
This has the benefit that every element in one set is associated to exactly one element in the other set, which is not the case for the Chamfer loss.


\paragraph{Responsibility problem}


A widely-used approach is to simply ignore the set structure of the problem.
A feature vector can be mapped to a set $\mhY$ by using an MLP that takes the vector as input and directly produces $\mhY$ with $n \times d$ outputs.
Since the order of elements in $\mhY$ does not matter, it appears reasonable to always produce them in a certain order based on the weights of the MLP.

While this seems like a promising approach, we pointed out \autoref{fspool sec:problem} that this results in a discontinuity issue: there are points where a small change in set space requires a large change in the neural network outputs.
The model needs to ``decide'' which of its outputs is responsible for producing which element, and this responsibility must be resolved discontinuously.


%
Here, we give a different example to give the intuition behind this problem.
Consider an MLP that detects the colour of two otherwise identical objects present in an image, so it has two outputs with dimensionality 3 (R, G, B) corresponding to those two colours.
We are given an image with a blue and red object, so let us say that output 1 predicts blue and output 2 predicts red; perhaps the weights of output 1 are more attuned to the blue channel and output 2 is more attuned to the red channel.
We are given another image with a blue and green object, so it is reasonable for output 1 to again predict blue and output 2 to now predict green.
When we now give the model an image with a red and green object, or two red objects, it is unclear which output should be responsible for predicting which object.
Output 2 ``wants'' to predict both red and green, but has to decide between one of them, and output 1 now has to be responsible for the other object while previously being a blue detector.
This responsibility must be resolved discontinuously, which makes modeling sets with MLPs difficult.

The main problem is that there is a notion of output 1 and output 2 -- an ordered output representation -- in the first place, which forces the model to give the set an order.
Instead, it would be better if the outputs of the model were freely interchangeable -- in the same way the elements of the set are interchangeable --  to not impose an order on the outputs.
This is exactly what our model accomplishes.

\sectionmark{Model}
\section{Deep set prediction networks}\label{dspn sec:model}
\sectionmark{Model}
This section contains our primary contribution: a model for decoding a feature vector into a set of feature vectors.
As we have previously established, it is important for the model to properly respect the set structure of the problem to avoid the responsibility problem.

Our main idea is based on the observation that the gradient of a set encoder with respect to the input set is permutation-equivariant (see \autoref{dspn app:proof}):
\emph{to decode a feature vector into a set, we can use gradient descent to find a set that \emph{encodes} to that feature vector}.
Since each update of the set using the gradient is permutation-equivariant, we always properly treat it as a set and avoid the responsibility problem.
This gives rise to a nested optimisation: an inner loop that changes a set to encode more similarly to the input feature vector, and an outer loop that changes the weights of the encoder to minimise a loss over a dataset.

With this idea in mind, we build up models of increasing usefulness for predicting sets.
We start with the simplest case of auto-encoding fixed-size sets (\autoref{dspn sec:fixed}), where a latent representation is decoded back into a set.
This is modified to support variable-size sets, which is necessary for most sets encountered in the real-world.
Lastly and most importantly, we extend our model to general set prediction tasks where the input no longer needs to be a set (\autoref{dspn sec:predict}).
This gives us a model that can predict a set of feature vectors from a single feature vector.
We give the pseudo-code of this method in \autoref{alg:dspn}.

\newlength{\insidetextwidth}
\setlength{\insidetextwidth}{\textwidth}
\begin{tcolorbox}[blanker, float=t, oversize=2cm]
    \begin{algorithm}[H]
        \caption[Forward pass of set prediction algorithm]{
        One forward pass of the set prediction algorithm within the training loop.
        }
        \label{alg:dspn}
        \centering
        \begin{algorithmic}[1]

            \State $\vz = F(x)$
            \Comment{encode input with a model}

            \State $\mhY^{(0)} \gets \text{init}$
            \Comment{initialise set}

            \For{$\text{t} \gets 1, T$}
                \State $l \gets \Lrepr(\mhY^{(t - 1)}, \vz)$
                \Comment{compute representation loss}

                \State $\mhY^{(t)} \gets \mhY^{(t - 1)} - \eta \frac{\partial l}{\partial \mhY^{(t - 1)}}$
                \Comment{gradient descent step on the set}
            \EndFor

            \State predict $\mhY^{(T)}$

            \State $\Ls = \frac{1}{T} \sum_{t=0}^T \Lset(\mhY^{(t)}, \mY) + \lambda \Lrepr(\mY, \vz)$
            \Comment{outer optimisation loss}
        \end{algorithmic}
    \end{algorithm}
\end{tcolorbox}

\subsection{Proof of permutation-equivariance}\label{dspn app:proof}

Recall definitions of permutation-invariance and equivariance:

\begin{definition}
    A function $f: \R^{n \times d} \to \R^{c}$ is permutation-invariant iff it satisfies:

    \begin{equation}
        f(\mX) = f(\mP\mX)
    \end{equation}

    for all permutation matrices $\mP$.
\end{definition}

\begin{definition}
    A function $g: \R^{n \times d} \to \R^{n \times c}$ is permutation-equivariant iff it satisfies:

    \begin{equation}
        \mP g(\mX) = g(\mP\mX)
    \end{equation}
    
    for all permutation matrices $\mP$.
\end{definition}

The definitions here use the transposed set representations to match the exposition in this chapter.
With these definitions, we can prove the following:

\begin{thm}
    The gradient of a permutation-invariant function $f: \R^{n \times d} \to \R^c$ with respect to its input is permutation-equivariant:

    \begin{equation}
        \mP \frac{\partial f(\mX)}{\partial \mX}
        = \frac{\partial f(\mP\mX)}{\partial \mP\mX}
    \end{equation}

\end{thm}

\begin{proof}
    Using Definition 1, the chain rule, and the orthogonality of $\mP$:

    \begin{align}
        \mP \frac{\partial f(\mX)}{\partial \mX}
        &= \mP \frac{\partial f(\mP\mX)}{\partial \mX} \\
        &= \mP \frac{\partial \mP\mX}{\partial \mX} \frac{\partial f(\mP\mX)}{\partial \mP\mX} \\
        &= \mP \mP^\T \frac{\partial f(\mP\mX)}{\partial \mP\mX} \\
        &= \frac{\partial f(\mP\mX)}{\partial \mP\mX}
    \end{align}
\end{proof}

This property ensures that our model is permutation-equivariant.

\subsection{Auto-encoding fixed-size sets}\label{dspn sec:fixed}
In a set auto-encoder, the goal is to turn the input set $\mY$ into a small latent space $\vz = \enc(\mY)$ with the encoder $\enc$ and turn it back into the predicted set $\mhY = \dec(\vz)$ with the decoder $\dec$.
Using our main idea, we define a \emph{representation loss} and the corresponding decoder as:

\begin{equation}
    \Lrepr(\mhY, \vz) = \norm{ \enc(\mhY) - \vz }^2
\end{equation}%
\begin{equation}\label{dspn eq:dec}
    \dec(\vz) = \argmin_{\mhY} \Lrepr(\mhY, \vz)
\end{equation}

In essence, $\Lrepr$ compares $\mhY$ to $\mY$ in the latent space.
To understand what the decoder does, first consider the simple, albeit not very useful case of the identity encoder $\enc(\mY) = \mY$.
Solving $\dec(\vz)$ simply means setting $\mhY = \mY$, which perfectly reconstructs the input as desired.

When we instead choose $\enc$ to be a set encoder, the latent representation $\vz$ is a permutation-invariant feature vector.
If this representation is ``good'', $\mhY$ will only encode to similar latent variables as $\mY$ if the two sets themselves are similar.
Thus, the minimisation in \autoref{dspn eq:dec} should still produce a set $\mhY$ that is the same (up to permutation) as $\mY$, except this has now been achieved with $\vz$ as a bottleneck.

Since the problem is non-convex when $\enc$ is a neural network, it is infeasible to solve \autoref{dspn eq:dec} exactly.
Instead, we perform gradient descent to approximate a solution.
Starting from some initial set $\mhY^{(0)}$, gradient descent is performed for a fixed number of steps $T$ with the update rule:

\begin{equation}
    \mhY^{(t+1)} = \mhY^{(t)} - \eta \cdot \frac{\partial \Lrepr(\mhY^{(t)}, \vz)}{\partial \mhY^{(t)}}
\end{equation}

with $\eta$ as the learning rate and the prediction being the final state, $\dec(\vz) = \mhY^{(T)}$.
This is the aforementioned inner optimisation loop.
In practice, we let $\mhY^{(0)}$ be a learnable $\R^{d \times n}$ matrix which is part of the neural network parameters.

\begin{figure}
    \centering
    \centerfloat{\resizebox{\figurewidth}{!}{%
        \fontsize{32}{32}\selectfont
        \begin{tikzpicture}
            \node[inner sep=0] (input) at (0, 0) {\includegraphics[scale=0.7, trim={3.5mm 3.5mm 3.5mm 3.5mm}, clip]{poster/resources/mnist-32-11.pdf}};
            \node[inner sep=0] (target) at (20.5, 0) {\includegraphics[scale=0.7, trim={3.5mm 3.5mm 3.5mm 3.5mm}, clip]{poster/resources/mnist-32-11.pdf}};
            \node[inner sep=0] (y0) at (0, -4) {\includegraphics[scale=0.7, trim={3.5mm 3.5mm 3.5mm 3.5mm}, clip]{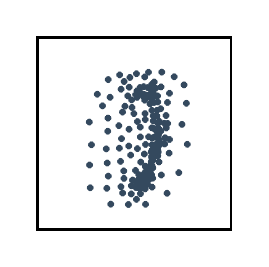}};
            \node[inner sep=0] (y1) at (8, -4) {\includegraphics[scale=0.7, trim={3.5mm 3.5mm 3.5mm 3.5mm}, clip]{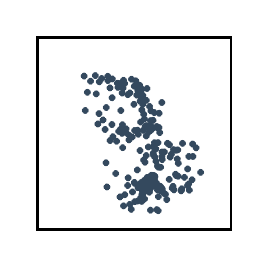}};
            \node[inner sep=0] (y2) at (16, -4) {\includegraphics[scale=0.7, trim={3.5mm 3.5mm 3.5mm 3.5mm}, clip]{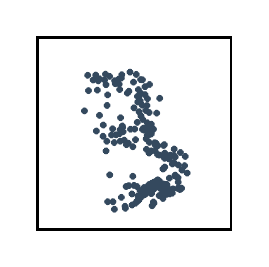}};
            \node[inner sep=0] (y10) at (20.5, -4) {\includegraphics[scale=0.7, trim={3.5mm 3.5mm 3.5mm 3.5mm}, clip]{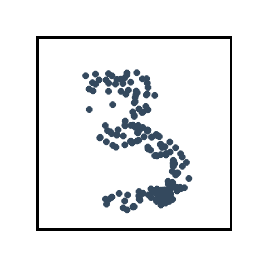}};

            \node (fvinput bottom)[box, scale=0.36, fill=Oranges-4-3] at (4, -0.5) {};
            \node (fvinput) [box, scale=0.36, fill=Blues-4-4] at (4, 0) {};
            \node [box, scale=0.36, fill=Blues-4-4] at (4, 0.5) {};

            \node (fvinput2 bottom)[box, scale=0.36, fill=Oranges-4-3] at (12, -0.5) {};
            \node (fvinput2) [box, scale=0.36, fill=Blues-4-4] at (12, 0) {};
            \node [box, scale=0.36, fill=Blues-4-4] at (12, 0.5) {};
            \node (fv0 top) [box, scale=0.36, fill=Oranges-4-1] at (4, -3.5) {};
            \node (fv0) [box, scale=0.36, fill=Purples-4-1] at (4, -4) {};
            \node [box, scale=0.36, fill=Oranges-4-1] at (4, -4.5) {};
            \node (fv1 top) [box, scale=0.36, fill=Blues-4-3!50!Blues-4-4] at (12, -3.5) {};
            \node (fv1) [box, scale=0.36, fill=Blues-4-3] at (12, -4) {};
            \node [box, scale=0.36, fill=Oranges-4-2] at (12, -4.5) {};

            \draw [sedge, in=200, out=-20] (y0.south east) to (y1.south west);
            \draw [sedge, in=200, out=-20] (y1.south east) to (y2.south west);
            
            \draw [ultra thick, color=colorTwo] (fvinput bottom) -- (fv0 top) node (mse0) [midway, scale=0.2, fill=white] {MSE};
            \draw [ultra thick, color=colorTwo] (fvinput2 bottom) -- (fv1 top) node (mse1) [midway, scale=0.2, fill=white] {MSE};

            \draw [sedge] (input) -- (fvinput) node [midway, scale=0.2, trapezium, rotate=-90, trapezium angle=70, inner ysep=12mm, draw, fill=colorThree!10!white] {Encoder};
            \draw [sedge] (y0) -- (fv0) node [midway, scale=0.2, trapezium, rotate=-90, trapezium angle=70, inner ysep=12mm, draw, fill=colorThree!10!white] {Encoder};
            \draw [sedge] (y1) -- (fv1) node [midway, scale=0.2, trapezium, rotate=-90, trapezium angle=70, inner ysep=12mm, draw, fill=colorThree!10!white] {Encoder};

            \draw [sedge, color=colorTwo] (mse0) -- (y1.north west)
            node [midway, scale=0.2, sloped, above] {$- \partial \text{ MSE}$}
            node [midway, scale=0.2, sloped, below] {$\partial \text{ Step 0}$};
            \draw [sedge, color=colorTwo] (mse1) -- (y2.north west)
            node [midway, scale=0.2, sloped, above] {$- \partial \text{ MSE}$}
            node [midway, scale=0.2, sloped, below] {$\partial \text{ Step 1}$};

            \node [scale=0.2, below = 0mm of y0] {Step 0};
            \node [scale=0.2, below = 0mm of y1] {Step 1};
            \node [scale=0.2, below = 0mm of y2] {Step 2};                        
            \node [scale=0.2, below = 0mm of y10] {Step 10};
            \node [scale=0.2, above = 0mm of input] {Input};
            \node [scale=0.2, above = 0mm of target] {Target};

            \draw [ultra thick, color=colorThree!50!black] (y10) -- (target) node [midway, fill=white, scale=0.2] {set loss};

            \node [scale=0.4] at ($(y2)!0.5!(y10)$) {\ldots};
        \end{tikzpicture}
    }}
    \caption[Overview of DSPN for auto-encoding]{Overview of our algorithm for auto-encoding an MNIST set.
    The initial set at step 0 is iteratively refined, at each step improving the current prediction.
    }
    \label{dspn fig:overview mnist}
\end{figure}

To obtain a good representation $\vz$, we still have to train the weights of $\enc$.
For this, we compute the auto-encoder objective $\Lset(\mhY^{(T)}, \mY)$ -- with $\Lset=\Lcha$ or $\Lhun$ -- and differentiate with respect to the weights as usual, backpropagating through the steps of the inner optimisation.
This is the aforementioned outer optimisation loop.

In summary, each forward pass of our auto-encoder first encodes the input set to a latent representation as normal.
To decode this back into a set, gradient descent is performed on an initial guess with the aim to obtain a set that encodes to the same latent representation as the input.
The same set encoder is used in the encoding and decoding stages.

\paragraph{Variable-size sets}
To extend this from fixed- to variable-size sets, we make a few modifications to this algorithm.
First, we pad all sets to a fixed maximum size to allow for efficient batch computation.
We then concatenate an additional mask feature $m_i$ to each set element $\vhy_i$ that indicates whether it is a regular element ($m_i=1$) or padding ($m_i=0$).
With this modification to $\mhY$, we can optimise the masks in the same way as the set elements are optimised.
To ensure that masks stay in the valid range between 0 and 1, we simply clamp values above 1 to 1 and values below 0 to 0 after each gradient descent step.
This performed better than using a sigmoid in our initial experiments, possibly because it allows exact 0s and 1s to be recovered.

\begin{figure}
    \centering
    \centerfloat{\resizebox{\figurewidth}{!}{%
        \begin{tikzpicture}
            \fontsize{32}{32}\selectfont
            \node[inner sep=0] (input) at (0, 1) {\includegraphics[scale=0.7, trim={3.5mm 3.5mm 3.5mm 3.5mm}, clip]{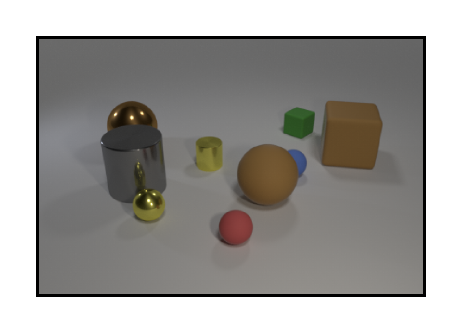}};
            \node[inner sep=0] (target) at (18, 1) {\includegraphics[scale=0.7, trim={3.5mm 3.5mm 3.5mm 3.5mm}, clip]{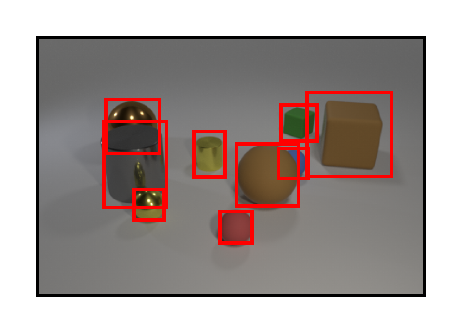}};
            \node[inner sep=0] (y0) at (0, -4) {\includegraphics[scale=0.7, trim={3.5mm 3.5mm 3.5mm 3.5mm}, clip]{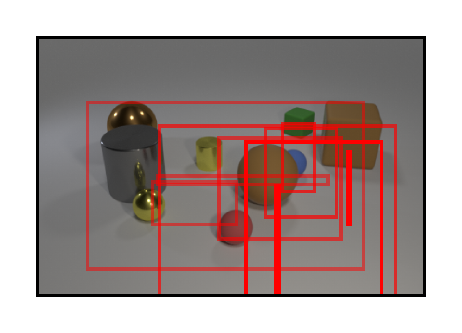}};
            \node[inner sep=0] (y1) at (11, -4) {\includegraphics[scale=0.7, trim={3.5mm 3.5mm 3.5mm 3.5mm}, clip]{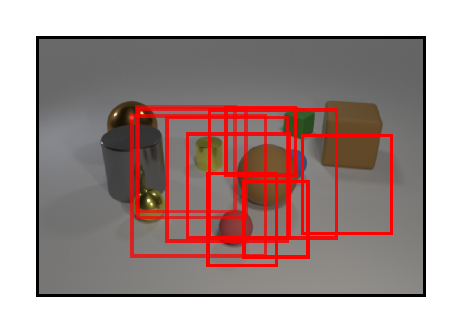}};
            \node[inner sep=0] (y10) at (18, -4) {\includegraphics[scale=0.7, trim={3.5mm 3.5mm 3.5mm 3.5mm}, clip]{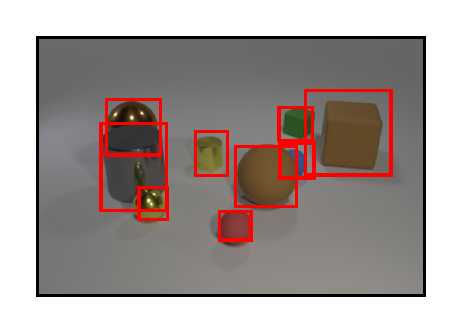}};

            \node (fvinput bottom)[box, scale=0.36, fill=Purples-4-4] at (6, 0.5) {};
            \node (fvinput) [box, scale=0.36, fill=Blues-4-4] at (6, 1) {};
            \node (fvinput top) [box, scale=0.36, fill=Oranges-4-3] at (6, 1.5) {};

            \node (fvtarget bottom)[box, scale=0.36, fill=Purples-4-4] at (11, 0.5) {};
            \node (fvtarget) [box, scale=0.36, fill=Blues-4-4] at (11, 1) {};
            \node [box, scale=0.36, fill=Oranges-4-3] at (11, 1.5) {};
            \node (fv0 top) [box, scale=0.36, fill=Oranges-4-1] at (6, -3.5) {};
            \node (fv0) [box, scale=0.36, fill=Blues-4-1] at (6, -4) {};
            \node [box, scale=0.36, fill=Purples-4-1] at (6, -4.5) {};

            \draw [sedge, in=200, out=-20] (y0.south east) to (y1.south west);
            
            \draw [ultra thick, color=colorTwo] (fvinput bottom) -- (fv0 top) node (mse0) [midway, scale=0.2, fill=white] {MSE};
            \draw [ultra thick, color=colorThree!50!black] (fvtarget) -- (fvinput) node [midway, scale=0.2, fill=white] {MSE loss};

            \draw [sedge] (input) -- (fvinput) node [midway, scale=0.2, trapezium, rotate=-90, trapezium angle=70, inner ysep=22mm, draw, fill=colorThree!10!white] {ResNet34};
            \draw [sedge] (y0) -- (fv0) node [midway, scale=0.2, trapezium, rotate=-90, trapezium angle=70, inner ysep=12mm, draw, fill=colorThree!10!white] {Encoder};
            \draw [sedge] (target) -- (fvtarget) node [midway, scale=0.2, trapezium, rotate=90, trapezium angle=70, inner ysep=12mm, draw, fill=colorThree!10!white] {Encoder};

            \draw [sedge, color=colorTwo] (mse0) -- (y1.north west)
            node [midway, scale=0.2, sloped, above] {$- \partial \text{ MSE}$}
            node [midway, scale=0.2, sloped, below] {$\partial \text{ Step 0}$};

            \node [scale=0.2, below = 0mm of y0] {Step 0};
            \node [scale=0.2, below = 0mm of y1] {Step 1};
            \node [scale=0.2, below = 0mm of y10] {Step 10};
            \node [scale=0.2, above = 0mm of input] {Input};
            \node [scale=0.2, above = 0mm of target] {Target};

            \draw [ultra thick, color=colorThree!50!black] (y10) -- (target) node [midway, fill=white, scale=0.2] {set loss};

            \node [scale=0.4] at ($(y1)!0.5!(y10)$) {\ldots};

        \end{tikzpicture}
    }}
    \caption[Overview of DSPN for supervised prediction]{Overview of DSPN for supervised prediction. The main difference is the MSE loss between the input encoding and target encoding.}
    \label{dspn fig:overview box}
\end{figure}
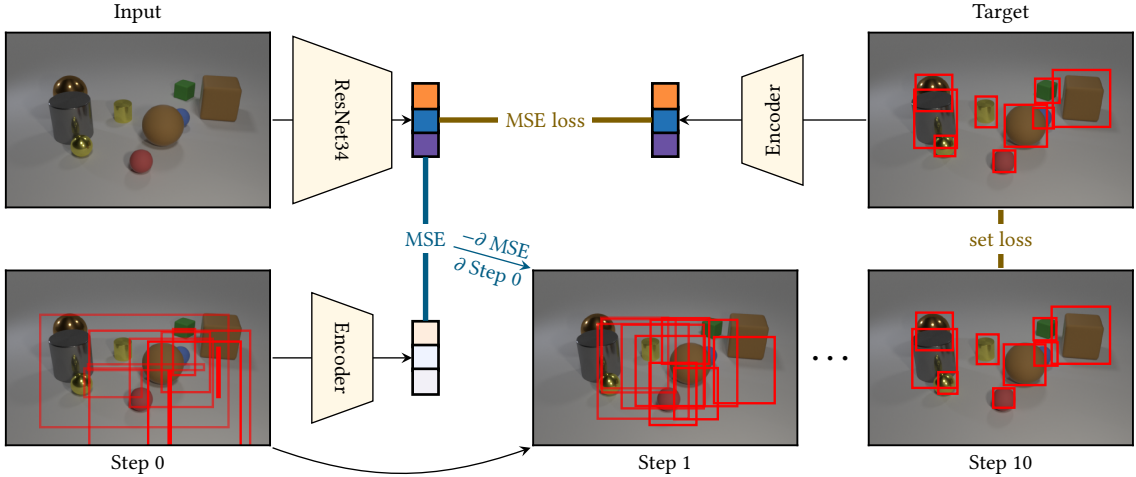

\subsection{Predicting sets from a feature vector}\label{dspn sec:predict}
In our auto-encoder, we used an encoder to produce both the latent representation as well as to decode the set.
This is no longer possible in the general set prediction setup, since the target representation $\vz$ can come from a separate model (for example an image encoder $F$ encoding an image $\vx$), so there is no longer a set encoder in the model.

When na\"ively using $\vz = F(\vx)$ as input to our decoder, our decoding process is unable to predict sets correctly from it.
Because the set encoder is no longer shared in our set decoder, there is no guarantee that optimising $\enc(\mhY)$ to match $\vz$ converges towards $\mY$ (or a permutation thereof).
To fix this, we simply add a term to the loss of the outer optimisation that encourages $\enc(\mY) \approx \vz$ again.
In other words, the target set should have a very low representation loss itself.
This gives us an additional $\Lrepr$ term in the loss function of the outer optimisation for supervised learning:

\begin{equation}
    \Ls = \Lset(\mhY, \mY) + \lambda \Lrepr(\mY, \vz)
\end{equation}

with $\Lset$ being either $\Lcha$ or $\Lhun$.
With this, minimising $\Lrepr(\mhY, \vz)$ in the inner optimisation will converge towards $\mY$.
The additional term is not necessary in the pure auto-encoder because $\vz = \enc(\mY)$, so $\Lrepr(\mY, \vz)$ is always 0 already.

\paragraph{Practical tricks}

For the outer optimisation, we can compute the set loss for not only $\mhY^{(T)}$, but all $\mhY^{(t)}$.
That is, we use the average set loss $\frac{1}{T} \sum_t \Lset(\mhY^{(t)}, \mY)$ as loss (similar to~\citet{Belanger2017}).
This encourages $\mhY$ to converge to $\mY$ quickly and not diverge with more steps, which increases the robustness of our algorithm.

We sometimes observed divergent training behaviour when the outer learning rate is set inappropriately.
By replacing the instances of $\norm{ \cdot }^2$ in $\Lset$ and $\Lrepr$ with the Huber loss (squared error for differences below 1 and absolute error above 1) -- as is commonly done in object detection models -- training became less sensitive to hyperparameter choices.

The inner optimisation can be modified to include a momentum term, which stops a prediction from oscillating around a solution.
This gives us slightly better results, but we did not use this for any experiments to keep our method as simple as possible.

It is possible to explicitly include the sum of masks as a feature in the representation $\vz$ for our model.
This improves our results on MNIST -- likely due to the explicit signal for the model to predict the correct set size -- but again, we do not use this for simplicity.

\section{Related work}\label{dspn sec:related}
The main approach we compare our method to is the simple method of using an MLP decoder to predict sets.
This has been used for predicting point clouds~\citep{Achlioptas2018, Fan2017}, bounding boxes~\citep{Rezatofighi2018, Balles2019}, and graphs (sets of nodes and edges) \citep{Cao2018, Simonovsky2018}.
These predict an ordered representation (list) and treat it as if it is unordered (set).
As we discussed in \autoref{dspn sec:background}, this approach runs into the responsibility problem.
Some works on predicting 3d point clouds make domain-specific assumptions such as independence of points within a set~\citep{Li2019} or grid-like structures~\citep{Yang2018}.
To avoid inefficient graph matching losses, \citet{Yang2019} compute a permutation-invariant loss between graphs by comparing them in the latent space (similar to our $\Lrepr$) in an adversarial setting.

An alternative approach is to use an RNN decoder to generate this list~\citep{Li2018a, Stewart2016, Vinyals2015}.
The problem can be made easier if it can be turned from a set into a sequence problem by giving a canonical order to the elements in the set through domain knowledge~\citep{Vinyals2015}.
For example, \citet{You2018} generate the nodes of a graph by ordering the set of nodes based on the node traversal order of a breadth-first search.

The closest work to ours is by \citet{Mordatch2018}.
They also iteratively minimise a function (their energy function) in each forward pass of the neural network and differentiate through the iteration to learn the weights.
They have only demonstrated that this works for modifying small sets of 2d elements in relatively simple ways, so it is unclear whether their approach scales to the harder problems such as object detection that we tackle in this chapter.
In particular, minimising $\Lrepr$ in our model has the easy-to-understand consequence of making the predicted set more similar to the target set, while it is less clear what minimising their learned energy function $E(\mhY, \vz)$ does.

In \autoref{cha:fspool}, we constructed an auto-encoder that pools a set into a feature vector where information from the encoder is shared with their decoder.
This is done to make the decoder permutation-equivariant, which we used to avoid the responsibility problem.
However, this strictly limits the decoder to usage in auto-encoders -- not set prediction -- because it requires an encoder to be present during inference.

\citet{Greff2019} construct an auto-encoder for images with a latent space that is set-structured.
They are able to find latent sets of variables to describe an image composed of a set of objects with some task-specific assumptions.
While interesting from a representation learning perspective, our model is immediately useful in practice because it works for general supervised learning tasks.

Our inspiration for using backpropagation through an encoder as a decoder comes from the line of introspective neural networks~\citep{Lazarow2017, Lee2018} for image modeling.
An important difference is that in these works, the two optimisation loops (generating predictions and learning the network weights) are performed in sequence, while ours are nested.
The nesting allows our outer optimisation to differentiate through the inner optimisation.
This type of nested optimisation to obtain structured outputs with neural networks was first studied by \citet{Belanger2016, Belanger2017}, of which our model can be considered an instance of.
Note that \citet{Greff2019} and \citet{Mordatch2018} also differentiate through an optimisation, which suggests that this approach is of general benefit when working with sets.By differentiating through a decoder rather than an encoder, \citet{Bojanowski2018} learn a representation instead of a prediction.+

It is important to clearly separate the vector-to-set setting in this chapter from some related works on set-to-set mappings, such as the equivariant version of Deep Sets~\citep{Zaheer2017} and self-attention~\citep{Vaswani2017}.
Tasks like object detection, where no set input is available, can not be solved with set-to-set methods alone;
the feature vector from the image encoder has to be turned into a set first, for which a vector-to-set model like ours is necessary.
Set-to-set methods do not have to deal with the responsibility problem, because the output usually has the same ordering as the input.
Methods like the one in~\citet{Mena2018} and our method in \autoref{cha:perm-optim} learn to predict a permutation matrix for a set (set-to-\emph{set-of-position-assignments}).
When this permutation is applied to the input set, the set is turned into a list (set-to-list).
Again, our model is about producing a set as \emph{output} while not necessarily taking a set as input.

\section{Experiments}\label{dspn sec:experiments}
In the following experiments, we compare our set prediction network to a model that uses an MLP or RNN (LSTM) as set decoder.
In all experiments, we fix the hyperparameters of our model to $T=10, \eta = 800, \lambda=0.1$.
Further details about the model architectures, training settings, and hyperparameters are given in \autoref{dspn app:details}.
We provide the PyTorch~\citep{Paszke2017} source code to reproduce all experiments at \url{https://github.com/Cyanogenoid/dspn}.

Note that there are two minor bugs (which do not affect the conclusions of the experiments) in the following results:
\begin{enumerate}
    \item the CLEVR experiments computed the set loss only on the last step $\mhY^{(T)}$,
    \item the bounding box results are consistently overestimated.
\end{enumerate}
\autoref{dspn app:fixed} shows our results with these bugs fixed.

\subsection{MNIST}\label{dspn sec:mnist}
We begin with the task of auto-encoding a set version of MNIST.
A set is constructed from each image by including all the pixel coordinates (x and y, scaled to the interval $[0, 1]$) of pixels that have a value above the mean pixel value.
The size of these sets varies from 32 to 342 across the dataset.

\paragraph{Model}
In our model, we use a set encoder that processes each element individually with a 3-layer MLP, followed by FSPool (\autoref{cha:fspool}) as pooling function to produce 64 latent variables.
These are decoded with our algorithm to predict the input set.
We compare this against a baseline model with the same encoder, but with a traditional MLP or LSTM as decoder.
This approach to decoding sets is used in models such as in~\citet{Achlioptas2018} (AE-CD variant) and~\citet{Stewart2016}; these baselines are representative of the best approaches for set prediction in the literature.
Note that these baselines have significantly more parameters than our model, since our decoder has almost no additional parameters by sharing the encoder weights (ours: $\sim$140 000 parameters, MLP: $\sim$530 000, LSTM: $\sim$470 000).
For the baselines, we include a mask feature with each element to allow for variable-size sets.
Due to the large maximum set size, use of Hungarian matching is too slow.
Instead, we use the Chamfer loss to compute the loss between predicted and target set in this experiment.

\begin{figure}
    \centering
    \centerfloat{
    \includegraphics[width=\figurewidth, trim=0 3mm 0 0, clip]{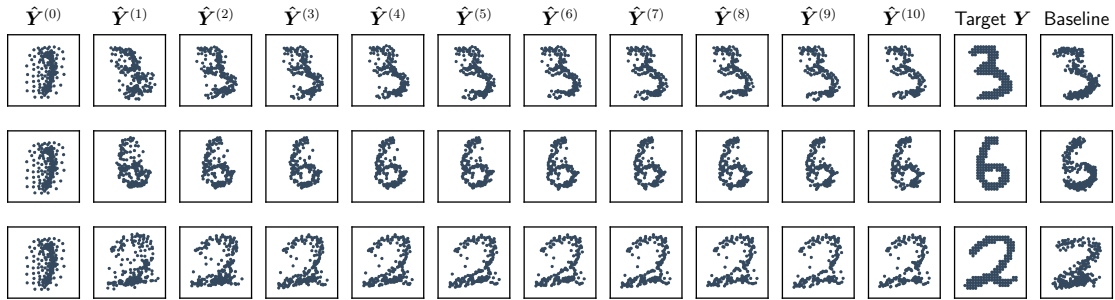}
    }
    \vspace{-3mm}
    \caption[Progression of set prediction algorithm on MNIST]{Progression of set prediction algorithm on MNIST ($\mhY^{(t)}$). Our predictions come from our model with $0.08 \times 10^{-3}$ loss, while the baseline predictions come from an MLP decoder model with $0.09 \times 10^{-3}$ loss.}
    \label{dspn fig:mnist}
\end{figure}

\begin{figure}
    \centering
    \centerfloat{
    \includegraphics[width=\figurewidth]{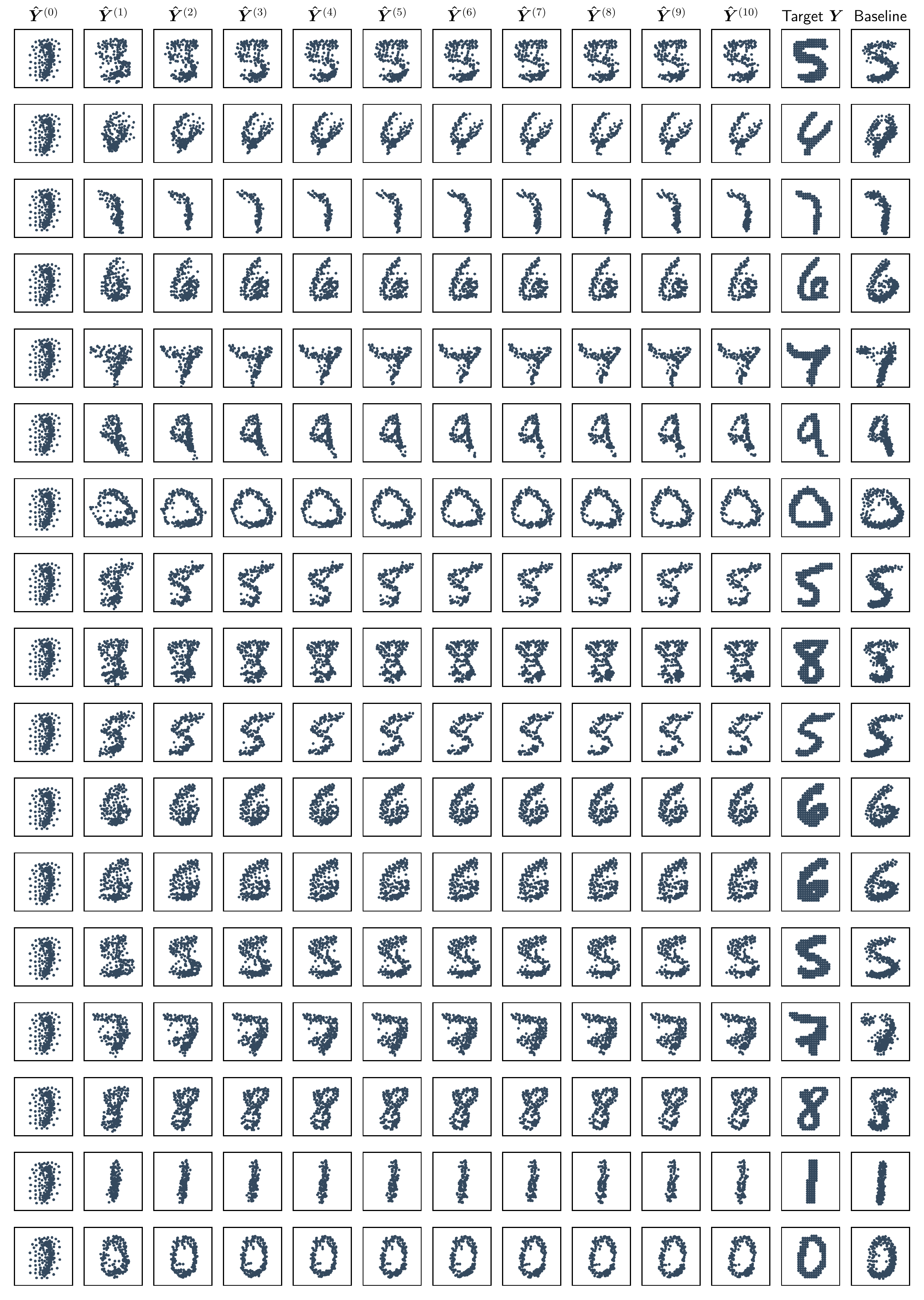}
    }
    \caption[More progression of set prediction algorithm on MNIST.]{Progression of set prediction algorithm on MNIST.}
    \label{dspn fig:mnist-full}
\end{figure}

\paragraph{Results}
\autoref{dspn tab:mnist} shows that our model improves over the two baselines.
In \autoref{dspn fig:mnist} and \autoref{dspn fig:mnist-full}, we show the progression of $\mhY$ throughout the minimisation with $\mhY^{(10)}$ as the final prediction, the ground-truth set, and the baseline prediction of an MLP decoder.
Observe how every optimisation starts with the same set $\mhY^{(0)}$, but is transformed differently depending on the gradient of $\enc$.
Through this minimisation of $\Lrepr$ by the inner optimisaton, the set is gradually changed into a shape that closely resembles the correct digit.

The types of errors of our model and the baseline are different, despite the use of models with similar losses in \autoref{dspn fig:mnist}.
Errors in our model are mostly due to scattered points outside of the main shape of the digit, which is particularly visible in the third row.
We believe that this is due to the limits of the encoder used: an encoder that is not powerful enough maps the slightly different sets to the same representation, so there is no $\Lrepr$ gradient to work with.
It still models the general shape accurately, but misses the fine details of these scattered points.
The MLP decoder has less of this scattering, but makes mistakes in the shape of the digit instead.
For example, in the third row, the baseline has a different curve at the top and a shorter line at the bottom.
This difference in types of errors is also present in the extended examples in \autoref{dspn fig:mnist-full}.

\begin{table}
    \centering
\caption[Reconstruction losses on MNIST]{Chamfer reconstruction loss on MNIST in 1000ths. Lower is better. Mean and standard deviation over 6 runs.}
    \label{dspn tab:mnist}
    \begin{tabular}{lccccc}
        \toprule
        Model & Loss\\
        \midrule
        MLP baseline & 0.21\tiny$\pm$0.18\\
        RNN baseline & 0.49\tiny$\pm$0.19\\
        Ours & \textbf{0.09}\tiny$\pm$0.01\\
        \bottomrule
    \end{tabular}
\end{table}

Note that reconstructions shown in (\autoref{fspool sec:mnist}) for the same auto-encoding task appear better because the decoder there uses additional information outside of the latent space: multiple $n \times n$ matrices are copied from the encoder into the decoder.
In contrast, all information about the set is completely contained in our permutation-invariant latent space.

\subsection{Bounding box prediction}\label{dspn sec:bbox}
Next, we turn to the task of object detection on the CLEVR dataset~\citep{Johnson2017a}, which contains 70,000 training and 15,000 validation images.
The goal is to predict the set of bounding boxes for the objects in an image.
The target set contains at most 10 elements with 4 dimensions each: the (normalised) x-y coordinates of the top-left and bottom-right corners of each box.
As the dataset does not contain bounding box information canonically, we use the processing method by \citet{Desta2018} to calculate approximate bounding boxes.
This causes the ground-truth bounding boxes to not always be perfect, which is a source of noise.

\paragraph{Model}
We encode the image with a ResNet34~\citep{He2016} into a 512d feature vector, which is fed into the set decoder.
The set decoder predicts the set of bounding boxes \emph{from this single feature vector} describing the whole image.
This is in contrast to existing region proposal networks~\citep{Ren2015} for bounding box prediction where the use of the entire feature map is required for the typical anchor-based approach.
As the set encoder in our model, we use a 2-layer relation network~\citep{Santoro2017} with FSPool (\autoref{cha:fspool}) as pooling.
This is stronger than the FSPool-only model (without RN) we used in the MNIST experiment.
We again compare this against a baseline that uses an MLP or LSTM as set decoder (matching AE-EMD~\citep{Achlioptas2018} and~\citep{Rezatofighi2018} for the MLP decoder, \citep{Stewart2016} for the LSTM decoder).
Since the sets are much smaller compared to our MNIST experiments, we can use the Hungarian loss as set loss.
We perform no post-processing (such as non-maximum suppression) on the predictions of the model.
The whole model is trained end-to-end.

\begin{figure}
    \centering
    \centerfloat{
    \includegraphics[width=\figurewidth, trim=0 3mm 0 0, clip]{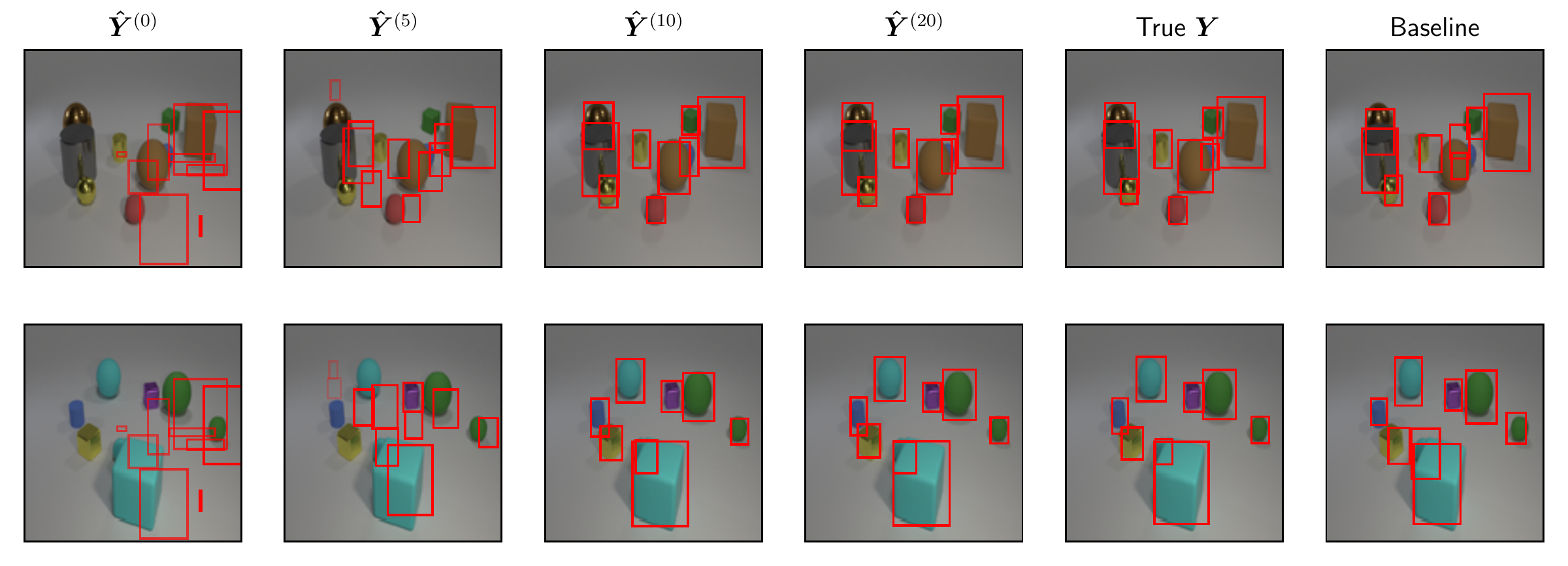}
    }
    \vspace{-3mm}
    \caption[Progression of set prediction algorithm on bounding box prediction]{Progression of set prediction algorithm for bounding boxes in CLEVR. The shown MLP baseline sometimes struggles with heavily-overlapping objects and often fails to centre the object in the boxes.}
    \label{dspn fig:box}
\end{figure}

\begin{figure}
    \centering
    \centerfloat{\includegraphics[width=\figurewidth]{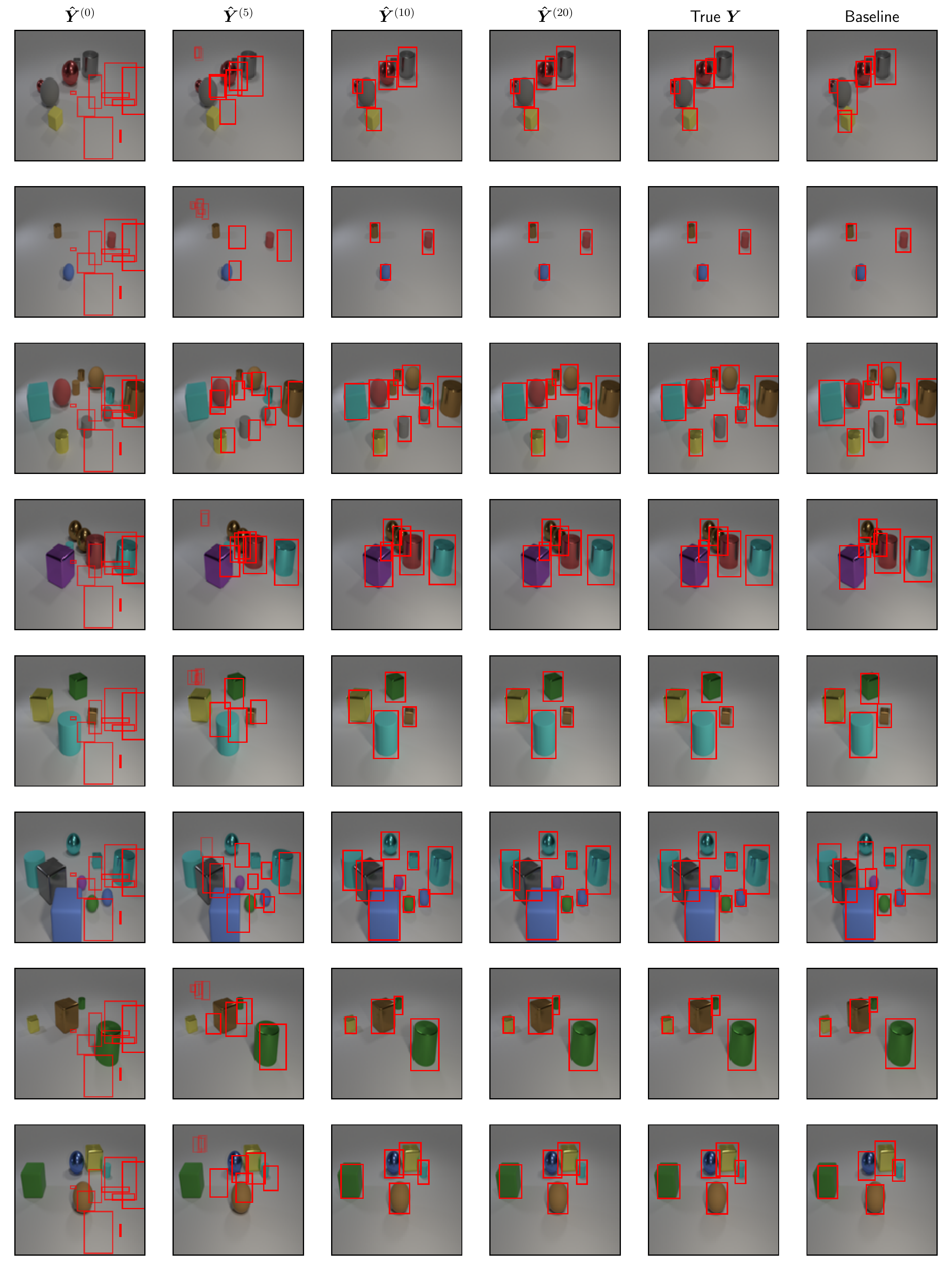}}
    \caption[More progression of set prediction algorithm on bounding box prediction.]{Progression of set prediction algorithm on CLEVR bounding boxes.}
    \label{dspn fig:box-full}
\end{figure}

\paragraph{Results}
We show our results in \autoref{dspn tab:box} using the standard average precision (AP) metric used in object detection with sample predictions in \autoref{dspn fig:box} and \autoref{dspn fig:box-full}.
Our model is able to very accurately localise the objects with high AP scores even when the intersection-over-union (IoU) threshold for a predicted box to match a ground truth box is very strict.
In particular, our model using 10 iterations (the same it was trained with) has much better AP\textsubscript{95} and AP\textsubscript{98} than the baselines.
The shown baseline model can predict bounding boxes in the close vicinity of objects, but fails to place the bounding box precisely on the object.
This is visible from the decent performance for low IoU thresholds, but bad performance for high IoU thresholds.

We can also run our model with more inner optimisation steps than the 10 it was trained with.
Many results improve when doubling the number of steps, which shows that further minimisation of $\Lrepr(\mhY, \vz)$ is still beneficial, even if it is unseen during training.
The model ``knows'' that its prediction is still suboptimal when $\Lrepr$ is high and also how to change the set to decrease it.
This confirms that the optimisation is reasonably stable and does not diverge significantly with more steps.
Being able to change the number of steps allows for a dynamic trade-off between prediction quality and inference time depending on what is needed for a given task.

\begin{table}
    \centering
    \caption[Average Precision results for bounding box prediction]{Average Precision (AP) for different intersection-over-union thresholds for a predicted bounding box to be considered correct. Higher is better. Mean and standard deviation over 6 runs.}
    \label{dspn tab:box}
    \begin{tabular}{lccccc}
        \toprule
        Model & AP\textsubscript{50} & AP\textsubscript{90} & AP\textsubscript{95} & AP\textsubscript{98} & AP\textsubscript{99} \\
        \midrule
        MLP baseline        & 99.3\tiny$\pm$0.2 & 94.0\tiny$\pm$1.9 & 57.9\tiny$\pm$7.9 &  0.7\tiny$\pm$0.2 & 0.0\tiny$\pm$0.0 \\
        RNN baseline        & 99.4\tiny$\pm$0.2 & 94.9\tiny$\pm$2.0 & 65.0\tiny$\pm$10.3 &  2.4\tiny$\pm$0.0 & 0.0\tiny$\pm$0.0 \\
        Ours (10 iters) & 98.8\tiny$\pm$0.3 & 94.3\tiny$\pm$1.5 & 85.7\tiny$\pm$3.0 & \textbf{34.5}\tiny$\pm$5.7 & \textbf{2.9}\tiny$\pm$1.2 \\
        Ours (20 iters) & \textbf{99.8}\tiny$\pm$0.0 & \textbf{98.7}\tiny$\pm$1.1 & \textbf{86.2}\tiny$\pm$7.2 & 24.3\tiny$\pm$8.0 & 1.4\tiny$\pm$0.9 \\
        Ours (30 iters) & \textbf{99.8}\tiny$\pm$0.1 & 96.7\tiny$\pm$2.4 & 75.5\tiny$\pm$12.3 & 17.4\tiny$\pm$7.7 & 0.9\tiny$\pm$0.7 \\
        \bottomrule
    \end{tabular}
\end{table}

\begin{table}
    \centering
    \caption[Set encoder ablations on bounding box prediction in CLEVR]{Ablation experiments with DSPN-RN-Sum (standard RN model) and DSPN-RN-Max. The DSPN-RN-FSPool results are the same as in the previous table.}
    \label{dspn tab:box fspool}
    \centerfloat{
    \begin{tabular}{lccccc}
        \toprule
        Model & AP\textsubscript{50} & AP\textsubscript{90} & AP\textsubscript{95} & AP\textsubscript{98} & AP\textsubscript{99} \\
        \midrule
        \textsc{DSPN-RN-FSPool} (10 iters) & 98.8\tiny$\pm$0.3 & 94.3\tiny$\pm$1.5 & 85.7\tiny$\pm$3.0 & \textbf{34.5}\tiny$\pm$5.7 & \textbf{2.9}\tiny$\pm$1.2 \\
        \textsc{DSPN-RN-FSPool} (20 iters) & \textbf{99.8}\tiny$\pm$0.0 & \textbf{98.7}\tiny$\pm$1.1 & \textbf{86.2}\tiny$\pm$7.2 & 24.3\tiny$\pm$8.0 & 1.4\tiny$\pm$0.9 \\
        \textsc{DSPN-RN-FSPool} (30 iters) & \textbf{99.8}\tiny$\pm$0.1 & 96.7\tiny$\pm$2.4 & 75.5\tiny$\pm$12.3 & 17.4\tiny$\pm$7.7 & 0.9\tiny$\pm$0.7 \\
        \midrule
        \textsc{DSPN-RN-Sum} (10 iters) & 88.3\tiny$\pm$3.7 & 43.4\tiny$\pm$14.4 & 10.0\tiny$\pm$7.4 & {0.1}\tiny$\pm$0.1 & {0.0}\tiny$\pm$0.0 \\
        \textsc{DSPN-RN-Sum} (20 iters) & {87.2}\tiny$\pm$3.0 & {42.9}\tiny$\pm$11.9 & {5.7}\tiny$\pm$3.5 & 0.0\tiny$\pm$0.0 & 0.0\tiny$\pm$0.0 \\
        \textsc{DSPN-RN-Sum} (30 iters) & {79.0}\tiny$\pm$11.9 & 32.5\tiny$\pm$12.4 & 3.4\tiny$\pm$2.2 & 0.0\tiny$\pm$0.0 & 0.0\tiny$\pm$0.0 \\
        \textsc{DSPN-RN-Max} (10 iters) & 68.0\tiny$\pm$4.3 & 4.0\tiny$\pm$2.2 & 0.1\tiny$\pm$0.1 & {0.0}\tiny$\pm$0.0 & {0.0}\tiny$\pm$0.0 \\
        \textsc{DSPN-RN-Max} (20 iters) & {66.6}\tiny$\pm$4.5 & {3.3}\tiny$\pm$1.8 & {0.1}\tiny$\pm$0.0 & 0.0\tiny$\pm$0.0 & 0.0\tiny$\pm$0.0 \\
        \textsc{DSPN-RN-Max} (30 iters) & {64.1}\tiny$\pm$5.0 & 2.3\tiny$\pm$1.1 & 0.0\tiny$\pm$0.0 & 0.0\tiny$\pm$0.0 & 0.0\tiny$\pm$0.0 \\
        \bottomrule
    \end{tabular}
    }
\end{table}

The less-strict AP metrics (which measure large mistakes) improve with more iterations, while the very strict AP\textsubscript{98} and AP\textsubscript{99} metrics consistently worsen.
This is a sign that the inner optimisation learned to reach its best prediction at exactly 10 steps, but slightly overshoots when run for longer.
The model has learned that it does not fully converge with 10 steps, so it is compensating for that by slightly biasing the inner optimisation to get a better 10 step prediction.
This is at the expense of the strictest AP metrics worsening with 20 steps, where this bias is not necessary anymore.



We also perform an ablation experiment where we replace FSPool in the encoder with sum or max pooling.
The results in \autoref{dspn tab:box fspool} show that using FSPool greatly benefits our algorithm, likely due to the better representation that it is able to learn.
The representation of the DSPN-RN-FSPool model is good enough that iterating our algorithm for more steps than the model was trained with can benefit the prediction, while for the baselines it generally only worsens.

Bear in mind that we do not intend to directly compete against traditional object detection methods.
Our goal is to demonstrate that our model can accurately predict a set from a single feature vector, which is of general use for set prediction tasks not limited to image inputs.


\subsection{Object attribute prediction}\label{dspn sec:state}
Lastly, we want to directly predict the full state of a scene from images on CLEVR.
This is the set of objects with their position in the 3d scene ($xyz$ coordinates), shape (sphere, cylinder, cube), colour (eight colours), size (small, large), and material (metal/shiny, rubber/matte) as features.
For example, an object can be a ``small cyan metal cube'' at position (0.95, -2.83, 0.35).
We encode the categorial features as one-hot vectors and concatenate them into an 18d feature vector for each object.
Note that we do not use bounding box information, so the model has to implicitly learn which object in the image corresponds to which set element with the associated properties.
This makes it different from usual object detection tasks, since bounding boxes are required for traditional object detection models that rely on anchors.

\paragraph{Model}
We use exactly the same model as for the bounding box prediction in the previous experiment with all hyperparameters kept the same.
The only difference is that it now outputs 18d instead of 4d set elements.
For simplicity, we continue using the Hungarian loss with the Huber loss as pairwise cost, as opposed to switching to cross-entropy for the categorical features.

\paragraph{Results}
We show our results in \autoref{dspn tab:state} and give sample outputs in \autoref{dspn fig:state-full}.
The evaluation metric is the standard average precision as used in object detection, with the modification that a prediction is considered correct if there is a matching groundtruth object with exactly the same properties and within a given Euclidean distance of the 3d coordinates.
Our model clearly outperforms the baselines.
This shows that our model is also suitable for modeling high-dimensional set elements.

When evaluating with more steps than our model was trained with, the difference in the more lenient metrics improves even up to 30 iterations.
This time, the results for 20 iterations are all better than for 10 iterations.
This suggests that 10 steps is too few to reach a good solution in training, likely due to the higher difficulty of this task compared to the bounding box prediction.
Still, the representation $\vz$ that the input encoder produces is good enough such that minimising $\Lrepr$ more at evaluation time leads to better results.
When going up to 30 iterations, the result for predicting the state only (excluding 3d position) improves further, but the accuracy of the 3d position worsens.
We believe that this is again caused by overshooting the target due to the bias of training the model with only 10 iterations.

\begin{table}
    \centering
\caption[Average Precision results for state prediction]{Average Precision (AP) in \% for different distance thresholds of a predicted set element to be considered correct. AP\textsubscript{$\infty$} only requires all attributes to be correct, regardless of 3d position. Higher is better. Mean and standard deviation over 6 runs.}
    \label{dspn tab:state}
    \begin{tabular}{lccccc}
        \toprule
        Model & AP\textsubscript{$\infty$} & AP\textsubscript{1} & AP\textsubscript{0.5} & AP\textsubscript{0.25} & AP\textsubscript{0.125}   \\
        \midrule
        MLP baseline & 3.6\tiny$\pm$0.5 & 1.5\tiny$\pm$0.4 & 0.8\tiny$\pm$0.3 & 0.2\tiny$\pm$0.1 & 0.0\tiny$\pm$0.0 \\
        RNN baseline & 4.0\tiny$\pm$1.9 & 1.8\tiny$\pm$1.2 & 0.9\tiny$\pm$0.5 & 0.2\tiny$\pm$0.1 & 0.0\tiny$\pm$0.0 \\
        Ours (10 iters) & 72.8\tiny$\pm$2.3 & 59.2\tiny$\pm$2.8 & 39.0\tiny$\pm$4.4 & 12.4\tiny$\pm$2.5 & 1.3\tiny$\pm$0.4 \\
        Ours (20 iters) & 84.0\tiny$\pm$4.5 & 80.0\tiny$\pm$4.9 & \textbf{57.0}\tiny$\pm$12.1 & \textbf{16.6}\tiny$\pm$9.0 & \textbf{1.6}\tiny$\pm$0.9 \\
        Ours (30 iters) & \textbf{85.2}\tiny$\pm$4.8 & \textbf{81.1}\tiny$\pm$5.2 & 47.4\tiny$\pm$17.6 & 10.8\tiny$\pm$9.0 & 0.6\tiny$\pm$0.7 \\
        \bottomrule
    \end{tabular}
\end{table}

\begin{table}
    \centering
    \caption[Set encoder ablations on state prediction in CLEVR]{Average Precision (AP, mean $\pm$ stdev) for different distance thresholds of the predicted state descriptions over 6 runs. All results are worse than the DSPN-RN-FSPool in the previous table. The DSPN-RN-FSPool results are the same as in the previous table.}
    \label{dspn tab:state fspool}
    \centerfloat{
    \begin{tabular}{lccccc}
        \toprule
        Model & AP\textsubscript{$\infty$} & AP\textsubscript{1} & AP\textsubscript{0.5} & AP\textsubscript{0.25} & AP\textsubscript{0.125}   \\
        \midrule
        \textsc{DSPN-RN-FSPool} (10 iters) & 72.8\tiny$\pm$2.3 & 59.2\tiny$\pm$2.8 & 39.0\tiny$\pm$4.4 & 12.4\tiny$\pm$2.5 & 1.3\tiny$\pm$0.4 \\
        \textsc{DSPN-RN-FSPool} (20 iters) & 84.0\tiny$\pm$4.5 & 80.0\tiny$\pm$4.9 & \textbf{57.0}\tiny$\pm$12.1 & \textbf{16.6}\tiny$\pm$9.0 & \textbf{1.6}\tiny$\pm$0.9 \\
        \textsc{DSPN-RN-FSPool} (30 iters) & \textbf{85.2}\tiny$\pm$4.8 & \textbf{81.1}\tiny$\pm$5.2 & 47.4\tiny$\pm$17.6 & 10.8\tiny$\pm$9.0 & 0.6\tiny$\pm$0.7 \\
        \midrule
        \textsc{DSPN-RN-Sum} (10 iters) & 44.6\tiny$\pm$3.8 & 21.9\tiny$\pm$4.8 & 7.1\tiny$\pm$2.7 & 1.0\tiny$\pm$0.5 & 0.0\tiny$\pm$0.0 \\
        \textsc{DSPN-RN-Sum} (20 iters) & 39.6\tiny$\pm$5.4 & 15.2\tiny$\pm$6.4 & 3.0\tiny$\pm$2.2 & 0.3\tiny$\pm$0.3 & 0.0\tiny$\pm$0.0 \\
        \textsc{DSPN-RN-Sum} (30 iters) & 30.2\tiny$\pm$9.2 & 7.1\tiny$\pm$3.8 & 0.9\tiny$\pm$0.8 & 0.1\tiny$\pm$0.1 & 0.0\tiny$\pm$0.0 \\
        \textsc{DSPN-RN-Max} (10 iters) & 3.0\tiny$\pm$0.2 & 0.9\tiny$\pm$0.1 & 0.5\tiny$\pm$0.2 & 0.1\tiny$\pm$0.1 & 0.0\tiny$\pm$0.0 \\
        \textsc{DSPN-RN-Max} (20 iters) & 3.1\tiny$\pm$0.1 & 1.2\tiny$\pm$0.1 & 0.8\tiny$\pm$0.2 & 0.3\tiny$\pm$0.2 & 0.0\tiny$\pm$0.0 \\
        \textsc{DSPN-RN-Max} (30 iters) & 3.1\tiny$\pm$0.1 & 1.2\tiny$\pm$0.1 & 0.9\tiny$\pm$0.2 & 0.3\tiny$\pm$0.2 & 0.0\tiny$\pm$0.0 \\
        \bottomrule
    \end{tabular}
    }
\end{table}

\begin{table}
    \centering
    \tiny
    \caption[Progression of set prediction algorithm on state prediction]{Progression of set prediction algorithm on CLEVR state prediction. Red text denotes a wrong attribute. Objects are sorted by x coordinate, so they are sometimes misaligned with wrongly-coloured red text (see third example: red entries in $\mhY^{(20)}$).}
    \label{dspn fig:state-full}
    
\includegraphics[width=0.33\linewidth]{img-0}
\centerfloat{\begin{tabular}{ccccc}
\toprule
$\hat{\bm{Y}}^{(5)}$ & $\hat{\bm{Y}}^{(10)}$ & $\hat{\bm{Y}}^{(20)}$ & True $\bm{Y}$ & Baseline\\
\midrule
(-0.14, 1.16, 3.57) & (-2.33, -2.41, 0.73) & (-2.33, -2.42, 0.78) & (-2.42, -2.40, 0.70) & (-1.65, -2.85, 0.69)\\
large \textcolor{red}{purple} \textcolor{red}{rubber} \textcolor{red}{sphere} & large yellow metal cube & large yellow metal cube & large yellow metal cube & large yellow metal cube\\
(0.01, 0.12, 3.42) & (-1.20, 1.27, 0.67) & (-1.21, 1.20, 0.65) & (-1.18, 1.25, 0.70) & (-0.95, 1.08, 0.68)\\
large \textcolor{red}{gray} \textcolor{red}{metal} \textcolor{red}{cube} & large purple rubber sphere & large purple rubber sphere & large purple rubber sphere & large \textcolor{red}{green} rubber sphere\\
(0.67, 0.65, 3.38) & (-0.96, 2.54, 0.36) & (-0.96, 2.59, 0.36) & (-1.02, 2.61, 0.35) & (-0.40, 2.14, 0.35)\\
small \textcolor{red}{purple} \textcolor{red}{metal} \textcolor{red}{cube} & small gray rubber sphere & small gray rubber sphere & small gray rubber sphere & small \textcolor{red}{red} rubber sphere\\
(0.67, 1.14, 2.96) & (1.61, 1.57, 0.36) & (1.58, 1.62, 0.38) & (1.74, 1.53, 0.35) & (1.68, 1.77, 0.35)\\
small purple \textcolor{red}{rubber} \textcolor{red}{sphere} & small \textcolor{red}{yellow} metal cube & small purple metal cube & small purple metal cube & small \textcolor{red}{brown} metal cube\\
\bottomrule
\end{tabular}}

\includegraphics[width=0.33\linewidth]{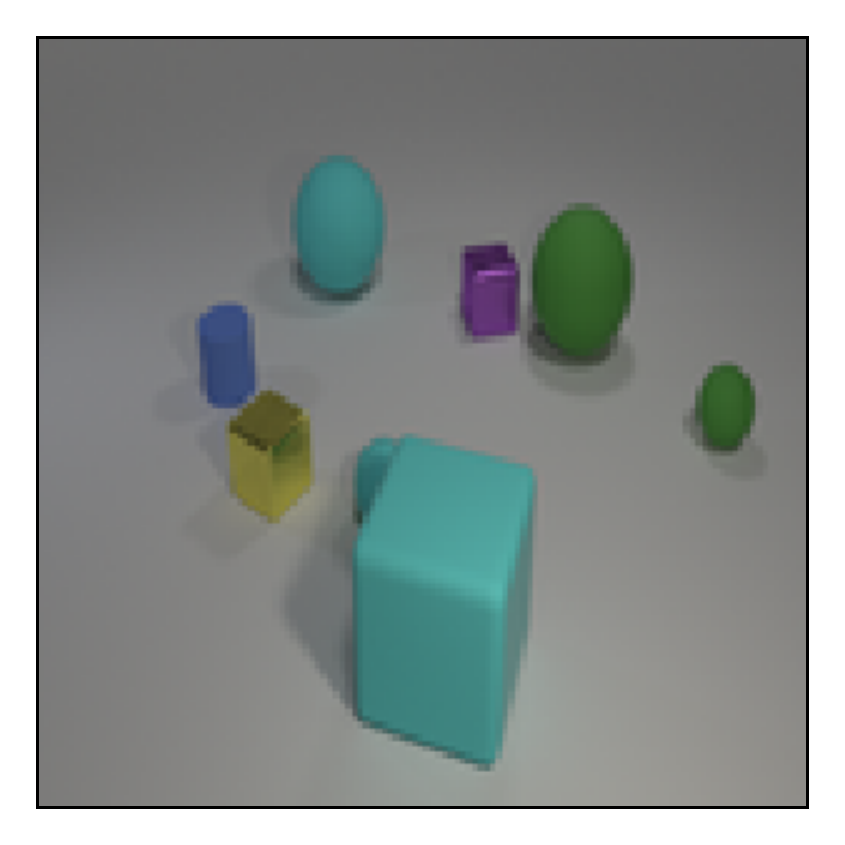}
\centerfloat{\begin{tabular}{ccccc}
\toprule
$\hat{\bm{Y}}^{(5)}$ & $\hat{\bm{Y}}^{(10)}$ & $\hat{\bm{Y}}^{(20)}$ & True $\bm{Y}$ & Baseline\\
\midrule
(-0.29, 1.14, 3.73) & (-2.78, 0.86, 0.72) & (-2.62, 0.83, 0.68) & (-2.88, 0.78, 0.70) & (-2.42, 0.63, 0.71)\\
\textcolor{red}{small} \textcolor{red}{purple} \textcolor{red}{metal} \textcolor{red}{cube} & large cyan rubber sphere & large cyan rubber sphere & large cyan rubber sphere & large \textcolor{red}{purple} rubber sphere\\
(-0.11, -0.37, 3.65) & (-2.17, -1.59, 0.38) & (-2.12, -1.58, 0.49) & (-2.14, -1.63, 0.35) & (-2.40, -2.07, 0.35)\\
small \textcolor{red}{brown} \textcolor{red}{metal} \textcolor{red}{cube} & small blue rubber cylinder & small blue rubber cylinder & small blue rubber cylinder & small \textcolor{red}{green} rubber cylinder\\
(0.08, 0.56, 3.84) & (-0.45, 2.19, 0.40) & (-0.60, 2.23, 0.29) & (-0.78, 1.97, 0.35) & (-0.74, 2.46, 0.33)\\
\textcolor{red}{large} \textcolor{red}{cyan} \textcolor{red}{rubber} cube & small purple metal cube & small purple metal cube & small purple metal cube & small \textcolor{red}{cyan} metal cube\\
(0.69, -0.43, 3.55) & (-0.14, -2.15, 0.38) & (-0.30, -1.99, 0.32) & (-0.38, -2.06, 0.35) & (0.30, -1.86, 0.34)\\
small \textcolor{red}{brown} \textcolor{red}{rubber} \textcolor{red}{sphere} & small yellow metal cube & small yellow metal cube & small yellow metal cube & small \textcolor{red}{gray} \textcolor{red}{rubber} \textcolor{red}{sphere}\\
(1.12, 0.21, 3.83) & (0.53, 2.56, 0.70) & (0.27, 2.46, 0.72) & (0.42, 2.56, 0.70) & (0.69, -2.10, 0.36)\\
large \textcolor{red}{cyan} rubber \textcolor{red}{cube} & large green rubber sphere & large green rubber sphere & large green rubber sphere & \textcolor{red}{small} \textcolor{red}{red} \textcolor{red}{metal} \textcolor{red}{cube}\\
(1.23, -0.25, 3.58) & (0.93, -1.41, 0.35) & (0.86, -1.31, 0.27) & (0.81, -1.30, 0.35) & (1.12, 2.28, 0.70)\\
small cyan rubber sphere & small cyan rubber sphere & small cyan rubber sphere & small cyan rubber sphere & \textcolor{red}{large} cyan rubber sphere\\
(1.73, 1.04, 3.57) & (2.50, -2.08, 0.76) & (2.64, -2.05, 0.76) & (2.56, -1.94, 0.70) & (2.55, -2.26, 0.73)\\
\textcolor{red}{small} cyan rubber \textcolor{red}{sphere} & large cyan rubber cube & large cyan rubber cube & large cyan rubber cube & large \textcolor{red}{yellow} rubber cube\\
(2.06, 1.94, 3.81) & (2.61, 2.59, 0.33) & (2.75, 2.73, 0.35) & (2.74, 2.64, 0.35) & (2.99, 2.59, 0.35)\\
\textcolor{red}{large} \textcolor{red}{brown} rubber sphere & small green rubber sphere & small green rubber sphere & small green rubber sphere & small \textcolor{red}{purple} rubber sphere\\
\bottomrule
\end{tabular}}

\includegraphics[width=0.33\linewidth]{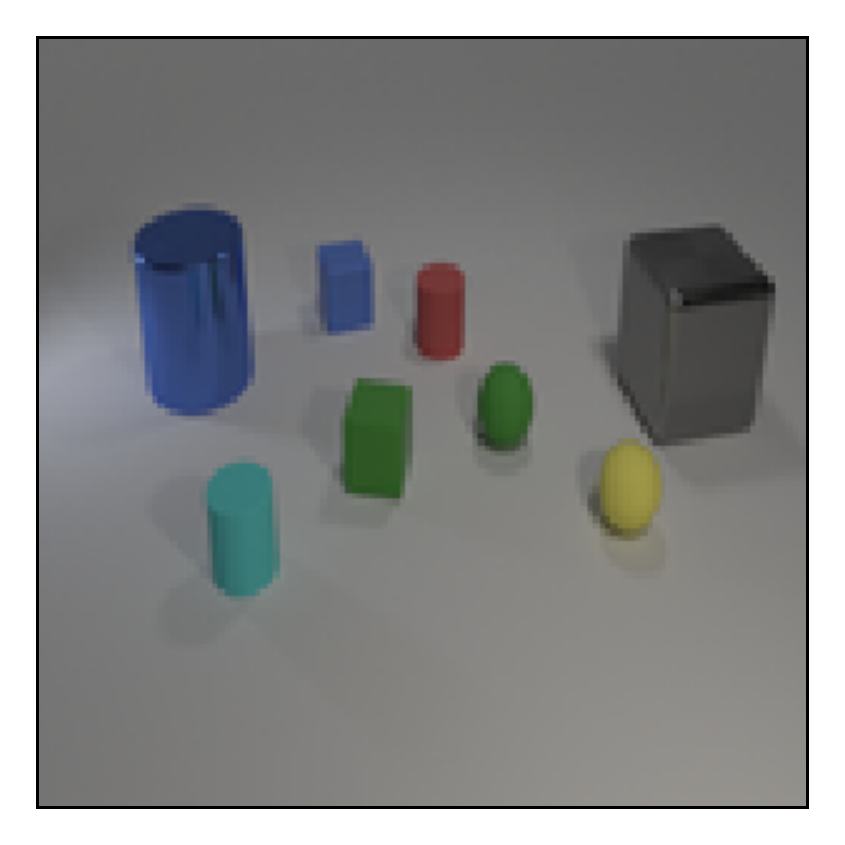}
\centerfloat{\begin{tabular}{ccccc}
\toprule
$\hat{\bm{Y}}^{(5)}$ & $\hat{\bm{Y}}^{(10)}$ & $\hat{\bm{Y}}^{(20)}$ & True $\bm{Y}$ & Baseline\\
\midrule
(0.22, 0.12, 3.47) & (-2.76, -1.42, 0.68) & (-2.68, -1.64, 0.77) & (-2.62, -1.76, 0.70) & (-2.47, -1.73, 0.70)\\
\textcolor{red}{small} \textcolor{red}{brown} \textcolor{red}{rubber} \textcolor{red}{cube} & large blue metal cylinder & large blue metal cylinder & large blue metal cylinder & large \textcolor{red}{cyan} metal cylinder\\
(0.41, 0.11, 3.77) & (-1.56, -0.61, 0.35) & (-2.43, 0.03, 0.34) & (-2.29, 0.49, 0.35) & (-2.42, 0.09, 0.36)\\
\textcolor{red}{large} \textcolor{red}{gray} \textcolor{red}{metal} cube & small blue rubber \textcolor{red}{cylinder} & small blue rubber cube & small blue rubber cube & small blue rubber \textcolor{red}{cylinder}\\
(0.50, 0.44, 3.61) & (-1.08, 0.23, 0.33) & (-1.00, 1.18, 0.33) & (-0.93, 1.15, 0.35) & (-1.24, 1.16, 0.36)\\
small \textcolor{red}{gray} rubber \textcolor{red}{cube} & small \textcolor{red}{green} rubber \textcolor{red}{cube} & small red rubber cylinder & small red rubber cylinder & small red rubber \textcolor{red}{cube}\\
(0.83, 0.53, 3.45) & (-0.07, 0.97, 0.36) & (-0.01, -1.00, 0.46) & (0.28, -2.84, 0.35) & (0.39, 0.20, 0.33)\\
small cyan rubber \textcolor{red}{sphere} & small \textcolor{red}{green} rubber cylinder & small \textcolor{red}{green} rubber \textcolor{red}{cube} & small cyan rubber cylinder & small \textcolor{red}{red} rubber \textcolor{red}{sphere}\\
(0.86, 0.85, 3.50) & (0.28, -2.44, 0.49) & (0.21, -2.88, 0.40) & (0.29, -0.98, 0.35) & (0.56, -3.11, 0.35)\\
small \textcolor{red}{gray} rubber \textcolor{red}{sphere} & small \textcolor{red}{cyan} rubber \textcolor{red}{cylinder} & small \textcolor{red}{cyan} rubber \textcolor{red}{cylinder} & small green rubber cube & small \textcolor{red}{yellow} rubber \textcolor{red}{cylinder}\\
(1.86, 2.34, 3.80) & (1.36, -0.63, 0.38) & (0.99, 0.17, 0.37) & (0.92, 0.54, 0.35) & (0.90, 0.64, 0.35)\\
\textcolor{red}{large} \textcolor{red}{gray} \textcolor{red}{metal} \textcolor{red}{cube} & small green rubber sphere & small green rubber sphere & small green rubber sphere & small green rubber sphere\\
(1.97, 0.55, 3.61) & (2.01, 3.07, 0.65) & (1.97, 2.89, 0.39) & (2.04, 2.78, 0.70) & (2.39, 0.27, 0.36)\\
\textcolor{red}{small} \textcolor{red}{green} \textcolor{red}{rubber} \textcolor{red}{sphere} & large gray metal cube & large gray metal cube & large gray metal cube & \textcolor{red}{small} \textcolor{red}{yellow} \textcolor{red}{rubber} \textcolor{red}{sphere}\\
 & (2.69, 0.63, 0.34) & (2.87, 0.51, 0.25) & (2.70, 0.67, 0.35) & (2.44, 2.55, 0.68)\\
\textcolor{red}{} & small yellow rubber sphere & small yellow rubber sphere & small yellow rubber sphere & \textcolor{red}{large} \textcolor{red}{gray} \textcolor{red}{metal} \textcolor{red}{cube}\\
\bottomrule
\end{tabular}}

\end{table}

The results of the ablation experiment where we replace FSPool in the encoder with sum or max pooling are shown in \autoref{dspn tab:state fspool}.
This time, the difference between FSPool and the other encoders is even larger than for the bounding box experiments.
These results confirm that a better set encoder leads to better prediction results.
They also show that a way to improve RNs is to simply replace the sum pooling with FSPool.


\section{Discussion}\label{dspn sec:discussion}
In this chapter we showed how to predict sets with a deep neural network in a way that respects the set structure of the problem.
We demonstrated in our experiments that this works for small (size 10) and large sets (up to size 342), as well as low-dimensional (2d) and higher-dimensional (18d) set elements.
Our model is consistently better than the baselines across all experiments by predicting sets properly, rather than predicting a list and pretending that it is a set.


The improved results of our approach come at a higher computational cost.
Each evaluation of the network requires time for $O(T)$ passes through the set encoder, which makes training take about 75\% longer on CLEVR with $T=10$.
Keep in mind that this only involves the set encoder (which can be fairly small), not the input encoder (such as a CNN or RNN) that produces the target $\vz$.
Further study into representationally-powerful and efficient set encoders such as RN~\citep{Santoro2017} and FSPool (\autoref{cha:fspool})~--~which we found to be critical for good results in our experiments~--~would be of considerable interest, as it could speed up the convergence and thus inference time of our method.
Another promising approach is to better initialise $Y^{(0)}$~--~perhaps with an MLP~--~so that the set needs to be changed less to minimise $\Lrepr$.
Our model would act as a set-aware refinement method of the MLP prediction.
Lastly, stopping criteria other than iterating for a fixed 10 steps can be used, such as stopping when $\Lrepr(\enc(\mhY), \vz)$ is below a fixed threshold: this would stop when the encoder thinks $\mhY$ is of a certain quality corresponding to that threshold.

Our algorithm may be suitable for generating samples under other invariance properties.
For example, we may want to generate images of objects where the rotation of the object does not matter (such as aerial images).
Using our decoding algorithm with a rotation-invariant image encoder could predict images without forcing the model to choose a fixed orientation of the image, which could be a useful inductive bias.

In conclusion, we are excited about enabling a wider variety of set prediction problems to be tackled with deep neural networks.
Our main idea should be readily extensible to similar domains such as graphs to allow for better graph prediction, for example molecular graph generation or end-to-end scene graph prediction from images.
We hope that our model inspires further research into graph generation, stronger object detection models, and -- more generally -- a more principled approach to set prediction.

\chapter{Future work}\label{cha:conclusion}
In this thesis, we have developed a variety of techniques for modeling sets with deep neural networks.
All of our proposed methods are fully differentiable, which makes them readily usable in existing and new neural networks for sets.

Starting from our initial work on the specialised task of counting object proposals in visual question answering (\autoref{cha:vqa}), we developed two building blocks for set encoders and two building blocks for set decoders.
Our set encoders tackle the bottleneck problem (\autoref{background sec:bottleneck}) of traditional approaches, which only use simple pooling functions like sum and max pooling.
By learning how to permute the set elements with the Permutation-optimisation module (\autoref{cha:perm-optim}) or sorting the features within a set independently with FSPool (\autoref{cha:fspool}), the set is turned into an ordered representation, which is much easier to work with.

In \autoref{cha:fspool}, we also realised that existing approaches for predicting sets suffer from a responsibility problem (\autoref{fspool sec:problem}), which forces a set decoder to predict discontinuous outputs.
Since we show that this can majorly hinder successful learning (\autoref{fspool sec:polygons}), we tried to find ways to avoid this problem.
We first did this through FSUnpool in the limited auto-encoder setting, then through Deep Set Prediction Networks (DSPN) in the much more general supervised prediction setting (\autoref{cha:dspn}).
This is perhaps the most significant contribution of this thesis, since it is the first model for predicting sets that has the right properties for sets while being able to scale to complex tasks like object detection.

Of course, many open problems remain in the area of set encoders and set decoders, which we discuss in \autoref{future sec:encoders} and \autoref{future sec:decoders}.
Last but not least, we discuss the potential of latent sets (\autoref{future sec:latent}), which we would have loved to work on.

\section{Set encoders}\label{future sec:encoders}
A better encoder can learn better representations and therefore should also result in better performance for the downstream task.
This can even extend to improvements to set prediction through DSPN.
Recent progress in set encoders has been closely linked to better modeling of \emph{relationships} between set elements and we believe that this will continue to play an important role.
Due to the similarity between the domains of sets and graphs, improvements in either area are easily transferable, so researchers in either field should be closely aware of the other.

\paragraph{The permutation-invariant step}
We have primarily used the idea of ordering the set in some way in \autoref{cha:perm-optim} and \autoref{cha:fspool}.
There may be other algorithms similar to sorting that are able to turn a set into a list in a way that is easy to learn from.
An important consideration is the trade-off between the complexity of relationships within the set that can be modeled and the computational efficiency of that method.
Developing new set pooling methods with different trade-offs and theoretically characterising the limitations like in \citet{Murphy2019,Xu2019,Wagstaff2019} are therefore both useful.

\paragraph{The permutation-equivariant step}
We have explored various equivariant approaches in this thesis: the FSPool-FSUnpool combination (\autoref{cha:fspool}) and backpropagating through a set encoder (\autoref{cha:dspn}).
While we used them in the context of predicting sets, it would be interesting to see how well these work as a building block in a pure set encoder or in per-element set prediction tasks.

What about other permutation-equivariant ways to propagate information within the set?
A recent major advance is the building block of self-attention in set transformers~\citep{Vaswani2017}, which has been successfully used in contexts other than sets such as language modeling.
Here, the same trade-off between complexity of relations and efficiency exists.
Methods that strike a good balance between the two are needed, and gaining deeper theoretical understanding of this balance may help in this regard.

\paragraph{Applications}
As we have pointed out in \autoref{sec:background-other}, various models that already use a sum or max -- such as pooling in CNNs -- allow for set methods to be used as an alternative.
In those cases, lessons from the set literature can be applied to understand the existing models and potentially improve them.

For language modeling, set transformers have the benefit over traditional sequential models of being able to efficiently model long-distance relationships (words that are far apart).
To not lose information about the order of the words in a sentence when going from the sequence of words to the set of words, the words are augmented with positional information.
In essence, the existing sequence problem is transformed into an equivalent set problem.
Due to the versatility of thinking about problems in terms of sets, there is potential in transforming other problems into sets and using set-based approaches on them.

\section{Set decoders}\label{future sec:decoders}
Predicting sets in new ways is an especially promising direction due to our contributions of the responsibility problem, FSPool-FSUnpool, and DSPN.
One option is to make improvements to our DPSN algorithm like the ones we suggest in \autoref{dspn sec:discussion}.
There are also various other research directions for set decoders.

\paragraph{Set losses}
While the Hungarian loss seems ideal from an optimal transport standpoint, the computational complexity of $\Theta(n^3)$ is not good enough for large sets.
Alternatives like Sinkhorn-based methods~\citep{Genevay2018} or something else entirely~\citep{Balles2019} could prove to be just as good in terms of quality while being much faster and more easily parallelisable.
There is also potential in the use of k-dimensional (k-d) trees to reduce computation.
This would allow for prediction of much larger sets and thus greater applicability of set decoders.

\paragraph{Other approaches}
We argued throughout the thesis that it is important to avoid the responsibility problem.
While we showed one solution with our DSPN model and its iterative optimisation approach, other ways to avoid it may exist.
One potential direction is to find a more lightweight method than backpropagating through a set encoder to determine the update to the set, perhaps with a self-attentive decoder~\citep{Vaswani2017, Belli2019}.
It would also be interesting to find a non-iterative approach, since the iteration is a core component of DSPN.

\subsection{Applications}\label{future sec:applications}
There are a variety of existing tasks in machine learning where the output is a set, but none of the models for the task treat it as a set prediction problem.
This is particularly evident in Computer Vision with tasks like object detection, instance segmentation, and multi-object tracking.
All of these share the commonality of being about \emph{multiple objects}, for which a set would be ideal.
Models with good performances on these tasks exist, but they usually rely on various post-processing heuristics like thresholding and  non-maximum suppression to produce a set of outputs.
By using a proper model for predicting sets like our DSPN, results in these tasks can potentially be greatly improved.

\paragraph{Object detection}
Of course, this is not without its challenges.
We will take the Faster R-CNN \citep{Ren2015} as representative example of traditional object detectors.
Our DSPN model does not make any assumptions about its input feature vector, while Faster R-CNN assumes that the input feature map comes from an image encoder.
While this has the benefit of DSPN being applicable to non-image tasks, it also means that DSPN makes \emph{global} predictions:
the existence of an object in one corner of an image can affect the entire feature vector and therefore how a different object is predicted in a different corner.
This is something that is often undesirable.
The convolutional approach to predicting the anchors in Faster R-CNN means that its predictions are reasonably translation-invariant and mostly local.
An object in one part of the image rarely affects an object in a completely different part.

Another potential challenge is how the model handles noise in the target sets.
We have only tested DSPN on a synthetic dataset where perfect information about the scene is available, which is unrealistic for datasets with real images and human annotators.
We currently do not know how robust our method is to noisy labels.
A missing object in the labeling could again affect the entire input feature vector and therefore the quality of the entire prediction.

A hybrid approach that combines DSPN with Faster R-CNN and similar object detection architectures could combine the benefits of the two: proper set prediction with end-to-end training whilst maintaining reasonable translation-invariance.

\paragraph{Instance segmentation}
Instance segmentation models typically use a two-stage approach: first detect the objects, then segment each object individually.
By using our DSPN model, this can be turned into a single stage of predicting the set of masks, each mask corresponding to the segmentation of one object.
By applying a softmax function on every pixel of these masks across the set, we could ensure that each pixel in the image belongs to at most one object or the background.
This is arguably a more elegant approach to instance segmentation, which could also lead to better results since it can be trained end-to-end.

\paragraph{Graph prediction}
A domain that is very similar to sets is the domain of graphs.
With the set of nodes and the set of edges, or the set of nodes and -- for each node -- the set of neighbours, the connection to sets is quite clear.
Graph prediction methods have also only relied on MLPs and RNNs so far, so the responsibility problem and our DSPN method to avoid it are directly applicable.
In particular, replacing the set encoder in our model with a graph encoder should immediately enable the prediction of graphs.
The main problem here would be the choice of loss function between graphs, which in the most difficult case would have to solve the graph isomorphism problem.
\citet{Simonovsky2018} have used such a graph loss before, which has a large $\Theta(n^4)$ time complexity in the number of nodes.
Finding good, efficient graph losses is perhaps the bigger challenge with predicting graphs.

\section{Latent sets in neural networks}\label{future sec:latent}
Lastly, it would be fascinating to have sets used in more contexts than in neural networks specifically for sets.
This is a more speculative idea.

An interesting aspect about sets is that they are a unique combination between a symbolic, object-based representation (set elements are discrete and each set element corresponds to a different symbol or object), while also being grounded (each set elements is associated with a continuous, learned feature vector describing it).
Humans are able to think about objects in terms of its smaller parts, like a hand being made of five fingers and a palm. 
By letting a traditional neural network like an image classifier use sets of feature vectors as latent representation, we may start to see such object- and parts-based representations emerge.
Instead of modeling an image of a hand as a feature vector, it could learn to model it as a set of five fingers and a palm, or other decompositions of the hand into smaller parts.
We know from \autoref{cha:vqa} that an object-based representation can be much easier to work with, so being able to discover such a representation automatically through the set structure would be quite useful.
In some ways, this is an extension of the parts-based representation idea that Capsule Neural Networks~\citep{Sabour2017} try to achieve, without forcing the notion of pose in Capsule Nets onto the neural network.

Because sets are in this unique position between connectionist and symbolic Artificial Intelligence, they have the potential to combine the best of both worlds.
We hope that our thesis provides a stepping stone towards this goal.

\appendix
\chapter[Appendix: Experimental details]{Experimental details}\label{cha:appendix}
\chaptermark{Experimental details}
\section{Counting in visual question answering}\label{app sec:vqa}
Here, we detail our improved baseline model for visual question answering in \autoref{cha:vqa}.
We provide the corresponding source code to reproduce our experiments at \url{https://github.com/Cyanogenoid/vqa-counting}.

The most significant change to the baseline model~\citep{Kazemi2017} that we make is the use of object proposal features by \citet{Anderson2018} as previously mentioned.
The following tweaks were made without considering the performance impact on the counting module; only the validation accuracy of the baseline was optimised.
Details not mentioned here can be assumed to be the same as in their paper.

To fuse vision features $\vx$ and question features $\vy$, the baseline concatenates and linearly projects them, followed by a ReLU activation.
This is equivalent to $\text{ReLU}(\mW_x \vx + \mW_y \vy)$.
We include an additional term that measures how different the projected $\vx$ is from the projected $\vy$, changing the fusion mechanism to 
\begin{equation}
    \vx \diamond \vy= \text{ReLU}(\mW_x \vx + \mW_y \vy) - (\mW_x \vx - \mW_y \vy)^2
\end{equation}

The LSTM~\citep{Hochreiter1997} for question encoding is replaced with a GRU~\citep{Cho2014} with the same hidden size with dynamic per-example unrolling instead of a fixed 14 words per question.
We apply batch normalisation~\citep{Ioffe2015} before the last linear projection in the classifier to the 3000 classes.
The learning rate is increased from 0.001 to 0.0015 and the batch size is doubled to 256.
The model is trained for 100 epochs (1697 iterations per epoch to train on the training set, 2517 iterations per epoch to train on both training and validation sets) instead of 100,000 iterations, roughly in line with the doubling of dataset size when going from VQA v1 to VQA v2.

Note that this single-model baseline is regularised with dropout~\citep{Srivastava2014}, while the other current top models skip this and rely on ensembling to reduce overfitting.
This explains why our single-model baseline outperforms most single-model results of the state-of-the-art models.
We found ensembling of the regularised baseline to provide a much smaller benefit in preliminary experiments compared to the results of ensembling unregularised networks reported in \citet{Teney2017}.

\section{Permutation-optimisation}\label{perm app:details}

In this section, we describe the experimental set-up of \autoref{cha:perm-optim} in detail.
All of our experiments can be reproduced using our implementation at \url{https://github.com/Cyanogenoid/perm-optim} in PyTorch~\citep{Paszke2017} through the \texttt{experiments/all.sh} script.
For the former three experiments, we use the following hyperparameters throughout:

\begin{itemize}
    \item Optimiser: Adam~\citep{Kingma2015} (default settings in PyTorch: $\beta_1 = 0.9, \beta_2 = 0.999, \eps = 10^{-8}$)
    \item Initial step size $\eta$ in inner gradient descent: 1.0
\end{itemize}

All weights are initialised with Xavier initialisation~\citep{Glorot2010}.
We choose the $f$ within the ordering cost function $F$ to be a small MLP.
The input to $f$ has 2 times the number of dimensions of each element, obtained by concatenating the pair of elements. This is done for all pairs that can be formed from the input set.
This is linearly projected to some number of hidden units to which a ReLU activation is applied.
Lastly, this is projected down to 1 dimension for sorting numbers and VQA, and 2 dimensions for assembling image mosaics (1 output for row-wise costs, 1 output for column-wise costs).
These outputs are used for creating the ordering cost matrix $\mC$.

\subsection{Sorting numbers}
\begin{itemize}
    \item Inner gradient descent steps $T$: 6
    \item Adam learning rate: 0.1
    \item Batch size: 512
    \item Number of sets to sort in training set: $2^{18}$
    \item Set sizes: 5, 10, 15, 80, 100, 120, 512, 1024
    \item Evaluation intervals: $[0, 1], [0, 10], [0, 1000], [1, 2], [10, 11],$\\
    $[100, 101], [1000, 1001]$ (same as in \citet{Mena2018})
    \item $F$ size of hidden dimension: 16
\end{itemize}

The ordering cost function $F$ concatenates the two floats of each pair and applies a 2-layer MLP that takes the 2 inputs to 16 hidden units, ReLU activation, then to one output.

For evaluation, we switch to double precision floating point numbers.
This is because for the interval $[1000, 1001]$, as the set size increases, there are not enough unique single precision floats in that interval for the sets to contain only unique floats with high probability (the birthday problem).
Using double precision floats avoids this issue.
Note that using single precision floats is enough for the other intervals and smaller set sizes, and training is always done on the interval [0, 1] at single precision.

\subsection{Re-assembling image mosaics}
\begin{itemize}
    \item Adam learning rate: $10^{-3}$
    \item Inner gradient descent steps $T$: 4
    \item Batch size: 32
    \item Training epochs: 20 (MNIST, CIFAR10) or 1 (ImageNet)
    \item $F$ size of hidden dimension: 64 (MNIST, CIFAR10) or 128 (ImageNet)
\end{itemize}

For all three image datasets from which we take images (MNIST, CIFAR10, ImageNet), we first normalise the inputs to have zero mean and standard deviation one over the dataset as is common practice.
For ImageNet, we crop rectangular images to be square by reducing the size of the longer side to the length of the shorter side (centre cropping).
Images that are not exactly divisible by the number of tiles are first rescaled to the nearest bigger image size that is exactly divisible.
Following \citet{Mena2018}, we process each tile with a $5 \times 5$ convolution with padding and stride 1, $2 \times 2$ max pooling, and ReLU activation.
This is flattened into a vector to obtain the feature vector for each tile, which is then fed into our $F$.
Unlike \citet{Mena2018}, we decide not to arbitrarily upscale MNIST images by a factor of two, even when upscaling results in slightly better performance in general.

While we were able to mostly reproduce their MNIST results, we were not able to reproduce their ImageNet results for the $3 \times 3$ case.
In general, we observed that good settings for their model also improved the results of our PO-U and PO-LA models.
Better hyperparameters than what we used should improve all models similarly while keeping the ordering of how well they perform the same.

This task is also known as jigsaw puzzle~\citep{Noroozi2016}, but we decided on naming it image mosaics because the tiles are square which can lead to multiple solutions, rather than the typical unique solution in traditional jigsaw puzzles enforced by the different tile shapes.

\subsection{Implicit permutations through classification}
We use the same setting as for the image mosaics, but further process the output image with a ResNet-18.
For MNIST and CIFAR10, we replace the first convolutional layer with one that has a $3 \times 3$ kernel size and no striding.
This ResNet-18 is first trained on the original dataset for 20 epochs (1 for ImageNet), though images may be rescaled if the image size is not divisible by the number of tiles per side.
All weights are then frozen and the permutation method is trained for 20 epochs (1 for ImageNet).
As stated previously, this is necessary in order for the ResNet-18 to not use each tile individually and ignore the resulting artefacts from the permuted tiles.
This is also one of the reasons why we downscale ImageNet images to $64 \times 64$ pixels.
Because the resulting image tiles are so big while the receptive field of ResNet-18 is relatively small if we were to use $256 \times 256$ images, the permutation artefacts barely affect results because they are only a small fraction of the globally-pooled features.
The permutation permutes each set of tiles, which are reconstructed (without use of the Hungarian algorithm) into an image, which is then processed by the ResNet-18.

We observed that the LinAssign model by \citet{Mena2018} consistently results in NaN values after Sinkhorn normalisation in this set-up, despite our Sinkhorn implementation using the numerically-stable version of softmax with the exp-normalise trick.
We avoided this issue by clipping the outputs of their model into the [-10, 10] interval before Sinkhorn normalisation.
We did not observe these NaN issues with our PO-U model.

\subsection{Visual question answering}
We use the official implementation of BAN as baseline without changing any of the hyperparameters.
We thus refer to~\citep{Kim2018} for details of their model architecture and hyperparameters.
The only change to hyperparameters that we make is reducing the batch size from 256 to 112 due to the GPU memory requirements of the baseline model, even without our permutation mechanism.

The BAN model generates attention weights between all object proposals in a set and words of the question.
We take the attention weight for a single object proposal to be the maximum attention weight for that proposal over all words of the question, the same as in their integration of the counting module.
Each element of the set, corresponding to object proposals, is the concatenation of this attention logit, bounding box coordinates, and the feature vector projected from 2048 down to 8 dimensions.
We found this projection necessary to not inhibit learning of the rest of the model, which might be due to gradient clipping or other hyperparameters that are no longer optimal in the BAN model.
This set of object proposals is then permuted with $T = 3$ and a 2-layer MLP with hidden dimension 128 for $f$ to produce the ordering costs.
The elements in the permuted sequence are weighted by how relevant each proposal is (sigmoid of the corresponding attention logit) and the sequence is then fed into an LSTM with 128 units.
The last cell state of the LSTM is the set representation which is projected, ReLUd, and added back into the hidden state of the BAN model.
The remainder of the BAN model is now able to use information from this set representation.
There are 8 attention glimpses, so we process each of these with a PO-U module and an LSTM with shared parameters across these 8 glimpses.

\section{Featurewise sort pooling}\label{fspool app:details}
In this section, we describe the experimental set-up of \autoref{cha:fspool} in detail.
We provide the code to reproduce all experiments at \url{https://github.com/Cyanogenoid/fspool}.

For almost all experiments, we used FSPool and the unpooling version of it with $k=20$.
We guessed this value without tuning, and we did not observe any major differences when we tried to change this on CLEVR to $k=5$ and $k=40$.
$\mbW$ can be initialised in different ways, such as by sampling from a standard Gaussian.
However, for the purposes of starting the model as similarly as possible to the sum pooling baseline on CLEVR and on the graph classification datasets, we initialise $\mbW$ to a matrix of all 1s on them.

\subsection{Polygons}
The polygons are centred on 0 with a radius of 1.
The points in the set are randomly permuted to remove any ordering in the set from the generation process that a model that is not permutation-invariant or permutation-equivariant could exploit.
We use a batch size of 16 for all three models and train it for 10240 steps.
We use the Adam optimiser~\citep{Kingma2015} with 0.001 learning rate and their suggested values for the other optimiser parameters (PyTorch defaults).
Weights of linear and convolutional layers are initialised as suggested in \citet{Glorot2010}.
The size of every hidden layer is set to 16 and the latent space is set to 1 (it should only need to store the rotation as latent variable).
We have also tried much hidden and latent space sizes of 128 when we tried to get better results for the baselines.

\subsection{MNIST}
We train on the training set of MNIST for 10 epochs and the shown results come from the test set of MNIST.
For an image, the coordinate of a pixel is included if the pixel is above the mean pixel level of 0.1307 (with pixel levels ranging 0--1).
Again, the order of the points are randomised.
We did not include results of the Hungarian loss because we did not get the model to converge to results of similar quality to the direct MSE loss or Chamfer loss, and training time took too long ($>$ 1 day) in order to find better parameters.

The latent space is increased from 1 to 16 and the size of the hidden layers is increased from 16 to 32.
All other hyperparameters are the the same as for the Polygons dataset.

\subsection{CLEVR}
The architecture and hyperparameters come from the open-source implementation available at\\\url{https://github.com/mesnico/RelationNetworks-CLEVR}.\\
For the RN baseline, the set is first expanded into the set of all pairs by concatenating the 2 feature vectors of the pair for all pairs of elements in the set.
For the Janossy Pooling baseline, we use the model configuration from \citet{Murphy2019} that appeared best in their experiments, which uses $\pi$-SGD with an LSTM that has $|h|$ as neighbourhood size.

The question representation coming from the 256-unit LSTM, processing the question tokens in reverse with each token embedded into 32 dimensions, is concatenated to all elements in the set.
Each element of this new set is first processed by a 4-layer MLP with 512 neurons in each layer and ReLU activations.
The set of feature vectors is pooled with a pooling method like sum and the output of this is processed with a 3-layer MLP (hidden sizes 512, 1024, and number of answer classes) with ReLU activations.
A dropout rate of 0.05 is applied before the last layer of this MLP.
Adam is used with a starting learning rate of 0.000005, which doubles every 20 epochs until the maximum learning rate of 0.0005 is reached.
Weight decay of 0.0001 is applied.
The model is trained for 350 epochs.

\subsection{Graph classification}
The GIN architecture starts with 5 sequential blocks of graph convolutions.
Each block starts with summing the feature vector of each node's neighbours into the node's own feature vector.
Then, an MLP is applied to the feature vectors of all the nodes individually.
The details of this MLP were somewhat unclear in~\citep{Xu2019} and we chose Linear-ReLU-BN-Linear-ReLU-BN in the end.
We tried Linear-BN-ReLU-Linear-BN-ReLU as well, which gave us slightly worse validation results for both the baseline and the FSPool version.
The outputs of each of the 5 blocks are concatenated and pooled, either with a sum for the social network datasets, mean for the social network datasets (this is as specified in GIN), or with FSPool for both types of datasets.
This is followed by BN-Linear-ReLU-Dropout-Linear as classifier with a softmax output and cross-entropy loss.
We used the torch-geometric library~\citep{Fey2018} to implement this model.

The starting learning rate for Adam is 0.01 and is reduced every 50 epochs.
Weights are initialised as suggested in~\citep{Glorot2010}.
The hyperparameters to choose from are: dropout ratio $\in \{0, 0.5\}$, batch size $\in \{32, 128\}$, if bioinformatics dataset hidden sizes of all layers $\in \{16, 32\}$ and 500 epochs, if social network dataset the hidden size is 64 and 250 epochs.
Due to GPU memory limitations we used a batch size of 100 instead of 128 for social network datasets.
The best hyperparameters are selected based on best average validation accuracy across the 10-fold cross-validation, where one of the 9 training folds is used as validation set each time.
In other words, within one 10-fold cross-validation run the hyperparameters used for the test set are the same, while across the 10 repeats of this with different seeds the best hyperparameters may differ.

\begin{table}
    \centering
    \caption[Graph classification hyperparameters]{
        Average of best hyperparameters over 10 repeats.
    }
    \label{fspool tab:graph-params}
    \small
    \setlength{\tabcolsep}{3pt}
    \centerfloat{\begin{tabular}{lccccccccc}
    \toprule
    {} & \footnotesize IMDB-B & \footnotesize IMDB-M & \footnotesize RDT-B & \footnotesize RDT-M5K & \footnotesize COLLAB & \footnotesize MUTAG & \footnotesize PROTEINS & \footnotesize PTC & \footnotesize NCI1 \\
    \midrule
    \textsc{GIN-FSPool} \\
    - \textit{dimensionality} & 64.0 & 64.0 & 64.0 & 64.0 & 64.0 & 28.8 & 19.2 & 28.8 & 30.4 \\
    - \textit{batch size} & 66.0 & 100 & 45.6 & 32.0 & 86.4 & 89.6 & 60.8 & 41.6 & 128 \\
    - \textit{dropout} & 0.25 & 0.15 & 0.35 & 0.10 & 0.40 & 0.15 & 0.35 & 0.20 & 0.50 \\
    \midrule
    \textsc{GIN-Base} \\
    - \textit{dimensionality} & 64.0 & 64.0 & 64.0 & 64.0 & 64.0 & 27.2 & 20.8 & 25.6 & 28.8 \\
    - \textit{batch size} & 86.4 & 93.2 & 72.8 & 100 & 100 & 70.4 & 60.8 & 60.8 & 128 \\
    - \textit{dropout} & 0.30 & 0.15 & 0.25 & 0.45 & 0.40 & 0.25 & 0.45 & 0.20 & 0.35 \\
    \bottomrule
    \end{tabular}}
\end{table}

\section{Deep set prediction networks}\label{dspn app:details}
In this section, we describe the experimental set-up of \autoref{cha:dspn} in detail.
In our algorithm, $\eta$ was chosen in initial experiments and we did not tune it beyond that.
We did this by increasing $\eta$ until the output set visibly changed between inner optimisation steps when the set encoder is randomly initialised.
This makes it so that changing the set encoder weights has a noticeable effect rather than being stuck with $\mhY^{(T)}~\approx~\mhY^{(0)}$.

$T=10$ was chosen because it seemed to be enough to converge to good solutions on MNIST.
We simply kept this for the supervised experiments on CLEVR.

In the supervised experiments, we would often observe large spikes in training that cause the model diverge when $\lambda = 1$.
By changing around various parameters, we found that reducing $\lambda$ eliminated most of this issue and also made training converge to better solutions.
Much smaller values than 0.1 converged to worse solutions.
This is likely because the issue of not having the $\Lrepr(\mY, \vz)$ term in the outer loss in the first place ($\lambda=0$) is present again -- see \autoref{dspn sec:predict}.

For all experiments, we used Adam with the default momentum values and batch size 32 for the outer optimisation.
The only hyperparameter we tuned in the experiments is the learning rate of the outer optimisation.
Every individual experiment is run on a single 1080 Ti GPU.

The MLP decoder baseline has 3 layers with 256 (MNIST) or 512 (CLEVR) neurons in the first two layers and the number of channels of the output set in the task in the third layer.
The LSTM decoder linearly transforms the latent space into 256 (MNIST) or 512 (CLEVR) dimensions, which is used as initial cell state of the LSTM.
The LSTM is run for the same number of steps as the maximum set size, and the outputs of these steps is each linearly transformed into the output dimensionality.

\subsection{MNIST}
For MNIST, we train our model and the baseline model for 100 epochs to make sure that they have converged.
Both models have a 3-layer MLP with ReLU activations and 256, 256, and 64 neurons in the three layers.
For simplicity, sets are padded to a fixed size for FSPool.
FSPool has 20 pieces in its piecewise linear function.
We tried learning rates in $\{1.0, 0.1, 0.03, 0.01, 0.003, 0.001, 0.0003, 0.0001, 0.00001\}$ and chose $0.01$.
For the baselines, none of the other learning rates performed significantly better than the one we chose.

The baselines are trained slightly differently to our model.
They do not output mask values natively, so we have to train them with the mask values in the training target.
In other words, they are trained to predict x coordinate, y coordinate, and the mask for each point.
We found it crucial to explicitly add 1 to the mask in the baseline model for good results.
Otherwise, many of the baseline outputs get stuck in the local optimum of predicting the (0, 0, 0) point and the output is too sparse.

\subsection{CLEVR}
We train our model and the baselines models for 100 epochs on the training set of CLEVR and evaluate on the validation set, since no ground-truth scene information is available for the test set.
All images are resized to 128$\times$128 resolution.
The set encoder is a 2-layer Relation Network with ReLU activation between the two layers, wherein the sum pooling is replaced with FSPool.
The two layers have 512 neurons each.
Because we use the Hungarian loss instead of the Chamfer loss here, including the mask feature in the target set does not worsen results, so we include the mask target for both the baseline and our model for consistency.
To tune the learning rate, we started with the learning rate found for MNIST and decreased it similarly-sized steps until the training accuracy after 100 epochs worsened.
We settled on 0.0003 as learning rate for both the bounding box and the state prediction task.
All other hyperparameters are kept the same as for MNIST.
The ResNet34 that encodes the image is not pre-trained.

\subsection{Fixed results}\label{dspn app:fixed}

Due to a bug, the practical trick of computing the loss over all $\mhY^{(t)}$ instead of only the last $\mhY^{(T)}$ was not used for the experiments on CLEVR.
Also, all evaluation numbers on CLEVR with bounding boxes are overestimates, but consistent across the models.
These are minor problems that do not affect the conclusions of the experiments.
The first bug was found by Sangwoo Mo \footnote{\texttt{swmo@kaist.ac.kr}, Korea Advanced Institute of Science and Technology.}, the second bug was found by Chao Feng \footnote{\texttt{chf2018@med.cornell.edu}, Weill Cornell Graduate School of Medical Sciences.}.
We are very grateful to them for finding and letting us know of these bugs.
The following tables show the results with the bugs fixed when using exactly the same hyperparameters as before:
\begin{itemize}
    \item \autoref{dspn tab:box-fixed1} shows the results when only fixing the evaluation bug (note the different AP metrics compared to \autoref{dspn tab:box}) for bounding box prediction,
    \item \autoref{dspn tab:box-fixed2} shows the results when fixing the evaluation bug and also computing the loss on all $\mhY^{(t)}$ for bounding box prediction,
    \item \autoref{dspn tab:state-fixed} shows the results when computing the loss on all $\mhY^{(t)}$ for state prediction.
\end{itemize}

In summary, our model still handily beats the baseline models.
On bounding box prediction, interestingly the results with the set loss bug fixed tend to be worse than when only using the last prediction to compute the loss. 
On state prediction, fixing the set loss bug improves almost all results.
The largest improvement is on the AP\textsubscript{0.5} metric with an increase from 57.0 for the best result to 66.8 for the best result.
Notably, running more iterations now only results in strict improvements without worsening on any metric.

\begin{table}
    \centering
    \caption{Fixed bounding box evaluation, still computing loss only on the last $\mhY^{(T)}$. Average Precision (AP) for different intersection-over-union thresholds for a predicted bounding box to be considered correct. Higher is better. Mean and standard deviation over 6 runs.}
    \label{dspn tab:box-fixed1}
    \begin{tabular}{lccccc}
        \toprule
        Model & AP\textsubscript{50} & AP\textsubscript{60} & AP\textsubscript{70} & AP\textsubscript{80} & AP\textsubscript{90} \\
        \midrule
        MLP baseline & 87.6\tiny$\pm$2.4 & 73.6\tiny$\pm$3.9 & 48.1\tiny$\pm$5.0 & 15.1\tiny$\pm$2.8 & 0.1\tiny$\pm$0.1 \\
        RNN baseline & 85.8\tiny$\pm$2.5 & 70.0\tiny$\pm$2.8 & 43.9\tiny$\pm$2.4 & 13.3\tiny$\pm$1.5 & 0.2\tiny$\pm$0.0 \\
        Ours (10 iters) & 94.0\tiny$\pm$0.4 & 90.6\tiny$\pm$0.6 & {82.2}\tiny$\pm$1.4 & \textbf{58.9}\tiny$\pm$3.4 & \textbf{16.0}\tiny$\pm$2.4 \\
        Ours (20 iters) & \textbf{98.0}\tiny$\pm$0.5 & \textbf{94.4}\tiny$\pm$0.9 & \textbf{83.0}\tiny$\pm$1.6 & 55.0\tiny$\pm$3.3 & 12.3\tiny$\pm$2.6 \\
        Ours (30 iters) & {95.9}\tiny$\pm$2.1 & {89.4}\tiny$\pm$3.3 & 74.8\tiny$\pm$3.9 & 46.1\tiny$\pm$3.4 & 9.4\tiny$\pm$1.9 \\
        \bottomrule
    \end{tabular}
    
    \vspace{1cm}

    \centering
    \caption{Fixed bounding box evaluation and computing the loss on all intermediate sets. Average Precision (AP) for different intersection-over-union thresholds for a predicted bounding box to be considered correct. Higher is better. Mean and standard deviation over 6 runs. Baselines are the same as in \autoref{dspn tab:box-fixed1}.}
    \label{dspn tab:box-fixed2}
    \begin{tabular}{lccccc}
        \toprule
        Model & AP\textsubscript{50} & AP\textsubscript{60} & AP\textsubscript{70} & AP\textsubscript{80} & AP\textsubscript{90} \\
        \midrule
        Ours (10 iters) & 97.1\tiny$\pm$1.1 & \textbf{91.4}\tiny$\pm$2.8 & \textbf{73.4}\tiny$\pm$5.1 & \textbf{35.4}\tiny$\pm$4.8 & \textbf{3.3}\tiny$\pm$1.3 \\
        Ours (20 iters) & \textbf{97.5}\tiny$\pm$1.0 & {90.8}\tiny$\pm$2.6 & {69.4}\tiny$\pm$4.9 & 31.3\tiny$\pm$4.7 & 2.9\tiny$\pm$1.2 \\
        Ours (30 iters) & {97.1}\tiny$\pm$1.0 & {88.6}\tiny$\pm$2.9 & 64.4\tiny$\pm$5.3 & 27.7\tiny$\pm$5.2 & 2.6\tiny$\pm$1.2\\
        \bottomrule
    \end{tabular}
    
    \vspace{1cm}

    \centering
\caption{State predicton when computing the loss on all intermediate sets. Average Precision (AP) in \% for different distance thresholds of a predicted set element to be considered correct. AP\textsubscript{$\infty$} only requires all attributes to be correct, regardless of 3d position. Higher is better. Mean and standard deviation over 6 runs. Baselines are the same as in \autoref{dspn tab:state}.}
    \label{dspn tab:state-fixed}
    \begin{tabular}{lccccc}
        \toprule
        Model & AP\textsubscript{$\infty$} & AP\textsubscript{1} & AP\textsubscript{0.5} & AP\textsubscript{0.25} & AP\textsubscript{0.125}   \\
        \midrule
        Ours (10 iters) & 76.4\tiny$\pm$2.5 & 69.6\tiny$\pm$2.8 & 53.7\tiny$\pm$4.3 & 18.2\tiny$\pm$3.7 & 2.2\tiny$\pm$0.6 \\
        Ours (20 iters) & 80.9\tiny$\pm$1.9 & 77.3\tiny$\pm$2.3 & {63.9}\tiny$\pm$4.1 & {22.7}\tiny$\pm$4.4 & {2.7}\tiny$\pm$0.6 \\
        Ours (30 iters) & \textbf{82.5}\tiny$\pm$1.7 & \textbf{79.8}\tiny$\pm$2.0 & \textbf{66.8}\tiny$\pm$4.1 & \textbf{23.6}\tiny$\pm$4.6 & \textbf{2.8}\tiny$\pm$0.8 \\
        \bottomrule
    \end{tabular}
\end{table}

\backmatter
\begingroup
\setlength{\emergencystretch}{0.65em}
\printbibliography[heading=bibintoc]

@InProceedings{Zhang2018,
  author    = {Yan Zhang and Jonathon Hare and Adam Pr\"ugel-Bennett},
  title     = {Learning to Count Objects in Natural Images for Visual Question Answering},
  booktitle = {International Conference on Learning Representations},
  year      = {2018},
  eprint    = {1802.05766},
  url       = {https://openreview.net/forum?id=B12Js_yRb},
}

@InProceedings{Johnson2017a,
  author    = {Johnson, Justin and Hariharan, Bharath and van der Maaten, Laurens and Fei-Fei, Li and Lawrence Zitnick, C. and Girshick, Ross},
  title     = {{CLEVR}: A Diagnostic Dataset for Compositional Language and Elementary Visual Reasoning},
  booktitle = {CVPR},
  year      = {2017},
}

@InProceedings{Mena2018,
  author    = {Gonzalo Mena and David Belanger and Scott Linderman and Jasper Snoek},
  title     = {{Learning Latent Permutations with Gumbel-Sinkhorn Networks}},
  booktitle = {ICLR},
  year      = {2018},
}

@InProceedings{Vinyals2015,
  author    = {Oriol Vinyals and Samy Bengio and Manjunath Kudlur},
  title     = {{Order Matters: Sequence to sequence for sets}},
  booktitle = {ICLR},
  year      = {2015},
}

@InProceedings{Qi2017,
  author    = {Charles R. Qi and Hao Su and Kaichun Mo and Leonidas J. Guibas},
  title     = {{PointNet: Deep Learning on Point Sets for 3D Classification and Segmentation}},
  booktitle = {CVPR},
  year      = {2017},
}

@InCollection{Zaheer2017,
  author    = {Zaheer, Manzil and Kottur, Satwik and Ravanbakhsh, Siamak and Poczos, Barnabas and Salakhutdinov, Ruslan R and Smola, Alexander J},
  title     = {{Deep Sets}},
  booktitle = {NIPS},
  year      = {2017},
}

@InProceedings{Cruz2017,
  author    = {Rodrigo Santa Cruz and Basura Fernando and Anoop Cherian and Stephen Gould},
  title     = {{DeepPermNet: Visual Permutation Learning}},
  booktitle = {CVPR},
  year      = {2017},
}

@InProceedings{Noroozi2016,
  author    = {Mehdi Noroozi and Paolo Favaro},
  title     = {Unsupervised Learning of Visual Representations by Solving Jigsaw Puzzles},
  booktitle = {ECCV},
  year      = {2016},
}

@InCollection{Santoro2017,
  author    = {Santoro, Adam and Raposo, David and Barrett, David G and Malinowski, Mateusz and Pascanu, Razvan and Battaglia, Peter and Lillicrap, Tim},
  title     = {A simple neural network module for relational reasoning},
  booktitle = {NIPS},
  year      = {2017},
}

@InProceedings{Amos2017,
  author    = {Brandon Amos and J. Zico Kolter},
  title     = {OptNet: Differentiable Optimization as a Layer in Neural Networks},
  booktitle = {ICML},
  year      = {2017},
}

@InProceedings{Anderson2018,
  author    = {Peter Anderson and Xiaodong He and Chris Buehler and Damien Teney and Mark Johnson and Stephen Gould and Lei Zhang},
  title     = {Bottom-Up and Top-Down Attention for Image Captioning and Visual Question Answering},
  booktitle = {CVPR},
  year      = {2018},
}

@InProceedings{Antol2015,
  author    = {Stanislaw Antol and Aishwarya Agrawal and Jiasen Lu and Margaret Mitchell and Dhruv Batra and C. Lawrence Zitnick and Devi Parikh},
  title     = {{VQA: Visual Question Answering}},
  booktitle = {ICCV},
  year      = {2015},
}

@InProceedings{Goyal2017,
  author    = {Yash Goyal and Tejas Khot and Douglas Summers{-}Stay and Dhruv Batra and Devi Parikh},
  title     = {Making the {V} in {VQA} Matter: Elevating the Role of Image Understanding in {V}isual {Q}uestion {A}nswering},
  booktitle = {CVPR},
  year      = {2017},
}

@InProceedings{Kim2018,
  author    = {Jin-Hwa Kim and Jaehyun Jun and Byoung-Tak Zhang},
  title     = {Bilinear Attention Networks},
  booktitle = {NIPS},
  year      = {2018},
}

@InCollection{Vinyals2015a,
  author    = {Vinyals, Oriol and Fortunato, Meire and Jaitly, Navdeep},
  title     = {Pointer Networks},
  booktitle = {NIPS},
  year      = {2015},
}

@Article{Pardalos1991,
  author  = {Panos M. Pardalos and Stephen A. Vavasis},
  title   = {{Quadratic programming with one negative eigenvalue is NP-hard}},
  journal = {J. Global Optimization},
  year    = {1991},
  volume  = {1},
  number  = {1},
  pages   = {15--22},
  doi     = {10.1007/BF00120662},
}

@InProceedings{Domke2012,
  author    = {Justin Domke},
  title     = {{Generic Methods for Optimization-Based Modeling}},
  booktitle = {AISTATS},
  year      = {2012},
}

@InProceedings{Stoyanov2011,
  author    = {Veselin Stoyanov and Alexander Ropson and Jason Eisner},
  title     = {Empirical Risk Minimization of Graphical Model Parameters Given Approximate Inference, Decoding, and Model Structureeneric methods for optimization-based modelin},
  booktitle = {AISTATS},
  year      = {2011},
}

@InProceedings{Belanger2017,
  author    = {Belanger, David and Yang, Bishan and McCallum, Andrew.},
  title     = {End-to-End Learning for Structured Prediction Energy Networks},
  booktitle = {ICML},
  year      = {2017},
}

@InProceedings{Burges2005,
  author    = {Burges, Chris and Shaked, Tal and Renshaw, Erin and Lazier, Ari and Deeds, Matt and Hamilton, Nicole and Hullender, Greg},
  title     = {Learning to rank using gradient-descent},
  booktitle = {ICML},
  year      = {2005},
}

@InProceedings{Severyn2015,
  author    = {Severyn, Aliaksei and Moschitti, Alessandro},
  title     = {Learning to Rank Short Text Pairs with Convolutional Deep Neural Networks},
  booktitle = {SIGIR},
  year      = {2015},
}

@InProceedings{Fogel2013,
  author    = {Fogel, Fajwel and Jenatton, Rodolphe and Bach, Francis and D\textquotesingle Aspremont, Alexandre},
  title     = {Convex Relaxations for Permutation Problems},
  booktitle = {NIPS},
  year      = {2013},
}

@InProceedings{Kipf2017,
  author    = {Thomas N. Kipf and Max Welling},
  title     = {Semi-Supervised Classification with Graph Convolutional Networks},
  booktitle = {ICLR},
  year      = {2017},
}

@InProceedings{Paszke2017,
  author    = {Paszke, Adam and Gross, Sam and Chintala, Soumith and Chanan, Gregory and Yang, Edward and DeVito, Zachary and Lin, Zeming and Desmaison, Alban and Antiga, Luca and Lerer, Adam},
  title     = {Automatic differentiation in PyTorch},
  booktitle = {NIPS-W},
  year      = {2017},
}

@InProceedings{Glorot2010,
  author    = {Xavier Glorot and Yoshua Bengio},
  title     = {Understanding the difficulty of training deep feedforward neural networks},
  booktitle = {AISTATS},
  year      = {2010},
}

@InProceedings{Kingma2015,
  author    = {Diederik P. Kingma and Jimmy Ba},
  title     = {Adam: {A} Method for Stochastic Optimization},
  booktitle = {ICLR},
  year      = {2015},
}

@Article{Adams2011,
  author = {Ryan Prescott Adams and Richard S. Zemel},
  title  = {Ranking via Sinkhorn Propagation},
  year   = {2011},
  volume = {arXiv:1106.1925},
}

@Article{Sinkhorn1964,
  author    = {Sinkhorn, Richard},
  title     = {A Relationship Between Arbitrary Positive Matrices and Doubly Stochastic Matrices},
  journal   = {Ann. Math. Statist.},
  year      = {1964},
  volume    = {35},
  number    = {2},
  month     = {06},
  pages     = {876--879},
  doi       = {10.1214/aoms/1177703591},
  fjournal  = {The Annals of Mathematical Statistics},
  publisher = {The Institute of Mathematical Statistics},
}

@Article{Charon2007,
  author    = {Ir{\`{e}}ne Charon and Olivier Hudry},
  title     = {A survey on the linear ordering problem for weighted or unweighted tournaments},
  journal   = {4OR},
  year      = {2007},
  volume    = {5},
  number    = {1},
  month     = {mar},
  pages     = {5--60},
  doi       = {10.1007/s10288-007-0036-6},
  publisher = {Springer Nature},
}

@InProceedings{Murphy2019,
  author    = {Ryan L. Murphy and Balasubramaniam Srinivasan and Vinayak Rao and Bruno Ribeiro},
  title     = {Janossy Pooling: Learning Deep Permutation-Invariant Functions for Variable-Size Inputs},
  booktitle = {ICLR},
  year      = {2019},
}

@InProceedings{Xu2019,
  author    = {Keyulu Xu and Weihua Hu and Jure Leskovec and Stefanie Jegelka},
  title     = {How Powerful are Graph Neural Networks?},
  booktitle = {International Conference on Learning Representations (ICLR)},
  year      = {2019},
}

@InProceedings{Niepert2016,
  author    = {Mathias Niepert and Mohamed Ahmed and Konstantin Kutzkov},
  title     = {Learning Convolutional Neural Networks for Graphs},
  booktitle = {Proceedings of the 33rd International Conference on Machine Learning (ICML)},
  year      = {2016},
}

@InProceedings{Belanger2016,
  author    = {David Belanger and Andrew McCallum},
  title     = {Structured Prediction Energy Networks},
  booktitle = {Proceedings of the 33rd International Conference on Machine Learning (ICML)},
  year      = {2016},
}

@Article{Rezatofighi2018,
  author      = {S. Hamid Rezatofighi and Roman Kaskman and Farbod T. Motlagh and Qinfeng Shi and Daniel Cremers and Laura Leal-Taixé and Ian Reid},
  title       = {Deep Perm-Set Net: Learn to predict sets with unknown permutation and cardinality using deep neural networks},
  year        = {2018},
  date        = {2018-05-02},
  volume      = {arXiv:1805.00613},
  eprint      = {http://arxiv.org/abs/1805.00613v4},
  eprintclass = {cs.CV},
  eprinttype  = {arXiv},
}

@InProceedings{Ren2015,
  author    = {Ren, Shaoqing and He, Kaiming and Girshick, Ross and Sun, Jian},
  title     = {Faster {R-CNN}: Towards Real-Time Object Detection with Region Proposal Networks},
  booktitle = {Advances in Neural Information Processing Systems 28 (NeurIPS)},
  year      = {2015},
}

@Misc{Balles2019,
  author = {Lukas Balles and Thomas Fischbacher},
  title  = {Holographic and other Point Set Distances for Machine Learning},
  year   = {2019},
  url    = {https://openreview.net/forum?id=rJlpUiAcYX},
}

@InProceedings{Hu2018,
  author    = {Hu, Jie and Shen, Li and Albanie, Samuel and Sun, Gang and Vedaldi, Andrea},
  title     = {{Gather-Excite}: Exploiting Feature Context in Convolutional Neural Networks},
  booktitle = {Advances in Neural Information Processing Systems 31 (NeurIPS)},
  year      = {2018},
}

@InProceedings{Grover2019,
  author    = {Aditya Grover and Eric Wang and Aaron Zweig and Stefano Ermon},
  title     = {Stochastic Optimization of Sorting Networks via Continuous Relaxations},
  booktitle = {International Conference on Learning Representations (ICLR)},
  year      = {2019},
}

@Misc{Gao2019,
  author = {Hongyang Gao and Shuiwang Ji},
  title  = {Graph {U-Net}},
  year   = {2019},
  url    = {https://openreview.net/forum?id=HJePRoAct7},
}

@Article{Cangea2018,
  author      = {Cătălina Cangea and Petar Veličković and Nikola Jovanović and Thomas Kipf and Pietro Liò},
  title       = {Towards Sparse Hierarchical Graph Classifiers},
  journal     = {NeurIPS Workshop},
  year        = {2018},
  date        = {2018-11-03},
  volume      = {Relational Representation Learning},
  eprint      = {http://arxiv.org/abs/1811.01287v1},
  eprintclass = {stat.ML},
  eprinttype  = {arXiv},
}

@Article{Shi2016,
  author   = {Zenglin Shi and Yangdong Ye and Yunpeng Wu},
  title    = {Rank-based pooling for deep convolutional neural networks},
  journal  = {Neural Networks},
  year     = {2016},
  volume   = {83},
  pages    = {21 - 31},
  issn     = {0893-6080},
}

@InProceedings{Cisse2017,
  author    = {Moustapha Cisse and Piotr Bojanowski and Edouard Grave and Yann Dauphin and Nicolas Usunier},
  title     = {Parseval {Networks}: Improving Robustness to Adversarial Examples},
  booktitle = {Proceedings of the 34th International Conference on Machine Learning (ICML)},
  year      = {2017},
}

@InProceedings{Welleck2018,
  author    = {Welleck, Sean and Yao, Zixin and Gai, Yu and Mao, Jialin and Zhang, Zheng and Cho, Kyunghyun},
  title     = {Loss Functions for Multiset Prediction},
  booktitle = {Advances in Neural Information Processing Systems 31 (NeurIPS)},
  year      = {2018},
}

@InProceedings{You2018,
  author    = {You, Jiaxuan and Ying, Rex and Ren, Xiang and Hamilton, William and Leskovec, Jure},
  title     = {{G}raph{RNN}: Generating Realistic Graphs with Deep Auto-regressive Models},
  booktitle = {Proceedings of the 35th International Conference on Machine Learning (ICML)},
  year      = {2018},
}

@Misc{Messina2018,
  author = {Nicola Messina and Giuseppe Amato and Fabio Carrara and Fabrizio Falchi and Claudio Gennaro},
  title  = {Learning Relationship-aware Visual Features},
  year   = {2018},
  url    = {http://www.rcbir.org/learning-relationship-aware-preprint.pdf},
}

@Misc{Kersting2016,
  author = {Kristian Kersting and Nils M. Kriege and Christopher Morris and Petra Mutzel and Marion Neumann},
  title  = {Benchmark Data Sets for Graph Kernels},
  year   = {2016},
  url    = {http://graphkernels.cs.tu-dortmund.de},
}

@InProceedings{Perez2018,
  author    = {Ethan Perez and Florian Strub and Harm de Vries and Vincent Dumoulin and Aaron C. Courville},
  title     = {{FiLM}: Visual Reasoning with a General Conditioning Layer},
  booktitle = {Proceedings of the Thirty-Second AAAI Conference on Artificial Intelligence},
  year      = {2018},
}

@InProceedings{Johnson2017,
  author    = {Daniel D. Johnson},
  title     = {Learning Graphical State Transitions},
  booktitle = {International Conference on Learning Representations (ICLR)},
  year      = {2017},
}

@InProceedings{Long2015,
  author    = {Long, Jonathan and Shelhamer, Evan and Darrell, Trevor},
  title     = {Fully Convolutional Networks for Semantic Segmentation},
  booktitle = {The IEEE Conference on Computer Vision and Pattern Recognition (CVPR)},
  year      = {2015},
}

@InProceedings{Vaswani2017,
  author    = {Vaswani, Ashish and Shazeer, Noam and Parmar, Niki and Uszkoreit, Jakob and Jones, Llion and Gomez, Aidan N and Kaiser, \L ukasz and Polosukhin, Illia},
  title     = {Attention is All you Need},
  booktitle = {Advances in Neural Information Processing Systems 3 (NeurIPS) 30},
  year      = {2017},
}

@InProceedings{Fey2018,
  author    = {Fey, Matthias and Lenssen, Jan Eric and Weichert, Frank and M{\"u}ller, Heinrich},
  title     = {{SplineCNN}: Fast Geometric Deep Learning with Continuous {B}-Spline Kernels},
  booktitle = {The IEEE Conference on Computer Vision and Pattern Recognition (CVPR)},
  year      = {2018},
}

@InProceedings{Yang2018,
  author    = {Yang, Yaoqing and Feng, Chen and Shen, Yiru and Tian, Dong},
  title     = {{FoldingNet}: Point Cloud Auto-Encoder via Deep Grid Deformation},
  booktitle = {The IEEE Conference on Computer Vision and Pattern Recognition (CVPR)},
  year      = {2018},
  month     = {June},
}

@InProceedings{Fan2017,
  author    = {Fan, Haoqiang and Su, Hao and Guibas, Leonidas J.},
  title     = {A Point Set Generation Network for {3D} Object Reconstruction From a Single Image},
  booktitle = {The IEEE Conference on Computer Vision and Pattern Recognition (CVPR)},
  year      = {2017},
}

@InProceedings{Ying2018,
  author    = {Ying, Zhitao and You, Jiaxuan and Morris, Christopher and Ren, Xiang and Hamilton, Will and Leskovec, Jure},
  title     = {Hierarchical Graph Representation Learning with Differentiable Pooling},
  booktitle = {Advances in Neural Information Processing Systems 31 (NeurIPS)},
  year      = {2018},
}

@InProceedings{Atwood2016,
  author    = {Atwood, James and Towsley, Don},
  title     = {Diffusion-Convolutional Neural Networks},
  booktitle = {Advances in Neural Information Processing Systems 29 (NeurIPS)},
  year      = {2016},
}

@InProceedings{Achlioptas2018,
  author    = {Achlioptas, Panos and Diamanti, Olga and Mitliagkas, Ioannis and Guibas, Leonidas J},
  title     = {Learning Representations and Generative Models For {3D} Point Clouds},
  booktitle = {Proceedings of the 35th International Conference on Machine Learning (ICML)},
  year      = {2018},
}

@Article{Neumann2016,
  author     = {Neumann, Marion and Garnett, Roman and Bauckhage, Christian and Kersting, Kristian},
  title      = {Propagation Kernels: Efficient Graph Kernels from Propagated Information},
  journal    = {Machine Learning},
  year       = {2016},
  volume     = {102},
  number     = {2},
  pages      = {209--245},
  issn       = {0885-6125},
  acmid      = {2887323},
  address    = {Hingham, MA, USA},
  issue_date = {February 2016},
  numpages   = {37},
  publisher  = {Kluwer Academic Publishers},
}

@Article{Shervashidze2011,
  author     = {Shervashidze, Nino and Schweitzer, Pascal and van Leeuwen, Erik Jan and Mehlhorn, Kurt and Borgwardt, Karsten M.},
  title      = {{Weisfeiler-Lehman Graph Kernels}},
  journal    = {Journal of Machine Learning Research},
  year       = {2011},
  volume     = {12},
  pages      = {2539--2561},
  issn       = {1532-4435},
  acmid      = {2078187},
  issue_date = {2/1/2011},
  numpages   = {23},
  publisher  = {JMLR.org},
}

@Article{Li2018a,
  author      = {Yujia Li and Oriol Vinyals and Chris Dyer and Razvan Pascanu and Peter Battaglia},
  title       = {Learning Deep Generative Models of Graphs},
  year        = {2018},
  date        = {2018-03-08},
  volume      = {arXiv:1803.03324},
  eprint      = {1803.03324v1},
  eprintclass = {cs.LG},
  eprinttype  = {arXiv},
}

@Article{Zhang2019FSPool,
  author = {Yan Zhang and Jonathon Hare and Adam Pr\"ugel-Bennett},
  title  = {{FSPool}: Learning Set Representations with Featurewise Sort Pooling},
  year   = {2019},
  volume = {arXiv:1906.02795},
  eprint = {1906.02795},
}

@InProceedings{Hudson2018,
  author    = {Drew A. Hudson and Christopher D. Manning},
  title     = {{Compositional Attention Networks for Machine Reasoning}},
  booktitle = {International Conference on Learning Representations (ICLR)},
  year      = {2018},
}

@Article{Stewart2016,
  author      = {Russell Stewart and Mykhaylo Andriluka},
  title       = {End-to-end people detection in crowded scenes},
  journal     = {The IEEE Conference on Computer Vision and Pattern Recognition (CVPR)},
  year        = {2016},
  date        = {2015-06-16},
  eprint      = {1506.04878v3},
  eprintclass = {cs.CV},
  eprinttype  = {arXiv},
}

@InProceedings{Zhang2019DSPN,
  author    = {Yan Zhang and Jonathon Hare and Adam Pr\"ugel-Bennett},
  title     = {{Deep Set Prediction Networks}},
  booktitle = {Advances in Neural Information Processing Systems (NeurIPS)},
  year      = {2019},
  eprint    = {1906.06565},
}

@Article{Cybenko1989,
author="Cybenko, George",
title="Approximation by superpositions of a sigmoidal function",
journal="Mathematics of Control, Signals and Systems",
year="1989",
volume="2",
number="4",
pages="303--314",
issn="1435-568X",
doi="10.1007/BF02551274",
}

@InProceedings{Zhang2019PermOptim,
  author    = {Yan Zhang and Jonathon Hare and Adam Prügel-Bennett},
  title     = {Learning Representations of Sets through Optimized Permutations},
eprint = { 1812.03928 },
eprinttype = { arxiv },
  booktitle = {International Conference on Learning Representations (ICLR)},
  year      = {2019},
}

@InProceedings{Bengio2009,
  author    = {Bengio, Yoshua and Louradour, J{\'e}r\^{o}me and Collobert, Ronan and Weston, Jason},
  title     = {Curriculum Learning},
  booktitle = {Proceedings of the 26th International Conference on Machine Learning (ICML)},
  year      = {2009},
}

@inproceedings{Belli2019,
  author      = {David Belli and Tomas Kipf},
  title       = {Image-Conditioned Graph Generationfor Road Network Extraction},
eprint = { 1910.14388 },
eprinttype = { arxiv },
  booktitle     = {NeurIPS workshop on Graph Representation Learning},
  year        = {2019},
}

@Book{Goodfellow2016,
  author    = {Ian Goodfellow and Yoshua Bengio and Aaron Courville},
  title     = {Deep Learning},
eprint = { 1807.07987 },
eprinttype = { arxiv },
  year      = {2016},
  publisher = {MIT Press},
  isbn      = {9780262035613},
}

@InProceedings{Anil2019,
  author    = {Cem Anil and James Lucas and Roger Grosse},
  title     = {Sorting out {Lipschitz} function approximation},
eprint = { 1811.05381 },
eprinttype = { arxiv },
  booktitle = {Proceedings of the 36th International Conference on Machine Learning (ICML)},
  year      = {2019},
}

@TECHREPORT{Krizhevsky2009,
    author = {Alex Krizhevsky},
    title = {Learning multiple layers of features from tiny images},
    institution = {},
    year = {2009}
}

@InProceedings{Jaderberg2015,
  author    = {Jaderberg, Max and Simonyan, Karen and Zisserman, Andrew and Kavukcuoglu, Koray},
  title     = {Spatial Transformer Networks},
eprint = { 1506.02025 },
eprinttype = { arxiv },
  booktitle = {Advances in Neural Information Processing Systems 28 (NeurIPS)},
  year      = {2015},
}

@InProceedings{Mnih2014,
  author    = {Mnih, Volodymyr and Heess, Nicolas and Graves, Alex and Kavukcuoglu, Koray},
  title     = {Recurrent Models of Visual Attention},
eprint = { 1406.6247 },
eprinttype = { arxiv },
  booktitle = {Advances in Neural Information Processing Systems 27 (NeurIPS)},
  year      = {2014},
}

@InProceedings{Bahdanau2015,
  author    = {Dzmitry Bahdanau and Kyunghyun Cho and Yoshua Bengio},
  title     = {Neural Machine Translation by Jointly Learning to Align and Translate},
eprint = { 1409.0473 },
eprinttype = { arxiv },
  booktitle = {International Conference on Learning Representations (ICLR)},
  year      = {2015},
}

@InProceedings{Genevay2018,
  author    = {Aude Genevay and Gabriel Peyré and Marco Cuturi},
  title     = {Learning Generative Models with Sinkhorn Divergences},
eprint = { 1706.00292 },
eprinttype = { arxiv },
  booktitle = {Proceedings of the 21st International Conference on Artificial Intelligence and Statistics (AISTATS)},
  year      = {2010},
}

@InProceedings{Sabour2017,
title = {Dynamic Routing Between Capsules},
author = {Sabour, Sara and Frosst, Nicholas and Hinton, Geoffrey E},
  booktitle = {Advances in Neural Information Processing Systems 30 (NeurIPS)},
  year      = {2017},
}

@InProceedings{Finley2005,
  author    = {Thomas Finley and Thorsten Joachims},
  title     = {Supervised Clustering with Support Vector Machines},
  booktitle = {Proceedings of the 22nd International Conference on Machine Learning (ICML)},
  year      = {2005},
}

@inproceedings{Mordatch2018,
  author    = {Igor Mordatch},
  title     = {Concept Learning with Energy-Based Models},
eprint = { 1811.02486 },
eprinttype = { arxiv },
  year      = {2018},
  date      = {2018-11-06},
  booktitle = {ICLR workshop},
}

@inproceedings{Greff2019,
  author      = {Klaus Greff and Raphaël Lopez Kaufmann and Rishab Kabra and Nick Watters and Chris Burgess and Daniel Zoran and Loic Matthey and Matthew Botvinick and Alexander Lerchner},
  title       = {Multi-Object Representation Learning with Iterative Variational Inference},
eprint = { 1903.00450 },
eprinttype = { arxiv },
  year        = {2019},
  booktitle     = {Proceedings of the 36th International Conference on Machine Learning (ICML)},
}

@InProceedings{Lee2018,
  author      = {Kwonjoon Lee and Weijian Xu and Fan Fan and Zhuowen Tu},
  title       = {Wasserstein Introspective Neural Networks},
eprint = { 1711.08875 },
eprinttype = { arxiv },
  booktitle   = {The IEEE Conference on Computer Vision and Pattern Recognition (CVPR)},
  year        = {2018},
}

@InProceedings{Lazarow2017,
  author    = {Lazarow, Justin and Jin, Long and Tu, Zhuowen},
  title     = {Introspective neural networks for generative modeling},
  booktitle = {The IEEE International Conference on Computer Vision (ICCV)},
  year      = {2017},
}

@InProceedings{Rezatofighi2017,
  author    = {Hamid Rezatofighi, S. and Kumar B G, Vijay and Milan, Anton and Abbasnejad, Ehsan and Dick, Anthony and Reid, Ian},
  title     = {{DeepSetNet}: Predicting Sets With Deep Neural Networks},
eprint = { 1611.08998 },
eprinttype = { arxiv },
  booktitle = {The IEEE International Conference on Computer Vision (ICCV)},
  year      = {2017},
}

@inproceedings{Cao2018,
  author      = {Nicola De Cao and Thomas Kipf},
  title       = {{MolGAN}: An implicit generative model for small molecular graphs},
eprint = { 1805.11973 },
eprinttype = { arxiv },
  booktitle     = {ICML workshop on Deep Generative Models},
  year        = {2018},
}

@InProceedings{He2016,
  author    = {Kaiming He and Xiangyu Zhang and Shaoqing Ren and Jian Sun},
  title     = {Deep Residual Learning for Image Recognition},
eprint = { 1512.03385 },
eprinttype = { arxiv },
  booktitle = {The IEEE Conference on Computer Vision and Pattern Recognition (CVPR)},
  year      = {2016},
}

@InProceedings{Simonovsky2018,
  author      = {Martin Simonovsky and Nikos Komodakis},
  title       = {{GraphVAE}: Towards Generation of Small Graphs Using Variational Autoencoders},
eprint = { 1802.03480 },
eprinttype = { arxiv },
  booktitle   = {International Conference on Artificial Neural Networks (ICANN)},
  year        = {2018},
}

@inproceedings{Li2019,
  author      = {Chun-Liang Li and Manzil Zaheer and Yang Zhang and Barnabas Poczos and Ruslan Salakhutdinov},
  title       = {Point Cloud {GAN}},
eprint = { 1810.05795 },
eprinttype = { arxiv },
  year        = {2019},
  booktitle = {ICLR workshop on Deep Generative Models for Highly Structured Data},
}

@inproceedings{Desta2018,
  author      = {Mikyas T. Desta and Larry Chen and Tomasz Kornuta},
  title       = {Object-based reasoning in {VQA}},
eprint = { 1801.09718 },
eprinttype = { arxiv },
  booktitle   = {IEEE Winter Conference on Applications of Computer Vision (WACV)},
  year        = {2018},
  date        = {2018-01-29},
}

@InProceedings{Wagstaff2019,
  author        = {Edward Wagstaff and Fabian B. Fuchs and Martin Engelcke and Ingmar Posner and Michael A. Osborne},
  title         = {On the Limitations of Representing Functions on Sets},
eprint = { 1901.09006 },
eprinttype = { arxiv },
  booktitle     = {Proceedings of the 36th International Conference on Machine Learning (ICML)},
  year          = {2019},
  __markedentry = {[yan:]},
}

@InProceedings{Lee2019,
  author        = {Juho Lee and Yoonho Lee and Jungtaek Kim and Adam R. Kosiorek and Seungjin Choi and Yee Whye Teh},
  title         = {Set Transformer: A Framework for Attention-based Permutation-Invariant Neural Networks},
eprint = { 1810.00825 },
eprinttype = { arxiv },
  booktitle     = {Proceedings of the 36th International Conference on Machine Learning (ICML)},
  year          = {2019},
  __markedentry = {[yan:]},
}

@InProceedings{Zhang2018Counting,
  author        = {Yan Zhang and Jonathon Hare and Adam Pr\"ugel-Bennett},
  title         = {Learning to Count Objects in Natural Images for Visual Question Answering},
eprint = { 1802.05766 },
eprinttype = { arxiv },
  booktitle     = {International Conference on Learning Representations (ICLR)},
  year          = {2018},
  eprint        = {1802.05766},
  __markedentry = {[yan:]},
}

@InProceedings{Ioffe2015,
  author        = {Ioffe, Sergey and Szegedy, Christian},
  title         = {Batch Normalization: Accelerating Deep Network Training by Reducing Internal Covariate Shift},
eprint = { 1502.03167 },
eprinttype = { arxiv },
  booktitle     = {Proceedings of the 32nd International Conference on Machine Learning (ICML)},
  year          = {2015},
  __markedentry = {[yan:]},
}

@InProceedings{Lu2016,
  author        = {Jiasen Lu and Jianwei Yang and Dhruv Batra and Devi Parikh},
  title         = {Hierarchical Question-Image Co-Attention for Visual Question Answering},
eprint = { 1606.00061 },
eprinttype = { arxiv },
  booktitle     = {Advances in Neural Information Processing Systems (NeurIPS)},
  year          = {2016},
}

@InProceedings{You2017,
  author        = {Seungil You and David Ding and Kevin Canini and Jan Pfeifer and Maya Gupta},
  title         = {{Deep Lattice Networks and Partial Monotonic Functions}},
eprint = { 1709.06680 },
eprinttype = { arxiv },
  booktitle     = {Advances in Neural Information Processing Systems 30 (NeurIPS)},
  year          = {2017},
  __markedentry = {[yan:]},
  groups        = {VQA},
  primaryclass  = {stat.ML},
}

@inproceedings{Zhu2017,
  author        = {Chen Zhu and Yanpeng Zhao and Shuaiyi Huang and Kewei Tu and Yi Ma},
  title         = {Structured Attentions for Visual Question Answering},
eprint = { 1708.02071 },
eprinttype = { arxiv },
  year          = {2017},
  booktitle     = {The IEEE International Conference on Computer Vision (ICCV)},
}

@InProceedings{Hosang2017,
  author        = {Hosang, Jan and Benenson, Rodrigo and Schiele, Bernt},
  title         = {Learning Non-Maximum Suppression},
eprint = { 1705.02950 },
eprinttype = { arxiv },
  booktitle     = {The IEEE Conference on Computer Vision and Pattern Recognition (CVPR)},
  year          = {2017},
  __markedentry = {[yan:]},
  groups        = {VQA},
}

@InProceedings{Kim2017,
  author        = {Yoon Kim and Carl Denton and Luong Hoang and Alexander M. Rush},
  title         = {Structured Attention Networks},
eprint = { 1702.00887 },
eprinttype = { arxiv },
  booktitle     = {International Conference on Learning Representations (ICLR)},
  year          = {2017},
  __markedentry = {[yan:]},
  groups        = {VQA},
  timestamp     = {Wed, 07 Jun 2017 14:40:11 +0200},
}

@inproceedings{Brendel2018,
  title={Approximating {CNN}s with Bag-of-local-Features models works surprisingly well on ImageNet},
eprint = { 1904.00760 },
eprinttype = { arxiv },
  author={Wieland Brendel and Matthias Bethge},
  booktitle={International Conference on Learning Representations (ICLR)},
  year={2019},
}

@Article{Zhou2017,
  author        = {Zhou, Yu and Jun, Yu and Chenchao, Xiang and Jianping, Fan and Dacheng, Tao},
  title         = {Beyond Bilinear: Generalized Multi-modal Factorized High-order Pooling for Visual Question Answering},
eprint = { 1708.03619 },
eprinttype = { arxiv },
  year          = {2017},
  eprinttype={arXiv},
  eprint        = {1708.03619},
  __markedentry = {[yan:]},
  groups        = {VQA},
}

@inproceedings{Cohen2017,
  author        = {Joseph Paul Cohen and Henry Z. Lo and Yoshua Bengio},
  title         = {Count-ception: Counting by Fully Convolutional Redundant Counting},
eprint = { 1703.08710 },
eprinttype = { arxiv },
  booktitle       = {ICCV workshop},
  year          = {2017},
}

@InProceedings{Ba2015,
  author        = {Jimmy Ba and Volodymyr Mnih and Koray Kavukcuoglu},
  title         = {Multiple Object Recognition with Visual Attention},
eprint = { 1412.7755 },
eprinttype = { arxiv },
  booktitle     = {International Conference on Learning Representations (ICLR)},
  year          = {2015},
  __markedentry = {[yan:]},
  bibsource     = {dblp computer science bibliography, http://dblp.org},
  groups        = {VQA},
  timestamp     = {Wed, 07 Jun 2017 14:43:09 +0200},
}

@InProceedings{Jabri2016,
  author        = {Allan Jabri and Armand Joulin and Laurens van der Maaten},
  title         = {Revisiting Visual Question Answering Baselines},
eprint = { 1606.08390 },
eprinttype = { arxiv },
  booktitle     = {European Conference on Computer Vision (ECCV)},
  year          = {2016},
  __markedentry = {[yan:]},
  groups        = {VQA},
}

@InProceedings{Lempitsky2010,
  author        = {Lempitsky, Victor and Zisserman, Andrew},
  title         = {Learning To Count Objects in Images},
  booktitle     = {Advances in Neural Information Processing Systems 23 (NeurIPS)},
  year          = {2010},
  __markedentry = {[yan:]},
  groups        = {VQA},
}

@InProceedings{Yang2016,
  author        = {Zichao Yang and Xiaodong He and Jianfeng Gao and Li Deng and Alexander J. Smola},
  title         = {Stacked Attention Networks for Image Question Answering},
eprint = { 1511.02274 },
eprinttype = { arxiv },
  booktitle     = {The IEEE Conference on Computer Vision and Pattern Recognition (CVPR)},
  year          = {2016},
  __markedentry = {[yan:]},
  groups        = {VQA},
  timestamp     = {Thu, 25 May 2017 00:41:19 +0200},
}

@InProceedings{Yang2019,
  author    = {Yang, Carl and Zhuang, Peiye and Shi, Wenhan and Luu, Alan and Li, Pan},
  title     = {Conditional Structure Generation through Graph Variational Generative Adversarial Nets},
  booktitle = {Advances in Neural Information Processing Systems 32 (NeurIPS)},
  year      = {2019},
}

@InProceedings{Bojanowski2018,
  author    = {Piotr Bojanowski and Armand Joulin and David Lopez Paz and Arthur Szlam},
  title     = {Optimizing the Latent Space of Generative Networks},
  eprint = { 1707.05776 },
  eprinttype = { arxiv },
  booktitle = {Proceedings of the 35th International Conference on Machine Learning (ICML)},
  year      = {2018},
}

@InProceedings{Henderson2017,
  author        = {Henderson, Paul and Ferrari, Vittorio},
  title         = {End-to-End Training of Object Class Detectors for Mean Average Precision},
eprint = { 1607.03476 },
eprinttype = { arxiv },
  booktitle     = {Asian Conference on Computer Vision (ACCV)},
  year          = {2017},
  __markedentry = {[yan:]},
  groups        = {VQA},
}

@InProceedings{Chattopadhyay2017,
  author        = {Chattopadhyay, Prithvijit and Vedantam, Ramakrishna and Selvaraju, Ramprasaath R. and Batra, Dhruv and Parikh, Devi},
  title         = {Counting Everyday Objects in Everyday Scenes},
eprint = { 1604.03505 },
eprinttype = { arxiv },
  booktitle     = {The IEEE Conference on Computer Vision and Pattern Recognition (CVPR)},
  year          = {2017},
  __markedentry = {[yan:]},
  groups        = {VQA},
}

@InProceedings{Ren2017,
  author        = {Ren, Mengye and Zemel, Richard S.},
  title         = {End-To-End Instance Segmentation With Recurrent Attention},
eprint = { 1605.09410 },
eprinttype = { arxiv },
  booktitle     = {The IEEE Conference on Computer Vision and Pattern Recognition (CVPR)},
  year          = {2017},
  __markedentry = {[yan:]},
  groups        = {VQA},
}

@inproceedings{Teney2017,
  author        = {Damien Teney and Peter Anderson and Xiaodong He and Anton van den Hengel},
  title         = {Tips and Tricks for Visual Question Answering: Learnings from the 2017 Challenge},
eprint = { 1708.02711 },
eprinttype = { arxiv },
  year={2018},
  booktitle={The IEEE Conference on Computer Vision and Pattern Recognition (CVPR)},
}

@Article{Kazemi2017,
  author        = {Vahid Kazemi and Ali Elqursh},
  title         = {Show, Ask, Attend, and Answer: {A} Strong Baseline For Visual Question Answering},
eprint = { 1704.03162 },
eprinttype = { arxiv },
  year          = {2017},
  eprinttype={arXiv},
  eprint        = {1704.03162},
}

@InProceedings{Azadi2017,
  author        = {Azadi, Samaneh and Feng, Jiashi and Darrell, Trevor},
  title         = {Learning Detection With Diverse Proposals},
eprint = { 1704.03533 },
eprinttype = { arxiv },
  booktitle     = {The IEEE Conference on Computer Vision and Pattern Recognition (CVPR)},
  year          = {2017},
  __markedentry = {[yan:]},
  groups        = {VQA},
}

@InProceedings{Larochelle2010,
  author        = {Larochelle, Hugo and Hinton, Geoffrey E},
  title         = {Learning to combine foveal glimpses with a third-order Boltzmann machine},
  booktitle     = {Advances in Neural Information Processing Systems 23 (NeurIPS)},
  year          = {2010},
  __markedentry = {[yan:]},
}

@InProceedings{Mishkin2016,
  author        = {Dmytro Mishkin and Jiri Matas},
  title         = {All you need is a good init},
eprint = { 1511.06422 },
eprinttype = { arxiv },
  booktitle     = {International Conference on Learning Representations (ICLR)},
  year          = {2016},
  __markedentry = {[yan:]},
  timestamp     = {Wed, 07 Jun 2017 14:41:18 +0200},
}

@Article{Srivastava2014,
  author        = {Nitish Srivastava and Geoffrey Hinton and Alex Krizhevsky and Ilya Sutskever and Ruslan Salakhutdinov},
  title         = {Dropout: A Simple Way to Prevent Neural Networks from Overfitting},
  journal       = {Journal of Machine Learning Research},
  year          = {2014},
  __markedentry = {[yan:]},
   issn = {1532-4435},
 pages = {1929--1958},
 publisher = {JMLR.org},
}

@Article{Hochreiter1997,
  author        = {Sepp Hochreiter and J\"urgen Schmidhuber},
  title         = {Long Short-Term Memory},
  journal       = {Neural Computation},
  year          = {1997},
  __markedentry = {[yan:]},
   doi = {10.1162/neco.1997.9.8.1735},
 publisher = {MIT Press},
 issn = {0899-7667},
 pages = {1735--1780},
}

@inproceedings{Cho2014,
  author        = {Kyunghyun Cho and Bart {van Merrienboer} and Dzmitry Bahdanau and Yoshua Bengio},
  title         = {On the properties of neural machine translation: Encoder-decoder approaches},
eprint = { 1409.1259 },
eprinttype = { arxiv },
  booktitle     = {EMNLP workshop on Syntax, Semantics and Structure in Statistical Translation (SSST)},
  year          = {2014},
  __markedentry = {[yan:]},
}

@InProceedings{Trott2018,
  author        = {Alexander Trott and Caiming Xiong and Richard Socher},
  title         = {Interpretable Counting for Visual Question Answering},
eprint = { 1712.08697 },
eprinttype = { arxiv },
  booktitle     = {International Conference on Learning Representations (ICLR)},
  year          = {2018},
  __markedentry = {[yan:]},
  groups        = {VQA},
  timestamp     = {Wed, 07 Jun 2017 14:43:09 +0200},
}

@inproceedings{Palm2018,
  author    = {Palm, Rasmus and Paquet, Ulrich and Winther, Ole},
  title     = {Recurrent Relational Networks},
eprint = { 1711.08028 },
eprinttype = { arxiv },
  booktitle = {Advances in Neural Information Processing Systems 31 (NeurIPS)},
  year      = {2018},
}

@InProceedings{Hu2018a,
  author    = {Hu, Han and Gu, Jiayuan and Zhang, Zheng and Dai, Jifeng and Wei, Yichen},
  title     = {Relation Networks for Object Detection},
eprint = { 1711.11575 },
eprinttype = { arxiv },
  booktitle = {The IEEE Conference on Computer Vision and Pattern Recognition (CVPR)},
  year      = {2018},
}

@InProceedings{Zambaldi2018,
  author    = {Vinicius Zambaldi and David Raposo and Adam Santoro and Victor Bapst and Yujia Li and Igor Babuschkin and Karl Tuyls and David Reichert and Timothy Lillicrap and Edward Lockhart and Murray Shanahan and Victoria Langston and Razvan Pascanu and Matthew Botvinick and Oriol Vinyals and Peter Battaglia},
  title     = {Deep reinforcement learning with relational inductive biases},
  booktitle = {International Conference on Learning Representations (ICLR)},
  year      = {2019},
}

@inproceedings{Lecun1998,
author={Yann {LeCun} and L\'eon Bottou and Yoshua Bengio and Patrick Haffner},
title={Gradient-based learning applied to document recognition},
year={1998},
    booktitle = {Proceedings of the IEEE},
doi={10.1109/5.726791}
}
\endgroup

\end{document}